\documentclass[10pt,twocolumn,letterpaper]{article}

\usepackage{3dv}
\usepackage{times}
\usepackage{epsfig}
\usepackage{graphicx}
\usepackage{amsmath}
\usepackage{amssymb}
\usepackage{lipsum}
\usepackage{multicol}
\usepackage{cuted}
\usepackage{capt-of}
\usepackage{url}
\usepackage{footnote}
\usepackage{multirow}
\usepackage{rotating}
\usepackage{array}
\usepackage[export]{adjustbox}
\usepackage{booktabs}

\usepackage[pagebackref=true,breaklinks=true,letterpaper=true,colorlinks,bookmarks=false]{hyperref}

\threedvfinalcopy %

\def\threedvPaperID{****} %

\ifthreedvfinal\fi
\begin{document}

\title{Arbitrary Point Cloud Upsampling with Spherical Mixture of Gaussians}

\author{Anthony Dell'Eva\thanks{These authors contributed equally to the work.}\\
University of Bologna \\
Vislab Srl - Ambarella Inc\\
{\tt\small anthony.delleva2@unibo.it}
\and
Marco Orsingher\footnotemark[1]\\
University of Parma \\
Vislab Srl - Ambarella Inc\\
{\tt\small marco.orsingher@unipr.it}
\and
Massimo Bertozzi\\
University of Parma \\
{\tt\small bertozzi@ce.unipr.it}
}

\maketitle
\thispagestyle{empty}

\begin{strip}\centering
\vspace{-1cm}
\begin{tabular}{ c c c c c c }
 \includegraphics[width=0.145\textwidth]{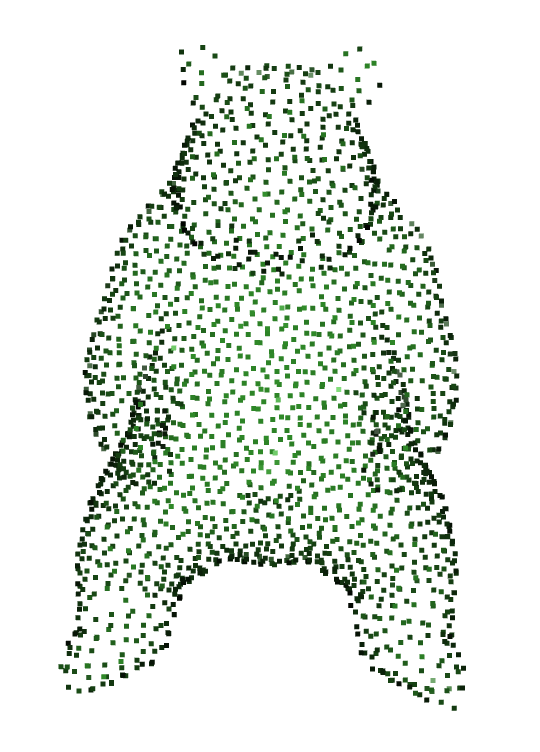} & 
 \includegraphics[width=0.145\textwidth]{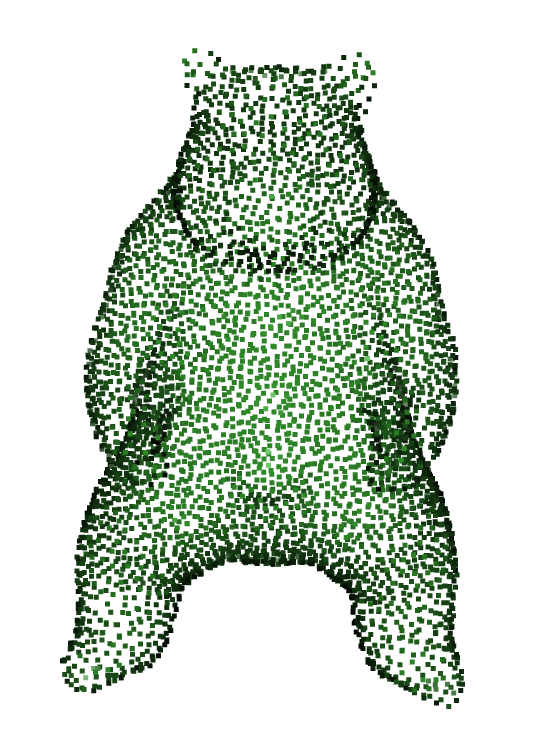} & 
 \includegraphics[width=0.145\textwidth]{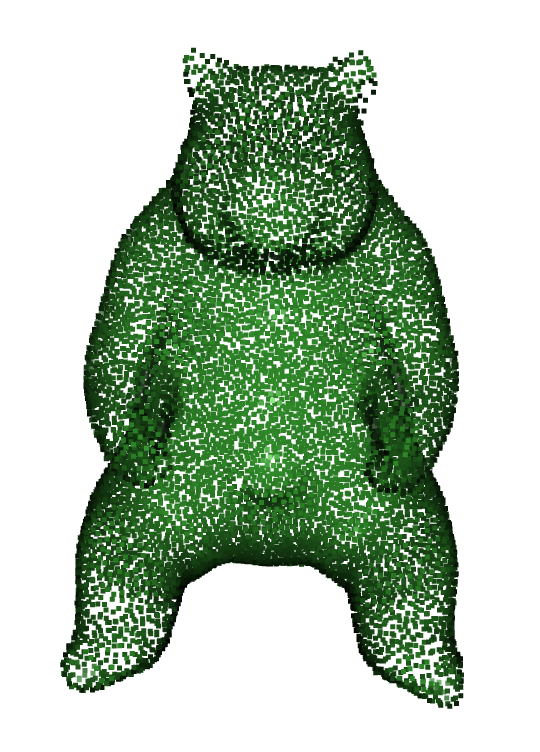} & 
 \includegraphics[width=0.145\textwidth]{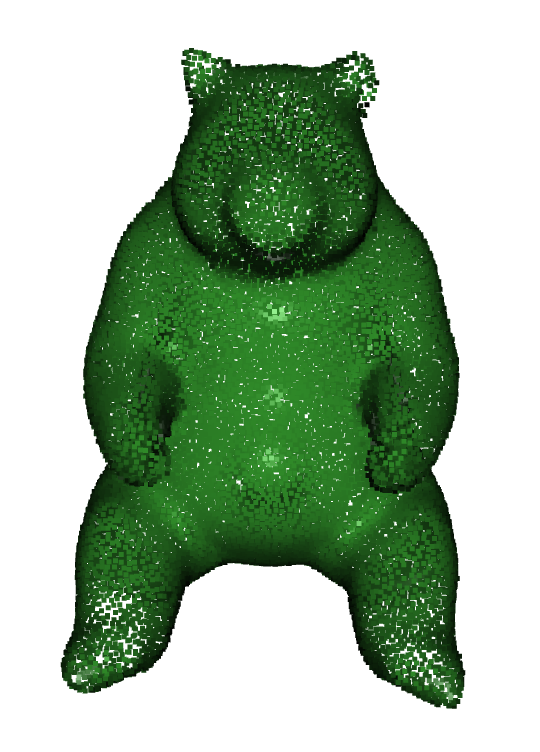} & 
 \includegraphics[width=0.145\textwidth]{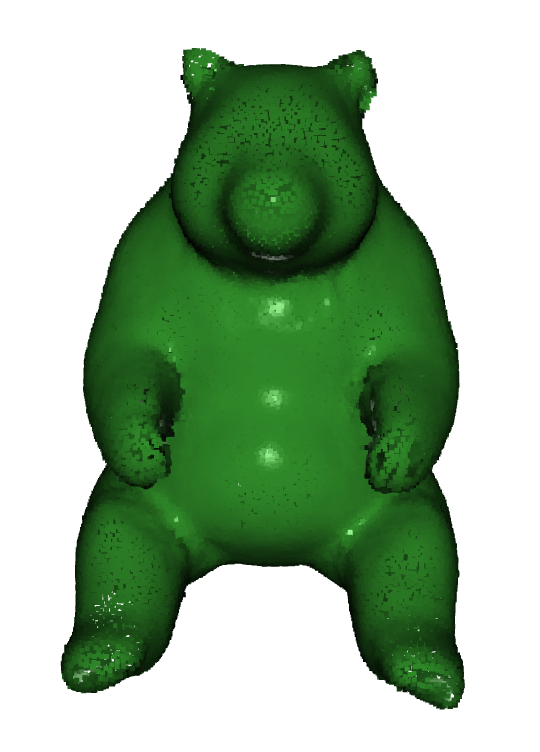} & 
 \includegraphics[width=0.145\textwidth]{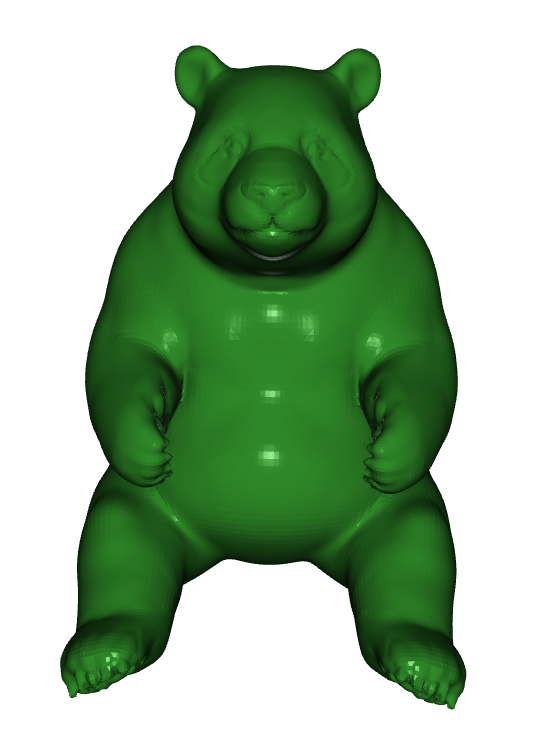} \\
 Input & $r = 2.34$ & $r = 4.93$ & $r = 9.51$ & $r = 22.86$ & GT Mesh
\end{tabular}
\vspace{0.2cm}
\captionof{figure}{The proposed approach can perform point cloud upsampling from a sparse input with $N$ points to a high-resolution output with $r \times N$ points. The upsampling ratio $r \in \mathbb{R}$ can be specified arbitrarily at test time, even if the model is trained a single time with $r = 4$.}
\label{fig:teaser}

\end{strip}

\begin{abstract}
    Generating dense point clouds from sparse raw data benefits downstream 3D understanding tasks, but existing models are limited to a fixed upsampling ratio or to a short range of integer values. In this paper, we present APU-SMOG, a Transformer-based model for Arbitrary Point cloud Upsampling (APU). The sparse input is firstly mapped to a Spherical Mixture of Gaussians (SMOG) distribution, from which an arbitrary number of points can be sampled. Then, these samples are fed as queries to the Transformer decoder, which maps them back to the target surface. Extensive qualitative and quantitative evaluations show that APU-SMOG outperforms state-of-the-art fixed-ratio methods, while effectively enabling upsampling with any scaling factor, including non-integer values, with a single trained model. The code is available at \url{https://github.com/apusmog/apusmog}.
\end{abstract}

\section{Introduction}

Point clouds are a common way to represent 3D data as an unordered list of points, which can be thought of as discrete samples from the underlying surface of the object. In recent years, the wide availability of low-cost scanning sensors has driven the research attention towards 3D point clouds analysis for several applications such as augmented reality, robotics and autonomous driving \cite{pointar, lidarslam, pillar, visualloc}. However, the sparsity and the noise level in raw data from such sensors pose key challenges in point cloud processing for downstream tasks, e.g. classification, segmentation and surface reconstruction. To this end, we focus on point cloud upsampling, which consists in generating a dense and uniform set of points from a sparse and noisy input.

Building on pioneering works for neural point processing~\cite{pointnet, pointnet++}, learning-based methods achieve state-of-the-art results in point cloud upsampling \cite{punet, pugan, pugcn, dispu} and significantly outperform classical optimization-based techniques~\cite{huang, lipman, ear}. One of the main limitations of current approaches is the tight coupling between the upsampling ratio and the network architecture. Typically, this value must be specified in advance and different models must be re-trained from scratch for different rates. Despite recent efforts on designing flexible networks with a user-defined upsampling factor at test time, existing works still limit its value to be integer and lower than a given bound~\cite{mafu,pueva,metapu}, or reconstruct the whole surface with ground truth normals as an intermediate step \cite{neuralpts}.

Our main goal is to remove these limitations and to enable arbitrary upsampling from raw point clouds with a single trained model, as shown in Fig.~\ref{fig:teaser}. The key intuition of the presented method is to split the upsampling procedure in two steps: (i) firstly, the input point cloud is mapped to a probability distribution on a canonical domain; (ii) then, an arbitrary number of points can be sampled from such distribution and mapped back to the target surface. 

Inspired by recent works on point cloud autoencoders~\cite{atlasnet, spgan, cpae}, we propose to use the unit sphere as intermediate representation and we define a Spherical Mixture of Gaussians (SMOG) distribution on such domain. In this way, each input point is associated with a mixture weight and a bivariate Gaussian in spherical coordinates, whose parameters are estimated by a neural network.

The inverse mapping from the samples on the unit sphere to the desired shape is implemented by querying the decoder of a Transformer model \cite{attention} with an arbitrary number of points, which effectively decouples the network architecture and the value of the upsampling ratio. Moreover, we design both a feature extraction backbone and a residual refinement block based on local self-attention, which has proved to learn powerful representations on point cloud data \cite{3detr, pointr, pointtransformer_1}. Therefore, our main contributions can be summarized as follows:
\begin{itemize}
    \item We present a novel approach for point cloud upsampling with arbitrary scaling factors, including non-integer values, with a single trained model.
    \item We propose to learn a mapping from the low-resolution input point cloud to a probability distribution on the unit sphere. This distribution can then be sampled arbitrarily and the inverse mapping is learned to generate the high-resolution output.
    \item We design a fully attention-based network architecture with state-of-the-art performances on multiple benchmarks and strong generalization capabilities.    %
\end{itemize}

\section{Related Work}

\paragraph{Canonical Primitives} In the context of point cloud auto-encoders, a common procedure to generate the output is to feed the decoder with a global latent vector encoding the input and a canonical primitive (e.g. a 2D grid~\cite{foldingnet}), which is deformed to match the target surface. Following the insights in~\cite{atlasnet}, several works use the unit sphere with \textit{uniform} sampling as an intermediate representation~\cite{spgan, cpae}. Recently, TearingNet~\cite{tearingnet} proposed to learn topology-friendly representations by additionally estimating pointwise offsets on the 2D domain. We take a step further by directly learning a mean vector on the unit sphere and 2D variances in spherical coordinates for each point in the input shape. This allows to define a distribution from which an arbitrary number of points are sampled and mapped back to the surface.

\paragraph{Learning-based Upsampling} Point cloud upsampling is an inherently ill-posed problem, since a finite number of samples correspond to many underlying surfaces and viceversa. For this reason, PU-Net~\cite{punet} pioneered the idea of learning geometric priors from data and outperformed previous classical methods~\cite{ear,lipman,huang}. Building on this seminal idea of learning and expanding multi-scale point features, MPU~\cite{mpu} proposes a patch-based progressive strategy for upsampling at different levels of detail, while PU-GAN~\cite{pugan} casts the upsampling procedure in a generative adversarial framework. PU-GCN~\cite{pugcn} introduces several modules built upon graph convolutional network that can be integrated into other architectures, whereas Dis-PU~\cite{dispu} disentangles the upsampling task into two cascaded subnetworks for dense point generation and spatial refinement. Despite showing promising results, all these works require a fixed upsampling ratio $r$ and train different models for varying values of $r$. This strategy does not adapt to real-world point clouds with different quality and it increases the training time significantly.

\paragraph{Arbitrary Upsampling} Recently, a few works~\cite{pueva, mafu, metapu} emerged with the goal of decoupling the upsampling ratio and the network architecture, thus achieving flexible upsampling. Meta-PU~\cite{metapu} employs meta-learning to predict the weights of residual graph convolution blocks dynamically for different values of $r$. However, the model first generates a set of $r_{max} \times N$ points, which are then downsampled to the desired ratio using farthest point sampling (FPS). MAFU~\cite{mafu} and PU-EVA~\cite{pueva} exploit the local geometry of the tangent plane at each point to sample a variable number of candidate points in its neighborhood, but they are limited to \textit{integer} upsampling factors within a predefined range $[r_{min}, r_{max}]$. In addition, the former requires normal vectors information to be trained. On the other hand, our approach is designed to support any value of $r \in \mathbb{R}$, without restrictions. Neural Points \cite{neuralpts} is a concurrent work that performs upsampling with unconstrained ratios by first encoding the continuous underlying surface with neural fields and then sampling an arbitrary number of points from it. Their main limitation is the requirement of ground truth normals for training, which are difficult to estimate with high accuracy, especially for noisy and sparse inputs. Conversely, our model operates in a discrete-to-discrete way on raw point clouds without normals. 

\begin{figure*}[t]
    \centering
    \includegraphics[width=\textwidth]{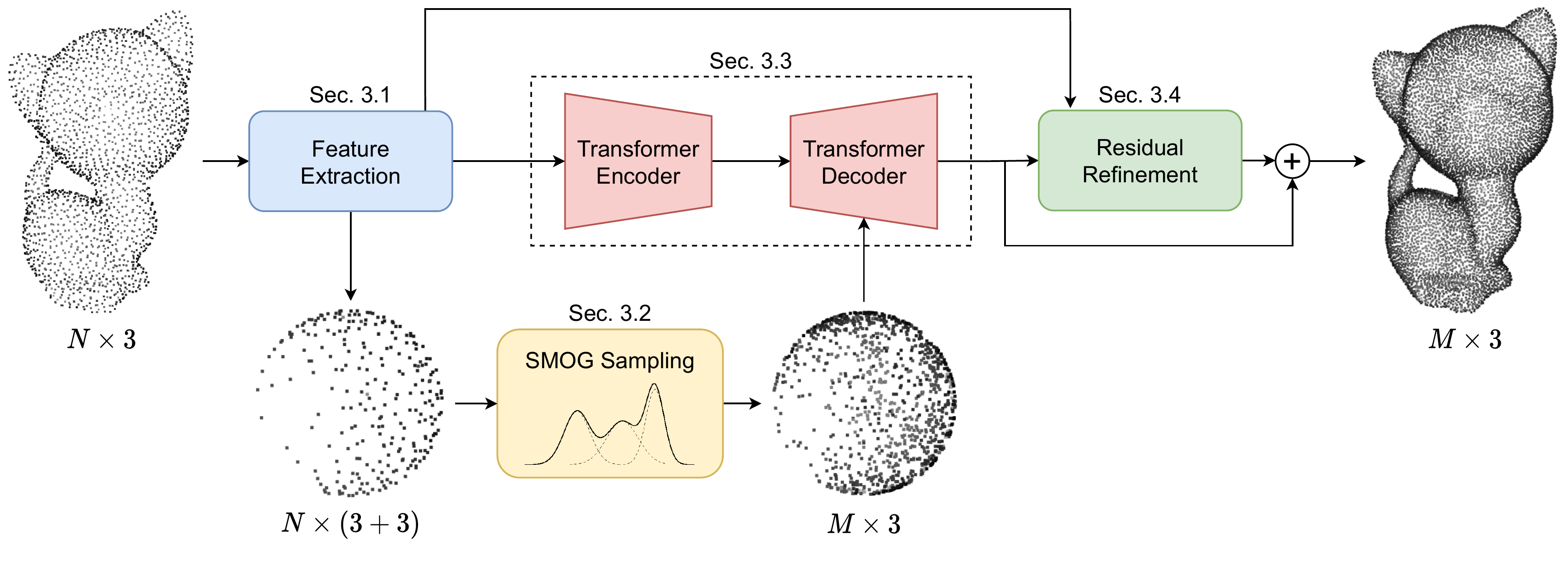}
    \caption{A high-level overview of the method. The low-resolution input point cloud with size $N \times 3$ is firstly mapped to a probability distribution on the unit sphere, with a mean vector and covariance matrix associated to each point. Then, this distribution can be sampled to produce the high-resolution output with size $M \times 3$, being $M = r \times N$ and $r$ the arbitrary upsampling ratio.}
    \label{fig:overview}
\end{figure*}

\paragraph{Transformers for Point Clouds} The Transformer model has revolutionized both natural language processing~\cite{attention} and computer vision~\cite{vit}, thanks to the attention mechanism at its core. Since the Transformer architecture is permutation invariant by design and thus naturally suited for 3D data, early works in the field focused on adapting its modules to point cloud processing~\cite{pointtransformer_1, pointtransformer_2, pct, cloudtransformer}. Attention-based methods have achieved state-of-the-art results in many 3D tasks, such as point cloud completion~\cite{pointr}, object detection~\cite{3detr} and classification~\cite{gumbel}. In this work, we propose to combine the Transformer architecture with an attention-based refinement module for point cloud upsampling. To the best of our knowledge, there is only one concurrent work on this aspect~\cite{putransformer}. However, they solely leverage the Transformer encoder, while we also exploit the possibility of querying the Transformer decoder for arbitrary upsampling.
\section{Method}
\label{method}

Denoting by ${\mathcal{P}=\{\mathbf{p}_i\in\mathbb{R}^3 \}_{i=1}^N}$ the unordered sparse input point cloud of $N$ 3D points, our objective is to generate an arbitrarily denser point set ${\mathcal{Q}_r=\{\mathbf{q}_i\in\mathbb{R}^3 \}_{i=1}^{M}}$ with $M = r \times N$ points, where $r \in \mathbb{R}$ is the upsampling factor. Note that point cloud upsampling is an ill-posed task, since there is not a single feasible correct output. This means that $\mathcal{Q}$ should represent the same underlying surface $\mathcal{S}$, while not being necessarily a superset of $\mathcal{P}$. To this end, we design a fully attention-based end-to-end network for arbitrary point cloud upsampling, taking advantage of Gaussian mixture sampling and Transformer queries to enable flexible ratios. An overview of our framework is shown in Fig.~\ref{fig:overview}. 

The details are explained in the following: in Sec.~\ref{sec:feature_extract} we present our lightweight feature extractor, while Sec.~\ref{sec:smog_sampling} describes the input mapping to a distribution on the unit sphere and the relative sampling. In Sec.~\ref{sec:transformer} we show how to leverage the Transformer model to obtain a flexible number of output points. Lastly, in Sec.~\ref{sec:refinement} we illustrate the refinement of the 3-dimensional coordinates to generate the final upsampled point cloud.

\subsection{Feature Extraction}
\label{sec:feature_extract}

The first step of our pipeline is to extract point-wise features from the $N \times 3$ input point cloud to obtain a $N \times D$ feature map. Differently from previous methods~\cite{pugcn, punet, pugan} that employ either PointNet \cite{pointnet,pointnet++} or DGCNN \cite{dgcnn} as backbone, we design a lightweight network based on Point Transformer \cite{pointtransformer_1}.
For each input point $\mathbf{p}_i \in \mathcal{P}$, its associated feature is computed by the Point Transformer Layer (PTL) as follows:
\begin{equation}
    \mathbf{f}_i = \sum_{\mathbf{p}_j \in \mathcal{N}(\mathbf{p}_i)} \rho(\gamma(\beta(\mathbf{p}_i) - \psi(\mathbf{p}_j) + \delta)) \odot (\alpha(\mathbf{p}_j) + \delta)
    \label{eq:pointtransf}
\end{equation}
where $\mathcal{N}(\mathbf{p}_i)$ is the set of $k$ nearest neighbors of $\mathbf{p}_i$, ${\delta = \eta(\mathbf{p}_i - \mathbf{p}_j)}$ is the positional encoding and the symbol $\odot$  denotes the element-wise product.
The mappings $\alpha$, $\beta$ and $\psi$ are simple linear layers, $\gamma$ and $\eta$ are two-layers MLP, while $\rho$ is the softmax function. Note that in the context of the classical interpretation of the Transformer model~\cite{attention}, $\alpha$ generates the \textit{values}, $\beta$ the \textit{queries} and $\psi$ the \textit{keys}.

\subsection{Spherical Mixture of Gaussians Sampling}
\label{sec:smog_sampling}

The core intuition of our approach is to map the input point cloud into a probability distribution on a canonical domain which can be conveniently sampled. To this end, we estimate the parameters of a Gaussian Mixture Model (GMM) on the unit sphere from the deep features extracted by the PTL. The choice of this domain over a 2D square is motivated by the fact that the unit sphere is a manifold without boundaries, which avoids the truncation of the Gaussian distributions in the GMM.

More specifically, we define a $K$ components GMM $\Gamma = \{w_i, \mu_i, \Sigma_i\}_{i=1}^K$, where $w_i$, $\mu_i$ and $\Sigma_i$ are the mixture weight, mean and covariance of the $i$-th Gaussian and $K = N$. Therefore, the likelihood of a point $\mathbf{z}$ on the unit sphere is given by a weighted combination of individual components $u_i(\mathbf{z})$:

\begin{equation}
    u_\Gamma(\mathbf{z}) = \sum_{i=1}^N w_i u_i(\mathbf{z})
\end{equation}
Each component is weighted equally (i.e. $w_i = \frac{1}{N}$) and its parameters $(\mu_i,\Sigma_i)$ are estimated by a MLP $\xi$:
\begin{equation}
    (\mu_i, \Sigma_i) = \xi(\mathbf{f}_i)
\end{equation}
The MLP predicts the mean vector $\mu_i$ in Cartesian coordinates and the covariance matrix $\Sigma_i$ in spherical coordinates. In particular, the covariance matrix is built by exploiting the constraint of symmetry:
\begin{equation}
    \mathbf{\Sigma}_i = \begin{bmatrix} \sigma_\theta^2 & \sigma_{\theta\phi} \\ \sigma_{\theta\phi} & \sigma_\phi^2 \end{bmatrix}
\end{equation}
where $\sigma_\theta^2$, $\sigma_\phi^2$ and $\sigma_{\theta\phi}$ are the outputs of the covariance head of the MLP corresponding to the azimuth angle $\theta \in [0, 2\pi]$ and the elevation angle $\phi \in [0, \pi]$, respectively. In order to generate a dense output point cloud, we sample an arbitrary number of points from the distribution and learn the inverse mapping from the spherical domain to the target surface. This process is illustrated in Fig.~\ref{fig:smog} and additional details are provided in the supplementary material.

\subsection{Transformer Model}
\label{sec:transformer}

The second core intuition which underpins our work is the possibility of querying the Transformer model~\cite{attention} with an arbitrary number of points. Inspired by the 3DETR architecture~\cite{3detr}, the set of $N\times D$ features produced by our backbone is fed to an encoder to produce a new feature map of dimension $N\times D$. The Transformer encoder has a single layer with a four-headed attention and an MLP with two layers. These features, along with a set of $r\times N$ queries, are processed by the two-layers Transformer decoder to generate $r \times N$ 3-dimensional Euclidean coordinates, which represent the coarse upsampled point cloud. The queries are obtained by embedding with Fourier positional encoding \cite{fourierencoding} the points sampled from the unit sphere manifold. 

\begin{figure}
    \centering
    \includegraphics[width=\linewidth]{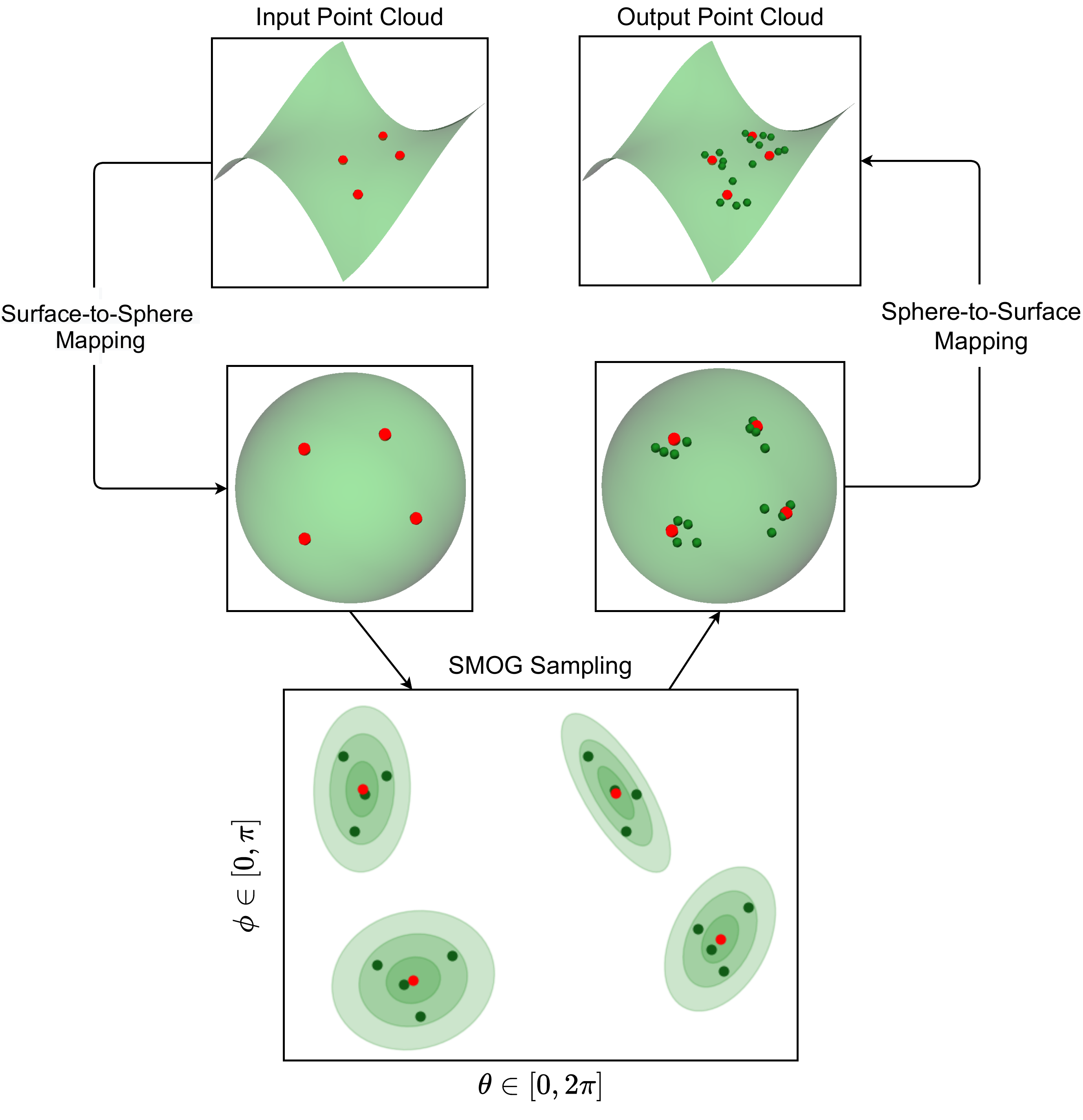}
    \caption{The sampling process shown in detail. Red points are the Gaussian means on the unit sphere associated to each input point, while green points are the arbitrary samples from the SMOG. Red points on the right branch are maintained for visualization.}
    \label{fig:smog}
\end{figure}

\subsection{Residual Refinement}
\label{sec:refinement}

The coarse output from the Transformer decoder might be noisy, non-uniform distributed and have several outliers. Motivated by other works~\cite{dispu, mafu}, we design an attention-based residual refinement with a similar architecture as the feature extractor described in Sec.~\ref{sec:feature_extract}. In particular, we employ a single Point Transformer Block (PTB)~\cite{pointtransformer_1} composed by a PTL, two linear layers, a ReLU activation function and a residual connection. Differently from Eq.~\ref{eq:pointtransf}, in this case the PTL takes as input the local features extracted from the backbone, while the positional encodings are computed with the 3D coarse coordinates.
The $D$-dimensional vector produced by the PTB is finally projected to a $3$-dimensional space with a two-layers MLP that computes the residual for each point, which is added to the coarse prediction to obtain the refined result.

\subsection{Loss Function}
\label{sec:loss}

Let ${\mathcal{\widetilde{Q}}_r=\{\widetilde{\mathbf{q}}_i\in\mathbb{R}^3 \}_{i=1}^{rN}}$ be the coarse prediction generated by the Transformer with upsampling rate $r$, ${\mathcal{Q}_r=\{\mathbf{q}_i\in\mathbb{R}^3 \}_{i=1}^{rN}}$ the refined prediction and ${\mathcal{Y}_r=\{\mathbf{y}_i\in\mathbb{R}^3 \}_{i=1}^{rN}}$ the ground-truth upsampled point cloud. For training, we adopt a similar strategy as in~\cite{neuralpts} to define a distance between two generic point clouds ${\mathcal{X}=\{\mathbf{x}_i\in\mathbb{R}^3 \}_{i=1}^{X}}$ and ${\mathcal{Z}=\{\mathbf{z}_i\in\mathbb{R}^3 \}_{i=1}^{Z}}$ as:
\begin{equation}
    d_{\Pi}(\mathcal{X}, \mathcal{Z}) = \frac{1}{X} \sum_{i=1}^{X} \|\mathbf{x}_i - \Pi(\mathbf{x}_i, \mathcal{Z})\|_2^2
\end{equation}
where $\Pi(\cdot, \cdot)$ is the projection of a 3D point to a point cloud. This term is computed as the weighted combination of the nearest points to $\mathbf{x}$ in $\mathcal{Z}$, with indices $\mathcal{N}(\mathbf{x}, \mathcal{Z})$:
\begin{equation}
    \Pi(\mathbf{x}, \mathcal{Z}) = \frac{\sum_{k \in \mathcal{N}(\mathbf{x},\mathcal{Z})}w_k \mathbf{z}_k}{\sum_{k \in \mathcal{N}(\mathbf{x},\mathcal{Z})}w_k}
\end{equation}
The weights $w_k$ are given by:
\begin{equation}
    w_k = e^{-\alpha \|\mathbf{x} - \mathbf{z}_k \|_2^2},\: k \in \mathcal{N}(\mathbf{x},\mathcal{Z}) 
\end{equation}
with $\alpha = 10^3$.
The corresponding loss is then defined by the following bidirectional sum:
\begin{equation}
    \mathcal{L}_{\Pi}(\mathcal{X}, \mathcal{Z}) = d_{\Pi}(\mathcal{X}, \mathcal{Z}) + d_{\Pi}(\mathcal{Z}, \mathcal{X}) 
\end{equation}
Our arbitrary upsampling network is trained with the sum of three losses:
\begin{equation}
 \begin{split}
    \mathcal{L} = \mathcal{L}_{\Pi}(\mathcal{\widetilde{Q}}_4, \mathcal{Y}_4) +
    \mathcal{L}_{\Pi}(\mathcal{Q}_4, \mathcal{Y}_4) +  \mathcal{L}_{ACD}(\mathcal{\widetilde{Q}}_1, \mathcal{P})
 \end{split}
\end{equation}
where the last term is the Augmented Chamfer Distance (ACD) between the reconstructed point cloud and the input, defined for two generic point clouds as follows:
\begin{equation}
 \begin{split}
    \mathcal{L}_{ACD}(\mathcal{X}, \mathcal{Z}) = \max \left\{\frac{1}{X}\sum_{\mathbf{x}\in \mathcal{X}}\min_{\mathbf{z}\in \mathcal{Z}}\|\mathbf{x} - \mathbf{z}\|_2^2, \right.\\
    \left.\frac{1}{Z}\sum_{\mathbf{z}\in \mathcal{Z}}\min_{\mathbf{x}\in \mathcal{X}}\|\mathbf{x} - \mathbf{z}\|_2^2 \right\}
 \end{split}
\end{equation}

In practice, each training iteration consists of two forward passes and a single backward pass. During the upsampling forward pass, the Transformer decoder is queried with $4 \times N$ points sampled from the SMOG and the projection loss components for both the coarse and refined upsampled outputs are computed. On the other hand, in the reconstruction phase, the estimated means of the SMOG components are fed to the decoder and the ACD with respect to the input point cloud is evaluated. This additional term is required to learn the proper positions of the Gaussian means on the unit sphere, which acts effectively as an intermediate probabilistic representation of the input shape. The ablation studies in Sec.~\ref{sec:ablation} prove this insight with numerical evidence.

\section{Experiments}

\subsection{Settings}

\paragraph{Datasets} In order to compare the proposed APU-SMOG with state-of-the-art \textit{fixed-ratio} methods, we employ the challenging PU1K dataset~\cite{pugcn}, which contains $1147$ synthetic 3D models of various shapes, split into $1020$ training samples and $127$ testing samples. Consistently with existing literature~\cite{pugcn, dispu, pugan, mpu}, $50$ patches are extracted from each training shape with $256$ points as input to the model and $1024$ points as ground truth ($r = 4$). During testing, $2048$ points are sampled from the original mesh with Poisson disk sampling. We employ a similar strategy as in ~\cite{ecnet} to extract overlapping patches with $256$ geodesically close points. The network predicts the upsampled outputs at patch level and the final result is given by combining the overlapping patches with FPS to obtain $8192$ points.

Moreover, we evaluate our approach against \textit{flexible-ratio} methods on the widely used PU-GAN dataset~\cite{pugan}, which is composed by $147$ shapes collected from the PU-Net~\cite{punet}, MPU~\cite{mpu} and Visionair~\cite{visionair} repositories. The same patch-based training and testing procedure is followed with $2048$ points as input, but ground truth point clouds with different sizes are generated to adapt to the values of the scaling factor $r$. Finally, we test the generalization capabilities of our approach on real-world LiDAR point clouds from the KITTI dataset~\cite{kitti}.

\paragraph{Evaluation Metrics} Following previous works, the results are quantitatively evaluated with the Chamfer Distance (CD), Hausdorff Distance (HD) and Point-to-Surface (P2F) distance metrics. The CD is the sum of squared distances between nearest neighbor correspondences of the input and ground truth point clouds, while the HD effectively measures the influence of outliers in the predicted results. On the other hand, the P2F distance is computed against the underlying surface, thus estimating the quality of the upsampled point cloud as an approximation of the real shape. All the metrics are averaged on the whole test set and a lower value indicates better upsampling performance.

\begin{table}
\centering
\setlength\extrarowheight{0.1pt}
\begin{tabular}{ l | c c c c c } 
 \toprule
 \textbf{Method} & \textbf{CD}$\downarrow$ & \textbf{HD}$\downarrow$ & \textbf{P2F $\mu$}$\downarrow$ & \textbf{P2F $\sigma$}$\downarrow$ \\
 \midrule
 PU-Net \cite{punet} & 1.155 & 11.626 & 4.834 & 6.799 \\ 
 MPU \cite{mpu} & 0.935 & 10.298 & 3.551 & 5.971 \\ 
 PU-GAN \cite{pugan} & 0.885 & 16.539 & 3.717 & 5.746 \\ 
 PU-GCN \cite{pugcn} & 0.584 & 5.822 & 2.499 & 4.441 \\ 
 Dis-PU \cite{dispu} & \textbf{0.511} & \underline{4.104} & \underline{2.013} & \underline{2.926}\\ 
 \midrule
 \textbf{Ours} & \underline{0.528} & \textbf{2.549} & \textbf{1.667} & \textbf{2.075} \\ 
 \bottomrule
 \end{tabular}
 \vspace{0.3cm}
 \caption{Quantitative comparison with state-of-the-art methods on the PU1K dataset and $r = 4$. The units are all $10^{-3}$ and lower is better. Best and second results are \textbf{bold} and \underline{underlined}.}
 \label{tab:quant}
\end{table}

\paragraph{Comparison Methods} For the task of upsampling with a fixed ratio $r = 4$ on the PU1K dataset, we provide quantitative comparison against the state-of-the-art models PU-Net~\cite{punet}, MPU~\cite{mpu}, PU-GAN~\cite{pugan}, PU-GCN~\cite{pugcn} and Dis-PU \cite{dispu}. On the other hand, the flexible methods MAFU~\cite{mafu}, PU-EVA~\cite{pueva} and Neural Points~\cite{neuralpts} are used as baselines for arbitrary upsampling on the PU-GAN dataset. For a fair comparison, we used the official pre-trained models when available and re-trained the other ones with the official published code.

\begin{table*}
\centering
\setlength{\tabcolsep}{5pt}
\setlength\extrarowheight{0.1pt}
\begin{tabular}{ c | c | c c c | c c c | c c c | c c c } 
 \toprule
 \multicolumn{2}{c|}{\multirow{2}*[-0.3em]{\textbf{Method}}}
  & \multicolumn{3}{c|}{$r = 4$}
  & \multicolumn{3}{c|}{$r = 8$}
  & \multicolumn{3}{c|}{$r = 12$}
  & \multicolumn{3}{c}{$r = 16$} \\
  \cmidrule{3-14}
  \multicolumn{2}{c|}{} & \textbf{CD}$\downarrow$ & \textbf{HD}$\downarrow$ & \textbf{P2F}$\downarrow$ & \textbf{CD}$\downarrow$ & \textbf{HD}$\downarrow$ & \textbf{P2F}$\downarrow$ & \textbf{CD}$\downarrow$ & \textbf{HD}$\downarrow$ & \textbf{P2F}$\downarrow$ & \textbf{CD}$\downarrow$ & \textbf{HD}$\downarrow$ & \textbf{P2F}$\downarrow$ \\
 \midrule
 {\multirow{3}{*}[-0.2em]{\rotatebox{90}{Fixed}}} & PU-GAN \cite{pugan} & 0.274 & 4.694 & 1.943 & 0.489 & 6.985 & 2.621 & 0.233 & 6.093 & 2.548 & 0.209 & 6.055 & 2.556 \\ 
 & PU-GCN \cite{pugcn} & 0.304 & 2.656 & 2.541 & 0.256 & 4.175 & 2.825 & 0.204 & 4.157 & 2.737 & 0.195 & 4.176 & 2.716 \\ 
 & Dis-PU \cite{dispu} & 0.360 & 5.133 & 2.868 & 0.352 & 7.028 & 3.338 & 0.291 & 6.694 & 3.258 & 0.271 & 6.645 & 3.240 \\ 
 \midrule
 {\multirow{4}{*}[-0.4em]{\rotatebox{90}{Flexible}}} & MAFU \cite{mafu} & \underline{0.322} & \underline{2.116} & \textbf{1.721} & \underline{0.195} & \underline{2.389} & \underline{2.037} & \underline{0.164} & \underline{2.392} & \underline{2.034} & \underline{0.158} & \underline{2.367} & \underline{1.971}  \\ 
 & NePs \cite{neuralpts} & 0.368 & 4.556 & \underline{1.875} & 0.254 & 10.146 & \textbf{1.928} & 0.203 &  10.018 &  \textbf{1.922} & 0.159 & 9.263 & \textbf{1.957}\\
 & PU-EVA \cite{pueva} & 0.394 & 7.676 & 2.915 & 0.322 & 7.951 & 3.148 & 0.290 & 8.191 & 3.234 & 0.286 & 8.390 & 3.269 \\ 
 \cmidrule{2-14}
 & \textbf{Ours} & \textbf{0.276} & \textbf{1.909} & 2.634 & \textbf{0.194} & \textbf{1.628} & 2.613 & \textbf{0.162} & \textbf{1.626} & 2.635 & \textbf{0.149} & \textbf{1.948} & 2.666\\ 
 \bottomrule
 \end{tabular}
 \vspace{0.3cm}
 \caption{Quantitative comparison with state-of-the-art methods on the PU-GAN dataset and flexible upsampling ratios. The units are all $10^{-3}$ and lower is better. Best and second results \textit{among flexible methods} are \textbf{bold} and \underline{underlined}. Fixed methods are included as reference.}
 \label{tab:flex}
\end{table*}

\paragraph{Implementation Details} Our framework is implemented in PyTorch~\cite{pytorch}. The features dimension is set to $D = 128$. The MLPs in the Transformer model has hidden dimension equal to $64$ in the encoder and $128$ in the decoder. For the local feature extractor and the refinement module, the number of neighboring points are set to $32$ and $4r$, respectively. The projection components in the loss function are weighted with $\lambda_1 = \lambda_2 = 0.01$, while the reconstruction component weight is set to $\lambda_3 = 1$.  In the $\Pi(\cdot, \cdot)$ computation, $k$ is equal to $4$.
The model is trained with a batch size of 64 for 100K iterations on a single V100 GPU with 32GB of RAM, using random rotation and perturbation as data augmentation techniques to avoid overfitting.
The AdamW~\cite{adamw} optimizer is used with a learning rate decay following a cosine schedule~\cite{cosineschedule} from $5\mathrm{e}{-4}$ to $1\mathrm{e}{-6}$, a weight decay of $0.1$ and a gradient $L_2$ norm clipping of $0.1$.

\begin{table*}
\centering
\addtolength{\tabcolsep}{-2pt} 
\begin{tabular}{ c c c c c c}
    \includegraphics[width=0.15\linewidth]{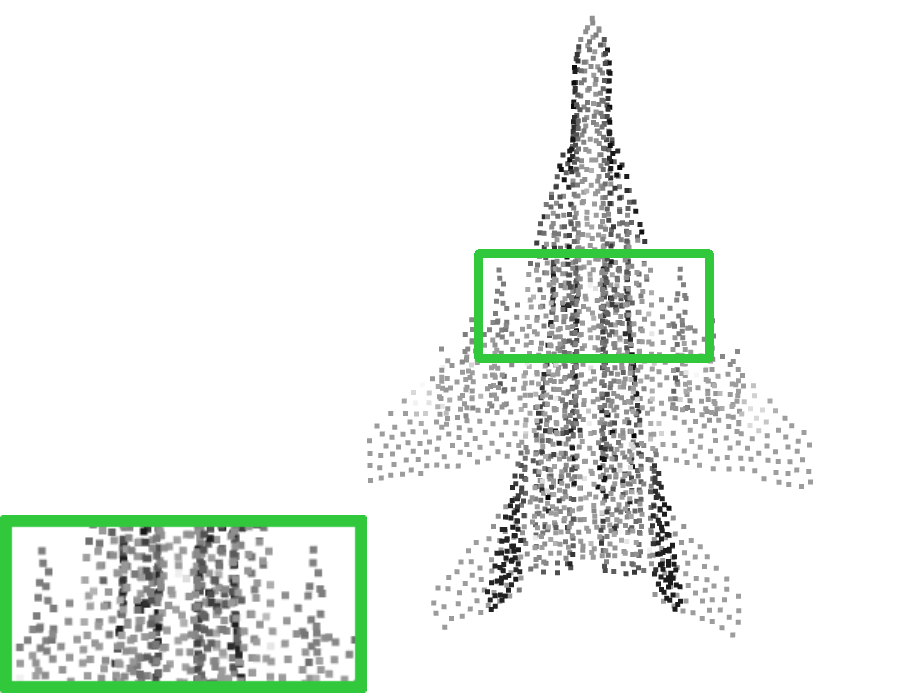} &
    \includegraphics[width=0.15\linewidth]{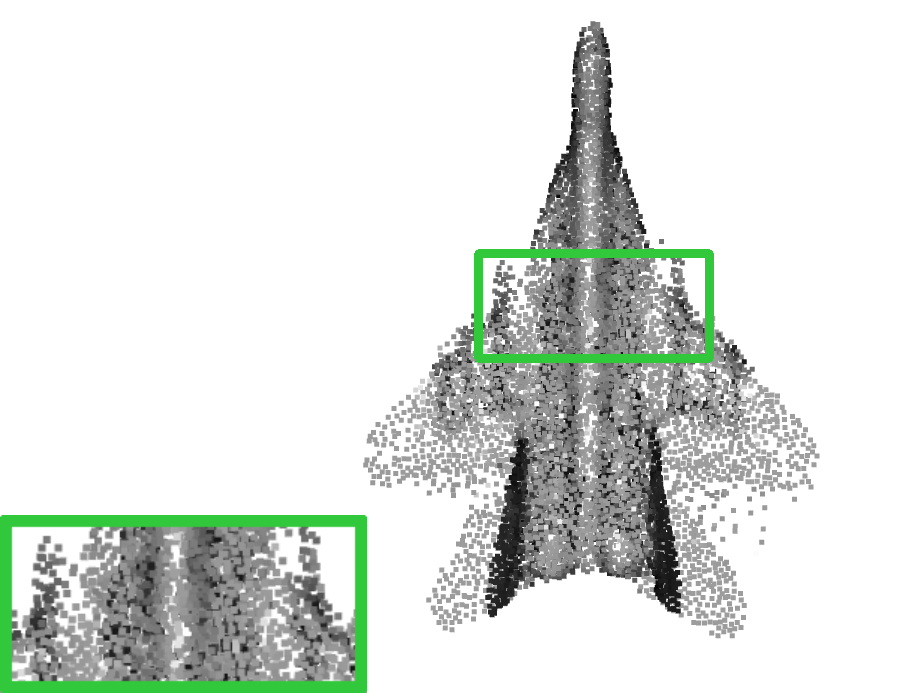} &
    \includegraphics[width=0.15\linewidth]{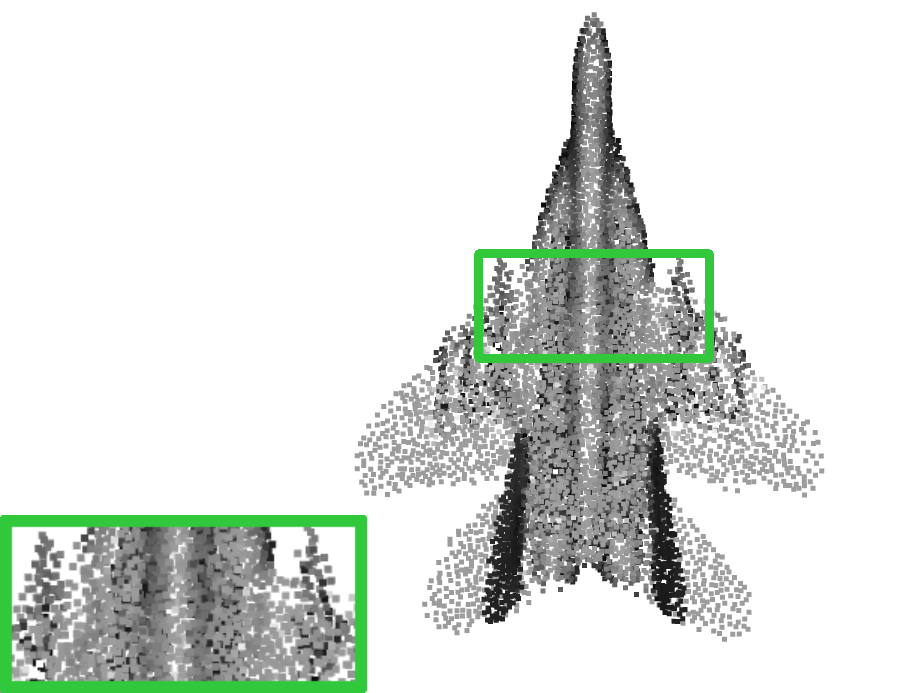} &
    \includegraphics[width=0.15\linewidth]{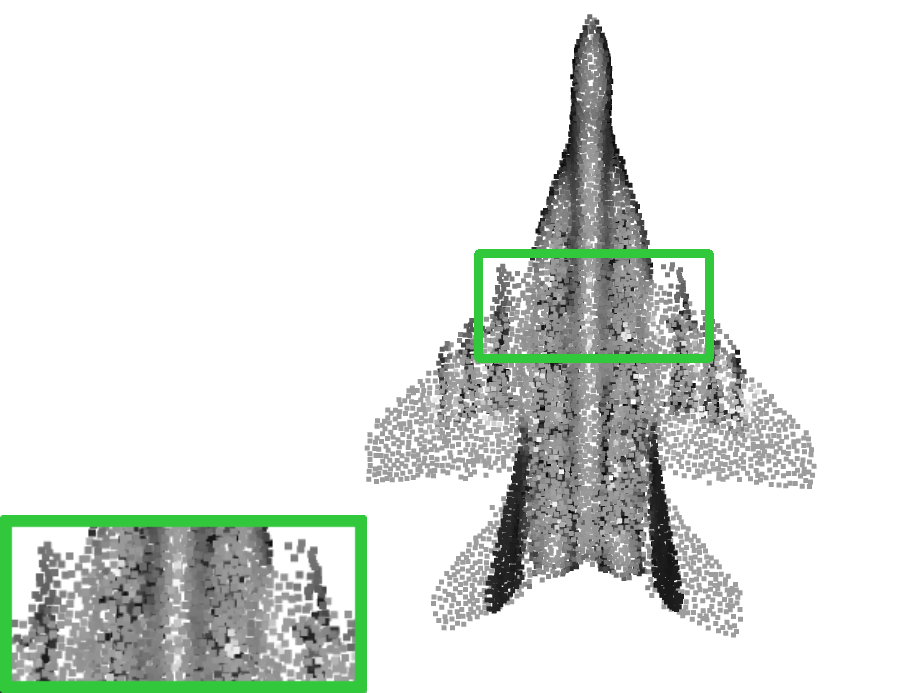} &
    \includegraphics[width=0.15\linewidth]{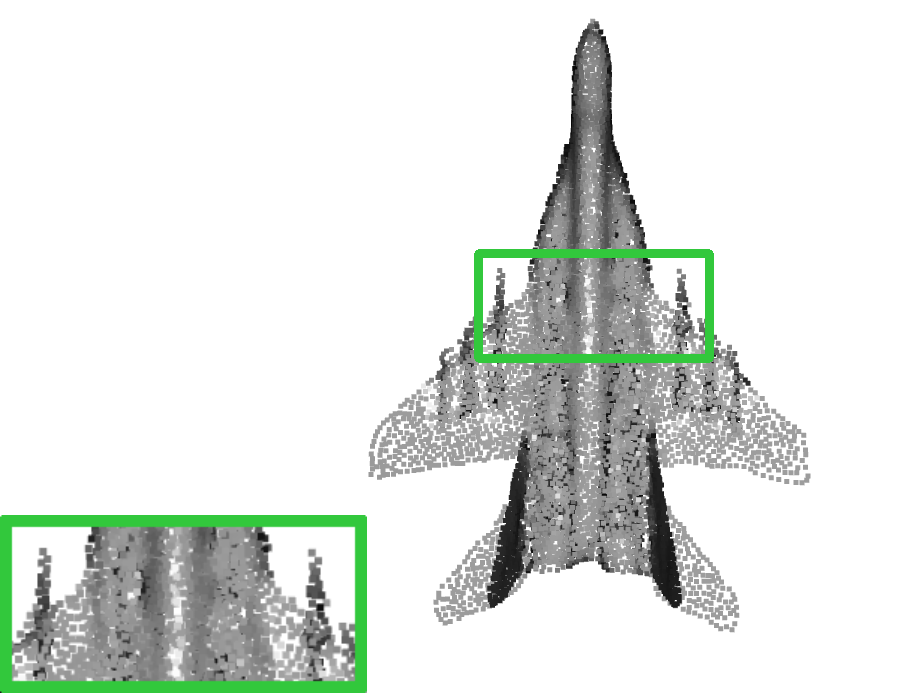} &
    \includegraphics[width=0.15\linewidth]{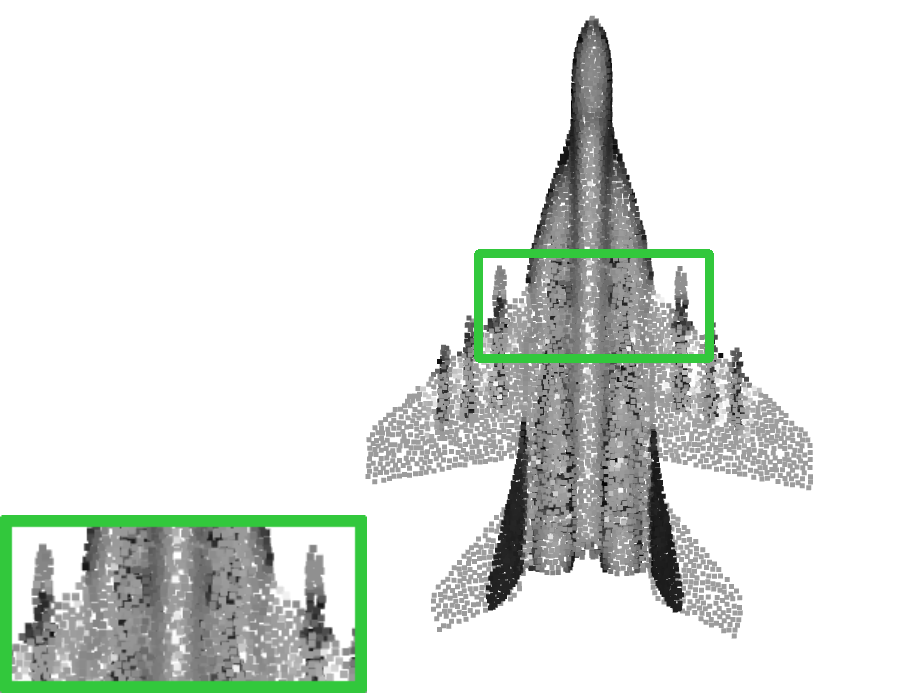} \\
    \includegraphics[width=0.15\linewidth]{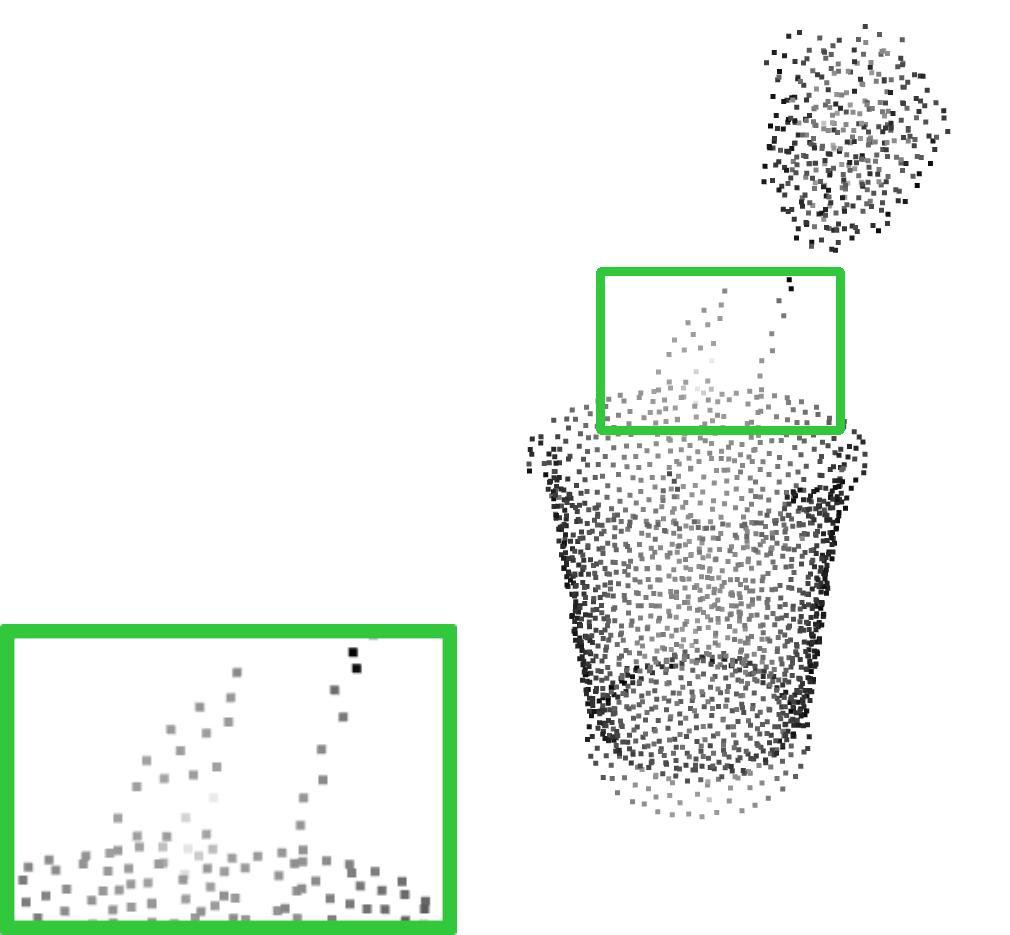} &
    \includegraphics[width=0.15\linewidth]{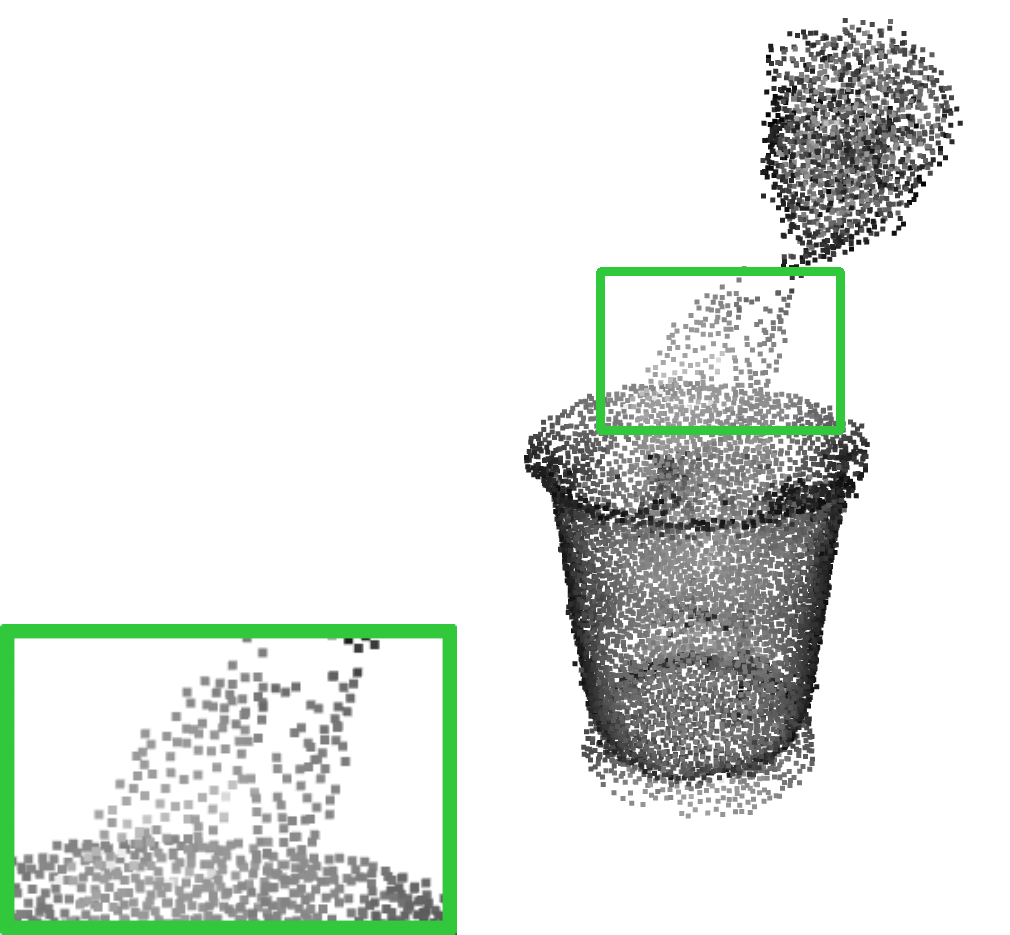} &
    \includegraphics[width=0.15\linewidth]{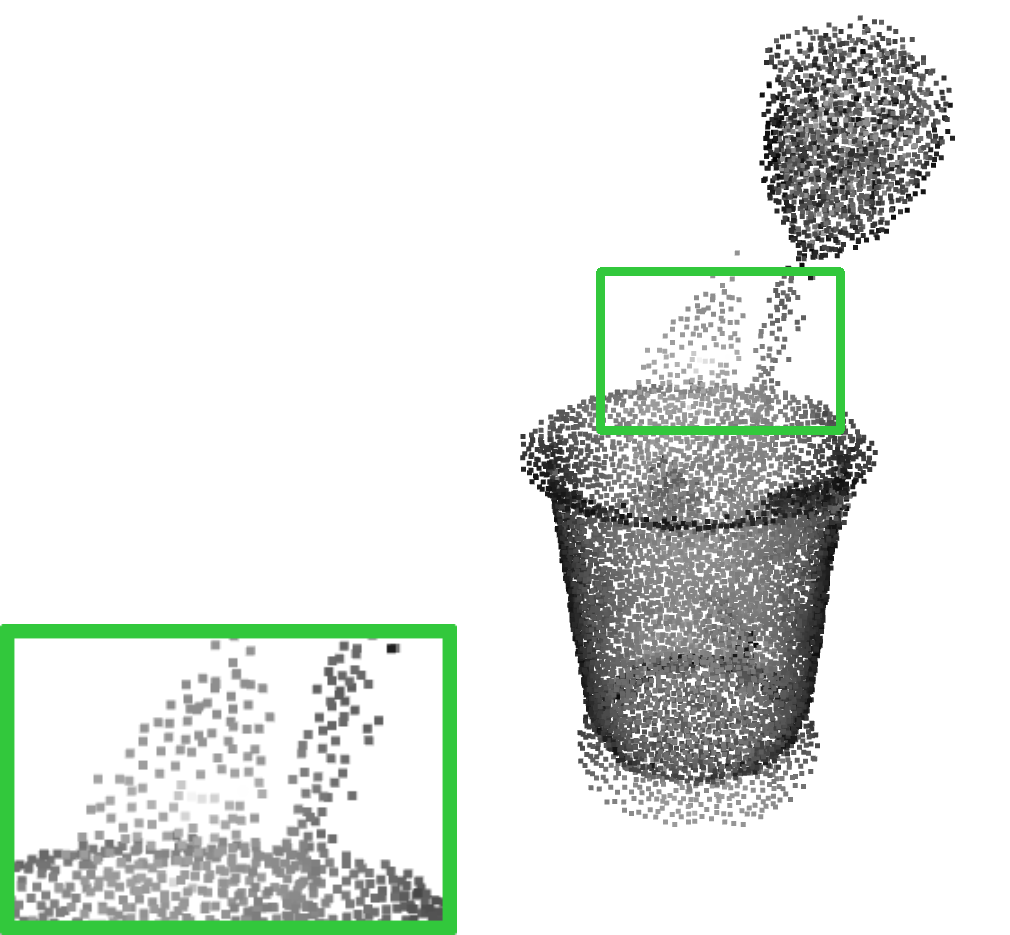} &
    \includegraphics[width=0.15\linewidth]{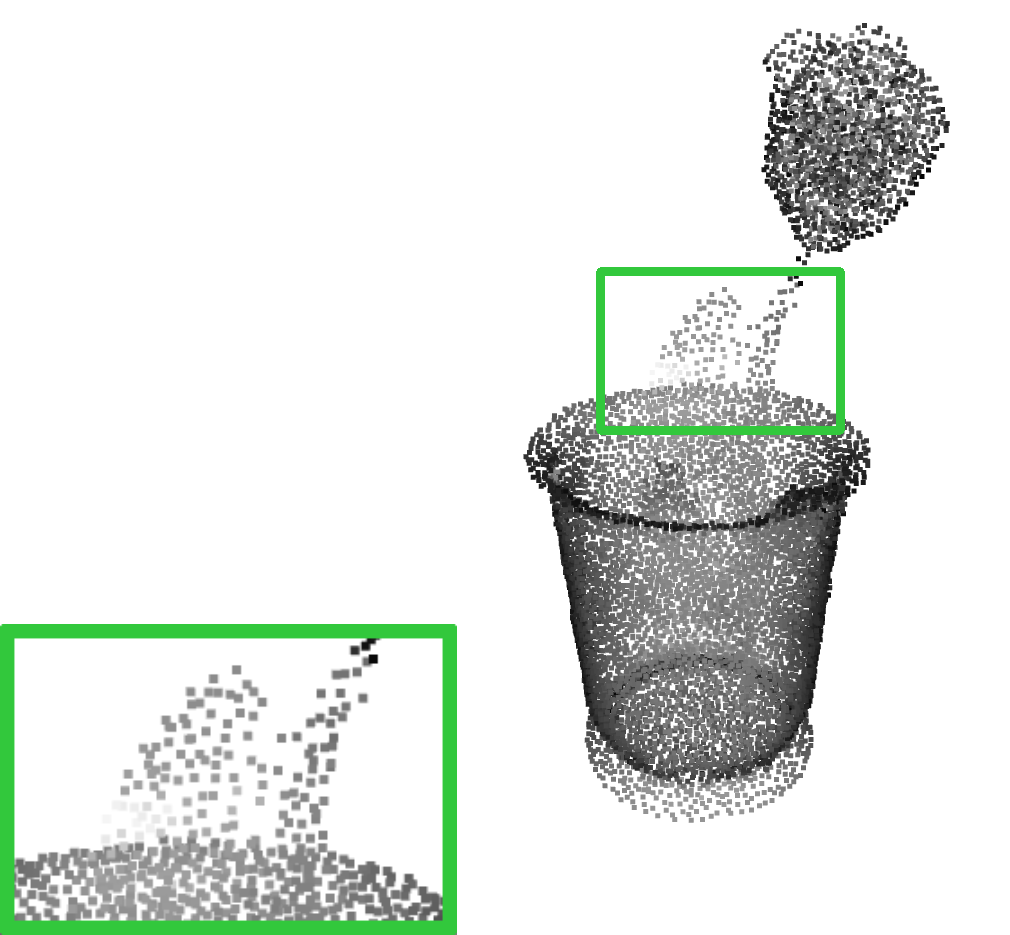} &
    \includegraphics[width=0.15\linewidth]{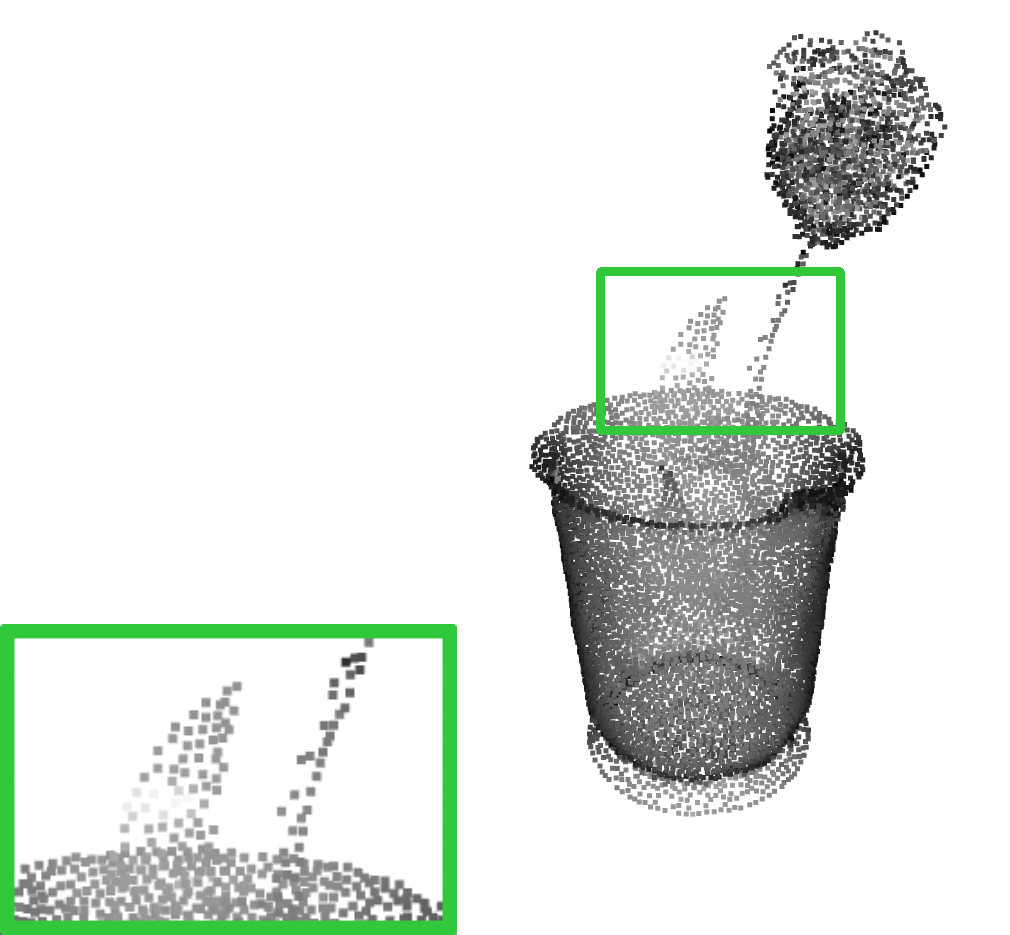} &
    \includegraphics[width=0.15\linewidth]{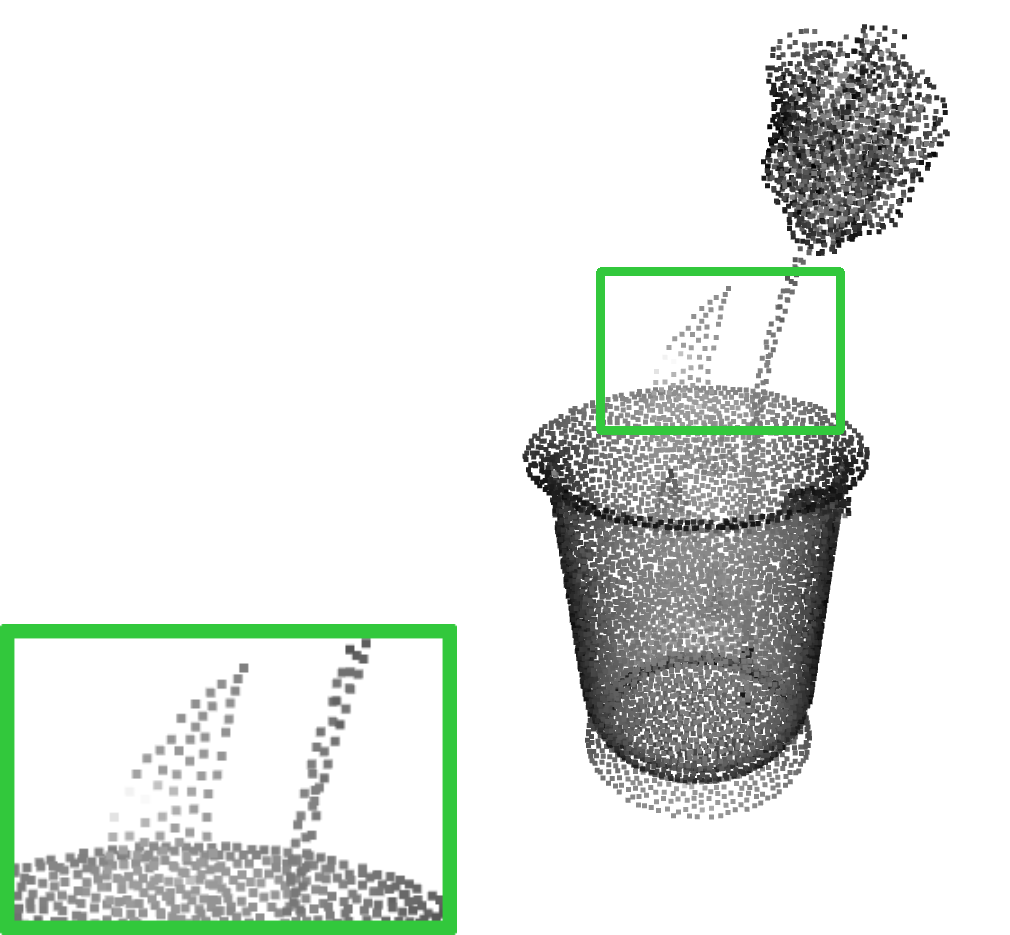} \\
    \includegraphics[width=0.15\linewidth]{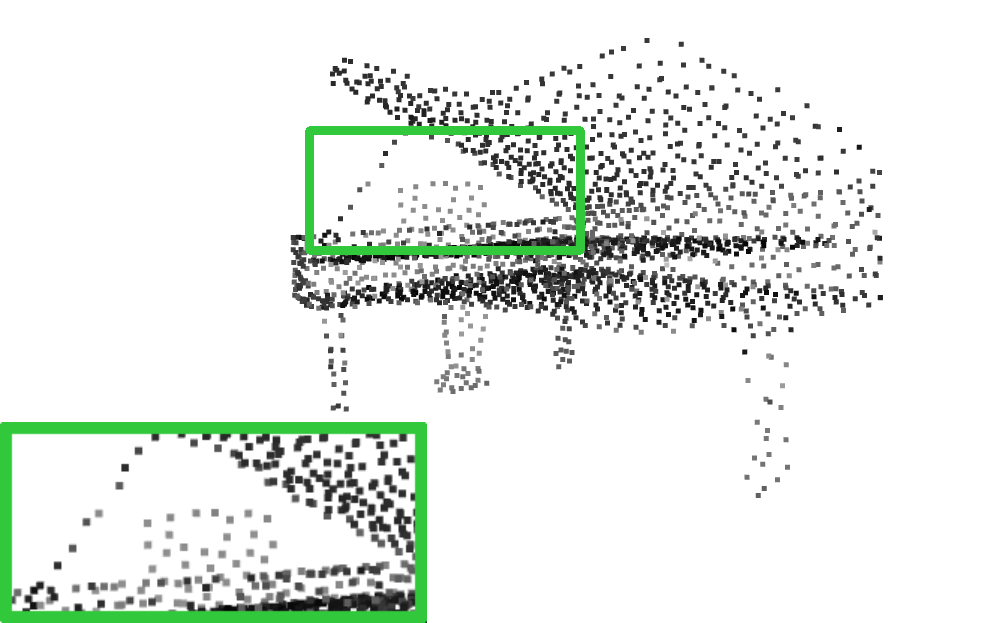} &
    \includegraphics[width=0.15\linewidth]{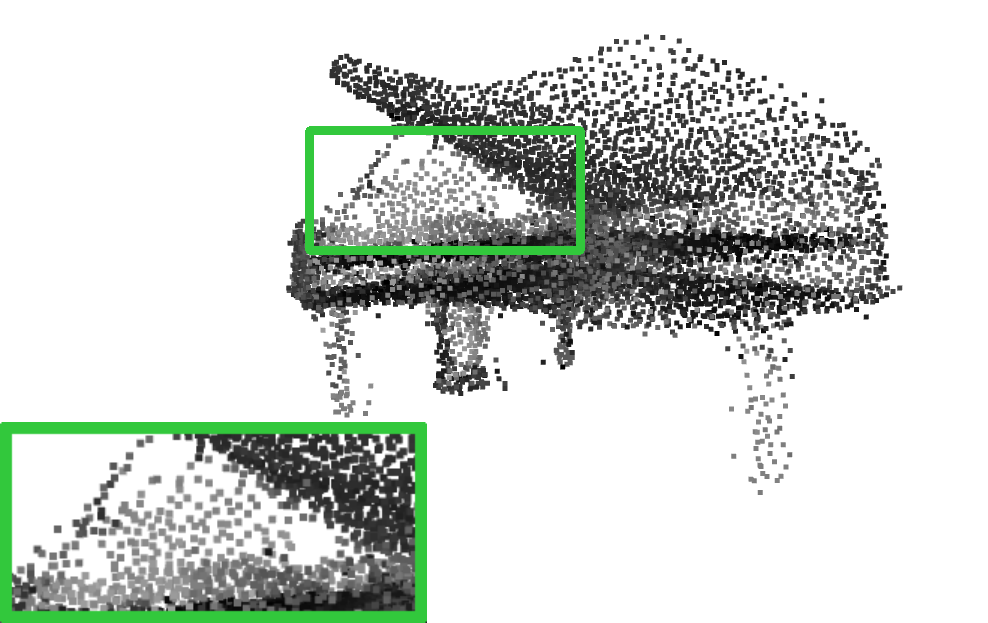} &
    \includegraphics[width=0.15\linewidth]{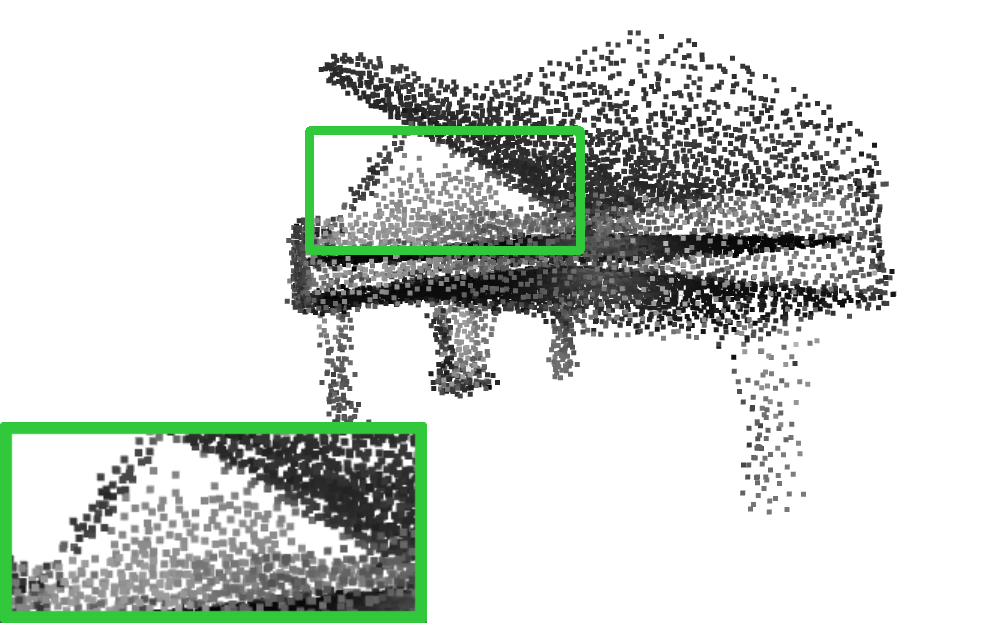} &
    \includegraphics[width=0.15\linewidth]{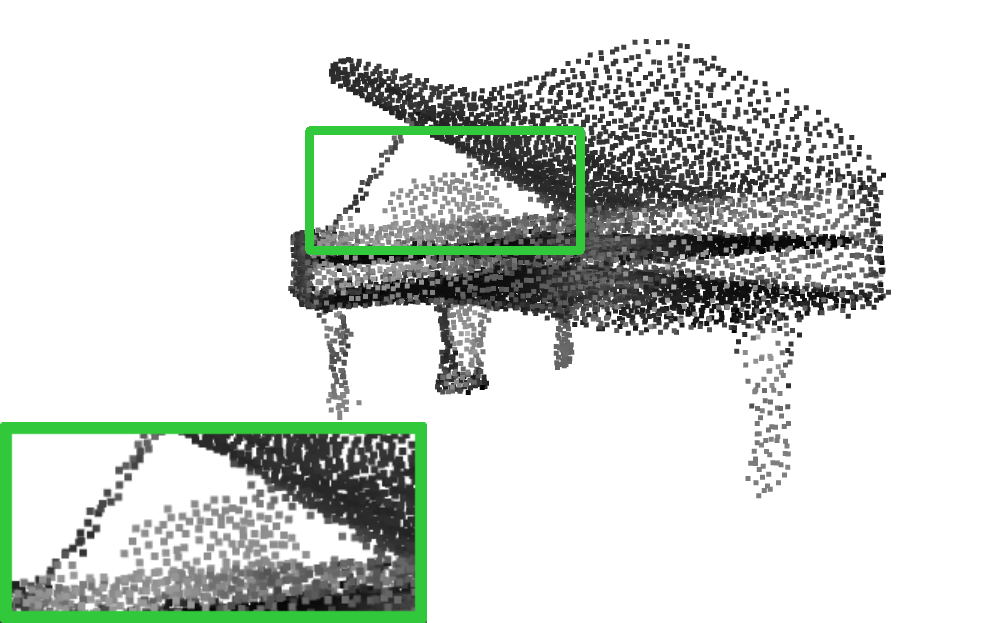} &
    \includegraphics[width=0.15\linewidth]{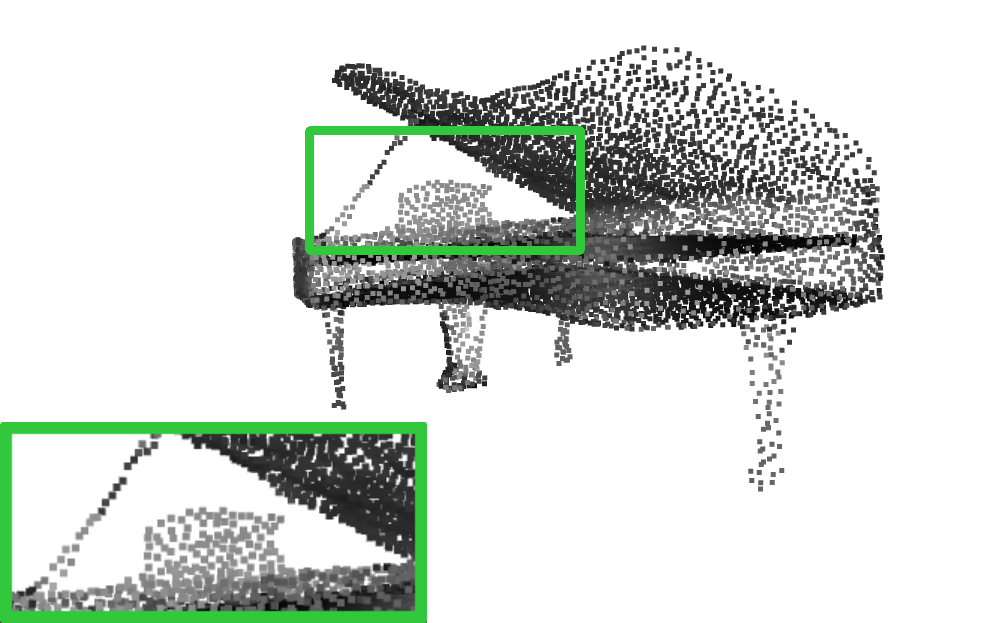} &
    \includegraphics[width=0.15\linewidth]{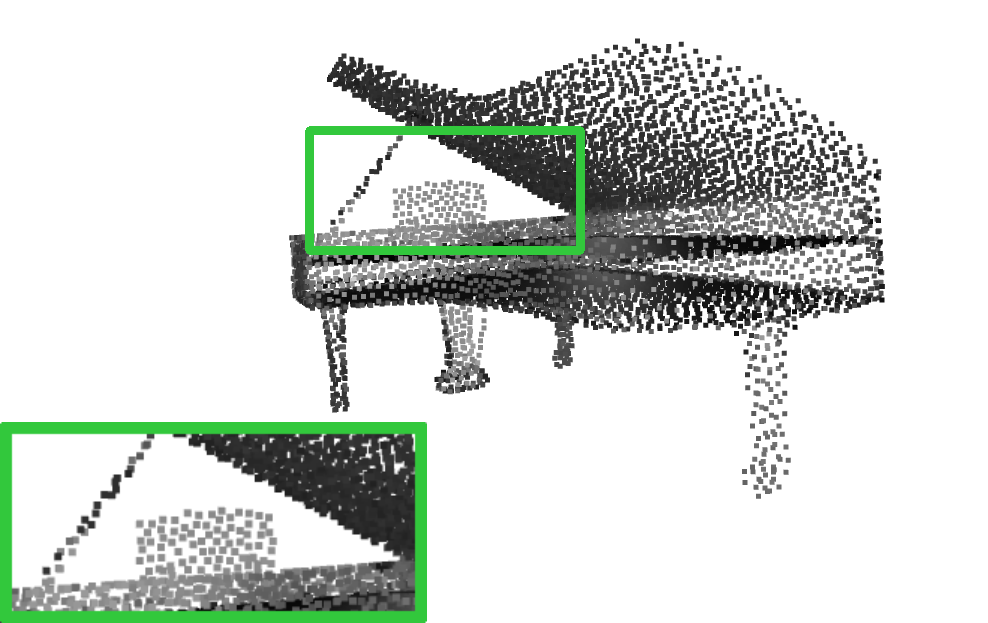} \\
    \includegraphics[width=0.15\linewidth]{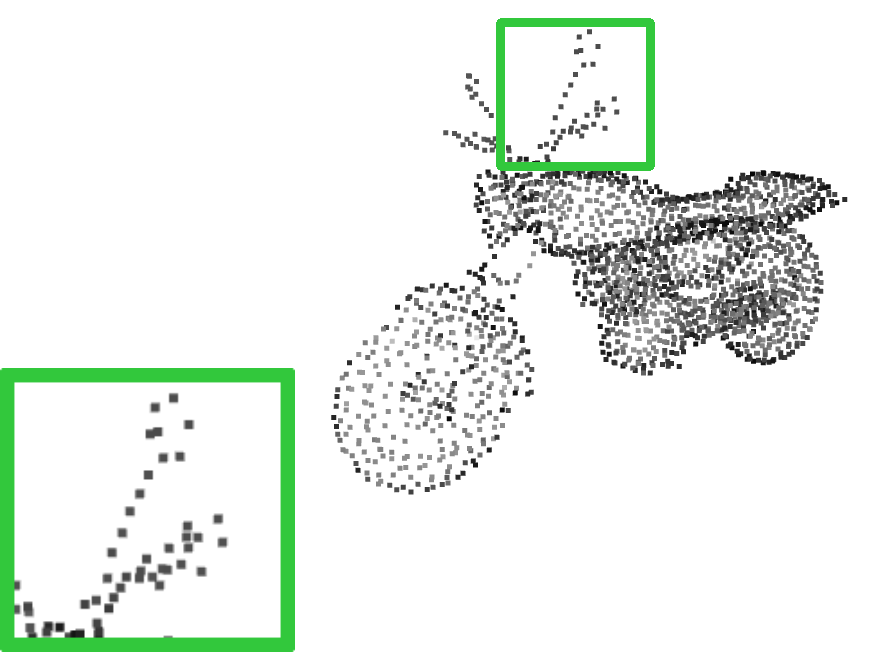} &
    \includegraphics[width=0.15\linewidth]{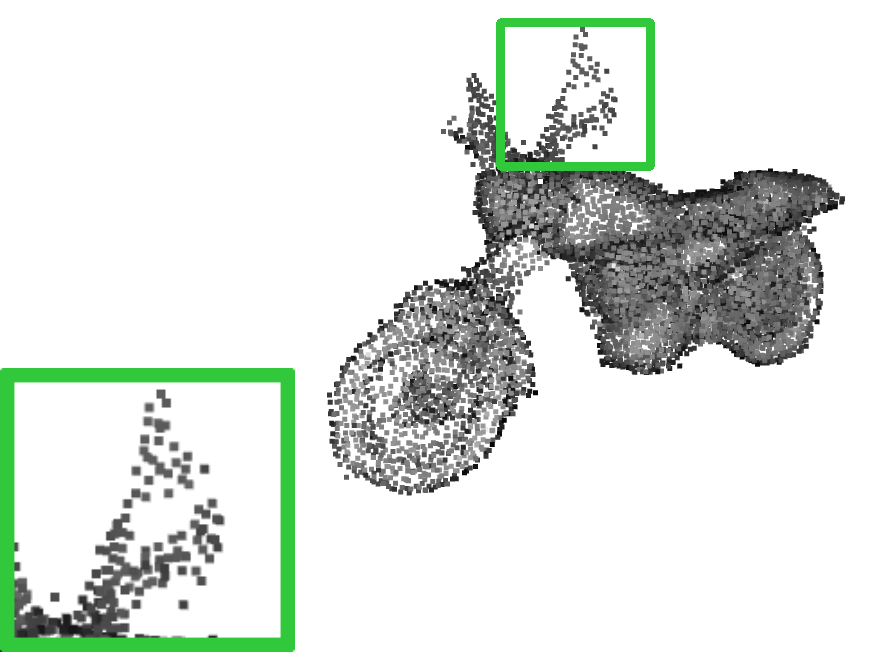} &
    \includegraphics[width=0.15\linewidth]{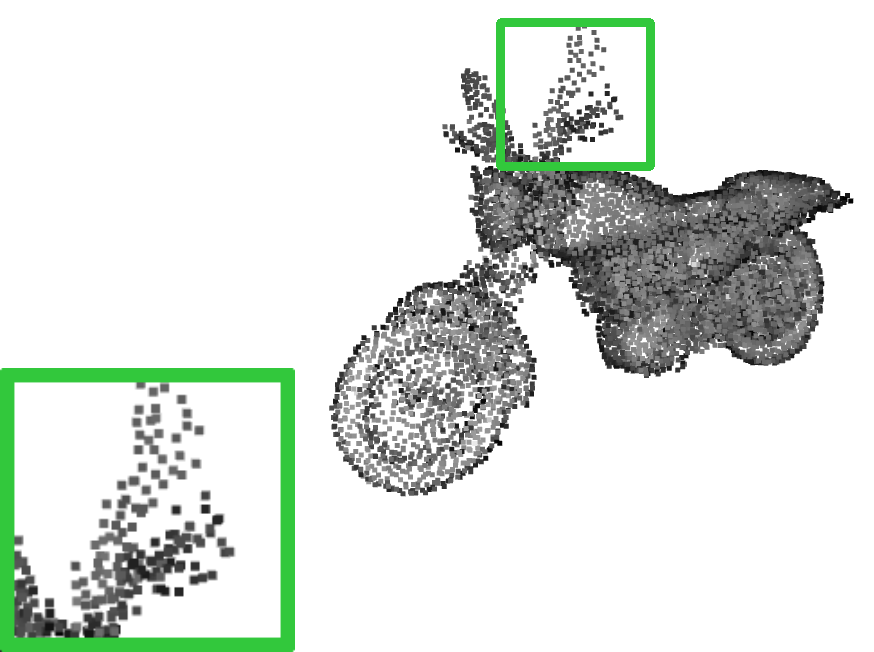} &
    \includegraphics[width=0.15\linewidth]{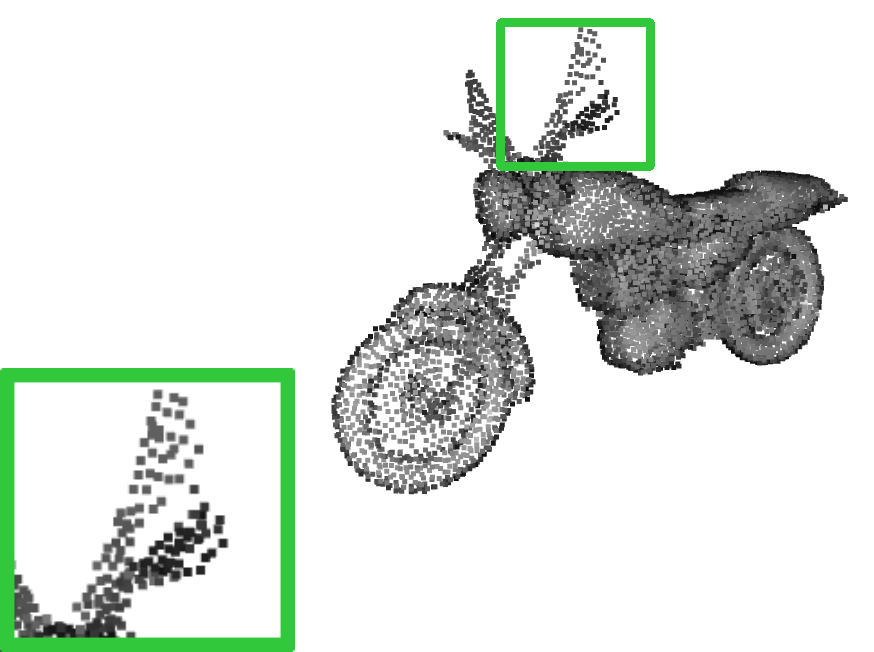} &
    \includegraphics[width=0.15\linewidth]{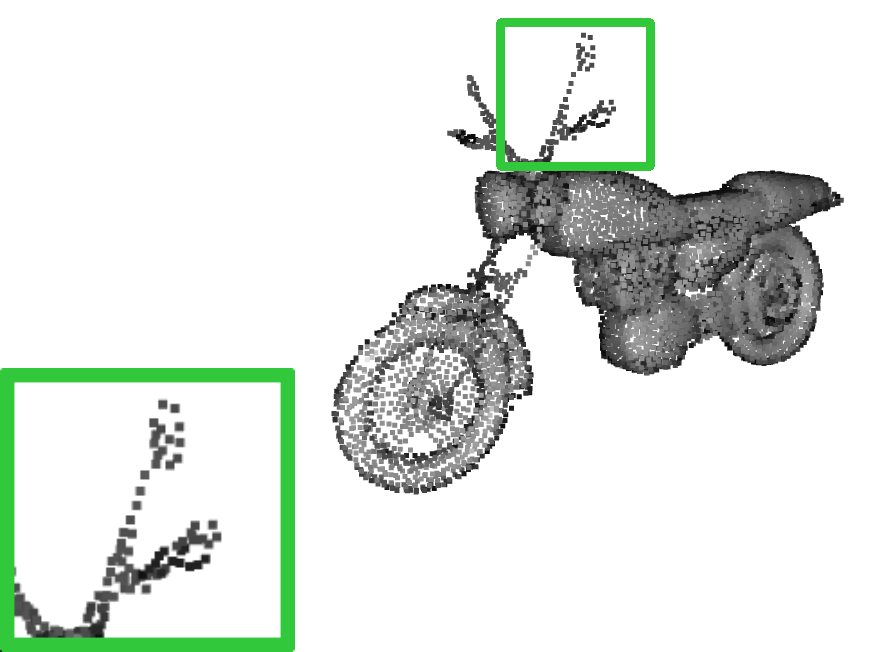} &
    \includegraphics[width=0.15\linewidth]{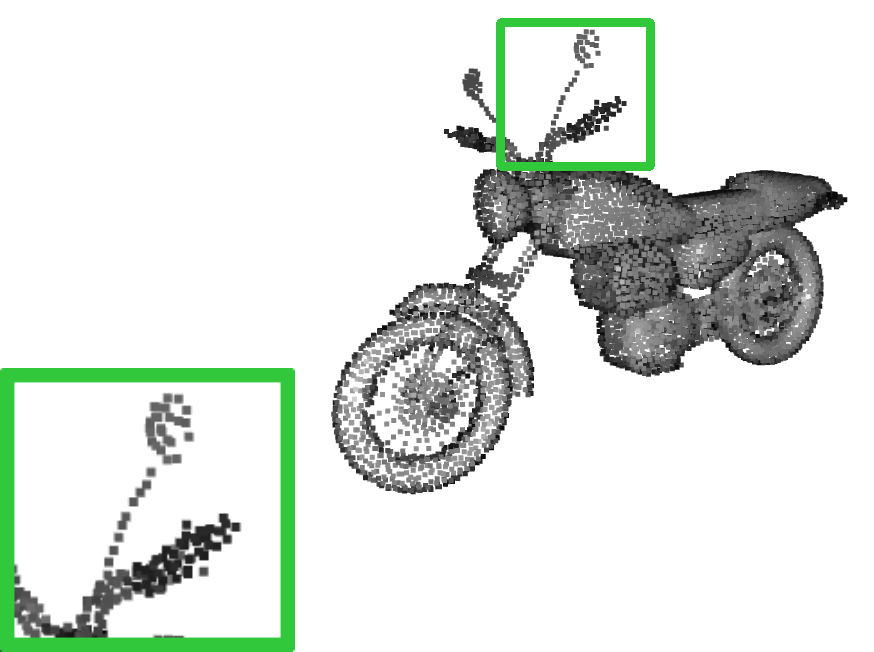} \\
    \includegraphics[width=0.15\linewidth]{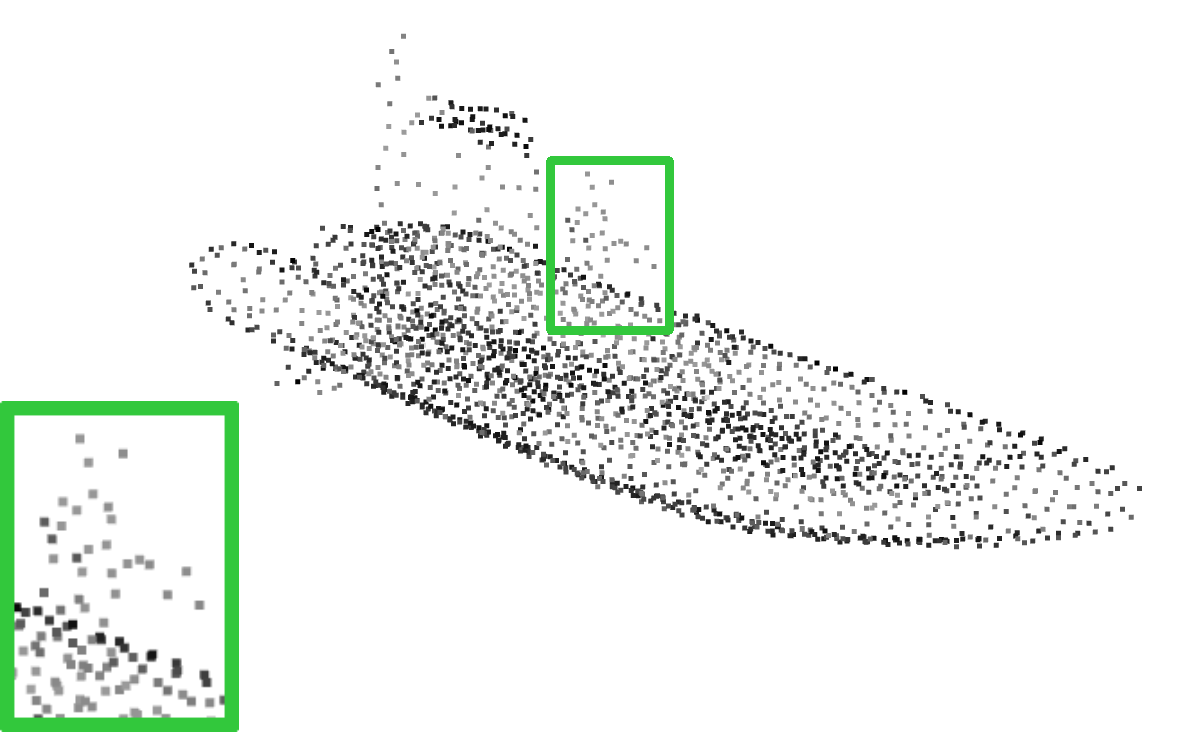} &
    \includegraphics[width=0.15\linewidth]{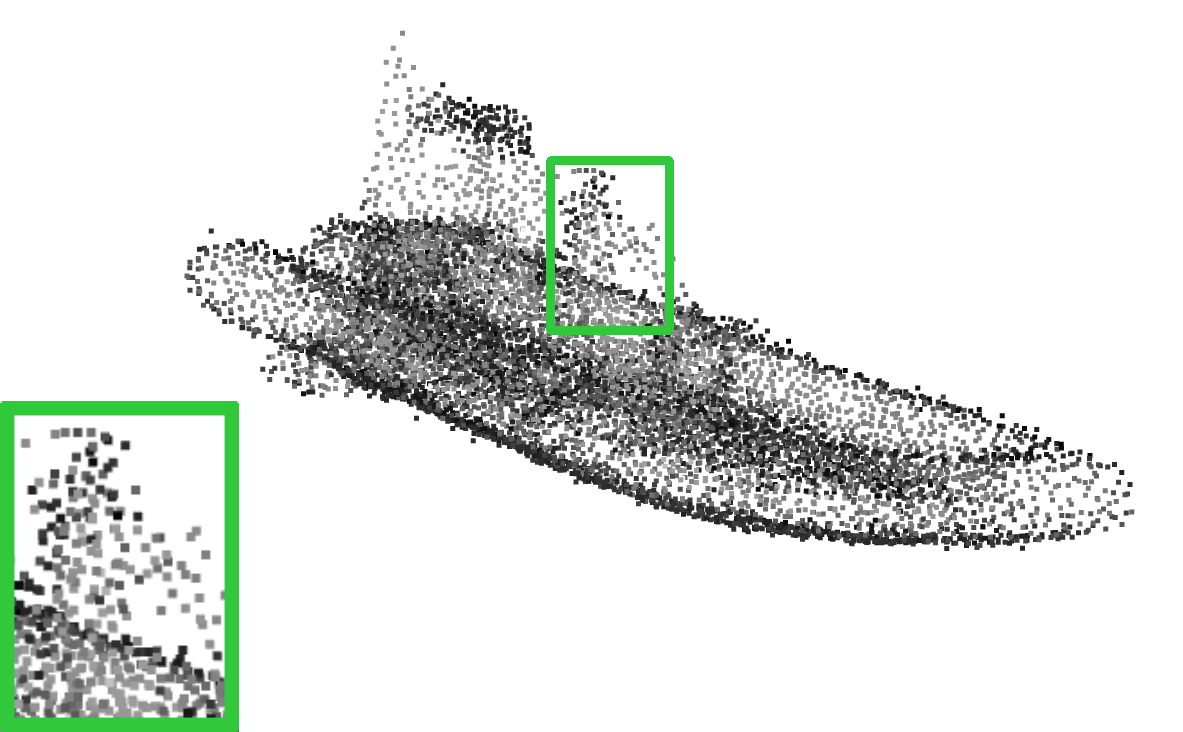} &
    \includegraphics[width=0.15\linewidth]{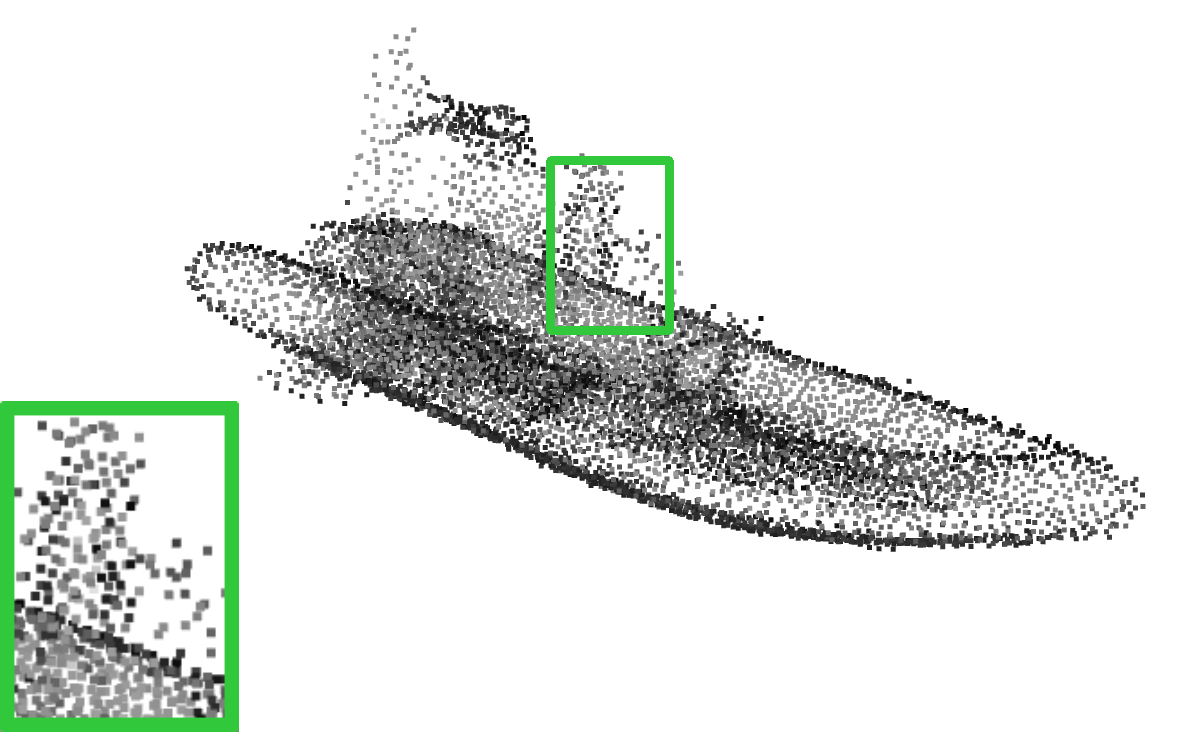} &
    \includegraphics[width=0.15\linewidth]{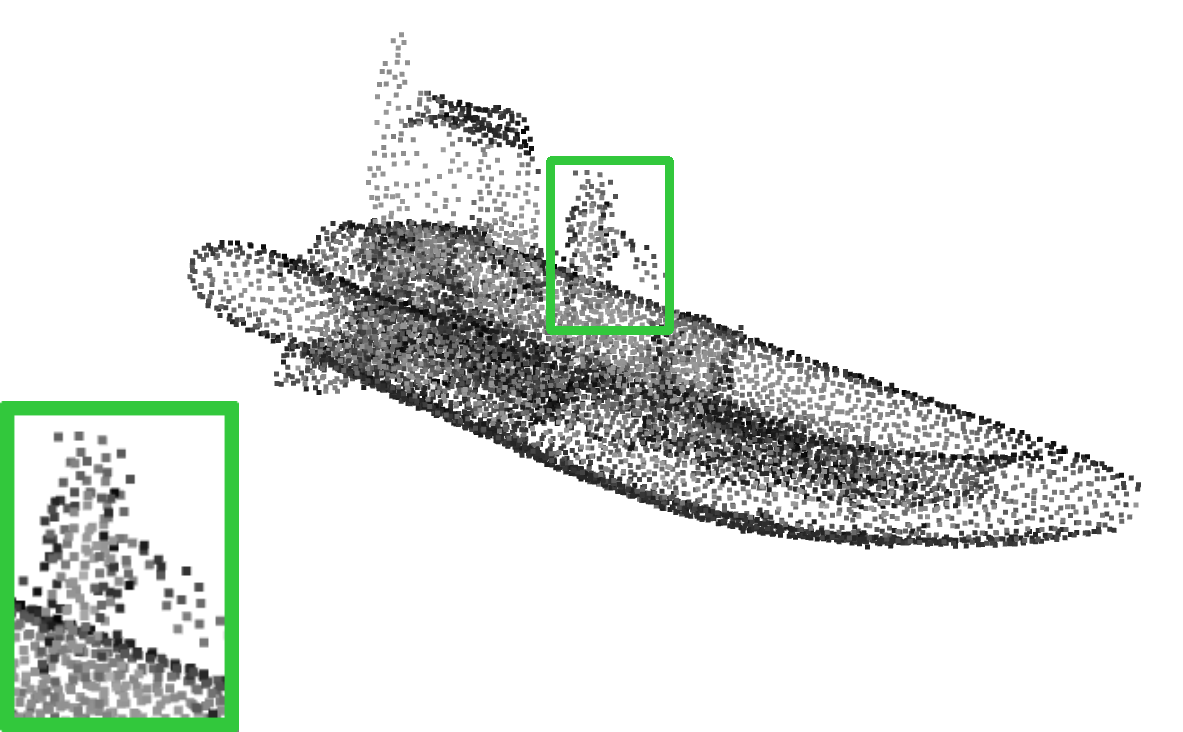} &
    \includegraphics[width=0.15\linewidth]{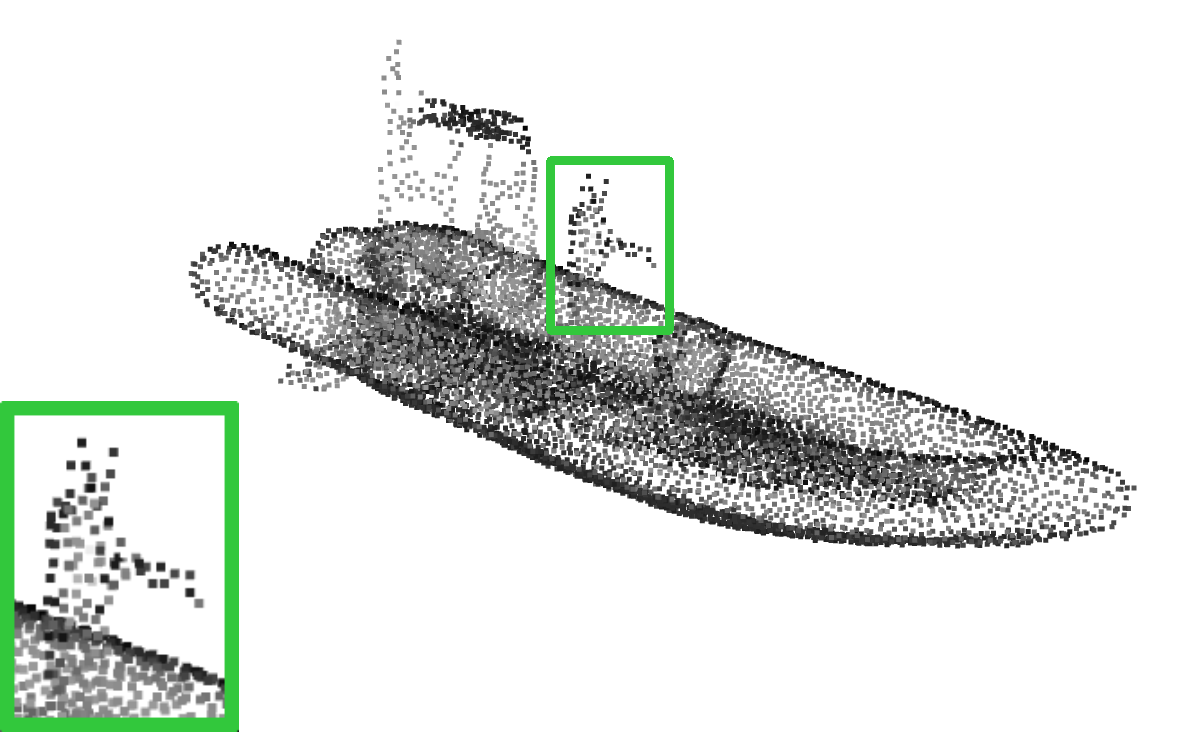} &
    \includegraphics[width=0.15\linewidth]{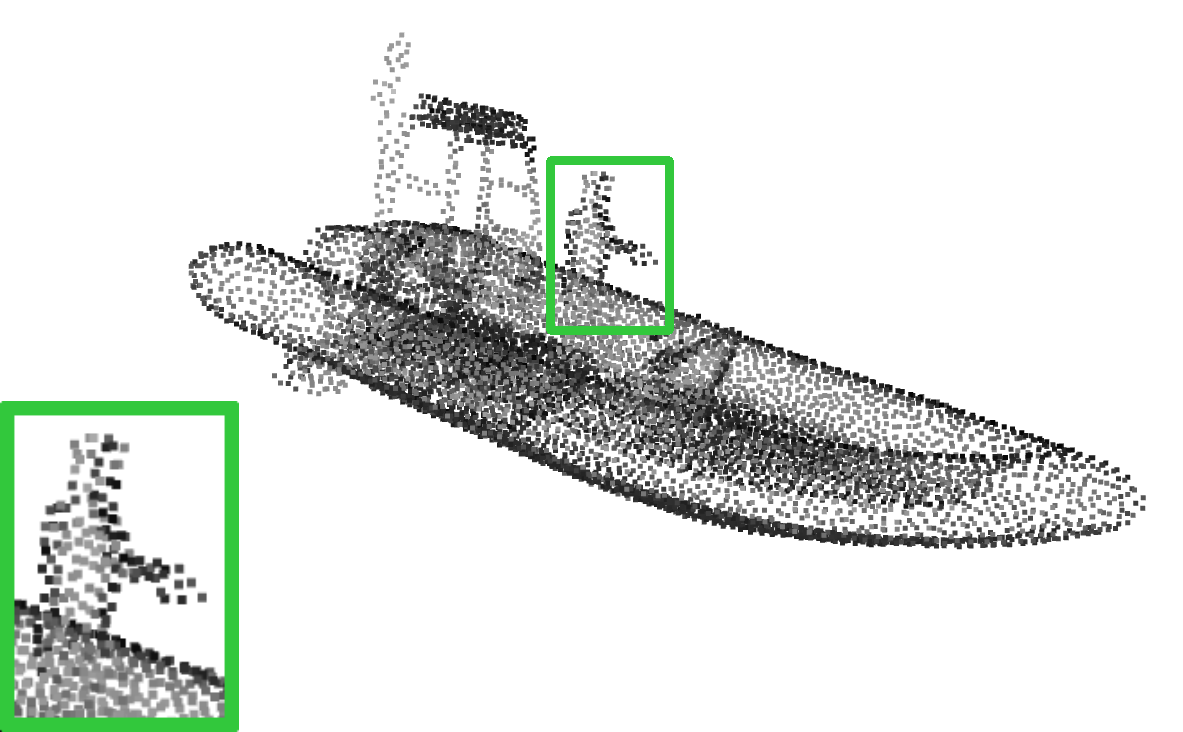} \\
    \includegraphics[width=0.15\linewidth]{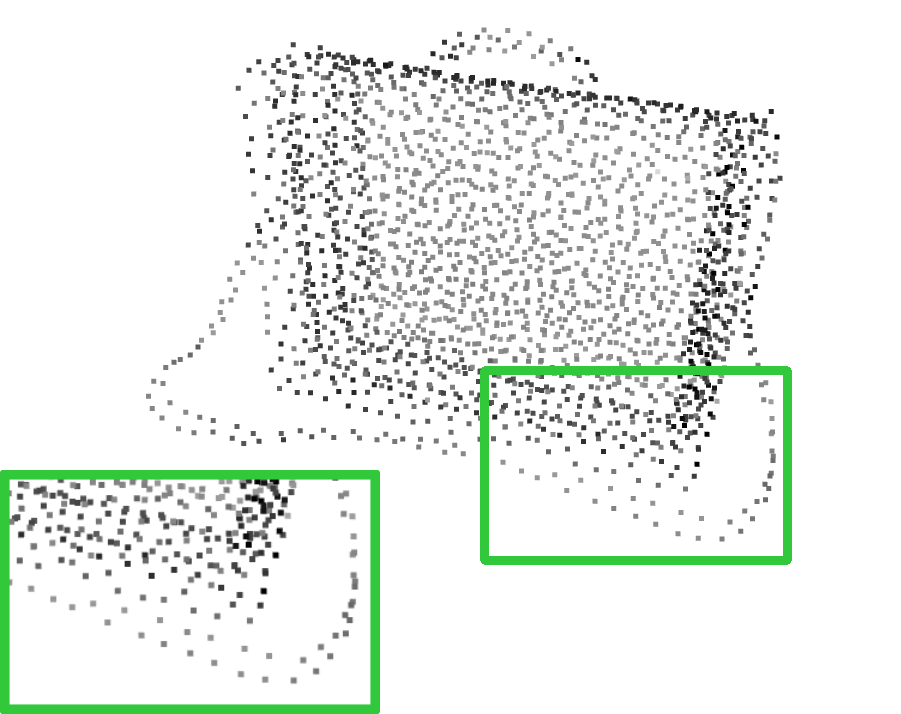} &
    \includegraphics[width=0.15\linewidth]{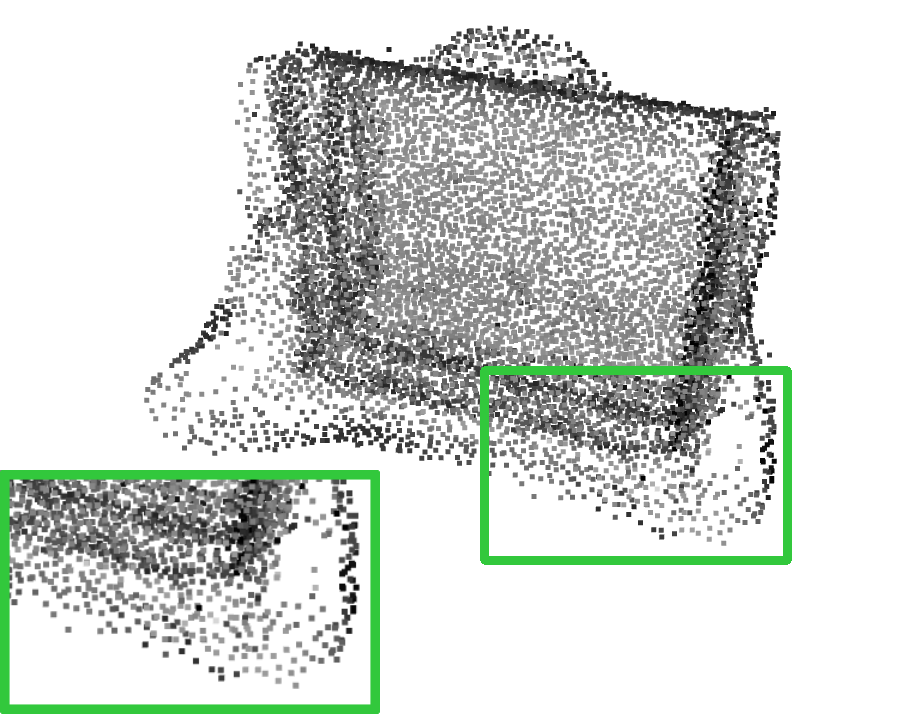} &
    \includegraphics[width=0.15\linewidth]{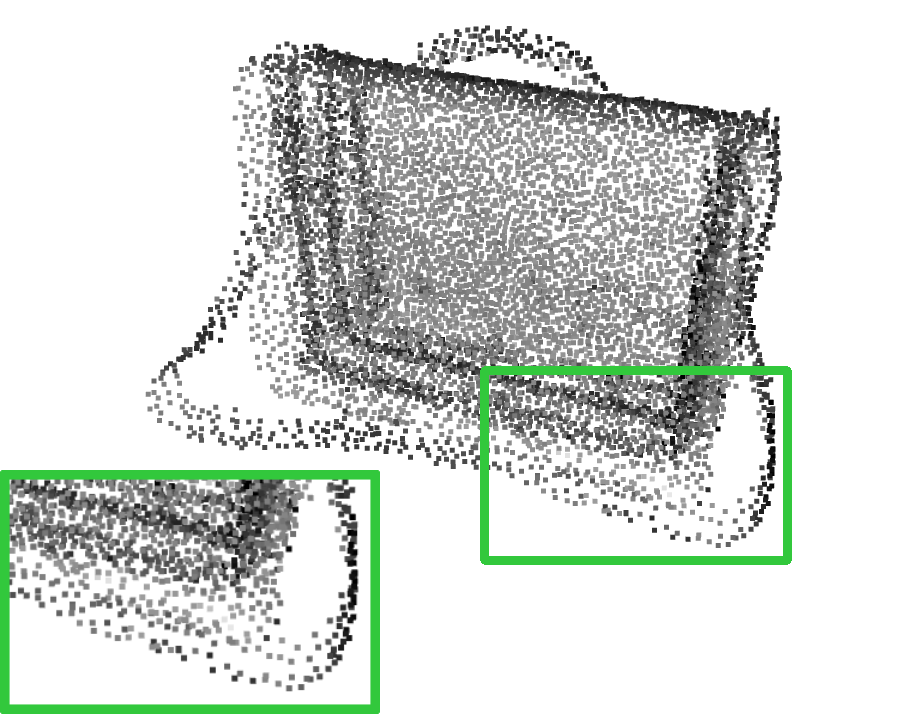} &
    \includegraphics[width=0.15\linewidth]{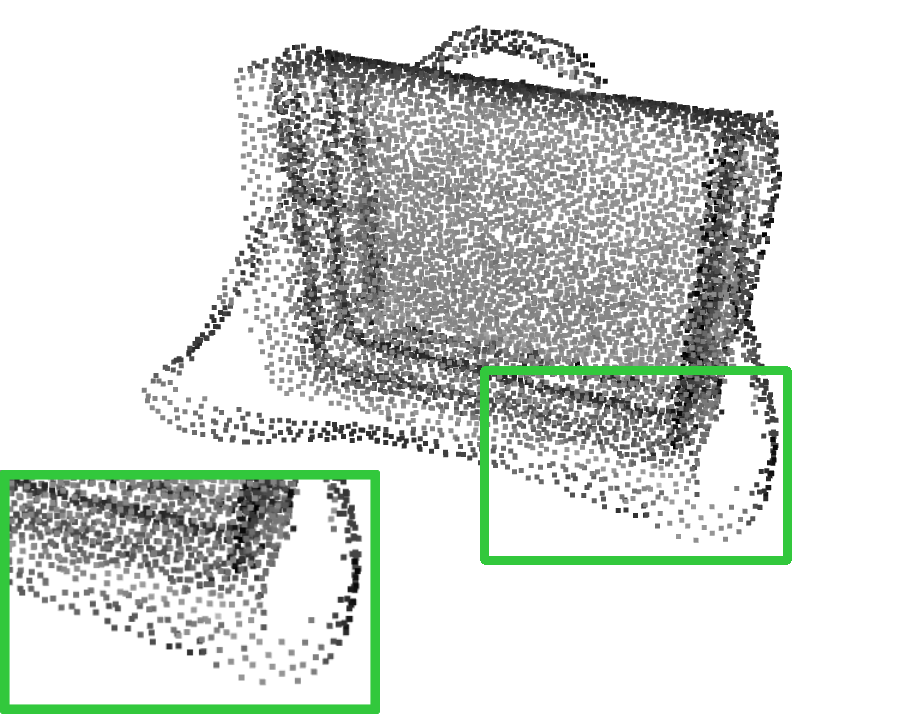} &
    \includegraphics[width=0.15\linewidth]{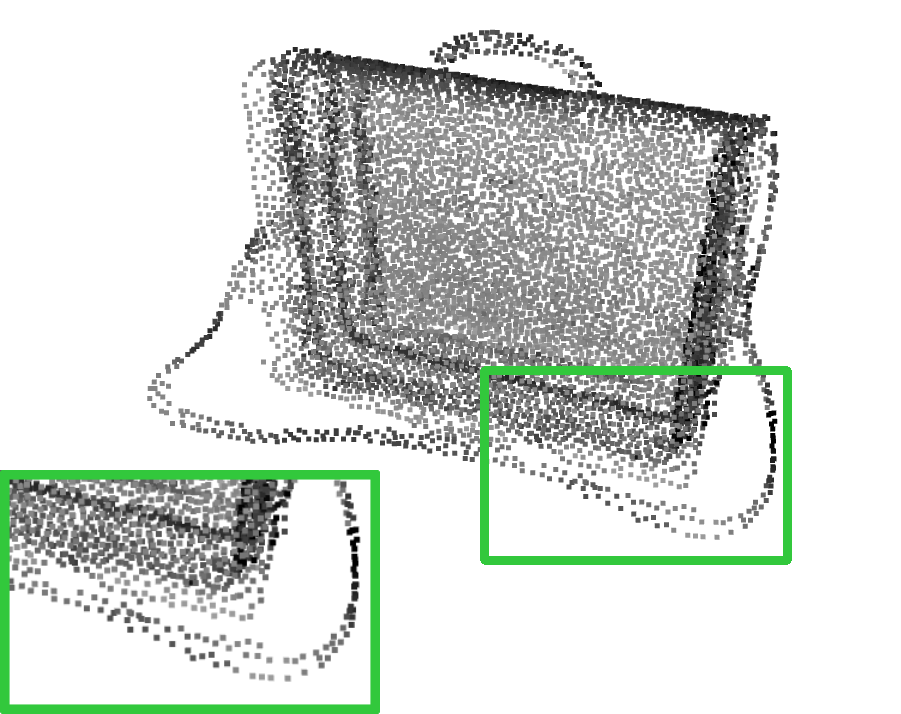} &
    \includegraphics[width=0.15\linewidth]{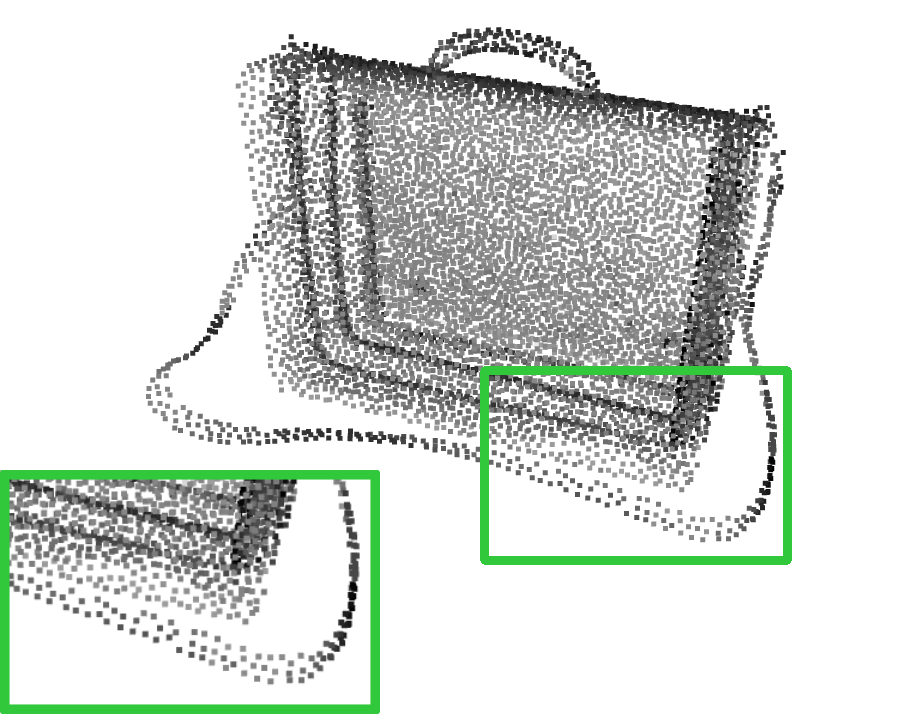} \\
    \includegraphics[width=0.15\linewidth]{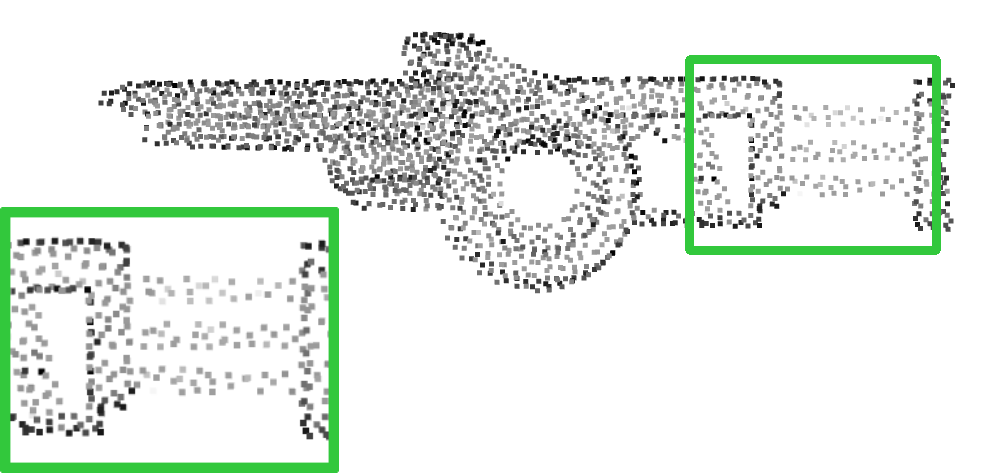} &
    \includegraphics[width=0.15\linewidth]{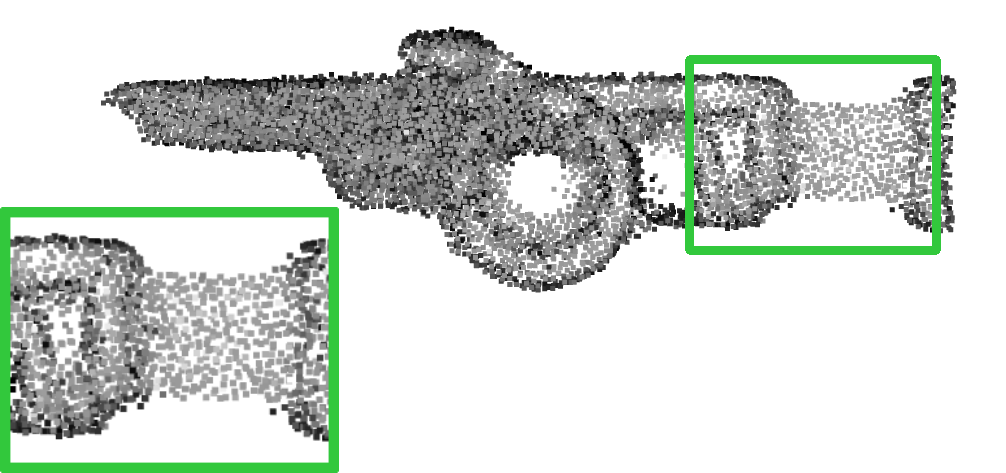} &
    \includegraphics[width=0.15\linewidth]{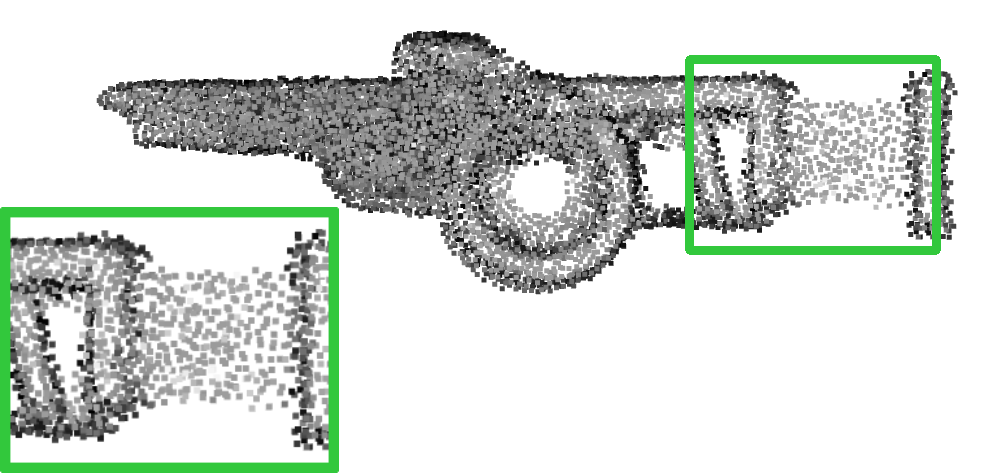} &
    \includegraphics[width=0.15\linewidth]{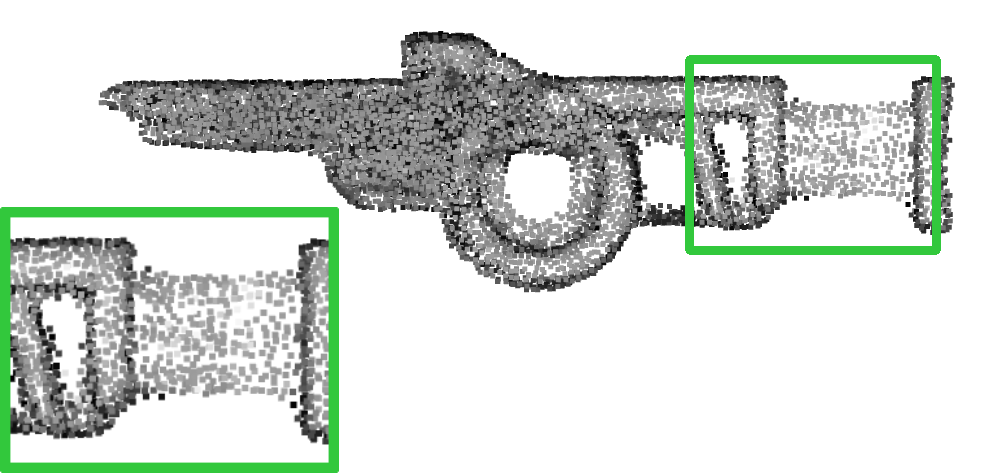} &
    \includegraphics[width=0.15\linewidth]{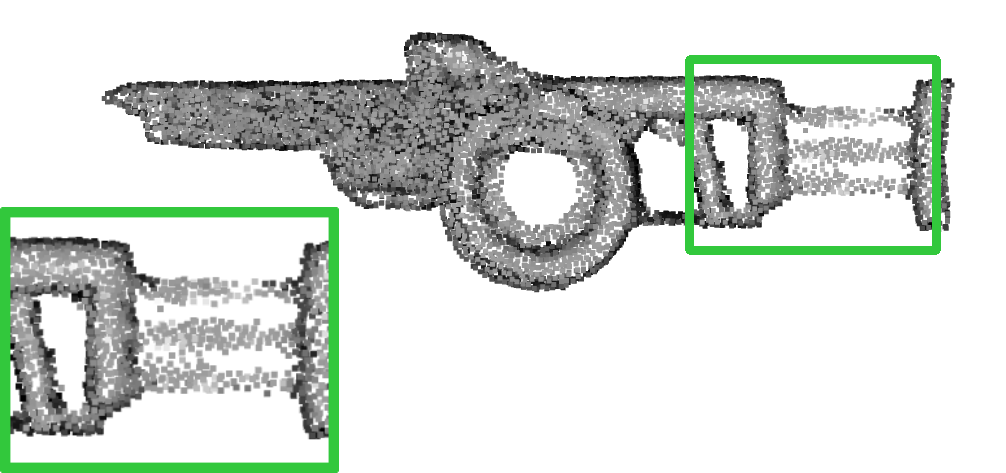} &
    \includegraphics[width=0.15\linewidth]{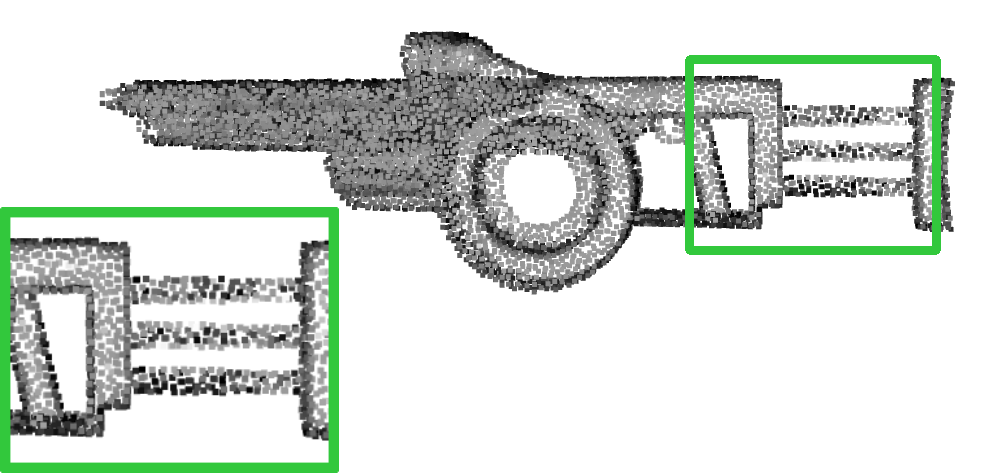} \\
    Input & PU-GAN \cite{pugan} & PU-GCN \cite{pugcn} & Dis-PU \cite{dispu} & Ours & Ground Truth
\end{tabular}
\addtolength{\tabcolsep}{2pt}
\vspace{0.2cm}
\captionof{figure}{Qualitative comparison with state-of-the-art methods on the PU1K dataset. Inputs with $2048$ points (left) are upsampled to $8192$ points (right), with upsampling ratio $r = 4$. Details are best viewed when zoomed in.}
\label{fig:qual}
\end{table*}

\subsection{Quantitative Results}

The quantitative evaluation of the method on the PU1K test set and a fixed upsampling ratio $r = 4$ for point clouds with $N = 2048$ points is presented in Tab.~\ref{tab:quant}. It can be noticed that we achieve the lowest HD value, indicating that APU-SMOG performs upsampling with fewer outliers with respect to state-of-the-art models, as well as the lowest P2F distance, showing a better approximation of the underlying surface. Our approach falls behind Dis-PU~\cite{dispu} in terms of the CD metric by a slight margin. Nevertheless, this quantity measures the consistency of the result with the ground truth \textit{point cloud}, rather than with the \textit{target shape}: two different sets of points sampled from the same mesh have a non-zero CD, despite belonging to the same surface.

Furthermore, the numerical performances as a function of the ratio $r$ on the PU-GAN test set are provided in Tab.~\ref{tab:flex}. In order to be able to compare against state-of-the-art flexible methods~\cite{mafu, pueva}, we constrain $r$ to be an integer value smaller than $16$, even if we can generate predictions with any $r \in \mathbb{R}$.
Our approach achieves the best CD value, as well as the best HD value by a significant margin, along the whole spectrum of ratios. Note that both MAFU \cite{mafu} and Neural Points \cite{neuralpts} have a lower P2F metric since they require ground truth normals at training time, which helps in finding the correct surface. On the other hand, our approach consistently outperforms PU-EVA \cite{pueva}, which is trained from raw point clouds.
For completeness, we include the comparison against \textit{fixed-ratio} models as reference. To generate predictions for $r \in \{8,12,16\}$, each network is queried iteratively with $r = 4$ and the desired number of points is obtained with FPS.

\subsection{Qualitative Results}

Fig.~\ref{fig:qual} shows qualitative upsampling results in comparison with state-of-the-art methods. These results have been obtained under the same settings as Tab.~\ref{tab:quant}, namely with all the models trained on the PU1K dataset with fixed upsampling ratio $r = 4$ for input point clouds having size ${N = 2048}$. Close-up views show that APU-SMOG is particularly effective in preserving fine-grained structures, such as the piano's pole and the motorcycle mirror, and disambiguating complex shape (see the bag's handle). It can be noticed that other models fail to distinguish between different details of the surface and tend to merge them together, thus producing noisy point clouds. Moreover, the proposed attention-based residual refinement block generates refined outputs with fewer outliers (e.g. the plane motors). Additional qualitative results, including more comparisons and reconstructed surfaces from predicted point clouds, can be found in the supplementary material.

\subsection{Generalization and Robustness}

We conduct further experiments to demonstrate the robustness and generalization capabilities of APU-SMOG.
Firstly, we provide qualitative results on real-world point clouds from the KITTI dataset~\cite{kitti}, without any fine-tuning. This task is particularly challenging, since street-level LiDAR data with noise and occlusions are very different from synthetic training samples. Fig.~\ref{fig:kitti} shows the generalization power of our approach on different urban elements such as cars, trucks and pedestrians. 
See the supplementaries for more examples on real-scanned point clouds \cite{uy-scanobjectnn-iccv19}.

Since the previous qualitative results in Fig.~\ref{fig:qual} have been generated with point clouds having ${N=2048}$ points, we show in Fig.~\ref{fig:size} that the predictions of our method closely follow the underlying surfaces for a wide variety of input sizes. Even for the sparsest input with $256$ points, thin structures such as the horse's legs are upsampled correctly. 
In order to simulate real noisy point clouds from scanning sensors, Fig.~\ref{fig:noise} shows the upsampling results for three different levels of additive Gaussian noise. It can be noticed that the duck shape is successfully maintained for both the clean and the corrupted inputs. 

Finally, the key novelty of our approach is the possibility to specify an arbitrary upsampling factor at test time, thus producing consistent results with the desired output resolution. This flexibility is shown in Fig.~\ref{fig:teaser} and additional examples with more details can be found in the supplementaries.

\begin{table}[t]
\centering
\setlength\extrarowheight{0.1pt}
\begin{tabular}{ l | c c c c } 
 \toprule
 \textbf{Ablation} & \textbf{CD}$\downarrow$ & \textbf{HD}$\downarrow$ & \textbf{P2F}$\downarrow$ \\
 \midrule
 FoldingNet-like \cite{foldingnet} & 0.558 & 2.761 & 1.700 \\ 
 $N/4$ SMOG components & 0.564 & 2.803 & 1.686\\
 w/o refinement & 0.572 & 8.536 &  2.182\\
 w/o rec. loss, $\mathcal{L}_{\Pi}$ ups. & 0.561 & 2.843 & 1.726\\ 
 w/o rec. loss, $\mathcal{L}_{ACD}$ ups. & 0.816 & 6.459 &  2.694\\ 
 \midrule
 \textbf{Ours} & \textbf{0.528} & \textbf{2.549} & \textbf{1.667}  \\ 
 \bottomrule
 \end{tabular}
 \vspace{0.3cm}
 \caption{Ablation studies on different versions of our model. The units are all $10^{-3}$ and \textbf{bold} denotes the best performance.}
 \label{tab:abl}
\end{table}

\begin{table*}
\centering
\begin{tabular}{ c c c c }
 \multirow{1}{*}[3em]{\rotatebox{90}{Input}} &
 \includegraphics[width=0.3\linewidth]{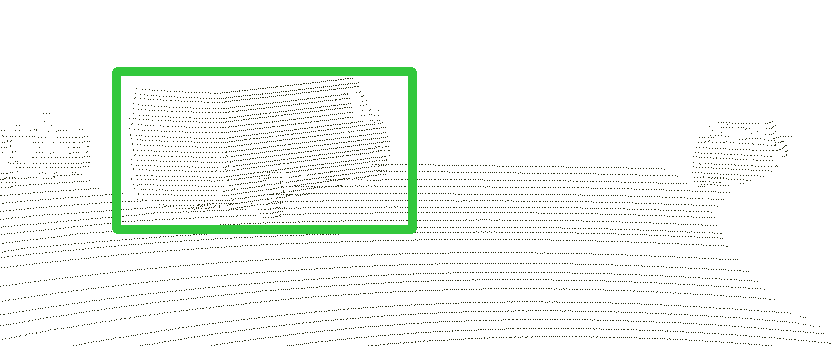} & 
 \includegraphics[width=0.3\linewidth]{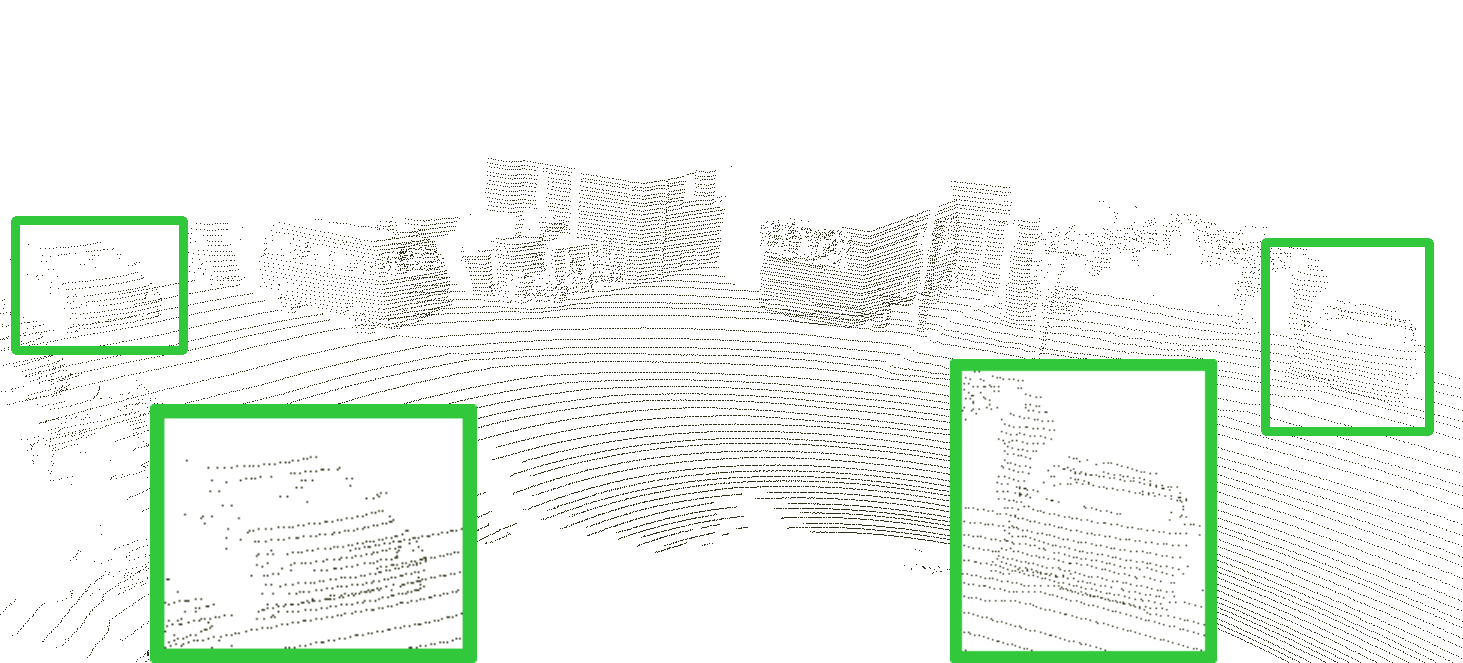} & 
 \includegraphics[width=0.3\linewidth]{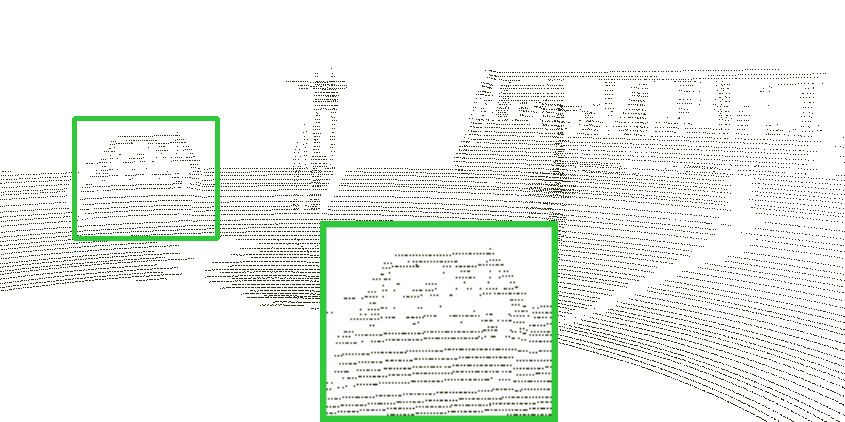} \\
 \multirow{1}{*}[4em]{\rotatebox{90}{Output}} &
 \includegraphics[width=0.3\linewidth]{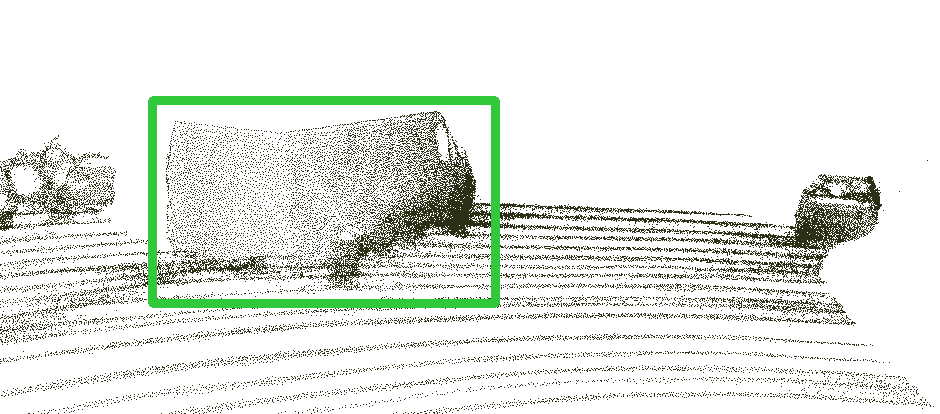} & 
 \includegraphics[width=0.3\linewidth]{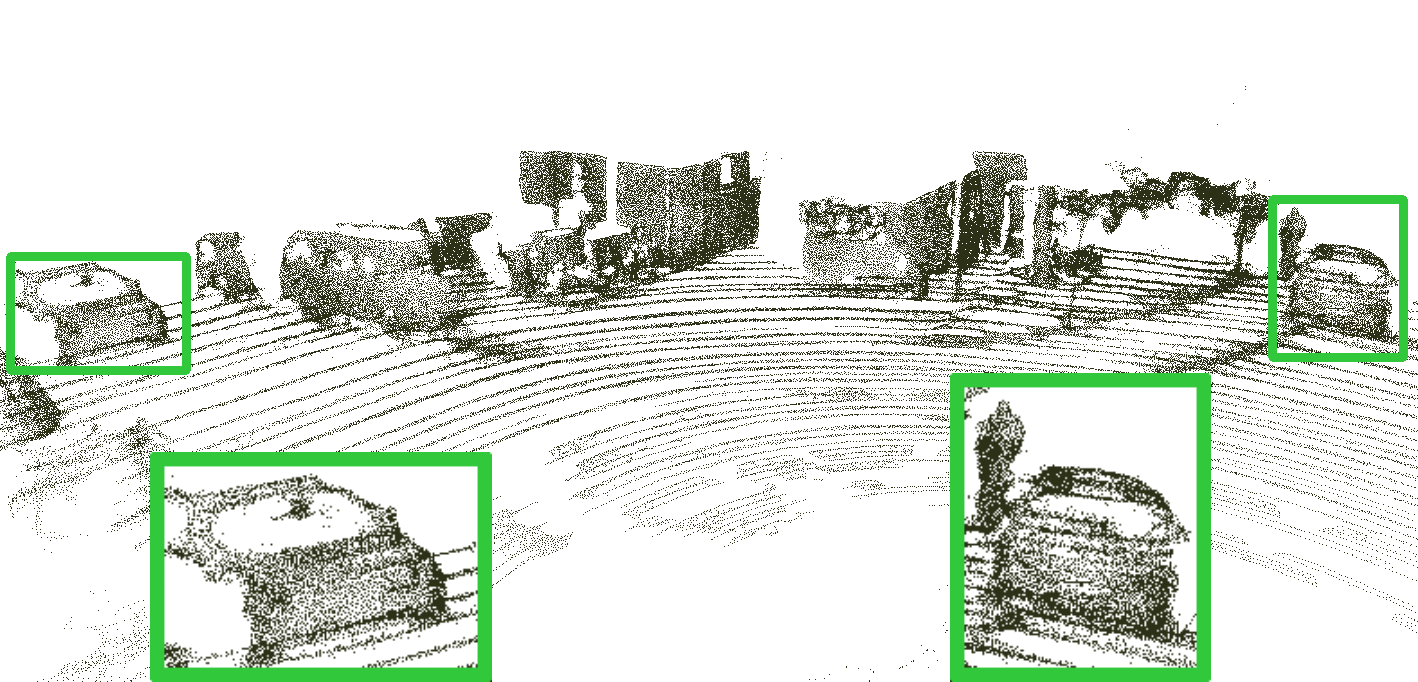} & 
 \includegraphics[width=0.3\linewidth]{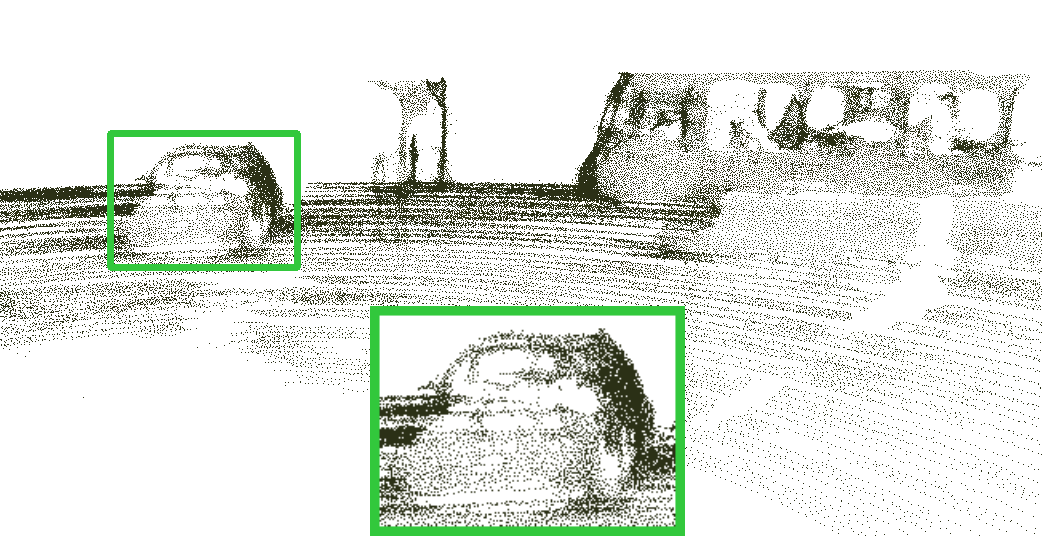}
\end{tabular}
\vspace{0.2cm}
\captionof{figure}{Generalization to real-world point clouds from the KITTI dataset \cite{kitti}. APU-SMOG can upsample successfully various urban instances, including fine-grained structures such as the car window. Details are best viewed when zoomed in.}
\label{fig:kitti}
\end{table*}

\begin{table}[h!]
\begin{tabular}{ c c c c }
 & 256 & 1024 & 4096 \\
 \multirow{1}{*}[3em]{\rotatebox{90}{Input}} &
 \includegraphics[width=0.25\linewidth]{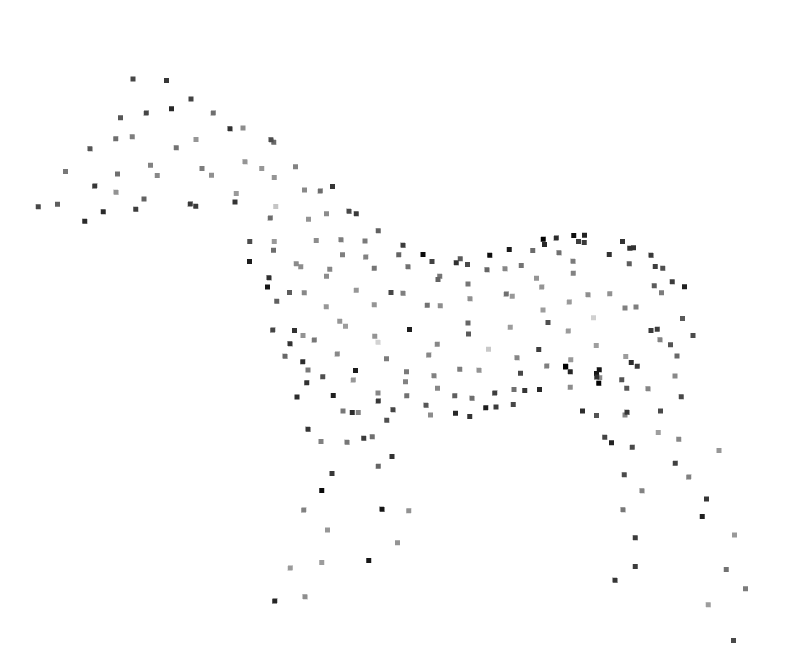} & 
 \includegraphics[width=0.25\linewidth]{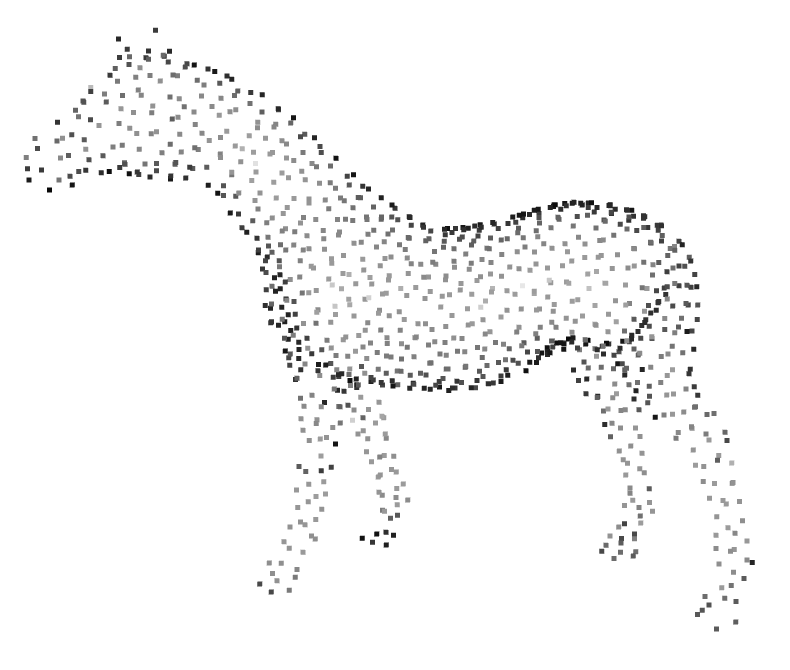} & 
 \includegraphics[width=0.25\linewidth]{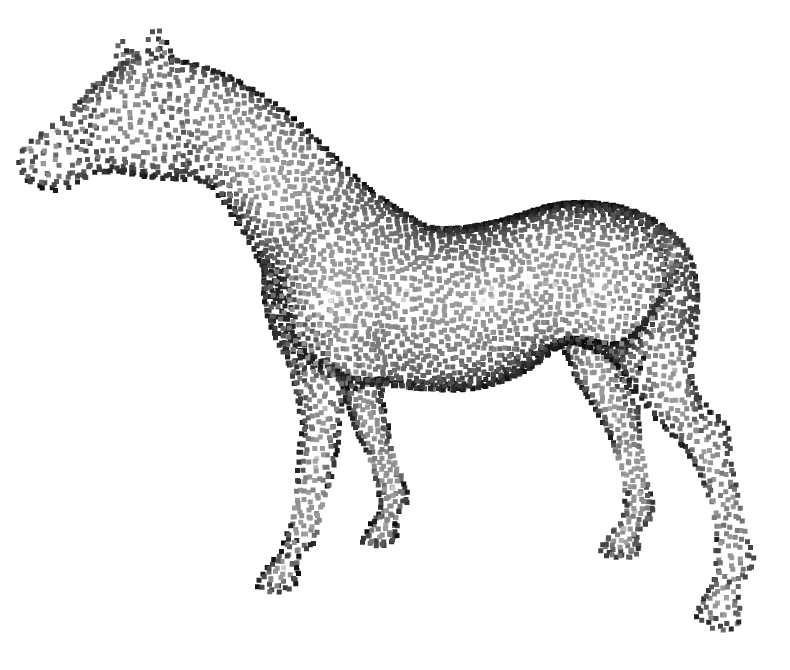} \\
 \multirow{1}{*}[3.3em]{\rotatebox{90}{Output}} &
 \includegraphics[width=0.25\linewidth]{figures/size/horse_1024.png} & 
 \includegraphics[width=0.25\linewidth]{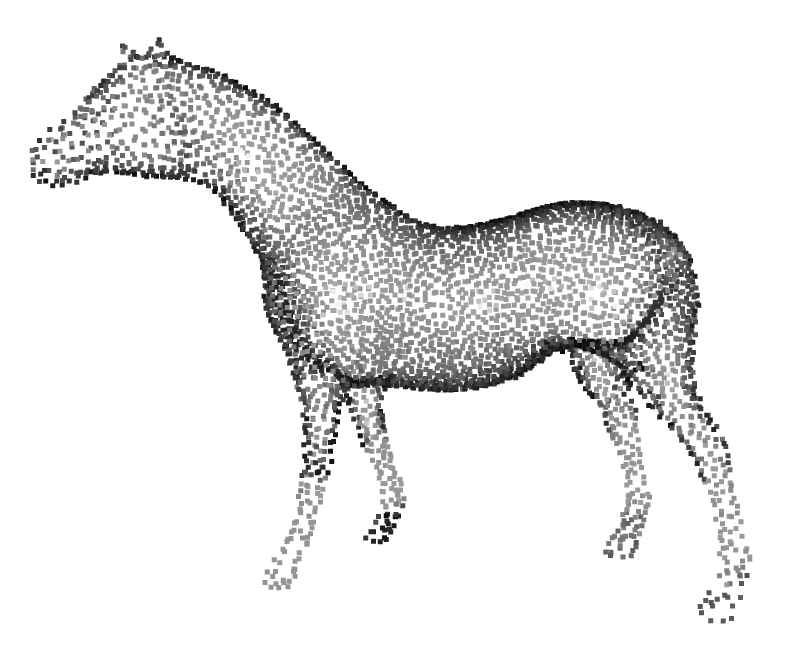} & 
 \includegraphics[width=0.25\linewidth]{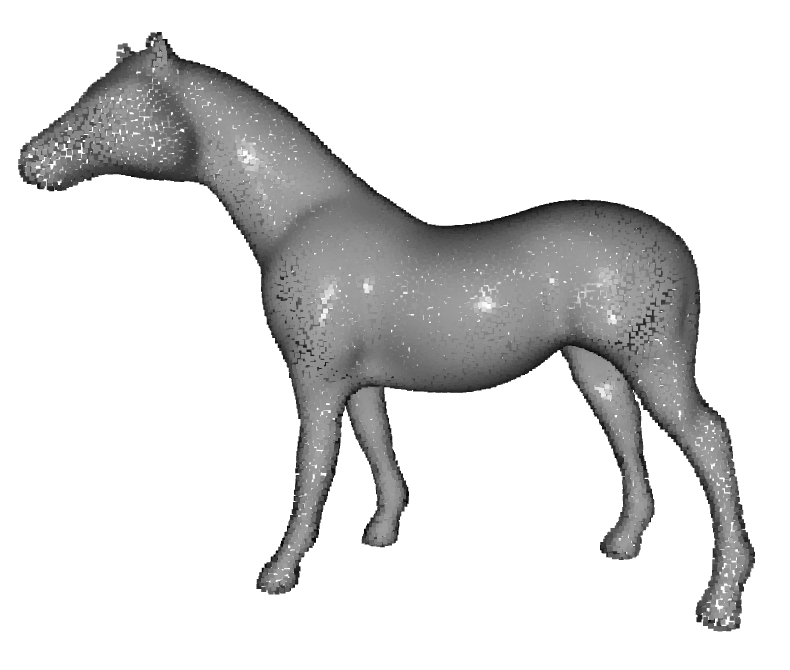}
\end{tabular}
\vspace{0.2cm}
\captionof{figure}{Effect of input point cloud size on the upsampling results.}
\label{fig:size}
\end{table}

\begin{table}[h!]
\begin{tabular}{ c c c c }
 & $\sigma = 0.00$ & $\sigma = 0.01$ & $\sigma = 0.02$ \\
 \multirow{1}{*}[3em]{\rotatebox{90}{Input}} &
 \includegraphics[width=0.25\linewidth]{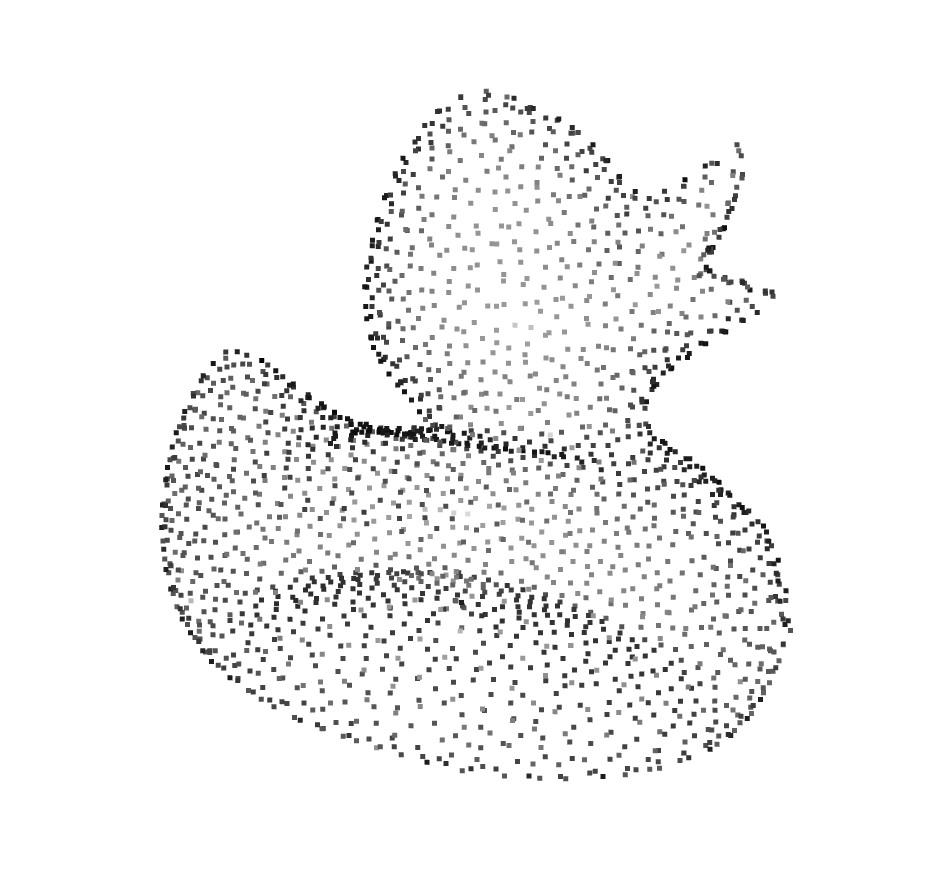} & 
 \includegraphics[width=0.25\linewidth]{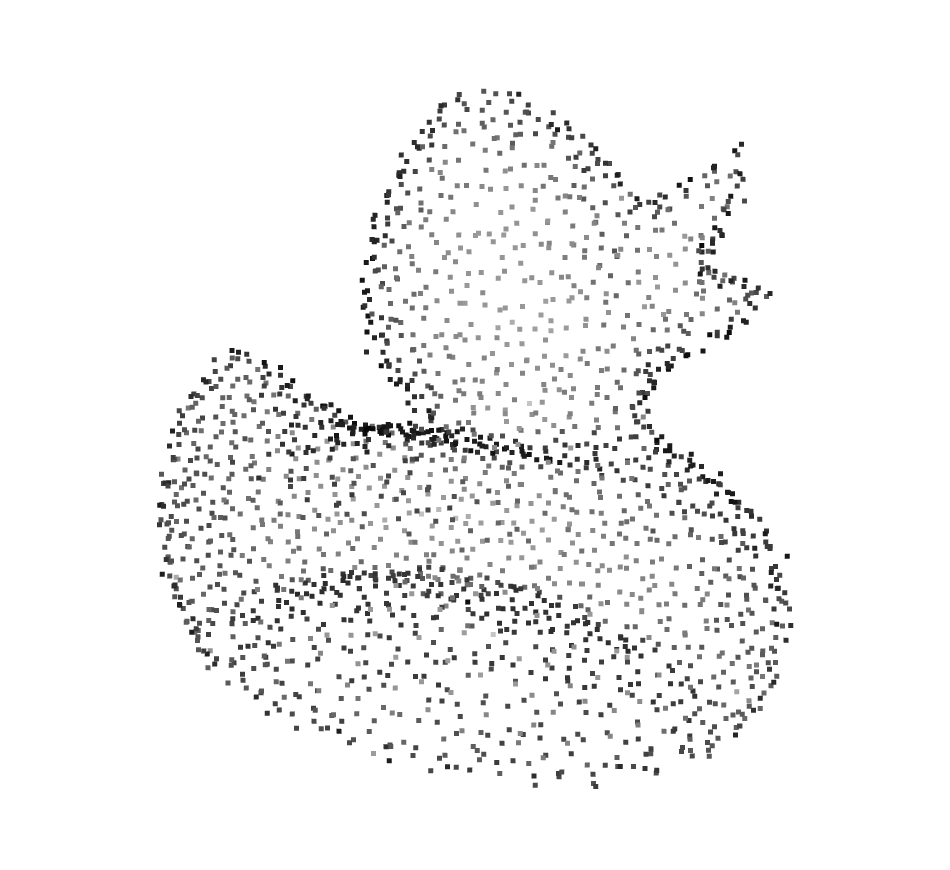} & 
 \includegraphics[width=0.25\linewidth]{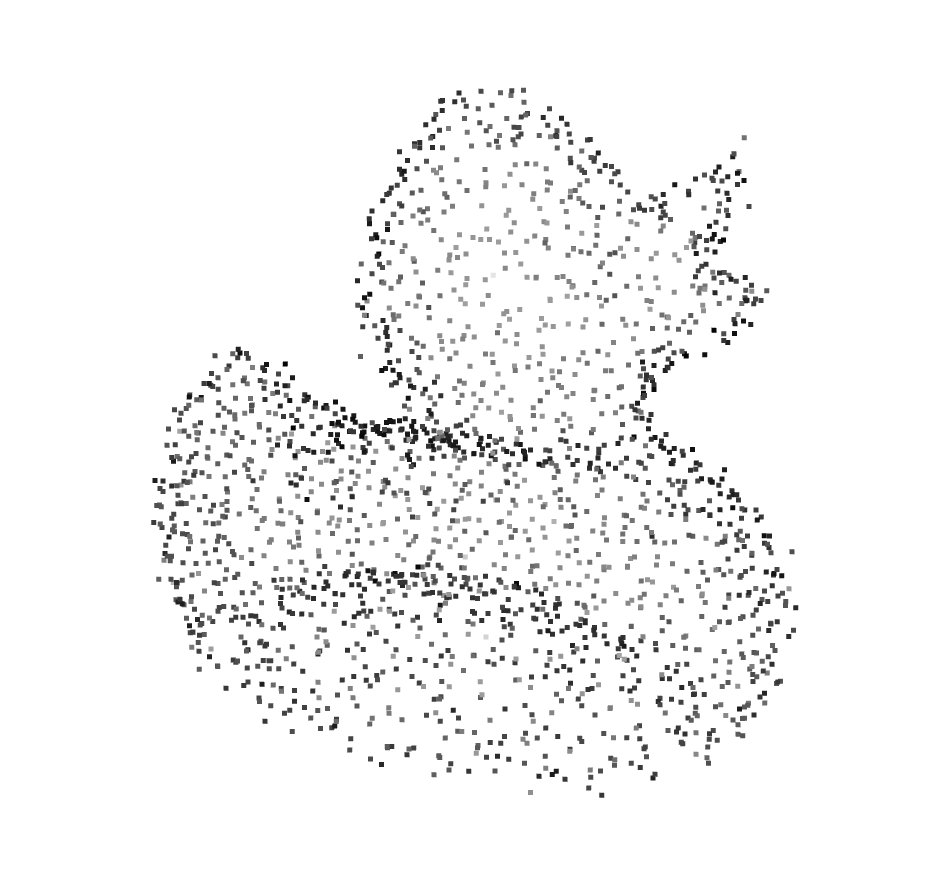} \\
 \multirow{1}{*}[3.3em]{\rotatebox{90}{Output}} &
 \includegraphics[width=0.25\linewidth]{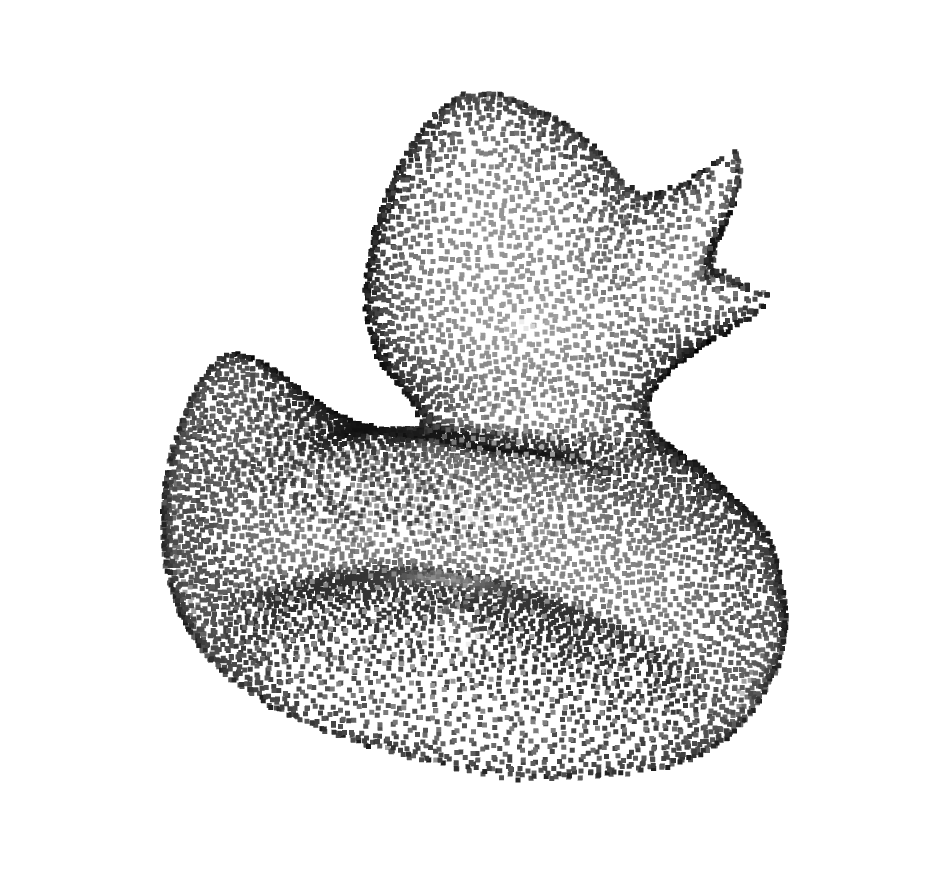} & 
 \includegraphics[width=0.25\linewidth]{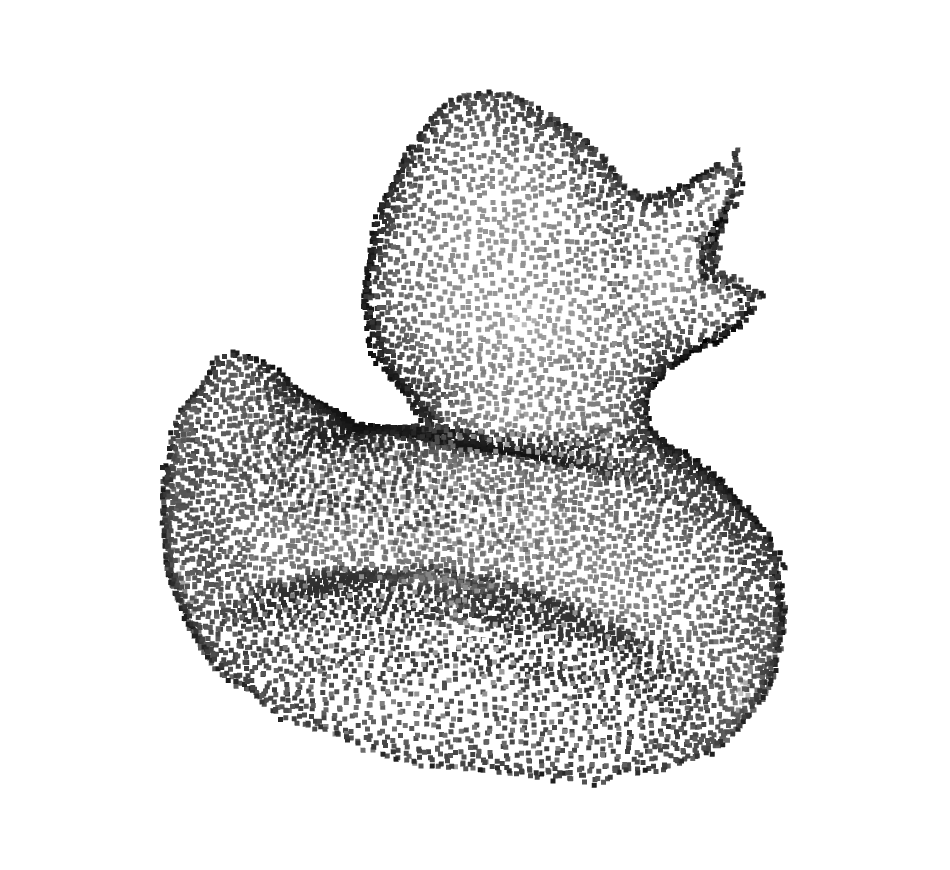} & 
 \includegraphics[width=0.25\linewidth]{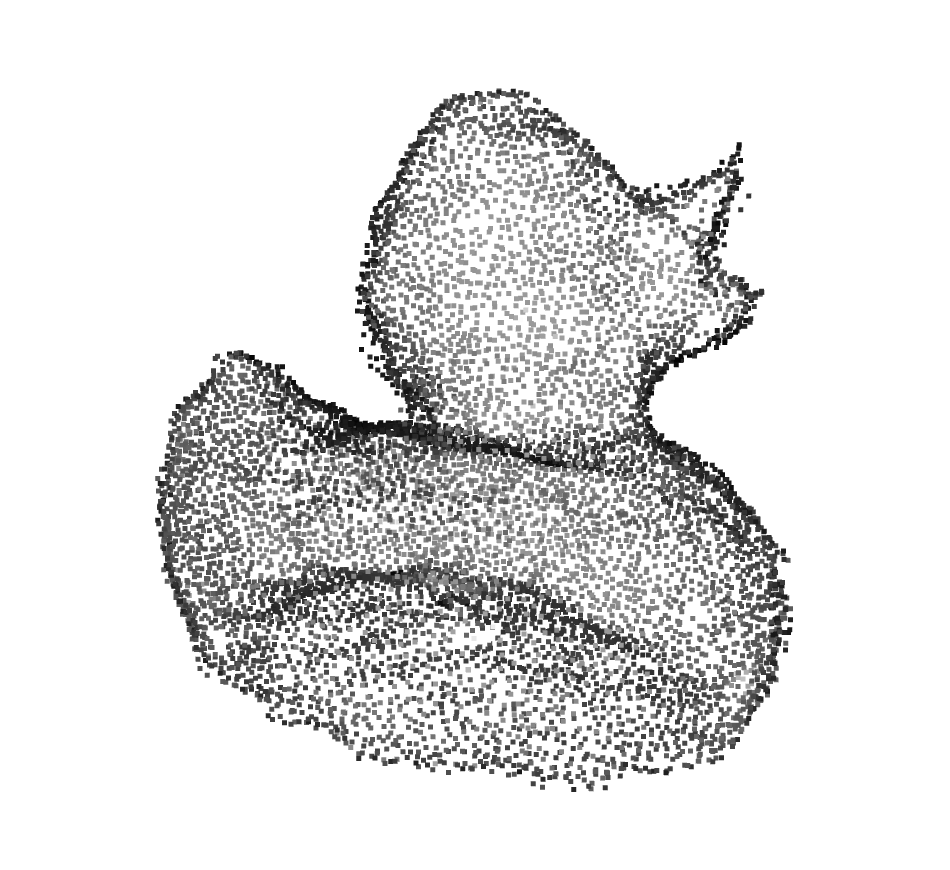}
\end{tabular}
\vspace{0.2cm}
\captionof{figure}{Effect of additive noise on the upsampling results.}
\label{fig:noise}
\end{table}

\subsection{Ablation Studies}
\label{sec:ablation}

We perform a set of ablation studies to evaluate the contribution of each module to the proposed pipeline. The presented results in Tab.~\ref{tab:abl} have been obtained with different versions of our model trained on the PU1K dataset with fixed upsampling ratio $r = 4$.

In order to measure the influence of the SMOG representation, we replace it with a FoldingNet-like~\cite{foldingnet} strategy, i.e. sampling the unit sphere uniformly. The quantitative results confirm that our approach is able to generate better predictions, as it adaptively learns a local probability distribution around each point. 

Furthermore, we investigate the influence of the number of GMM components on the output point clouds, reducing it to $K = N/4$. The numerical evaluation exhibits decent performances, suggesting that the model associates each Gaussian distribution to a local neighborhood of points. However, having a single component for each input point leads to the best results overall. 

The third row in Tab.~\ref{tab:abl} proves the effectiveness of the attention-based residual refinement step. The significantly higher HD value indicates that the raw output from the Transformer decoder contains several outliers, which are correctly positioned closer to the surface by this module.

Moreover, we train our model without the reconstruction loss and notice a performance drop. This implies that $\mathcal{L}_{ACD}$ is a strong bias towards learning the correct positions of the Gaussian means on the unit sphere. Finally, the advantage of the projection loss for upsampling over ACD is shown in the last row. The remarkable difference in all the metrics justifies the choice of $\mathcal{L}_{\Pi}$ in the final design.
\section{Conclusion}

In this paper, we present a novel approach for point cloud upsampling with arbitrary scaling factors. A Transformer-based architecture is designed to decouple the upsampling process in two key steps: (i) firstly, the sparse input is mapped to an intermediate representation as a Spherical Mixture of Gaussians; (ii) then, such distribution is sampled arbitrarily and the Transformer decoder learns to map each sample back to the surface. The predictions are further improved by an attention-based residual refinement module, which allows to achieve state-of-the-art results on different benchmarks. This strategy enables arbitrary upsampling, since the model is trained a single time with a fixed ratio and it can be queried at test time with any desired value.

The main limitation of our work is the patch-based training and testing procedure, which require a long inference time when upsampling large point clouds. As future work, we plan to tackle this issue and to learn the weights of the GMM jointly with the Gaussian parameters.

{\small
\bibliographystyle{ieee_fullname}
\bibliography{egbib}
}

\onecolumn
\setcounter{section}{0}
\renewcommand{\thesection}{\Alph{section}}

\begin{center}
\vspace*{1cm}
    \Large\textbf{Supplementary Material for \\ Arbitrary Point Cloud Upsampling with Spherical Mixture of Gaussians}
\end{center}

\vspace{1cm}

\section{Surface Reconstruction}

In this section, we present the qualitative results of our approach when the upsampled point clouds are used to extract a continuous mesh. This experiment is performed with the model trained on the PU1K dataset \cite{pugcn} with input size $N = 2048$, $r = 4$ and the open-source implementation of Poisson Surface Reconstruction (PSR) available in the Open3D library \cite{open3d}. Fig. \ref{fig:supp_mesh_1} shows that the outputs of our method allow to reconstruct better shapes with respect to other state-of-the-art methods (e.g. the index finger in the hand and the human arm in the statue). 

Moreover, we provide additional surface reconstruction results from APU-SMOG as a function of the upsampling ratio in Fig. \ref{fig:supp_mesh_2}. In this case, the input point cloud has size $N = 1024$ and the ratios are chosen randomly in a wide range to further demonstrate the ability of handling arbitrary factors $r \in \mathbb{R}$. Note that the ground truth mesh is the actual synthetic shape from the PU1K dataset, while the ground truth point cloud is generated for reference by Poisson disk sampling. The quality of the resulting mesh consistently improves with increasing values of the upsampling ratio, thus proving the effectiveness of our method for the downstream 3D reconstruction task.

\begin{table}[h!]
\centering
\addtolength{\tabcolsep}{-2pt} 
\begin{tabular}{ c c c c c c}
    \includegraphics[width=0.15\linewidth]{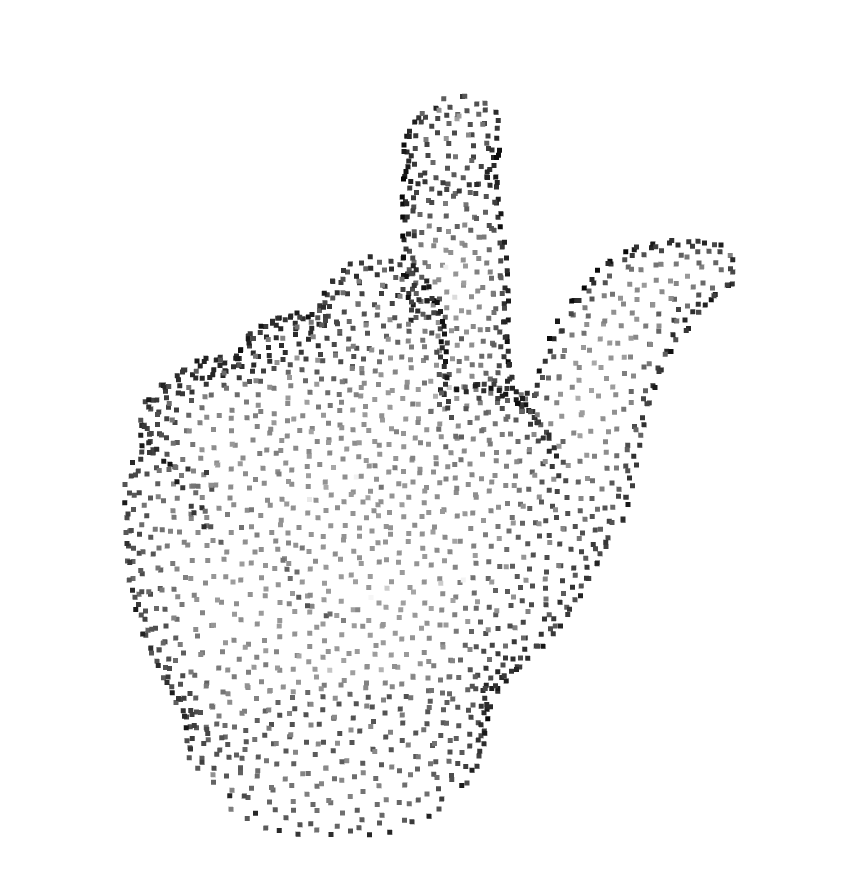} &
    \includegraphics[width=0.15\linewidth]{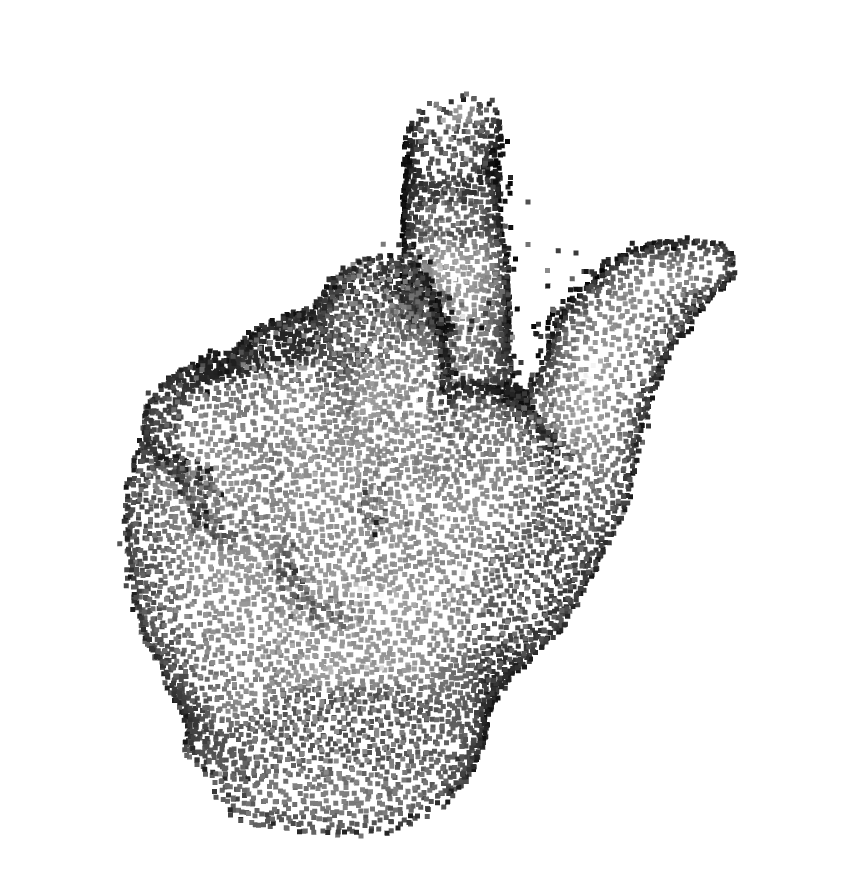} &
    \includegraphics[width=0.15\linewidth]{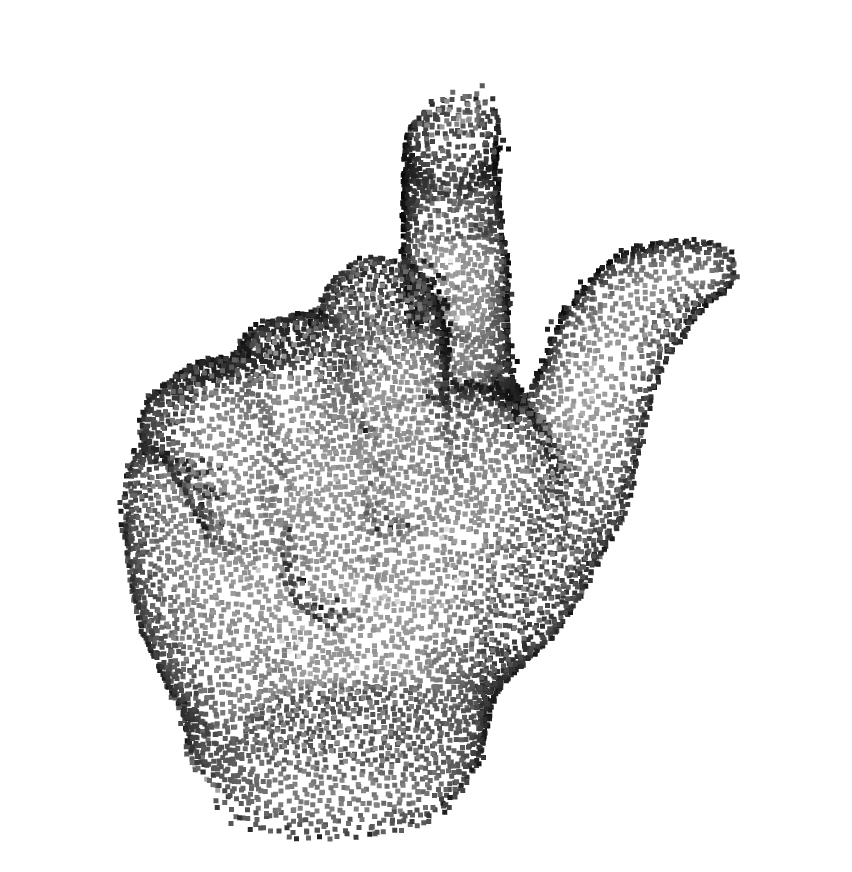} &
    \includegraphics[width=0.15\linewidth]{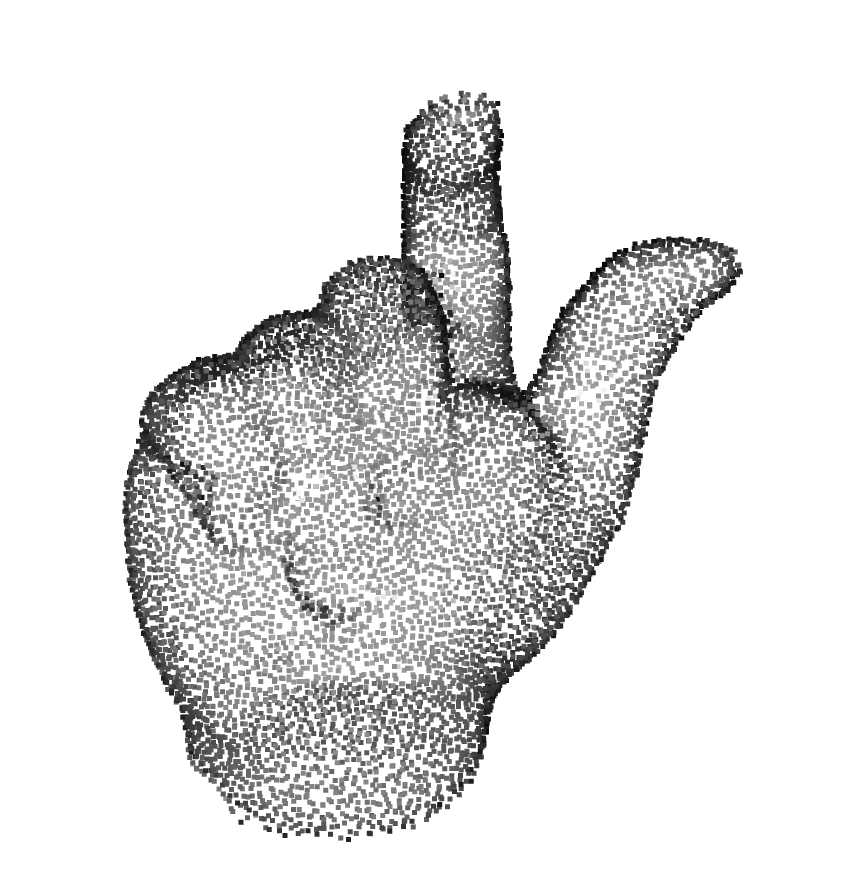} &
    \includegraphics[width=0.15\linewidth]{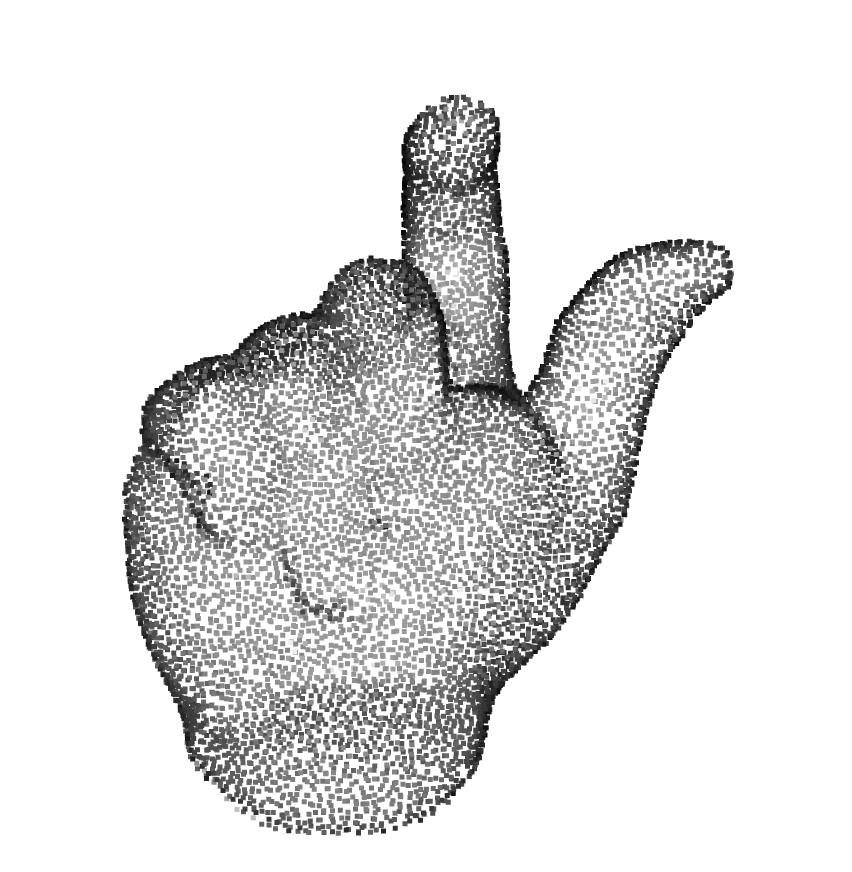} &
    \includegraphics[width=0.15\linewidth]{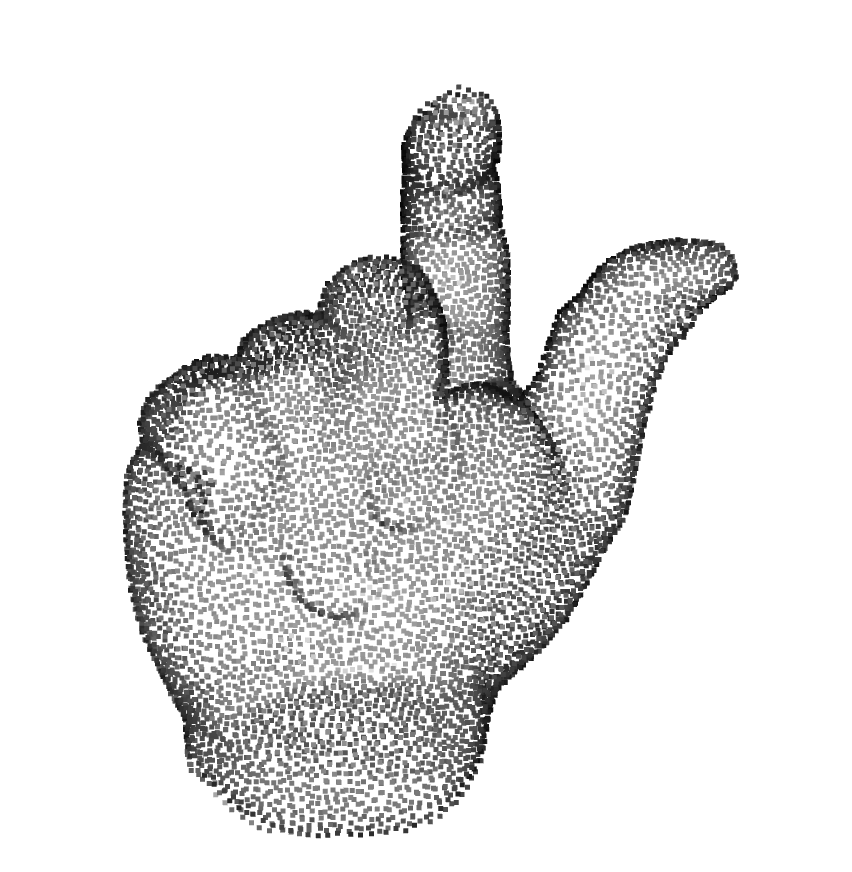} \\
    \includegraphics[width=0.15\linewidth]{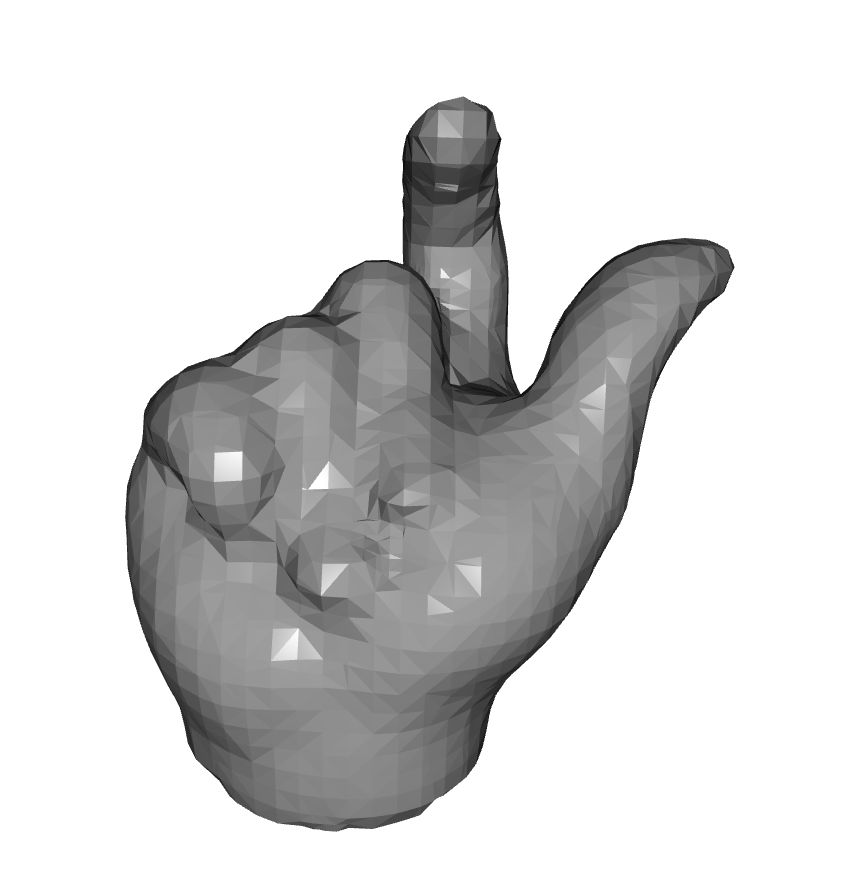} &
    \includegraphics[width=0.15\linewidth]{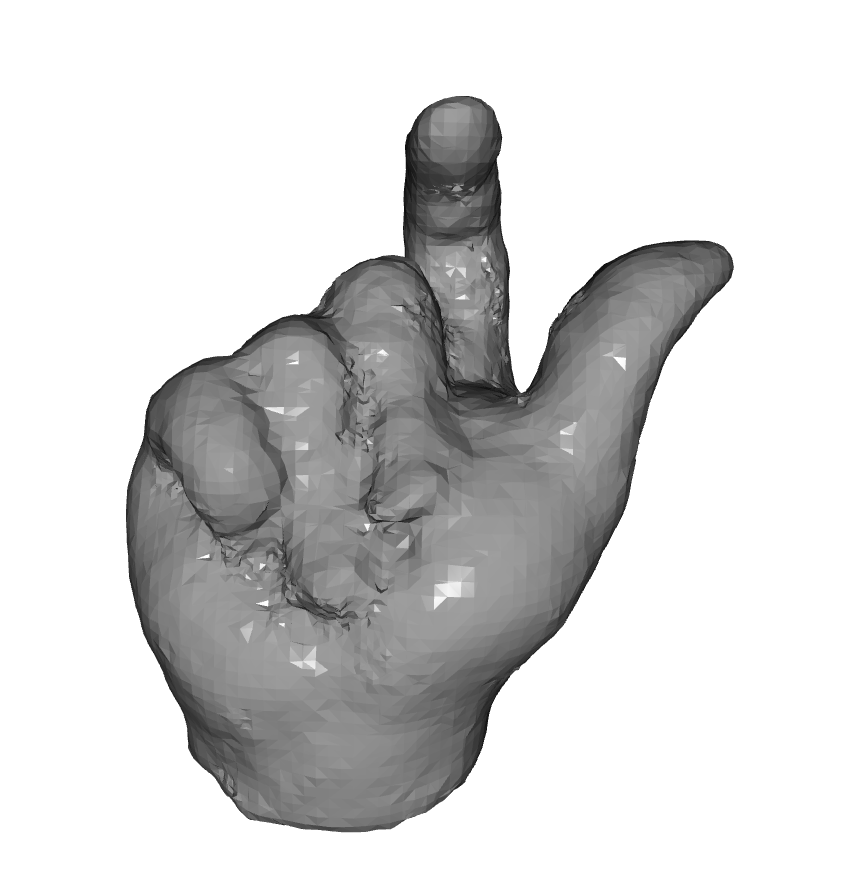} &
    \includegraphics[width=0.15\linewidth]{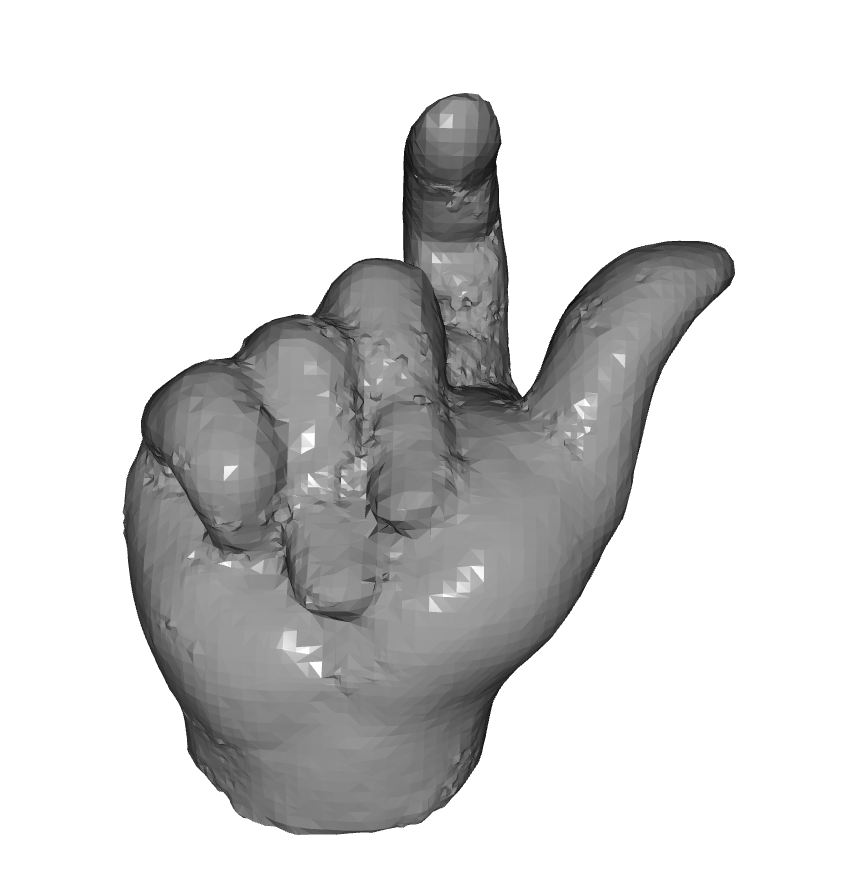} &
    \includegraphics[width=0.15\linewidth]{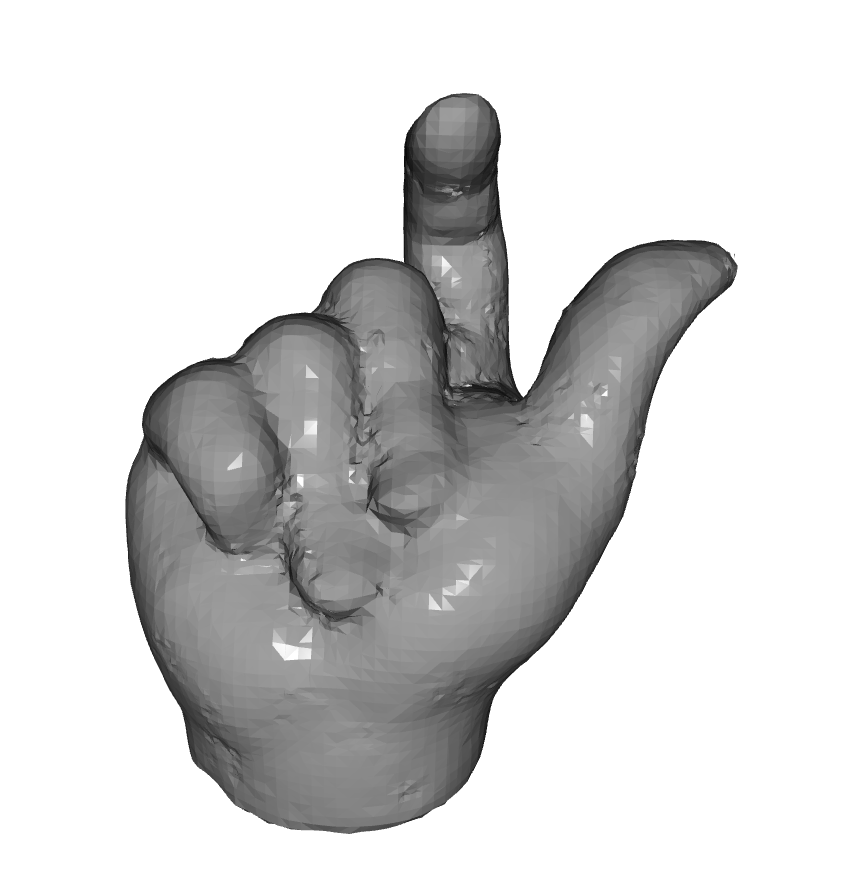} &
    \includegraphics[width=0.15\linewidth]{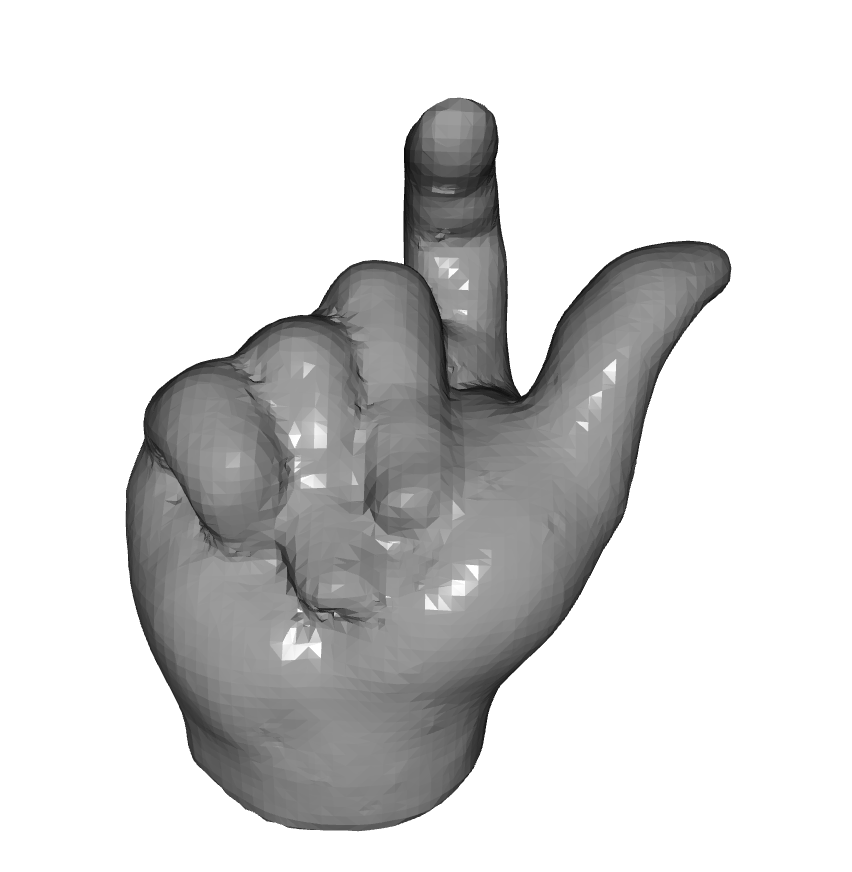} &
    \includegraphics[width=0.15\linewidth]{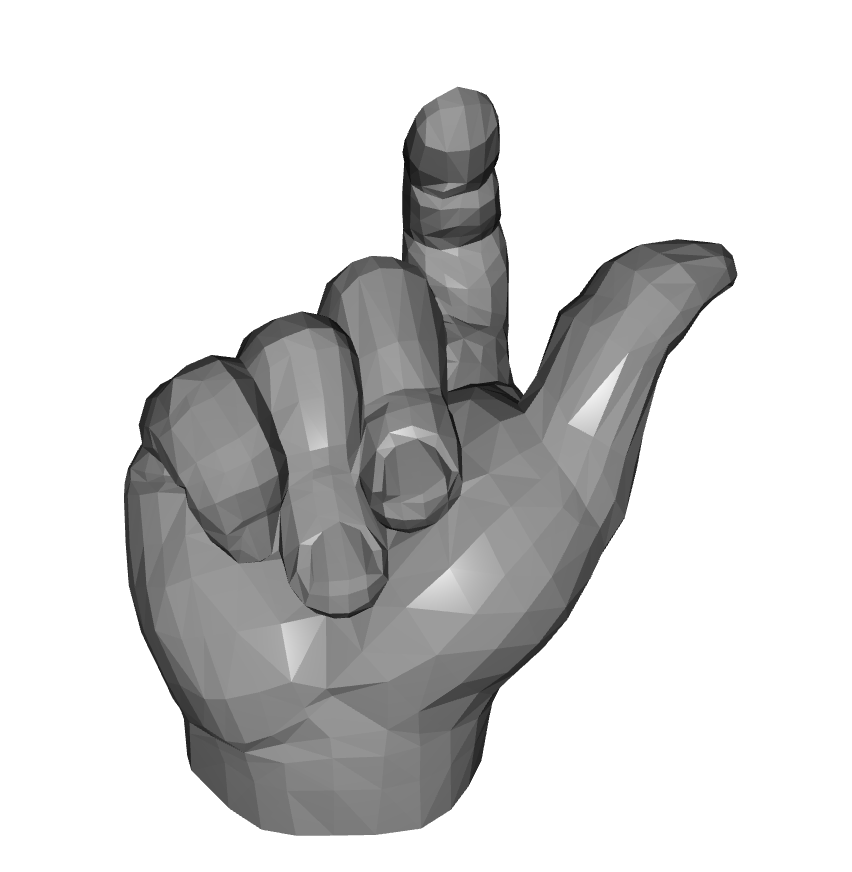} \\
    \includegraphics[width=0.15\linewidth]{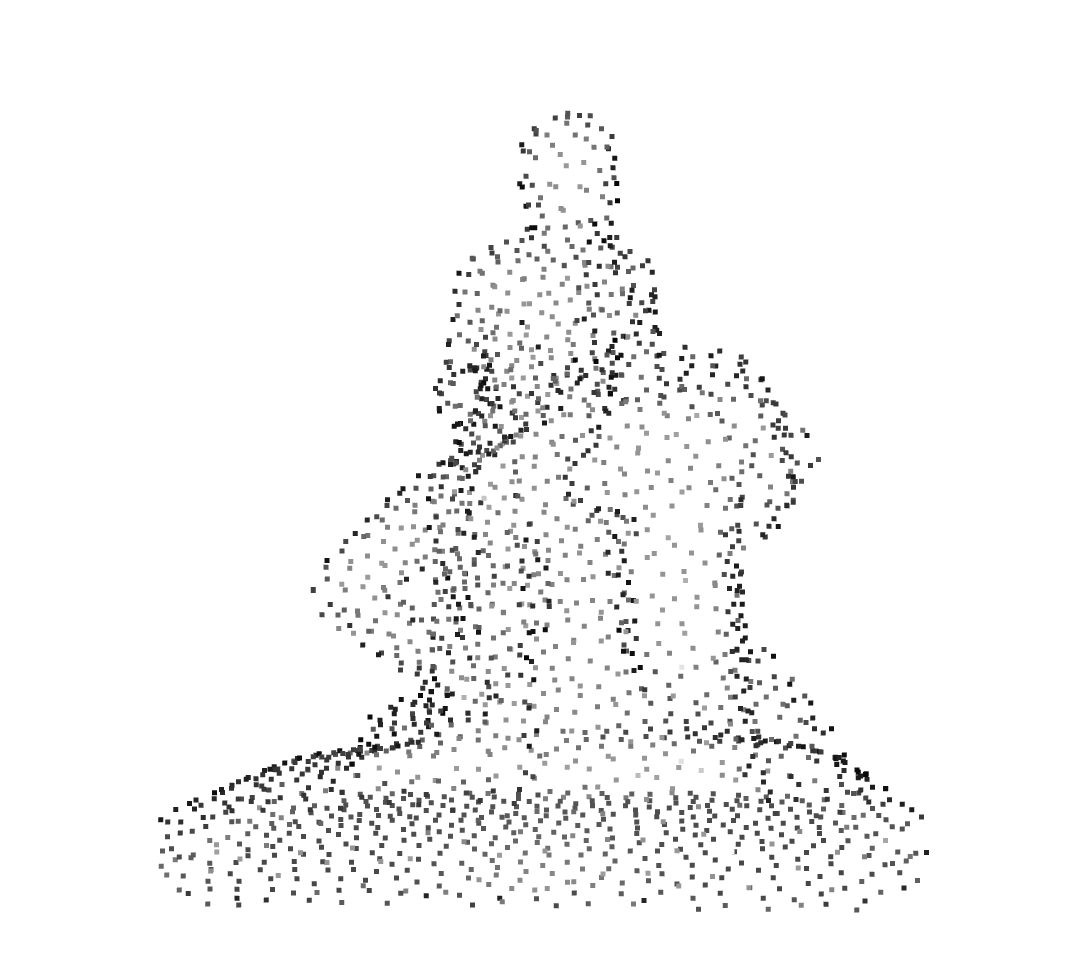} &
    \includegraphics[width=0.15\linewidth]{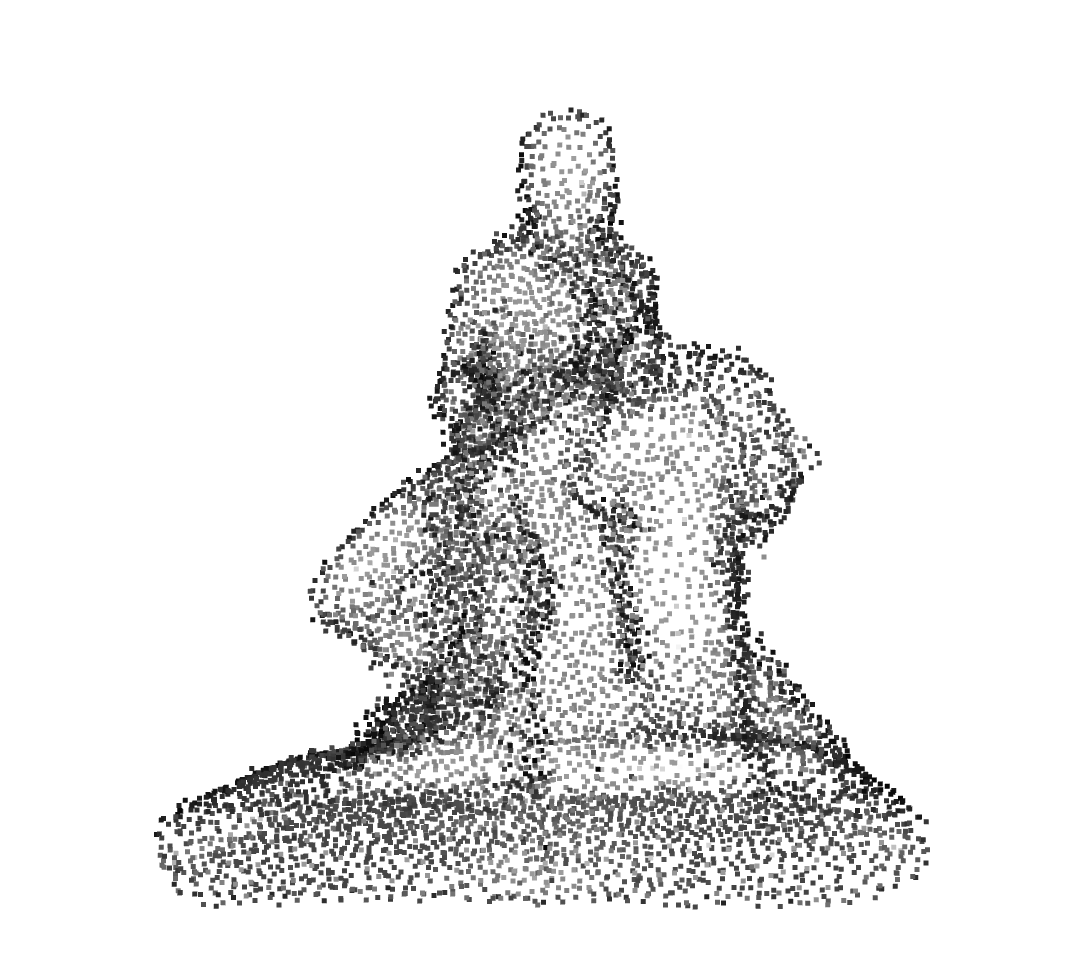} &
    \includegraphics[width=0.15\linewidth]{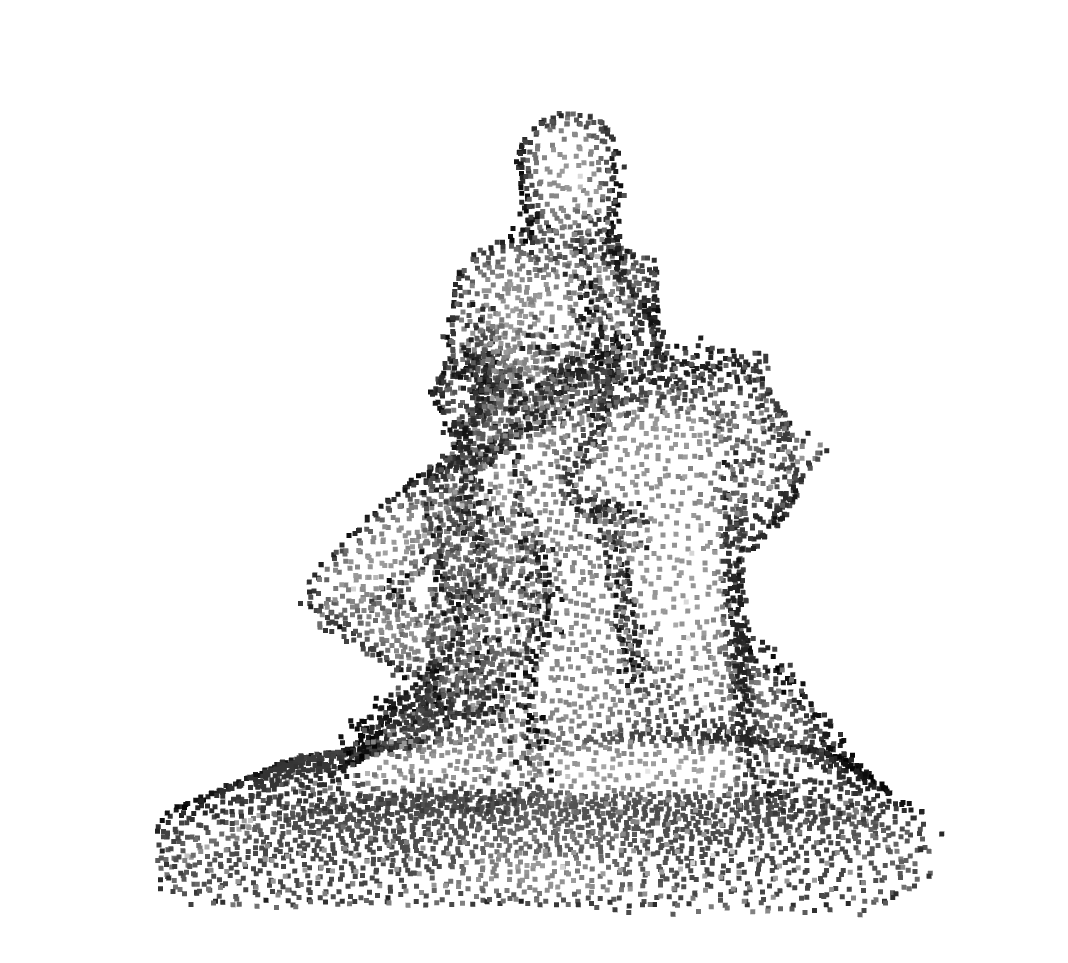} &
    \includegraphics[width=0.15\linewidth]{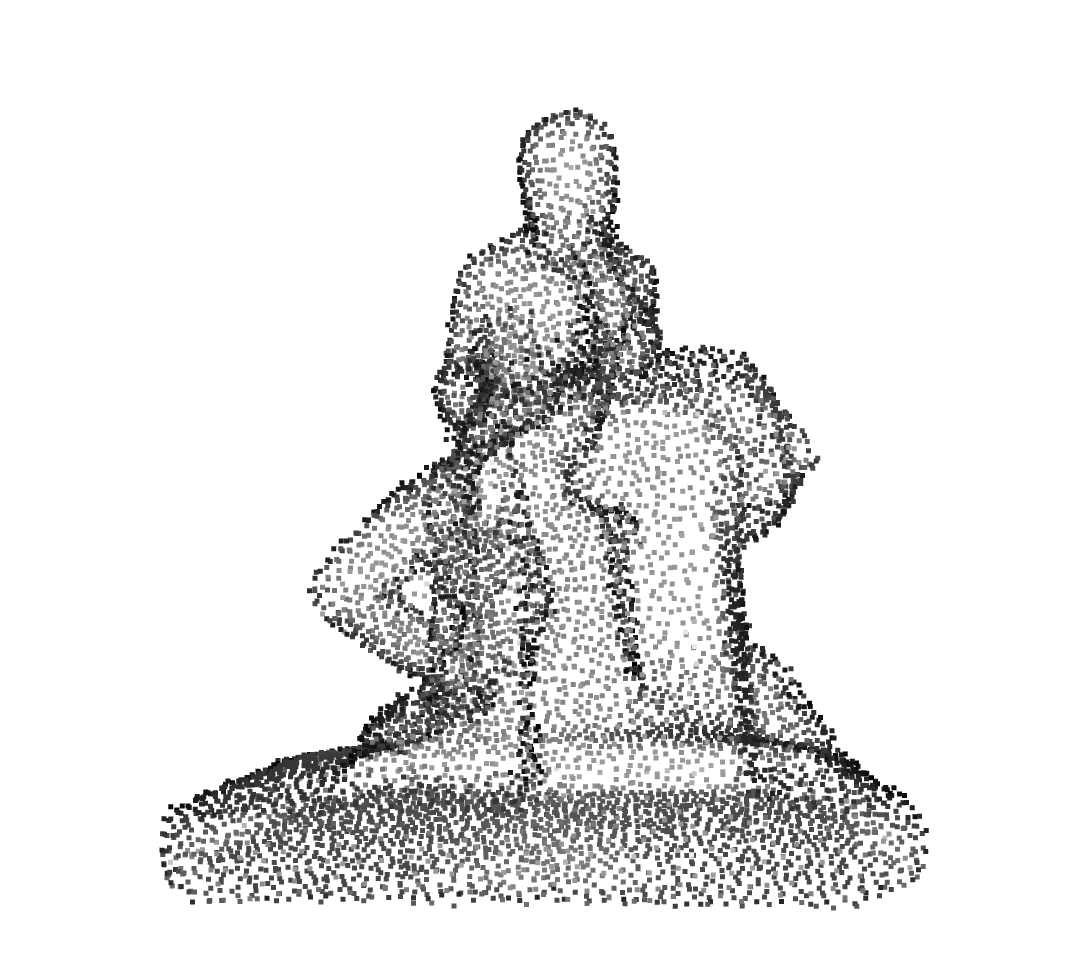} &
    \includegraphics[width=0.15\linewidth]{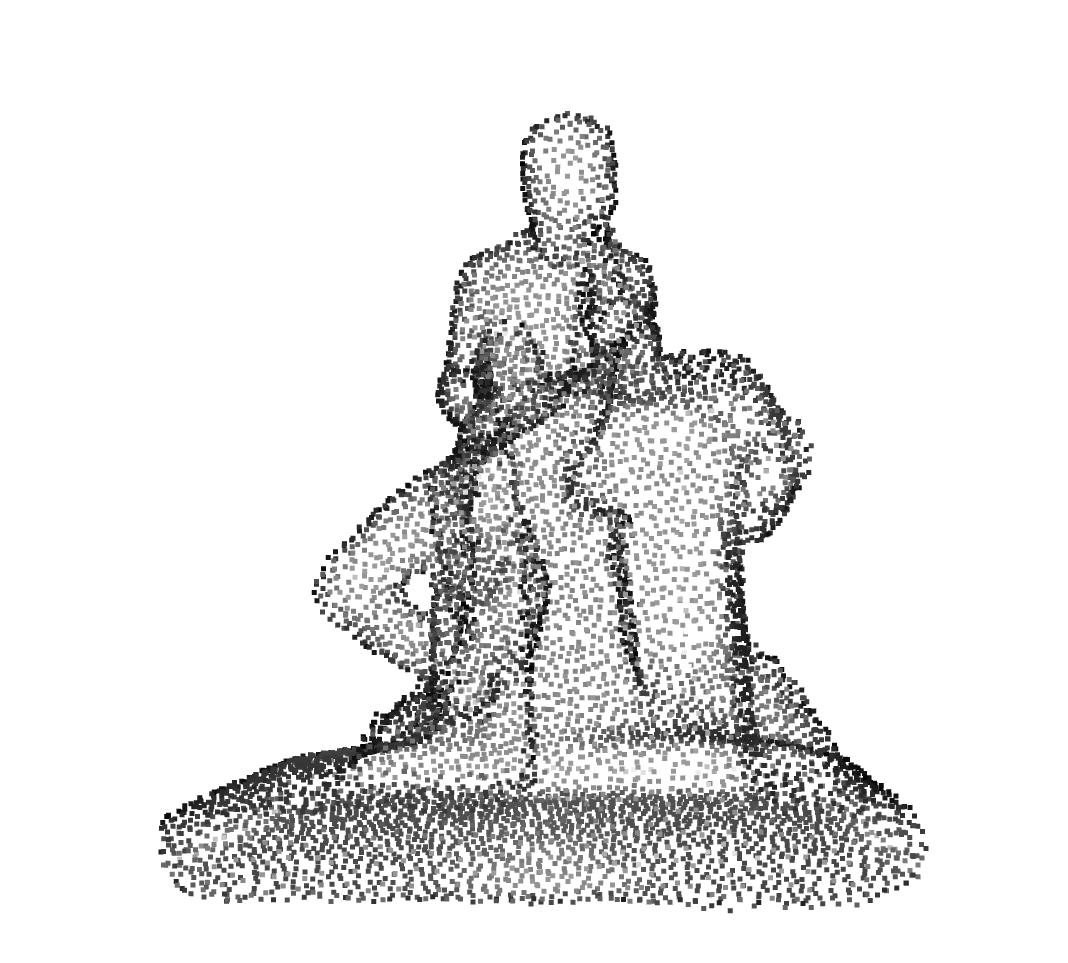} &
    \includegraphics[width=0.15\linewidth]{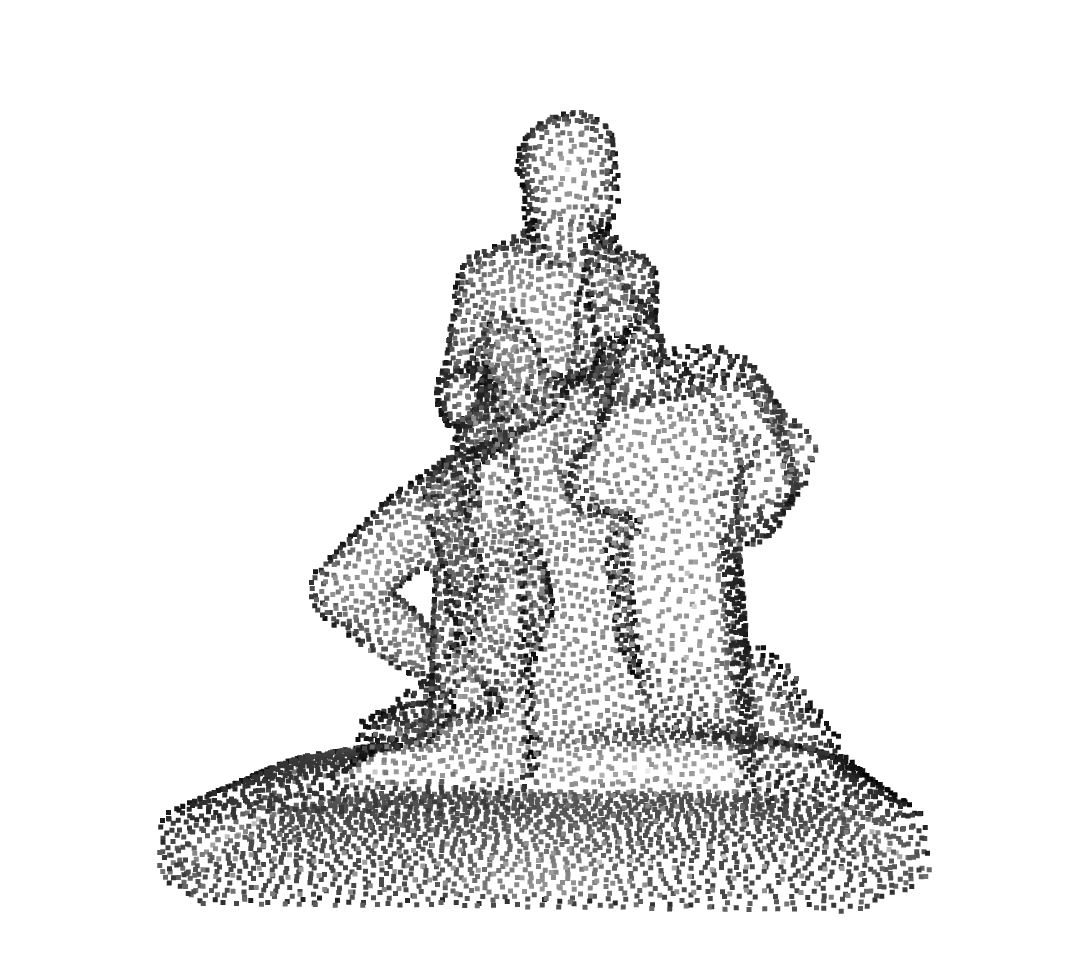} \\
    \includegraphics[width=0.15\linewidth]{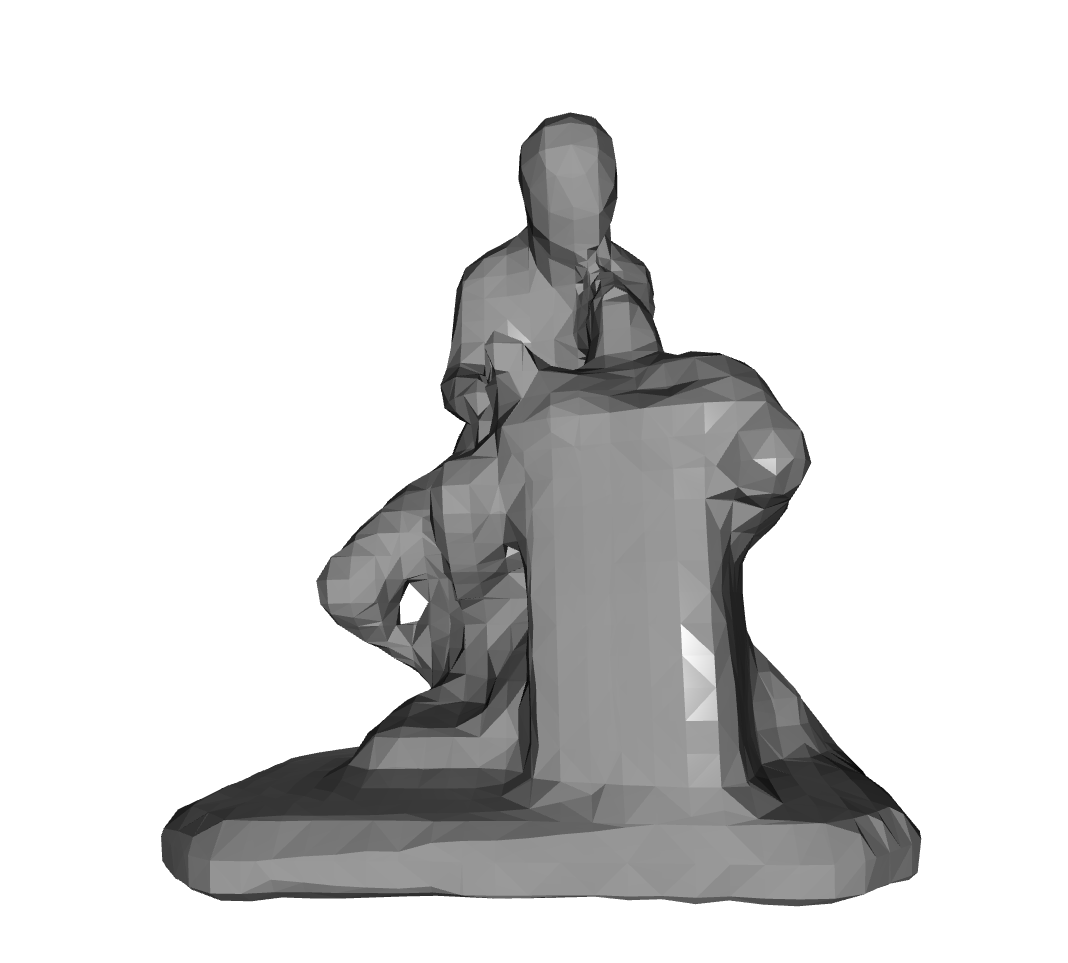} &
    \includegraphics[width=0.15\linewidth]{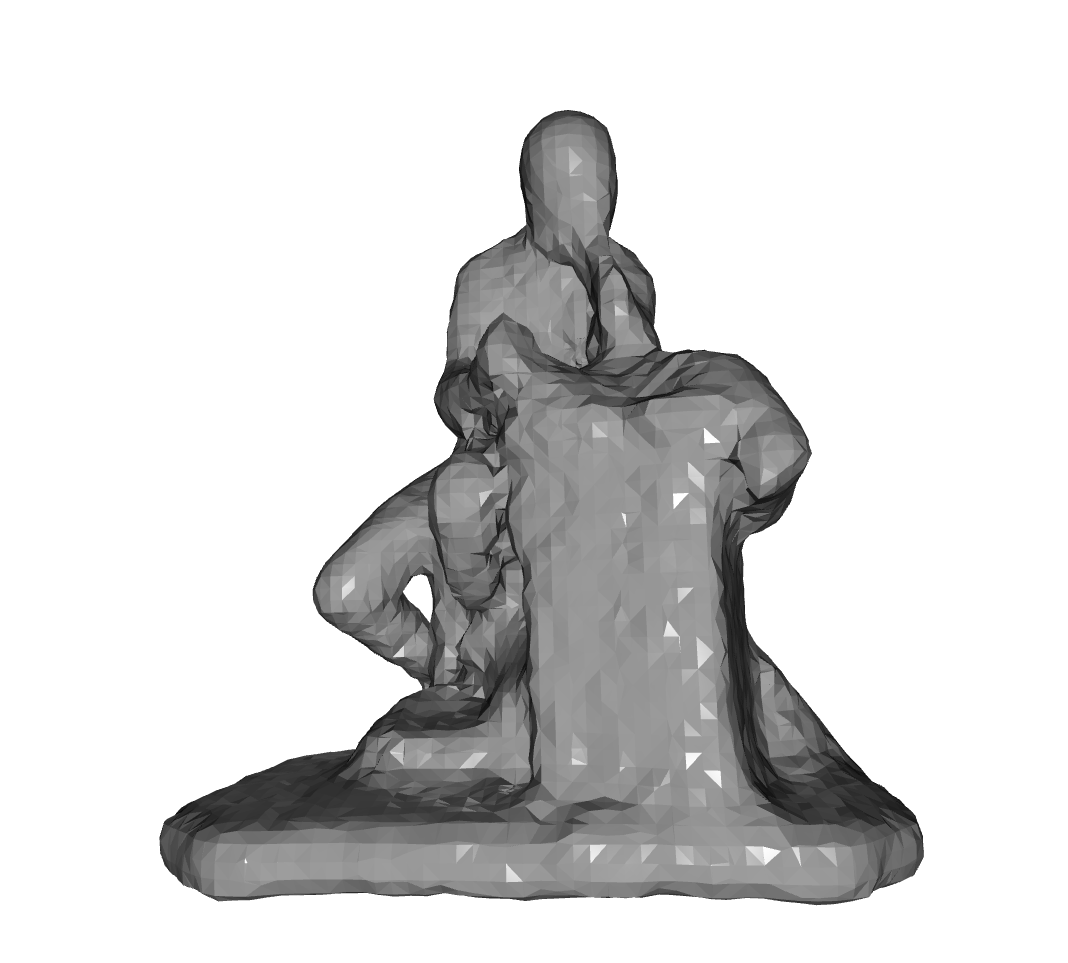} &
    \includegraphics[width=0.15\linewidth]{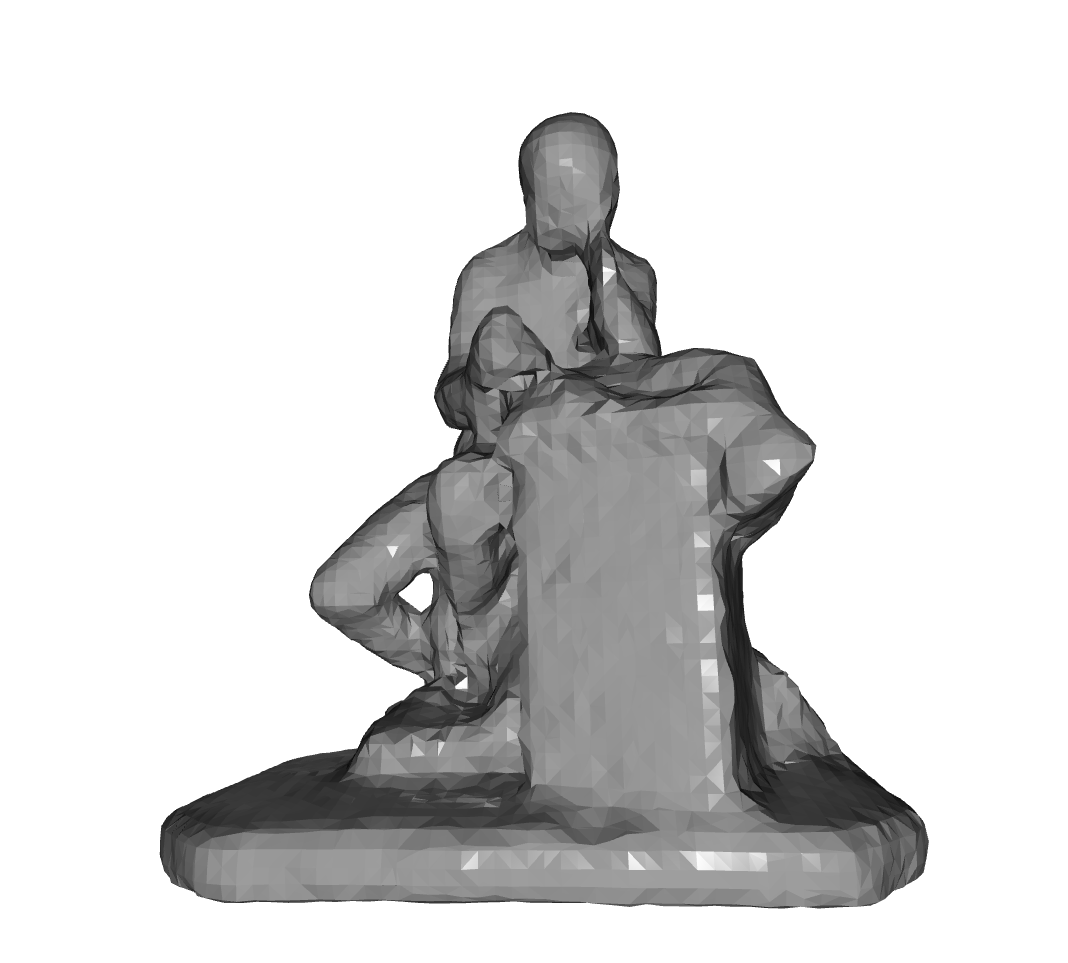} &
    \includegraphics[width=0.15\linewidth]{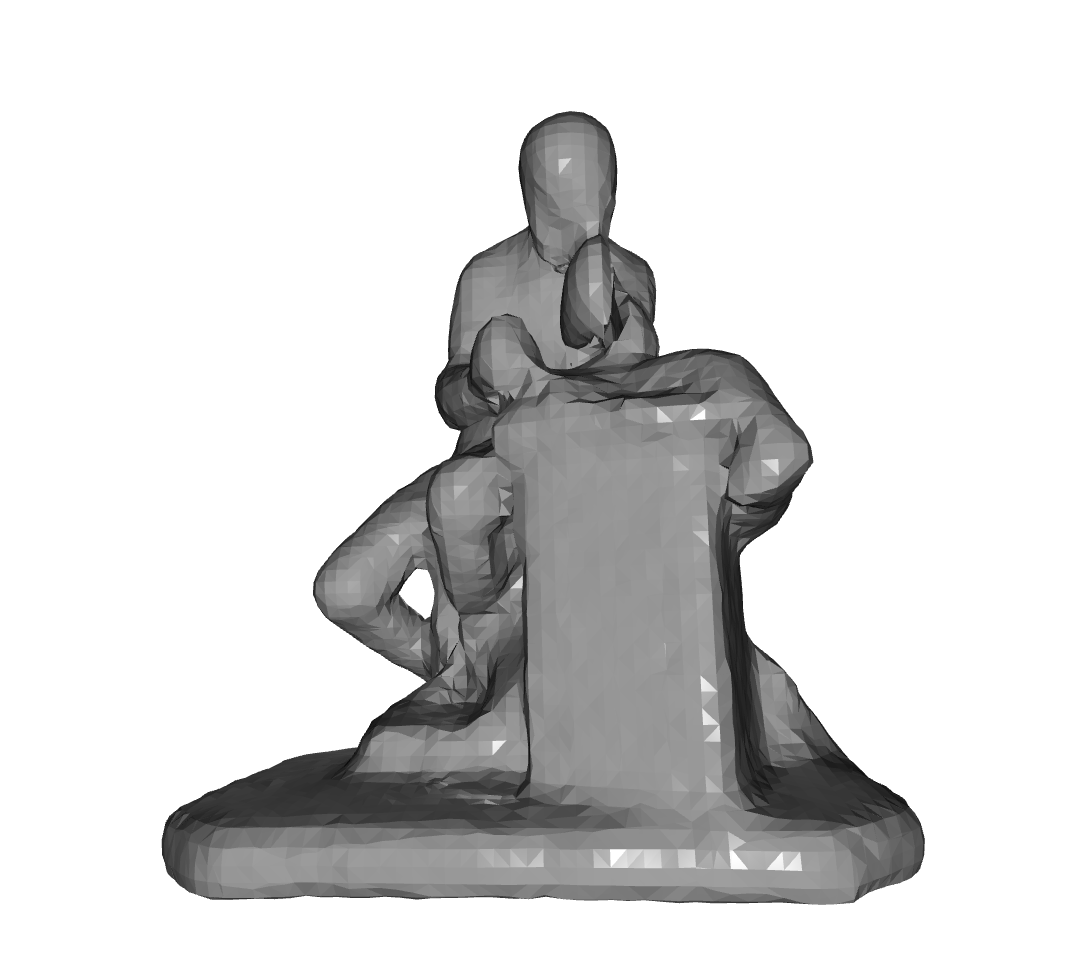} &
    \includegraphics[width=0.15\linewidth]{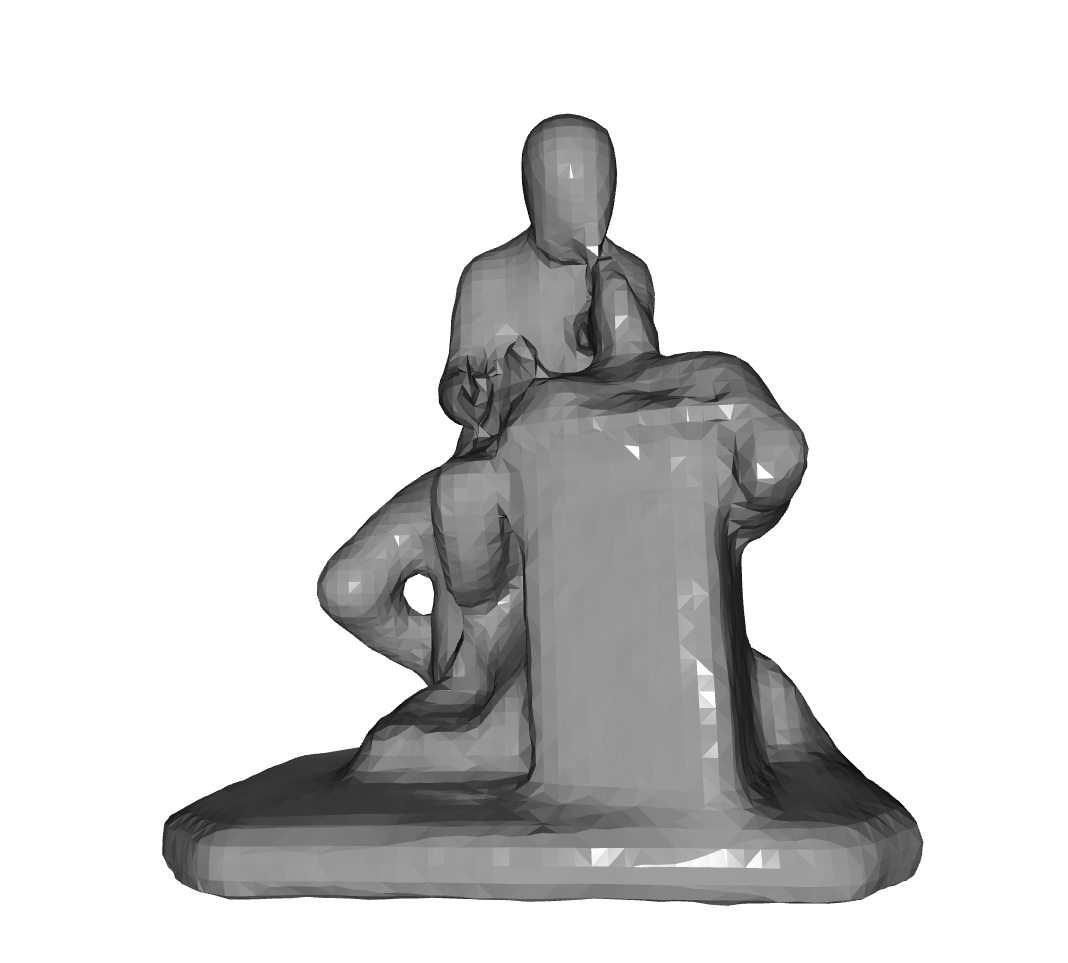} &
    \includegraphics[width=0.15\linewidth]{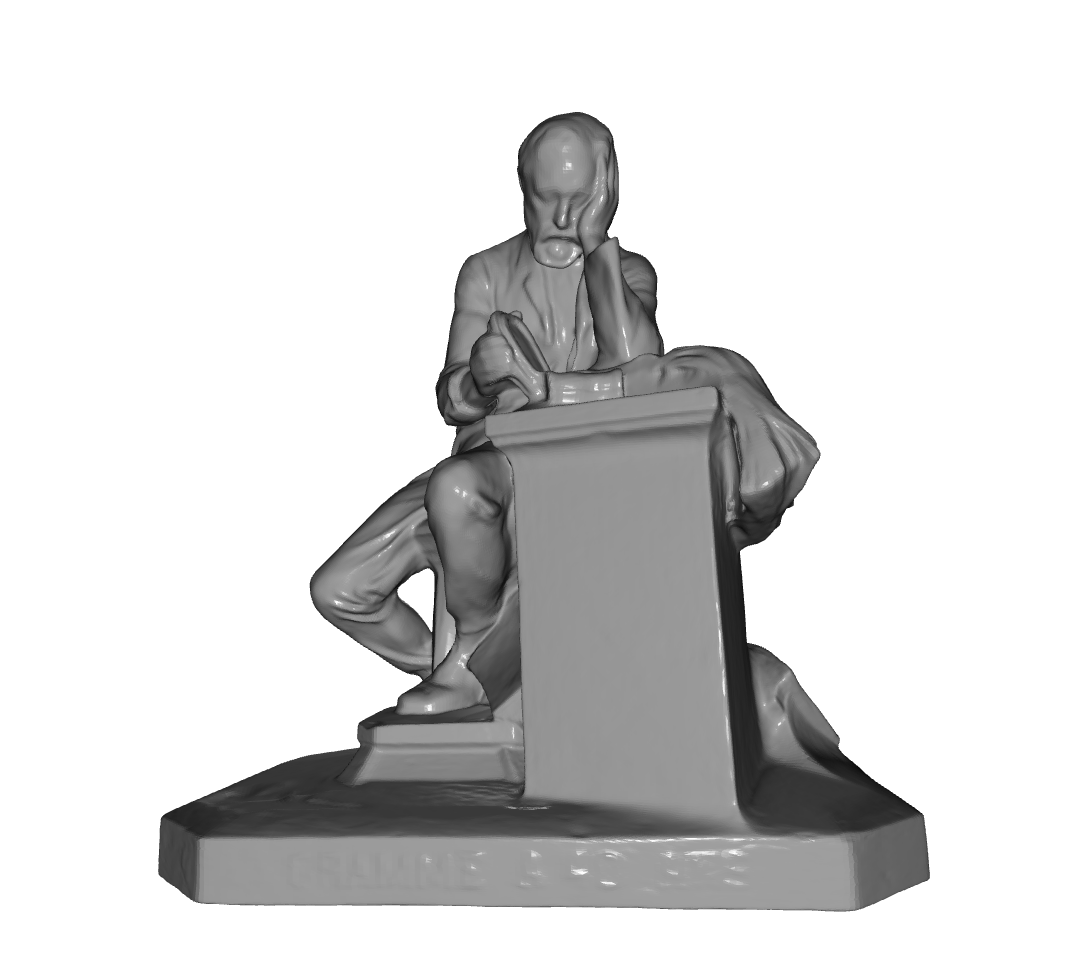} \\
    Input & PU-GAN \cite{pugan} & PU-GCN \cite{pugcn} & Dis-PU \cite{dispu} & Ours & Ground Truth
\end{tabular}
\addtolength{\tabcolsep}{2pt}
\vspace{0.2cm}
\captionof{figure}{Qualitative comparison with state-of-the-art methods when the upsampled point clouds are used to compute a mesh with PSR.}
\label{fig:supp_mesh_1}
\end{table}
\begin{table}[h!]
\centering
\addtolength{\tabcolsep}{-2pt} 
\begin{tabular}{ c c c c c c}
    \includegraphics[width=0.15\linewidth]{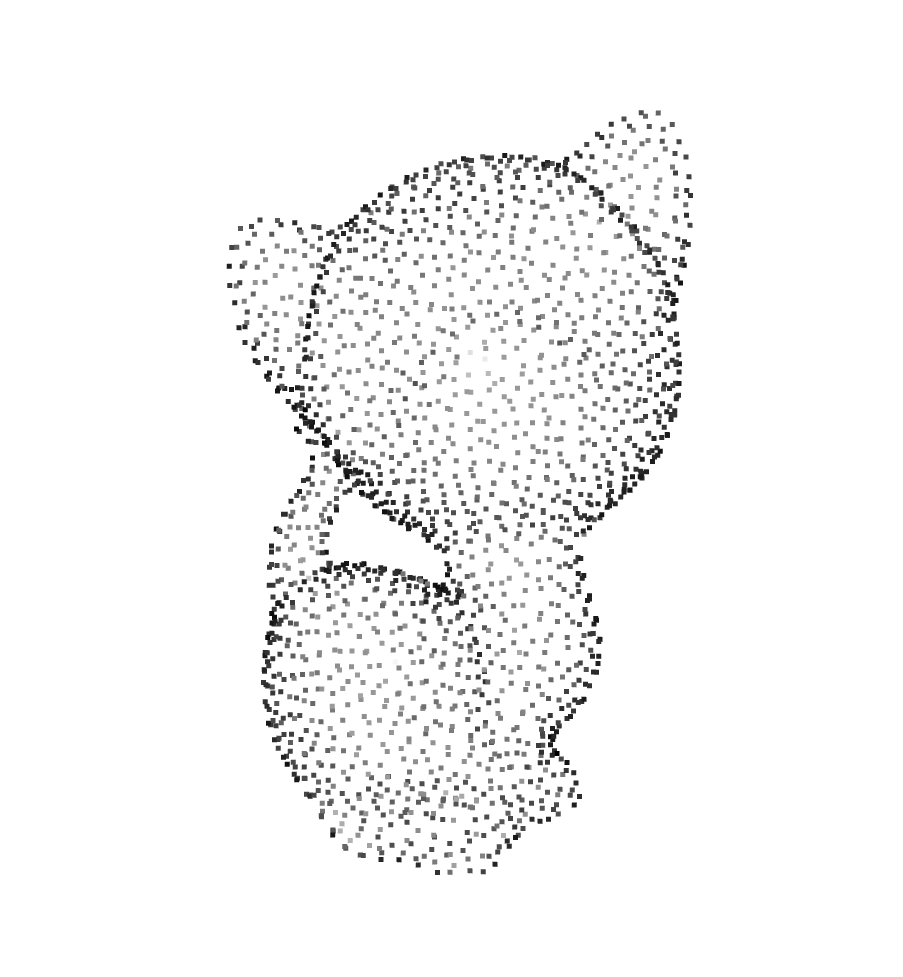} &
    \includegraphics[width=0.15\linewidth]{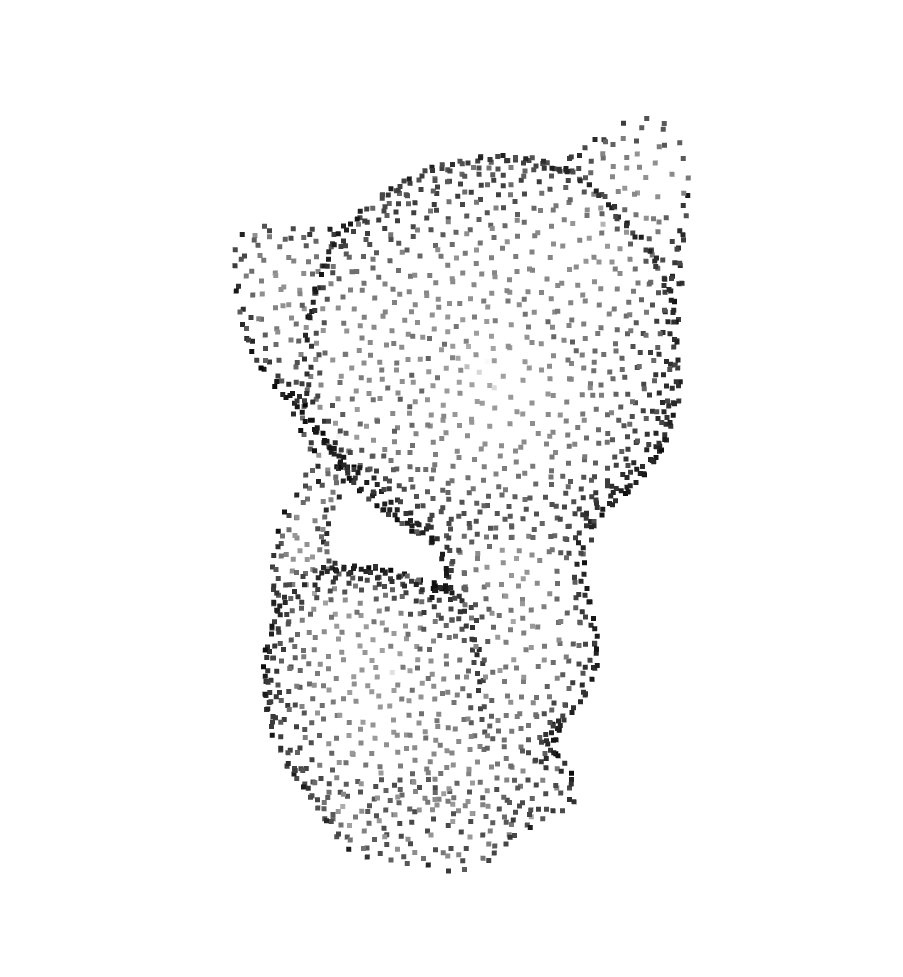} &
    \includegraphics[width=0.15\linewidth]{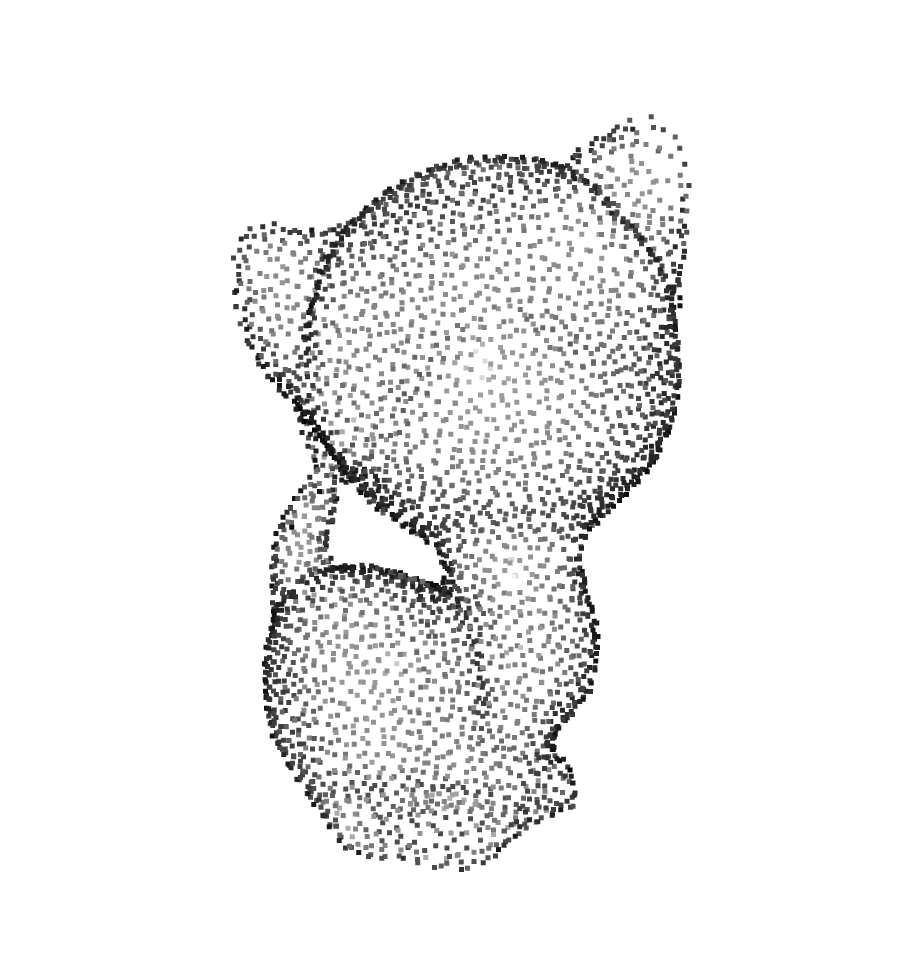} &
    \includegraphics[width=0.15\linewidth]{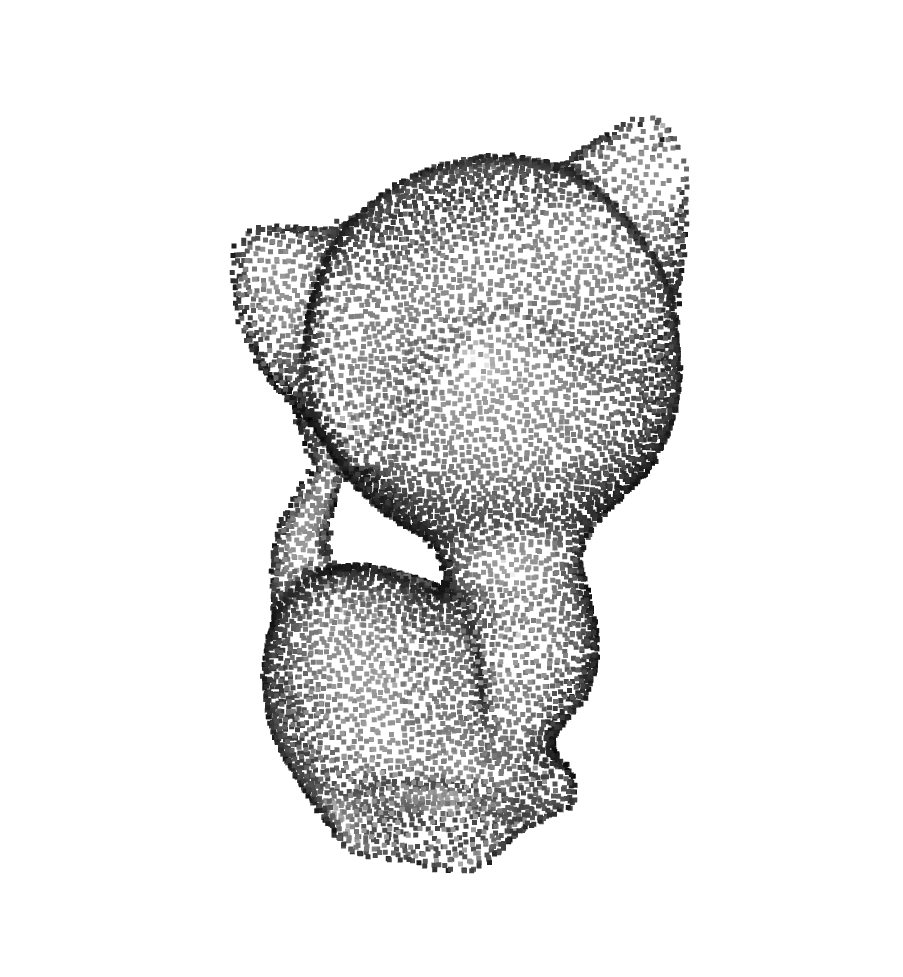} &
    \includegraphics[width=0.15\linewidth]{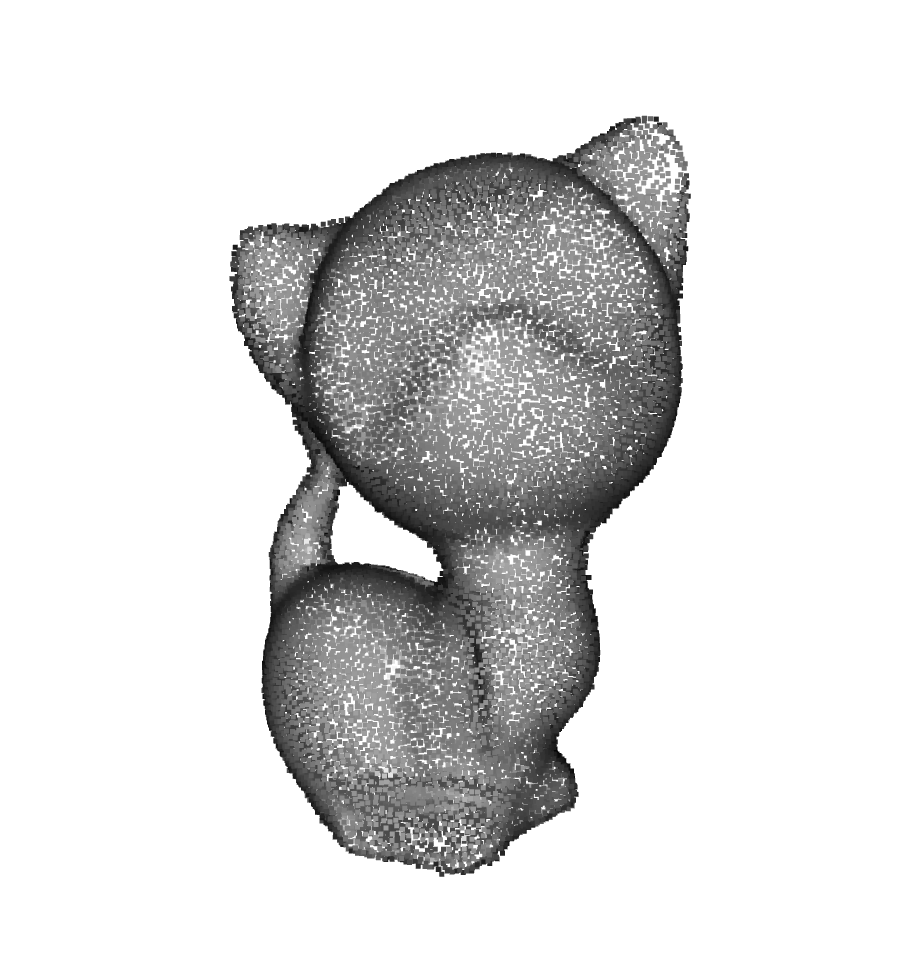} &
    \includegraphics[width=0.15\linewidth]{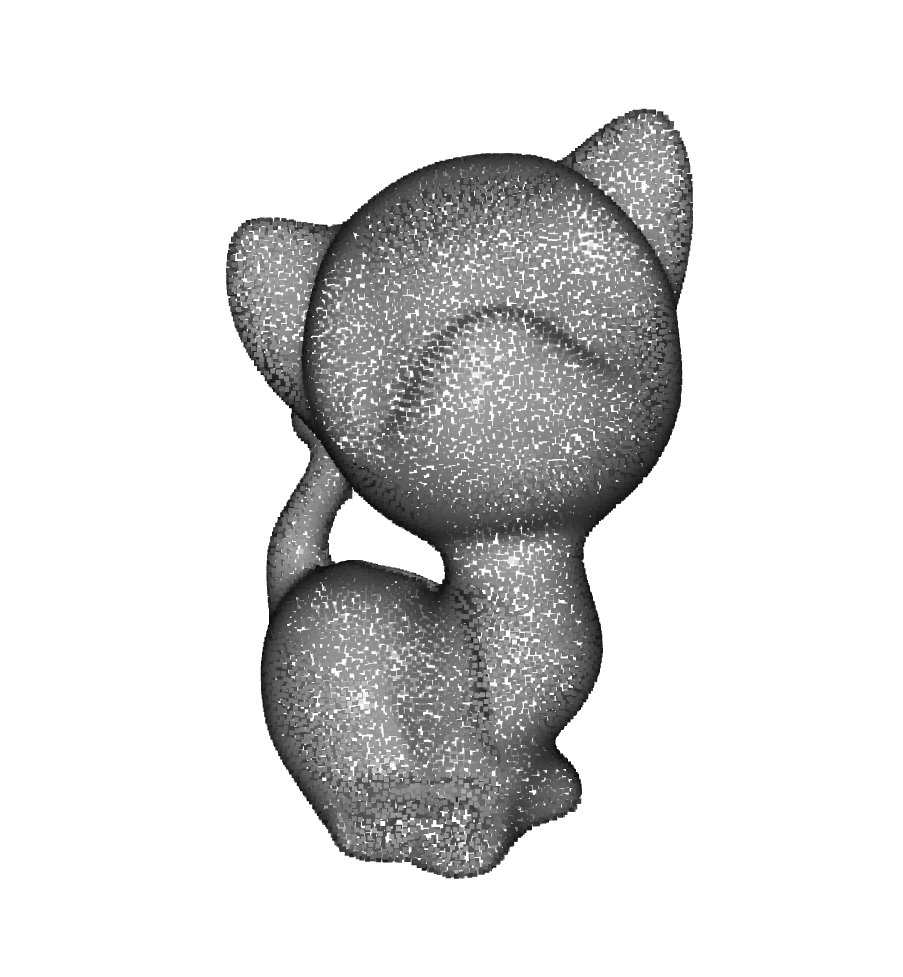} \\
    \includegraphics[width=0.15\linewidth]{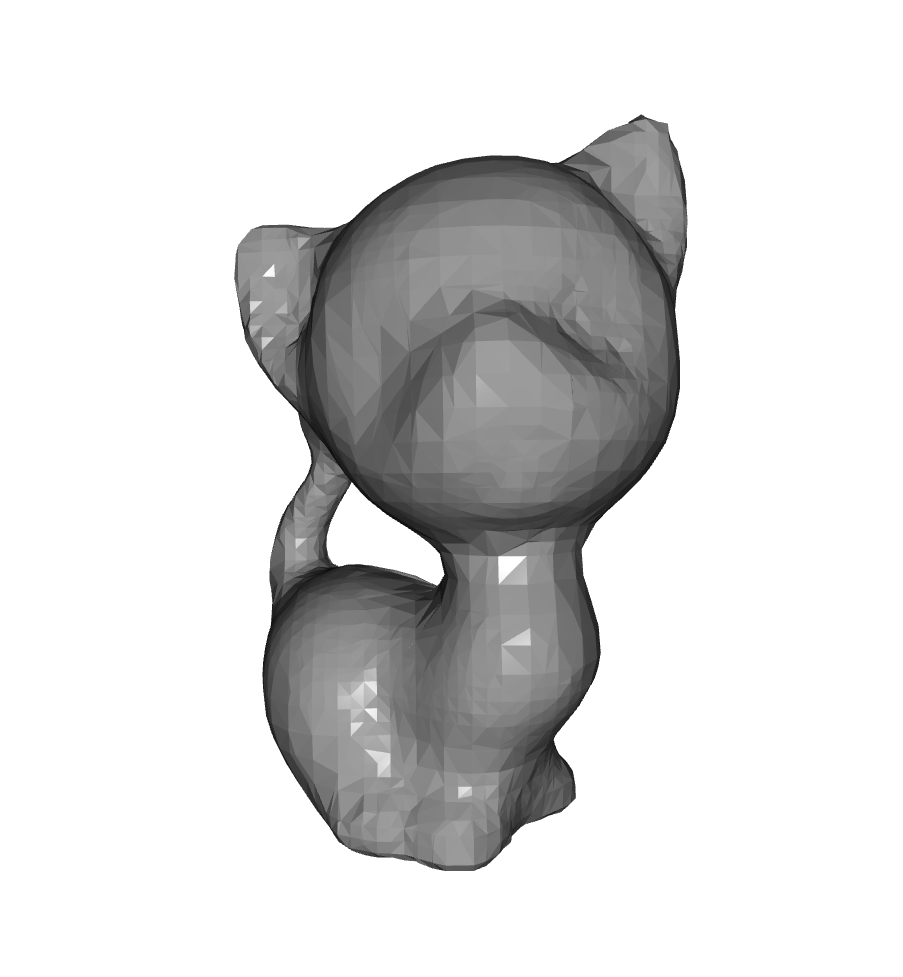} &
    \includegraphics[width=0.15\linewidth]{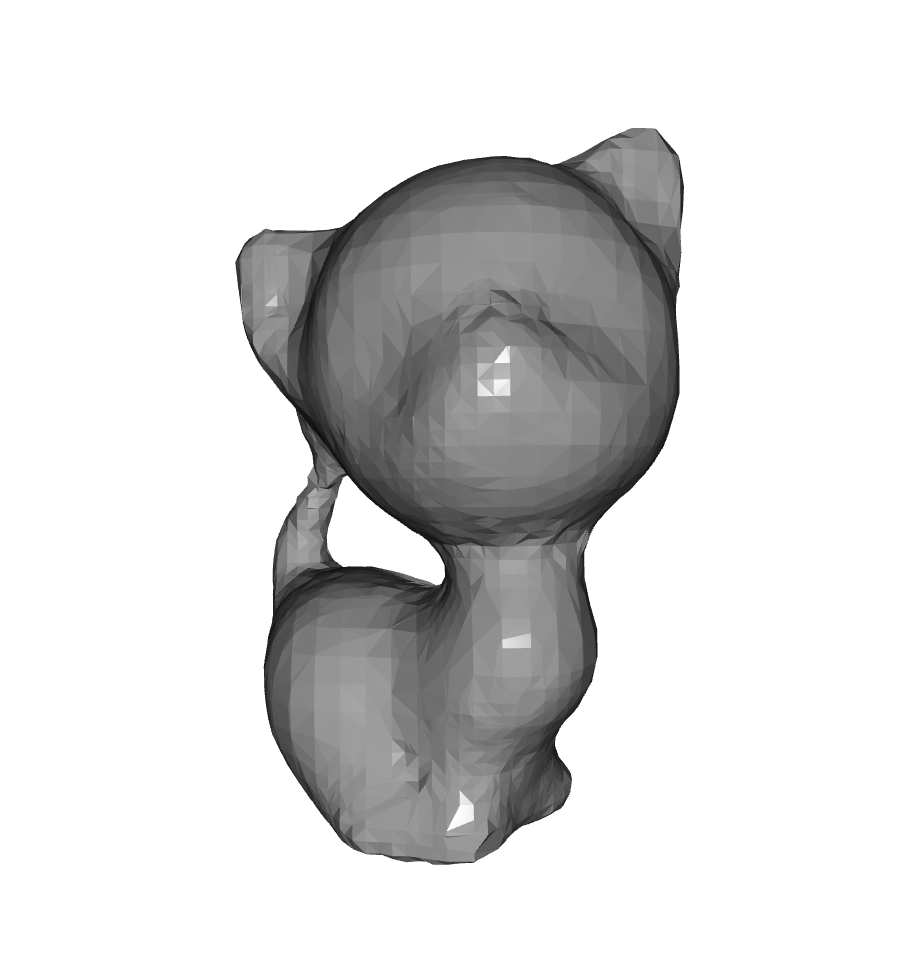} &
    \includegraphics[width=0.15\linewidth]{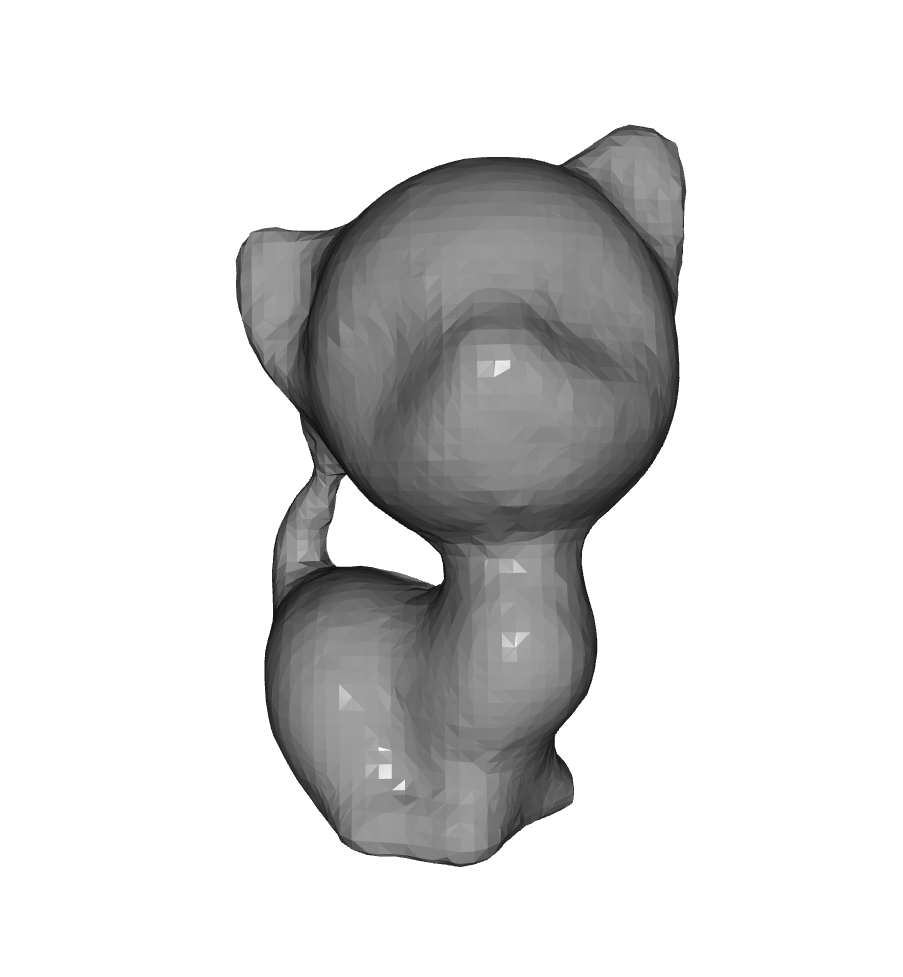} &
    \includegraphics[width=0.15\linewidth]{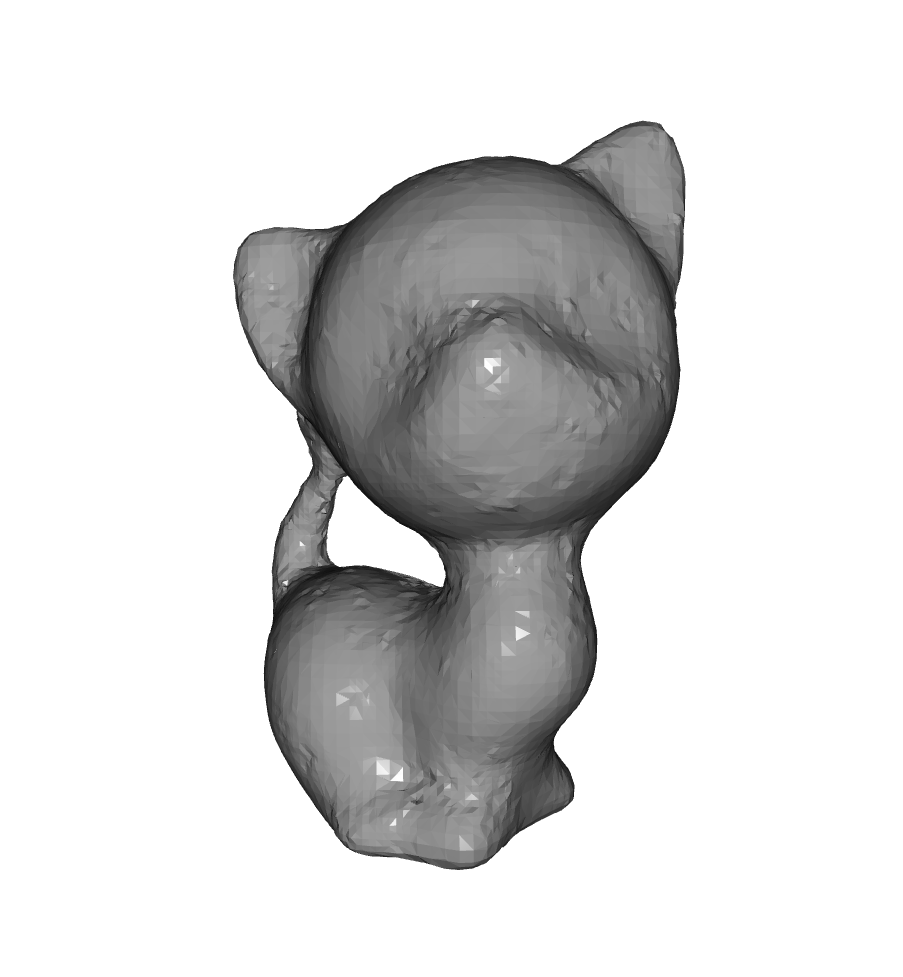} &
    \includegraphics[width=0.15\linewidth]{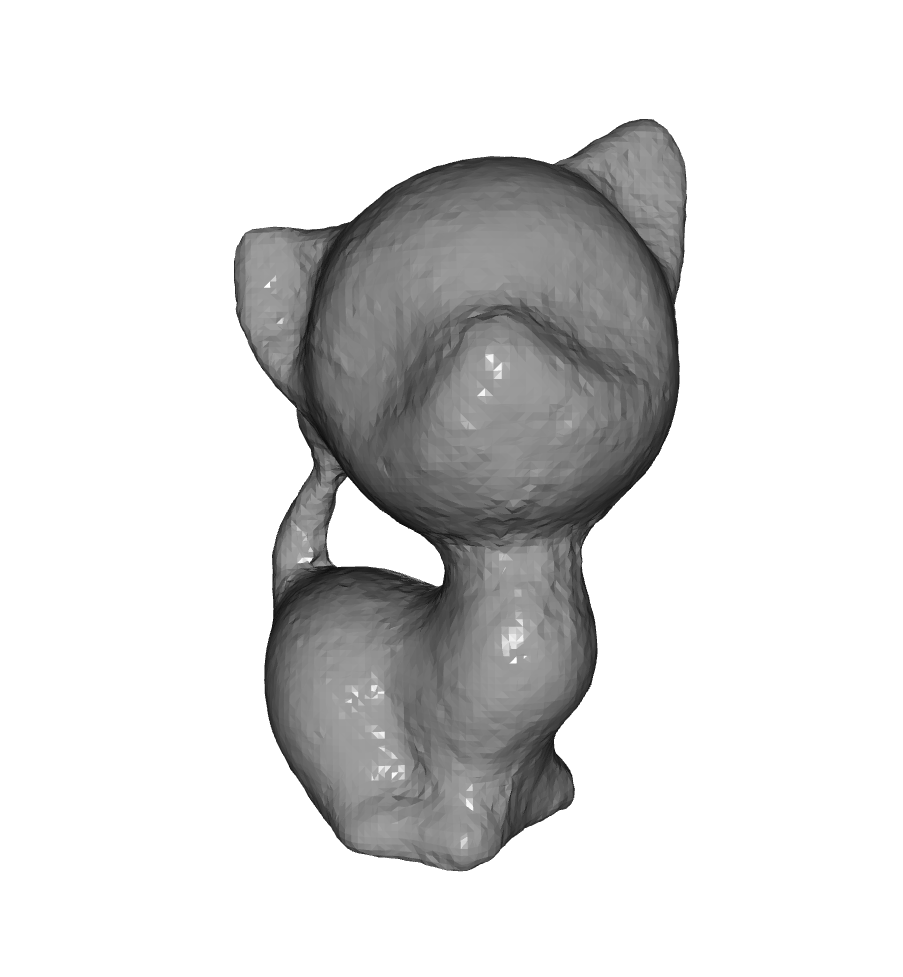} &
    \includegraphics[width=0.15\linewidth]{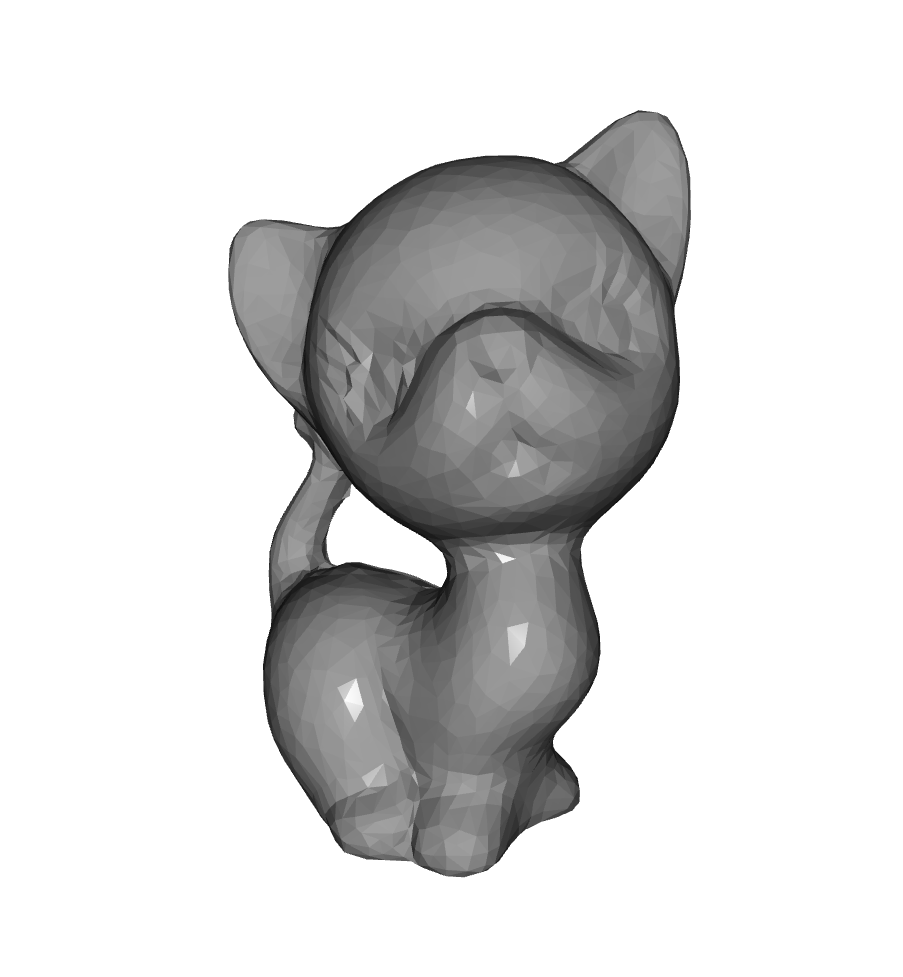} \\
    \includegraphics[width=0.15\linewidth]{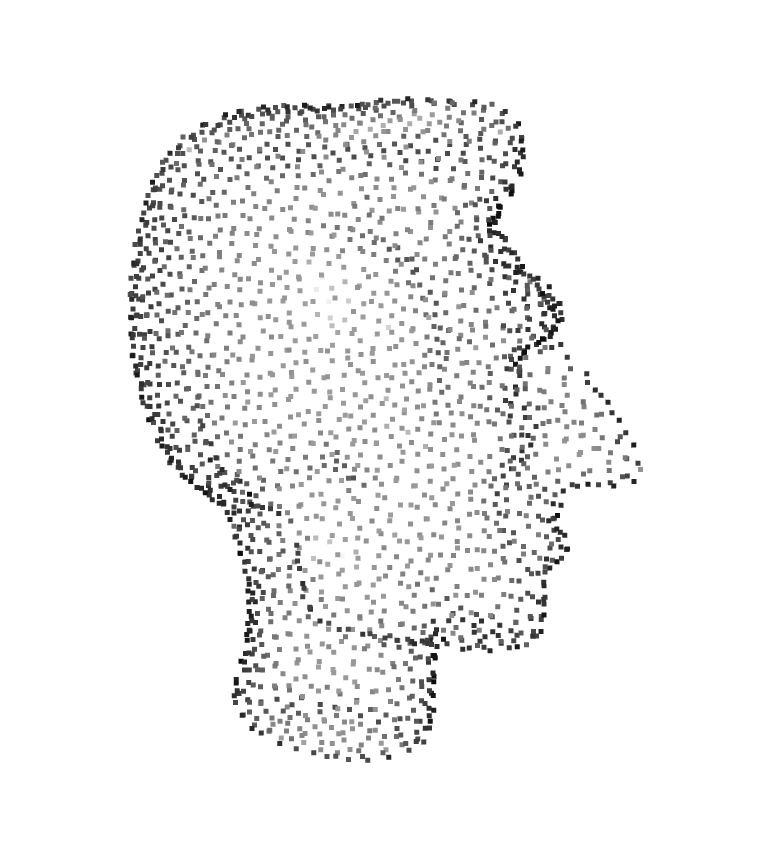} &
    \includegraphics[width=0.15\linewidth]{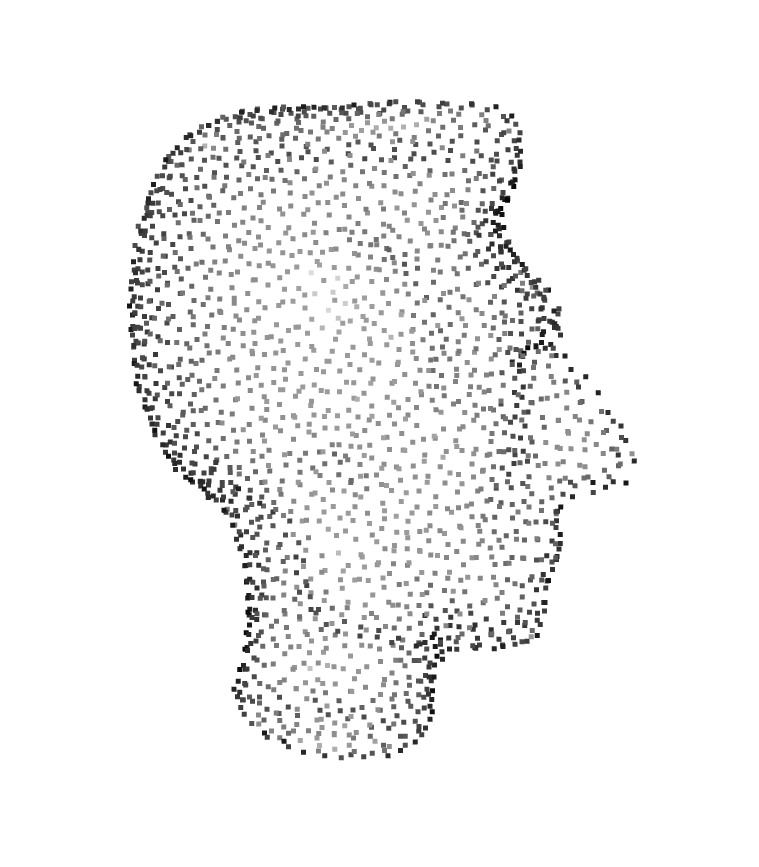} &
    \includegraphics[width=0.15\linewidth]{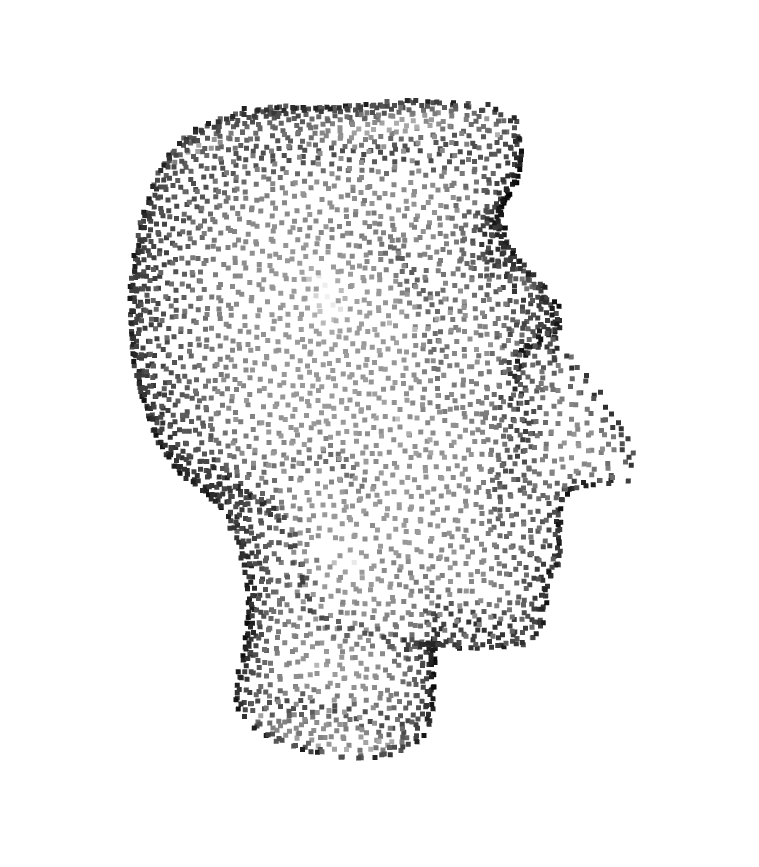} &
    \includegraphics[width=0.15\linewidth]{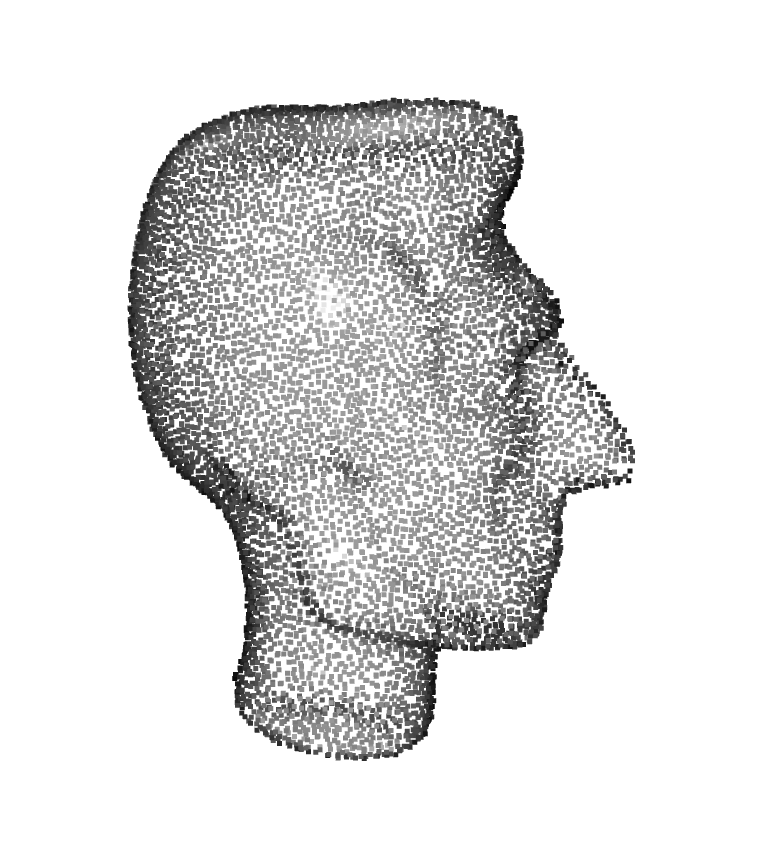} &
    \includegraphics[width=0.15\linewidth]{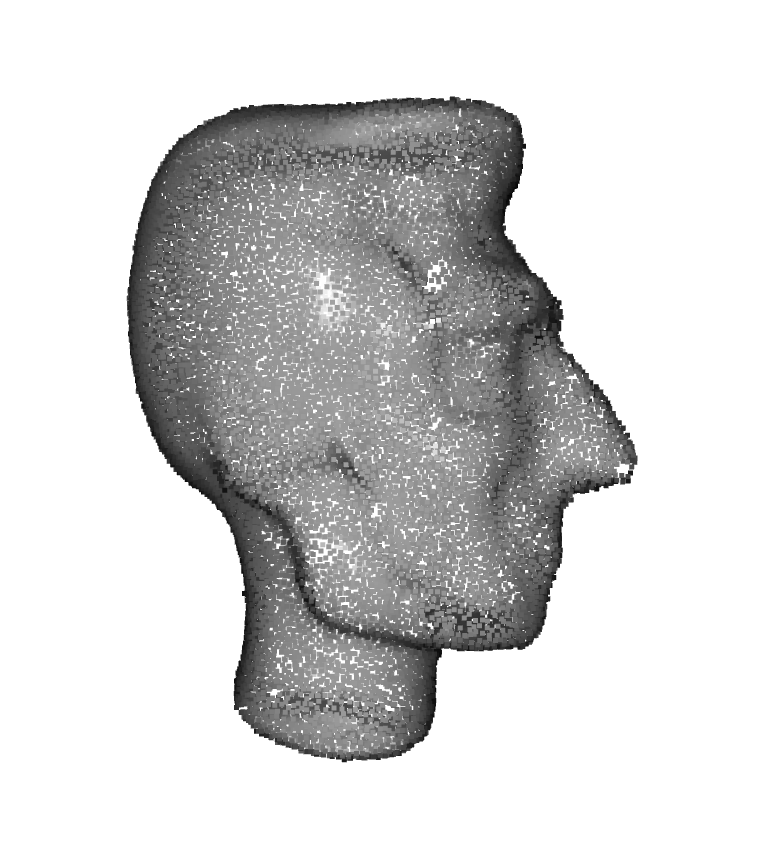} &
    \includegraphics[width=0.15\linewidth]{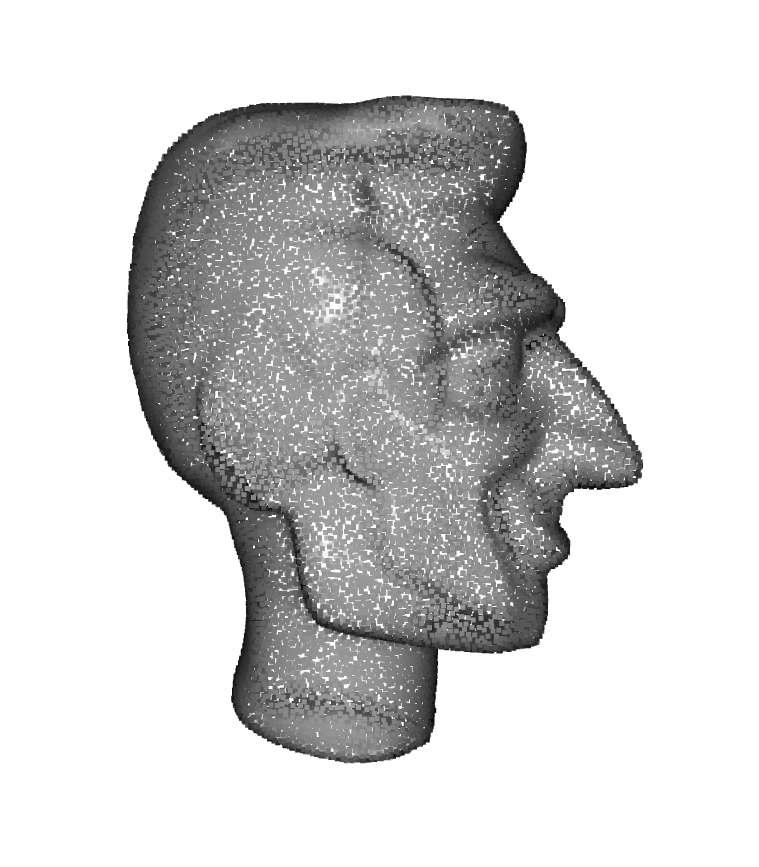} \\
    \includegraphics[width=0.15\linewidth]{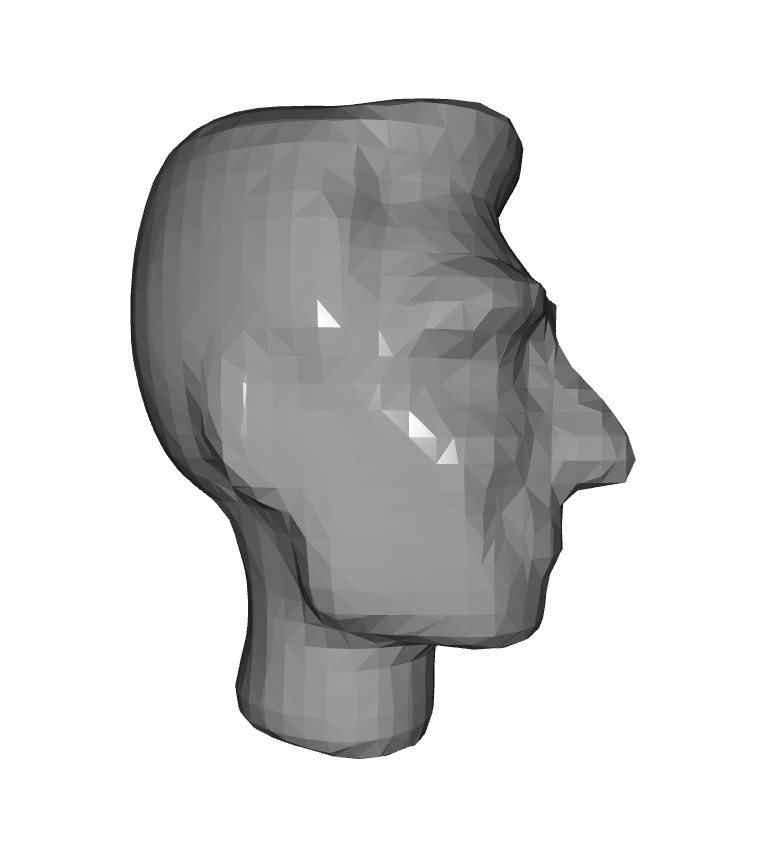} &
    \includegraphics[width=0.15\linewidth]{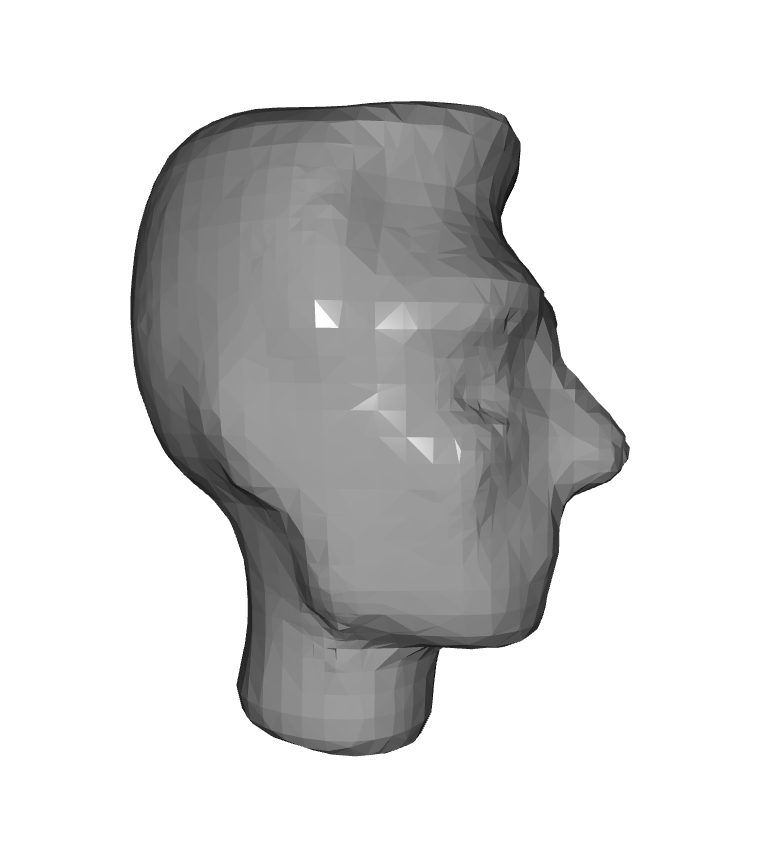} &
    \includegraphics[width=0.15\linewidth]{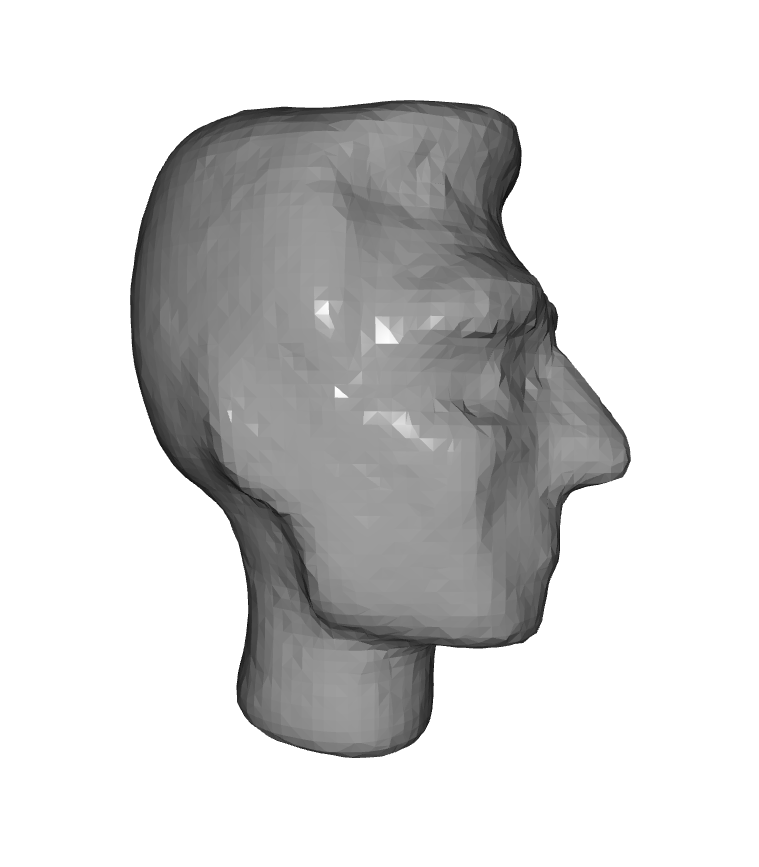} &
    \includegraphics[width=0.15\linewidth]{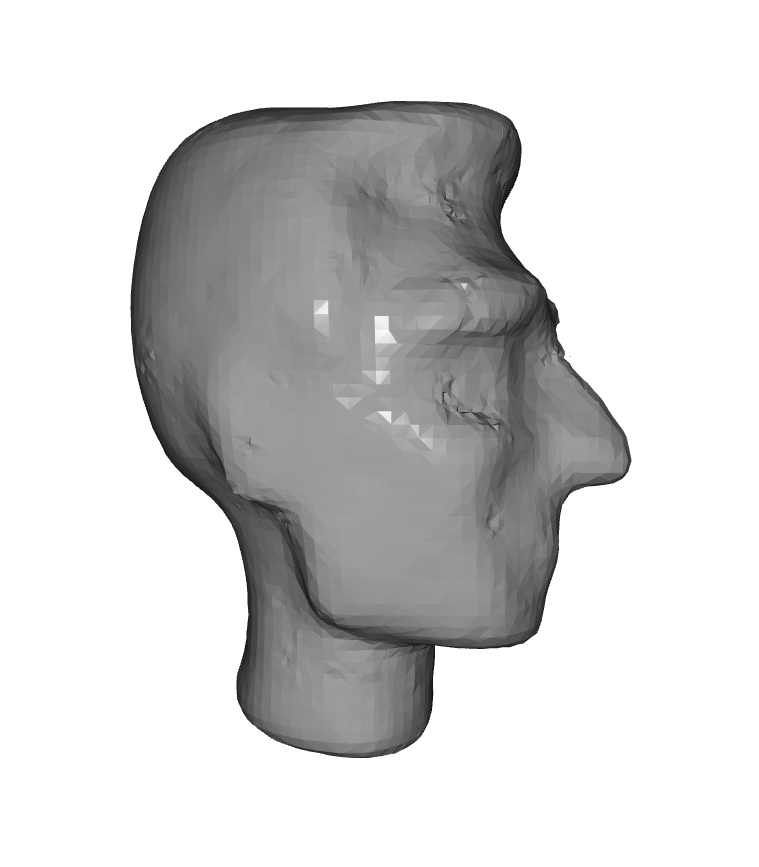} &
    \includegraphics[width=0.15\linewidth]{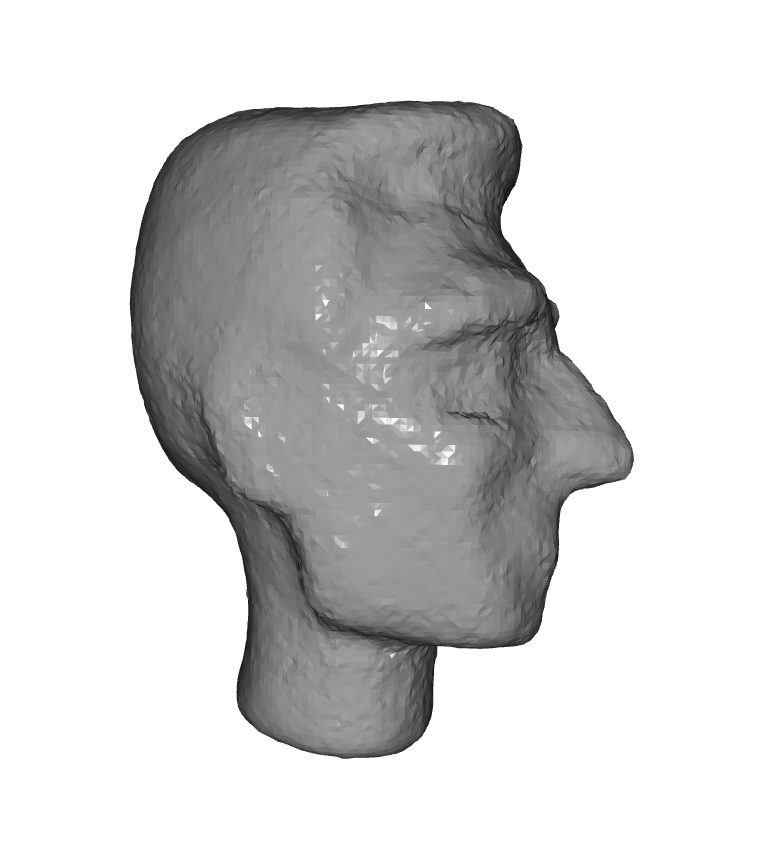} &
    \includegraphics[width=0.15\linewidth]{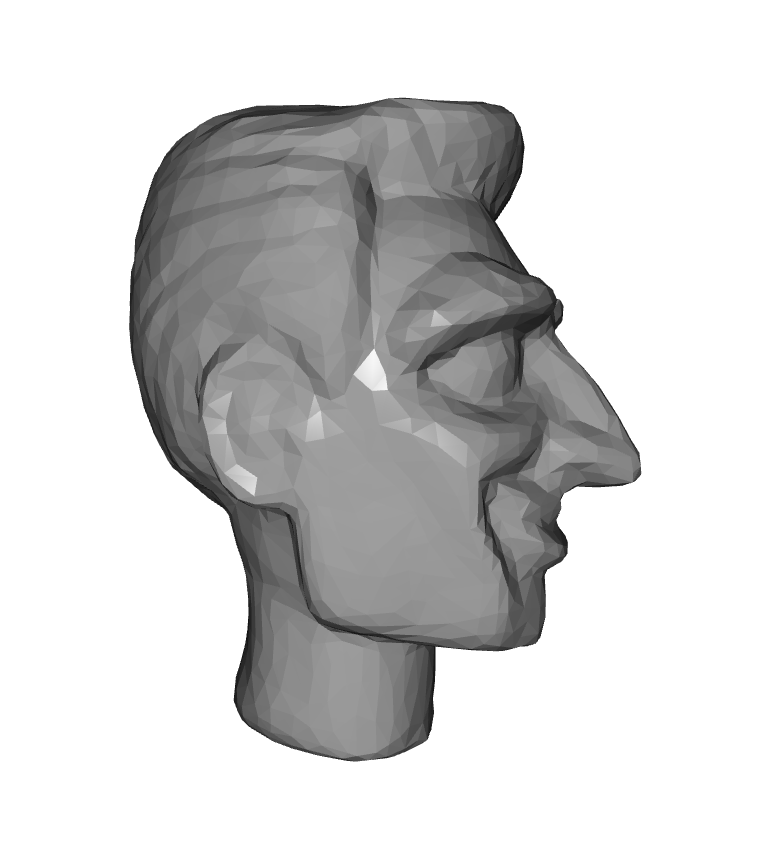} \\
    Input & $r = 1.98$ & $r = 3.45$ & $r = 7.72$ & $r = 18.11$ & Ground Truth
\end{tabular}
\addtolength{\tabcolsep}{2pt}
\vspace{0.2cm}
\captionof{figure}{Qualitative comparison with arbitrary upsampling ratios when the point clouds are used to compute a mesh with PSR.}
\label{fig:supp_mesh_2}
\end{table}

\section{Additional Qualitative Results}

In this section, we provide additional qualitative results of our method under different settings. Firstly, several upsampled point clouds from the PU1K dataset with $N = 2048$ and $r = 4$ are shown in Fig. \ref{fig:supp_qual_1}. These further examples confirm that APU-SMOG outperforms state-of-the-art methods in preserving fine-grained structures and reducing outliers.

Moreover, we investigate the upsampling results on the PU-GAN dataset \cite{pugan} against concurrent flexible architectures for different ratios $r \in \{4,8,12,16\}$. Close-up views in Fig. \ref{fig:supp_qual_2} show that our approach produces cleaner shapes along the whole range of upsampling factors, even when compared with methods requiring ground truth normals at training time \cite{mafu, neuralpts}.

Finally, Fig. \ref{fig:scanobj} shows additional examples on real-world objects point clouds from the ScanObjectNN dataset \cite{scanobj}. This figure is complementary to the visualization of upsampling results on real LiDAR data from KITTI \cite{kitti}, provided in the main paper. In both cases, our model is robust to domain shifts, without any fine-tuning.

\section{Robustness to Input Size}

We also study the effect of input point cloud size on the upsampling results from a different point of view. Whereas in the main paper we show the differences in the upsampled point clouds for a \textit{fixed} ratio $r = 4$, in Fig. \ref{fig:max_back} the output size is maintained fixed at $M = 8192$ and the upsampling factor varies in $r \in \{16, 8, 4\}$ for different input sizes. It can be seen that APU-SMOG achieves satisfying results also for the sparsest input $N = 512$. This robustness, combined with the flexibility of our method, provides a full control on the input and output sizes for real-world upsampling scenarios, where the number of points might need to adapt to computational and bandwidth constraints.

\begin{table}
\centering
\addtolength{\tabcolsep}{-2pt} 
\begin{tabular}{ c c c c c c}
    \includegraphics[width=0.15\linewidth]{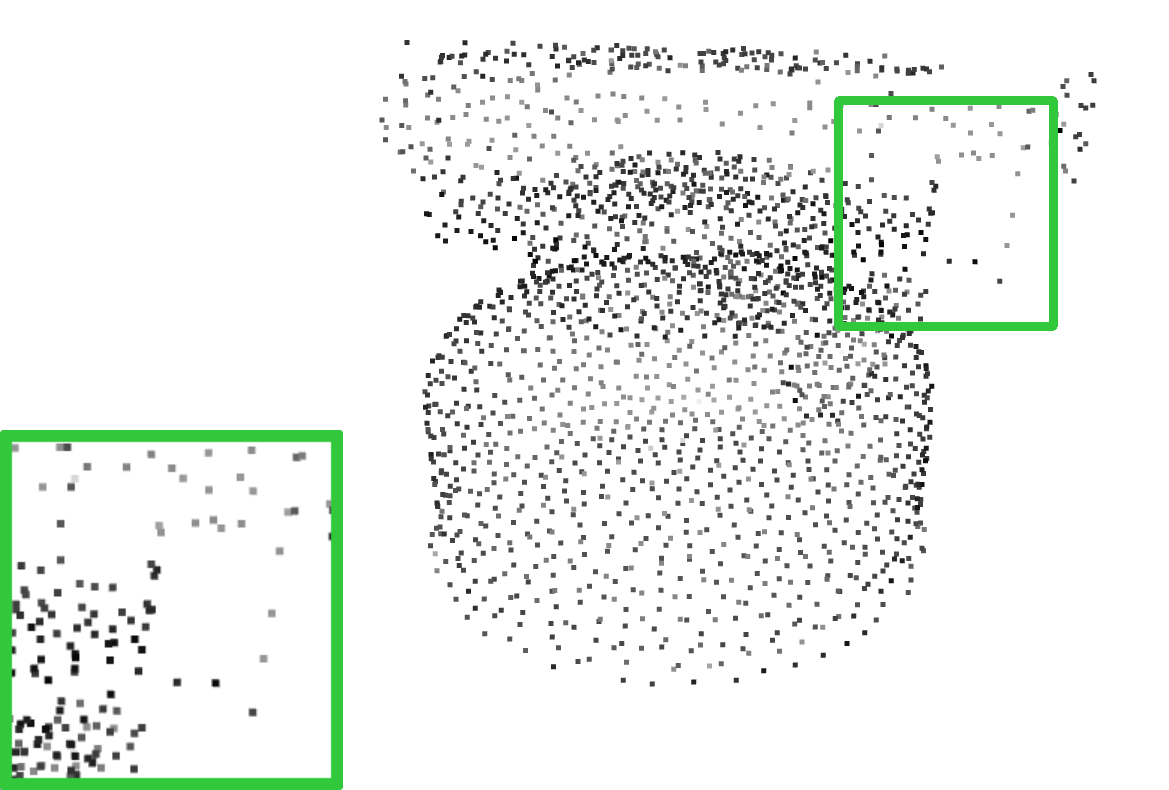} &
    \includegraphics[width=0.15\linewidth]{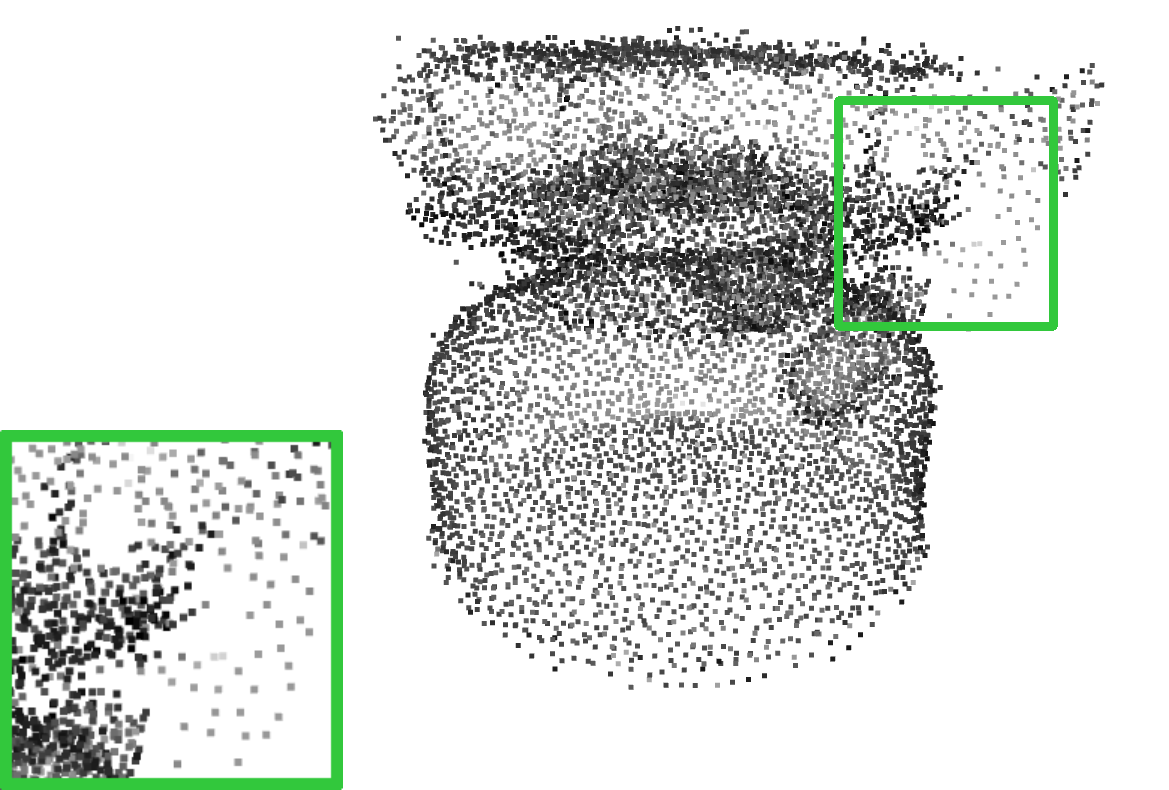} &
    \includegraphics[width=0.15\linewidth]{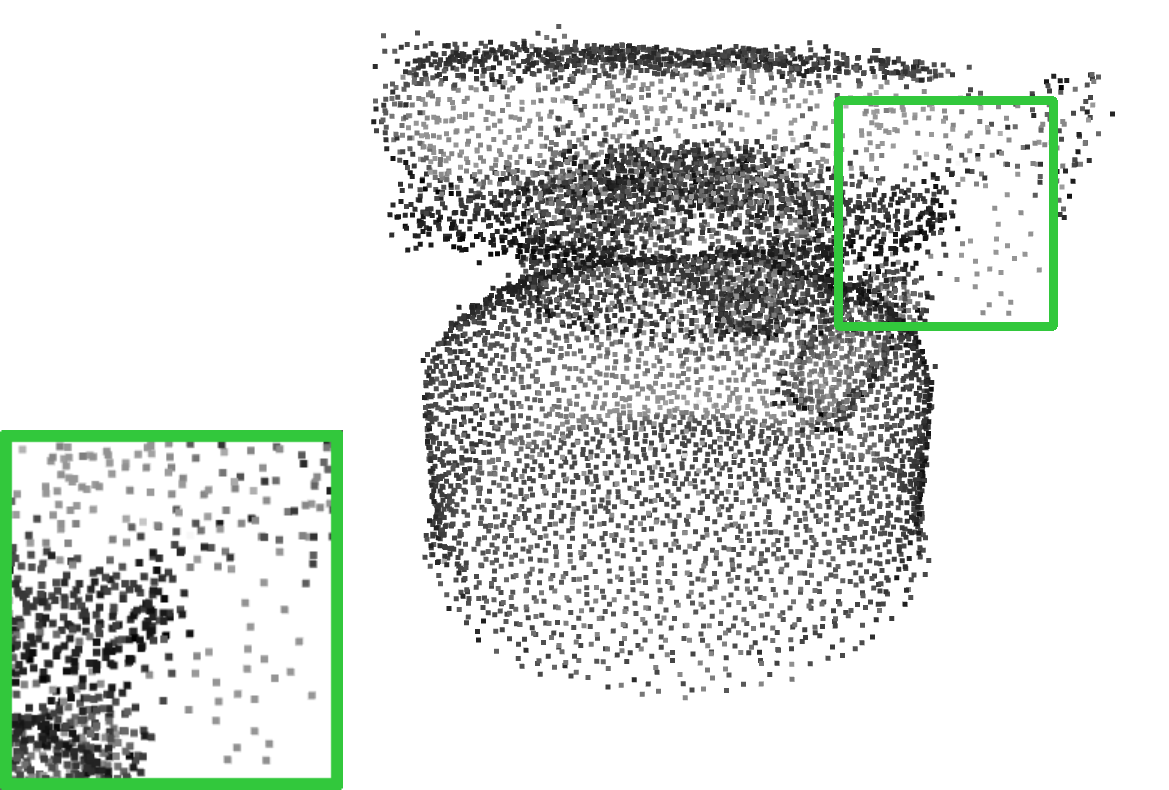} &
    \includegraphics[width=0.15\linewidth]{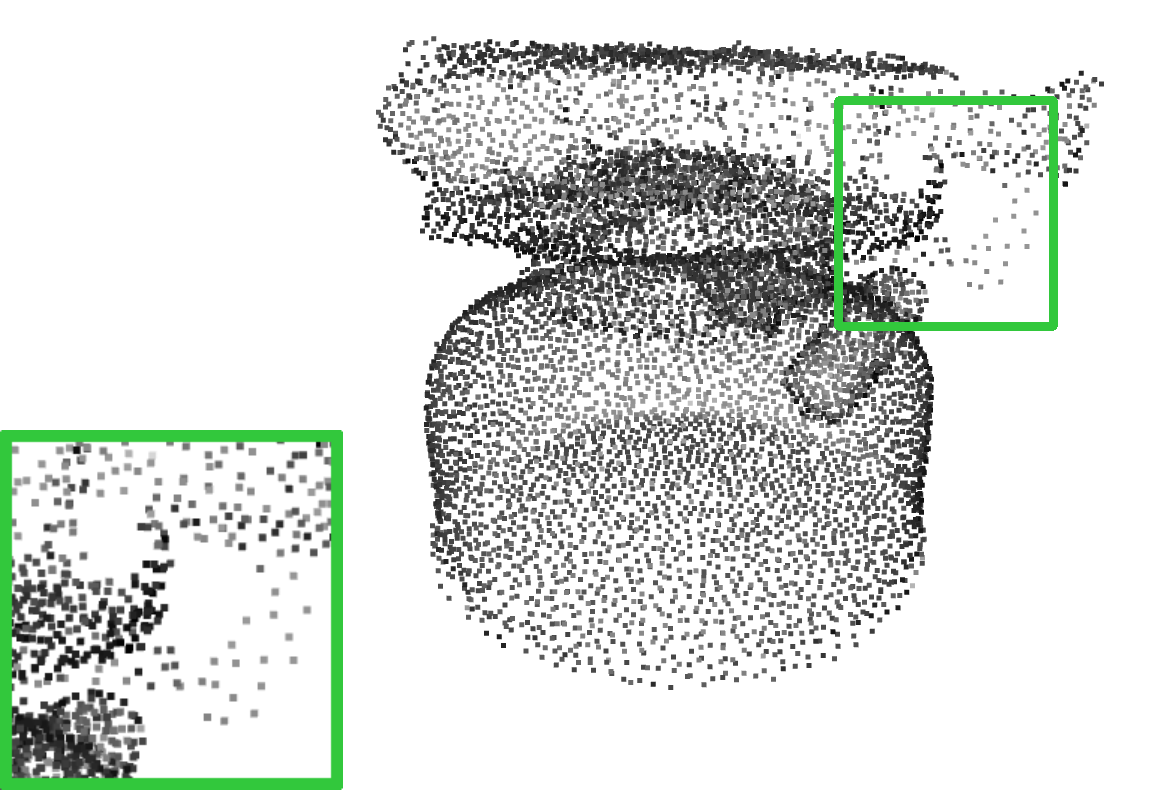} &
    \includegraphics[width=0.15\linewidth]{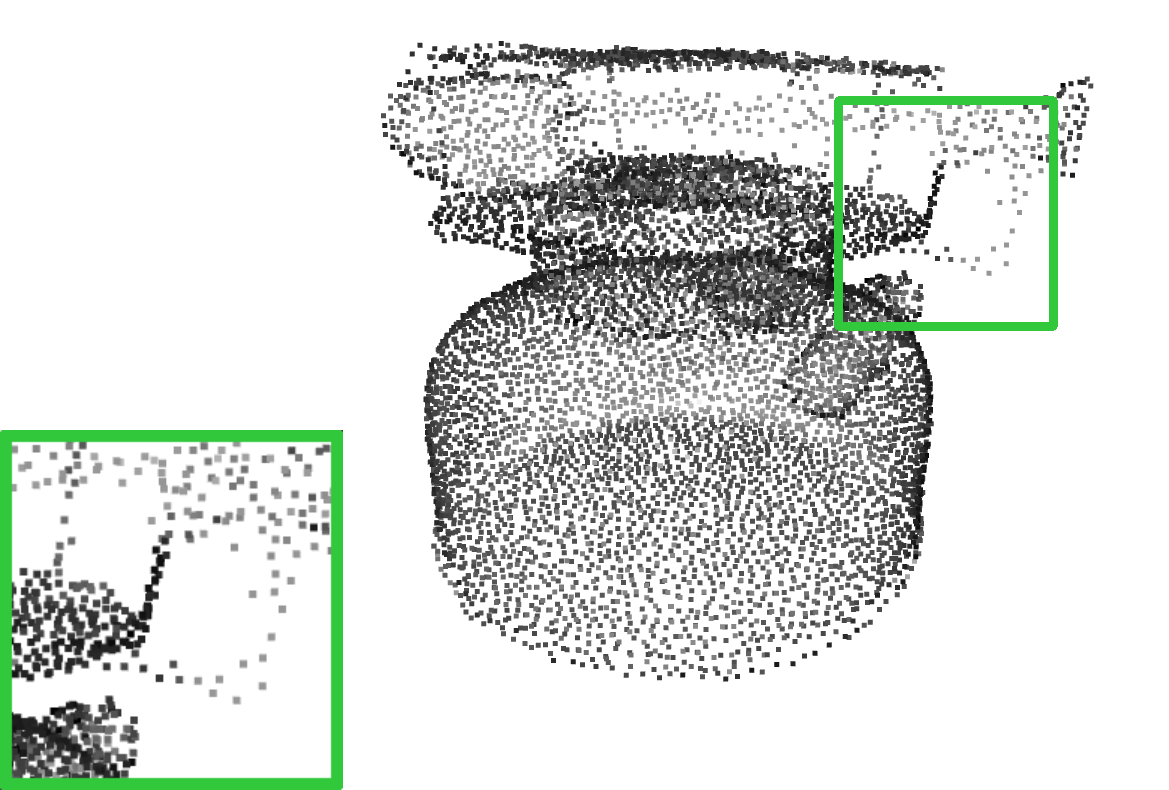} &
    \includegraphics[width=0.15\linewidth]{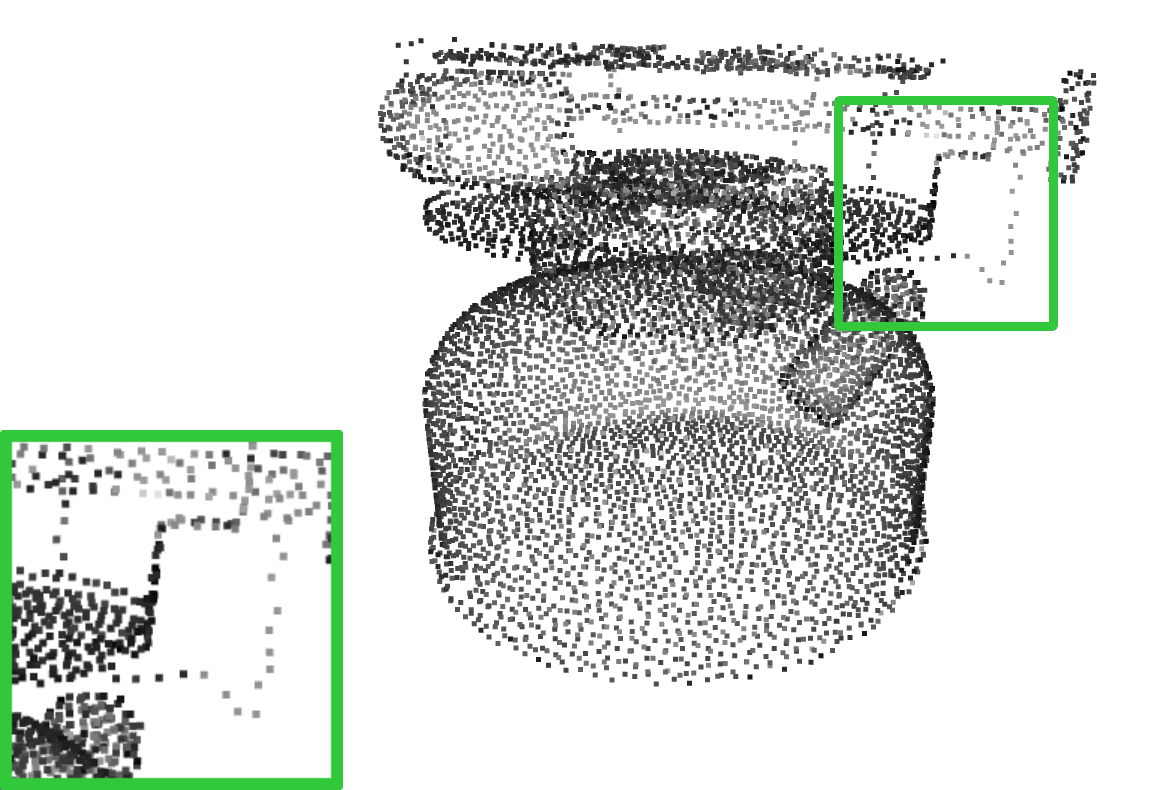} \\
    \includegraphics[width=0.15\linewidth]{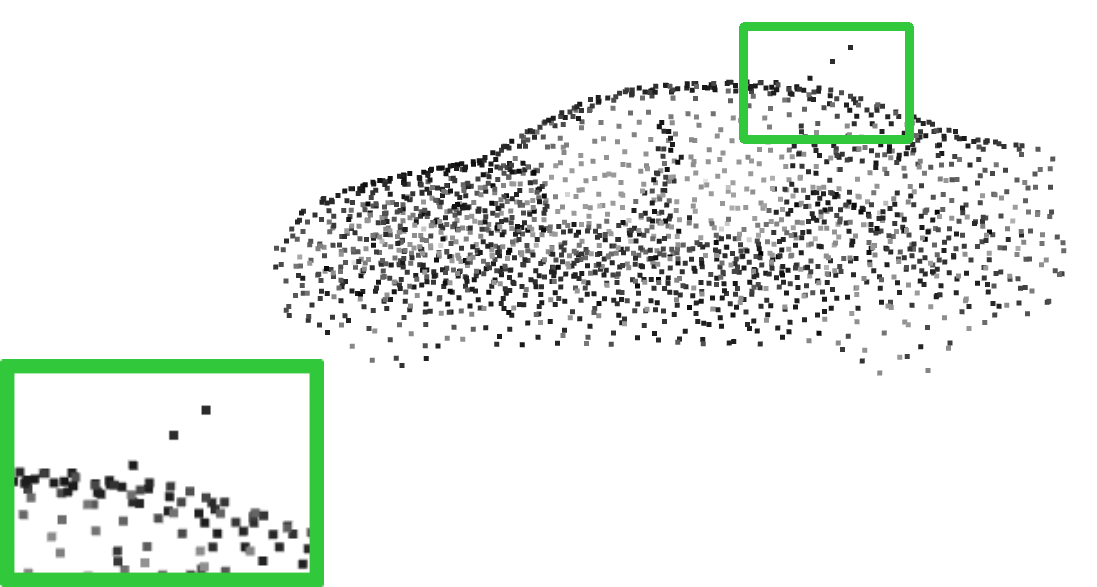} &
    \includegraphics[width=0.15\linewidth]{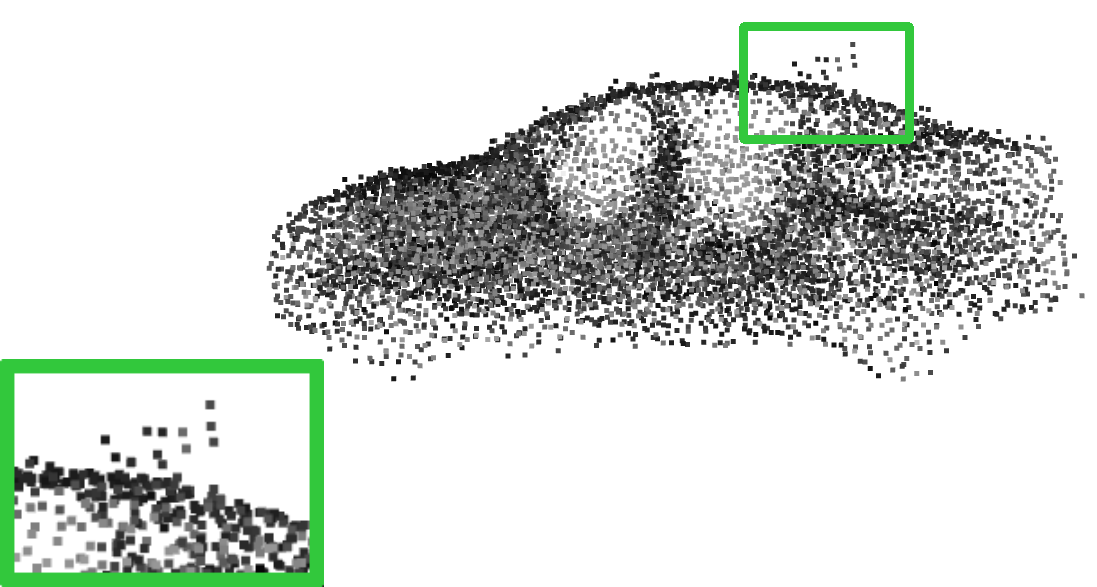} &
    \includegraphics[width=0.15\linewidth]{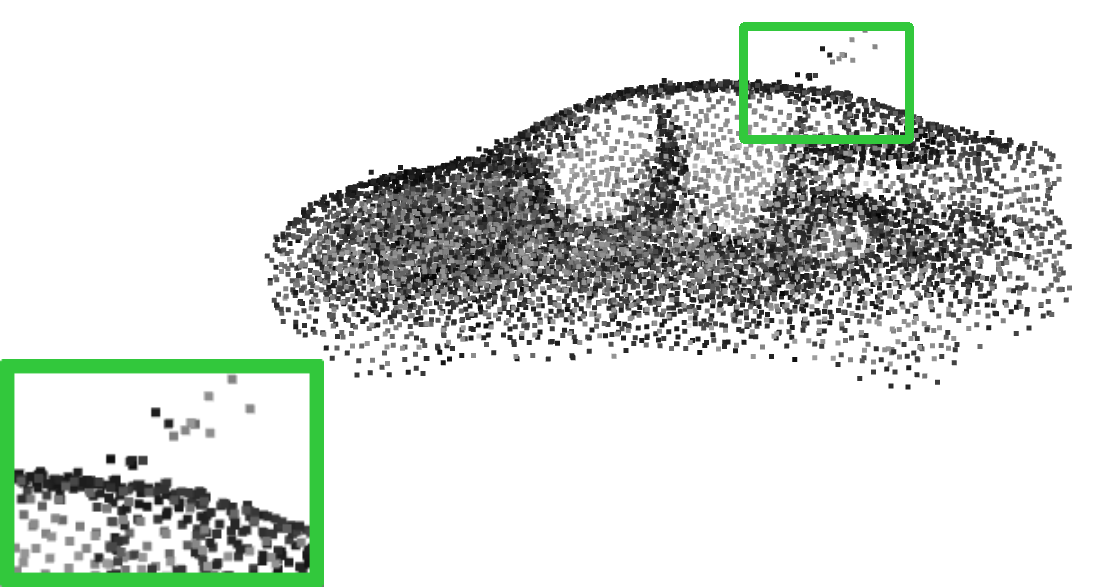} &
    \includegraphics[width=0.15\linewidth]{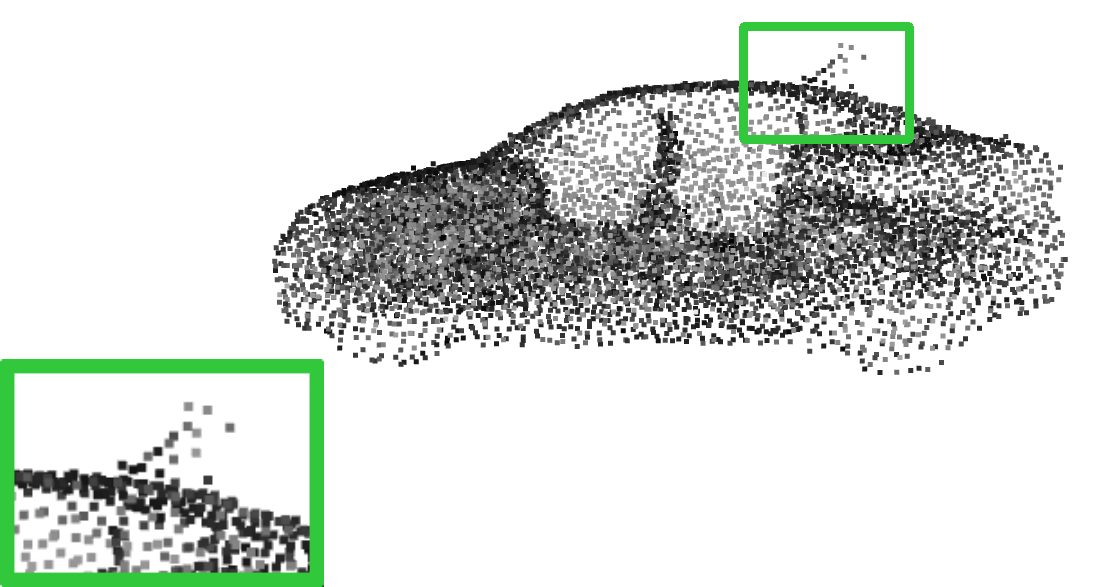} &
    \includegraphics[width=0.15\linewidth]{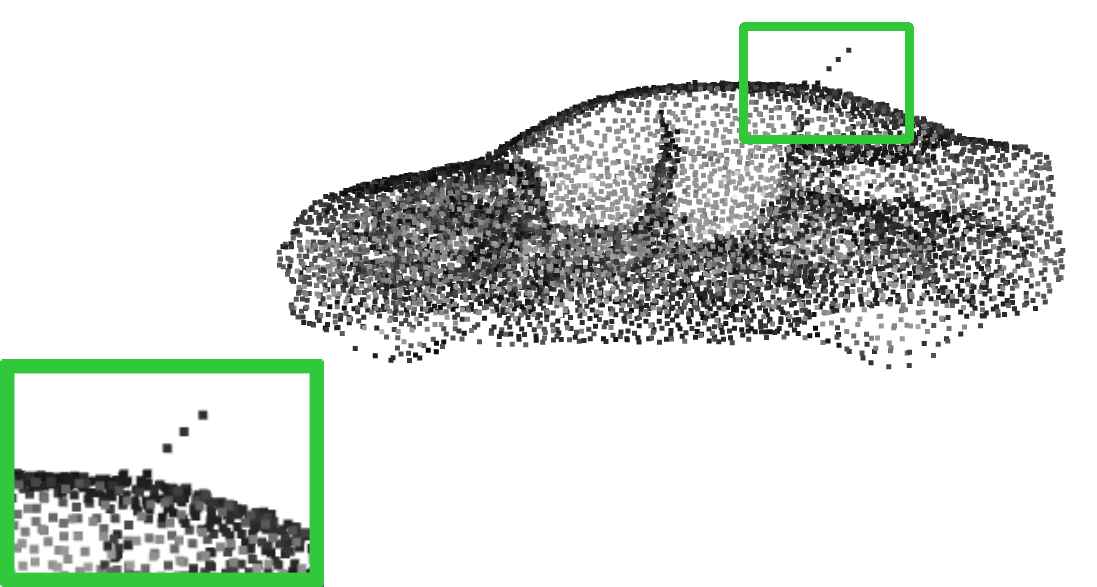} &
    \includegraphics[width=0.15\linewidth]{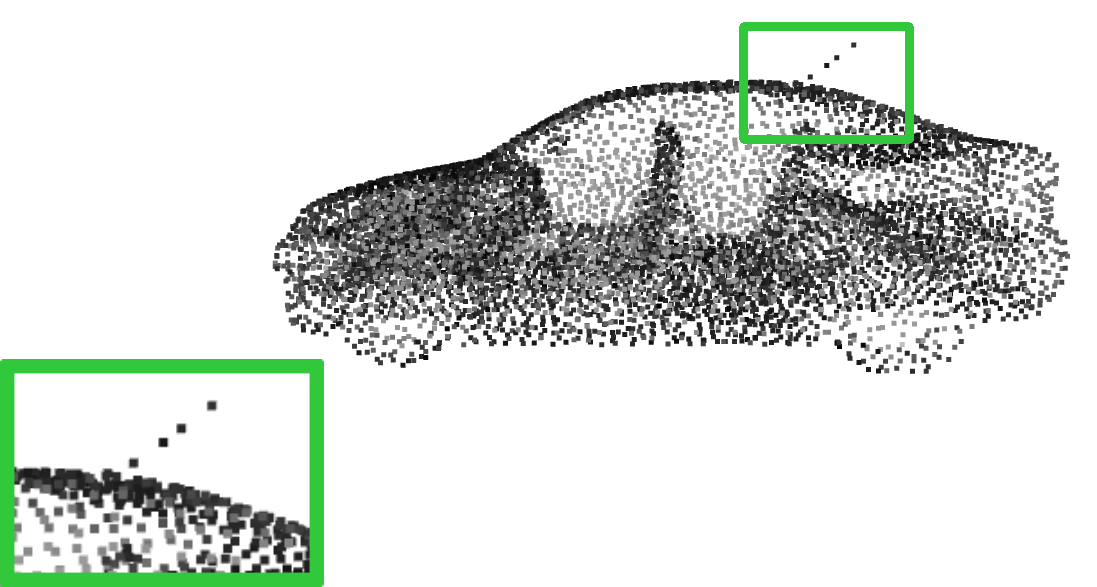} \\
    \includegraphics[width=0.15\linewidth]{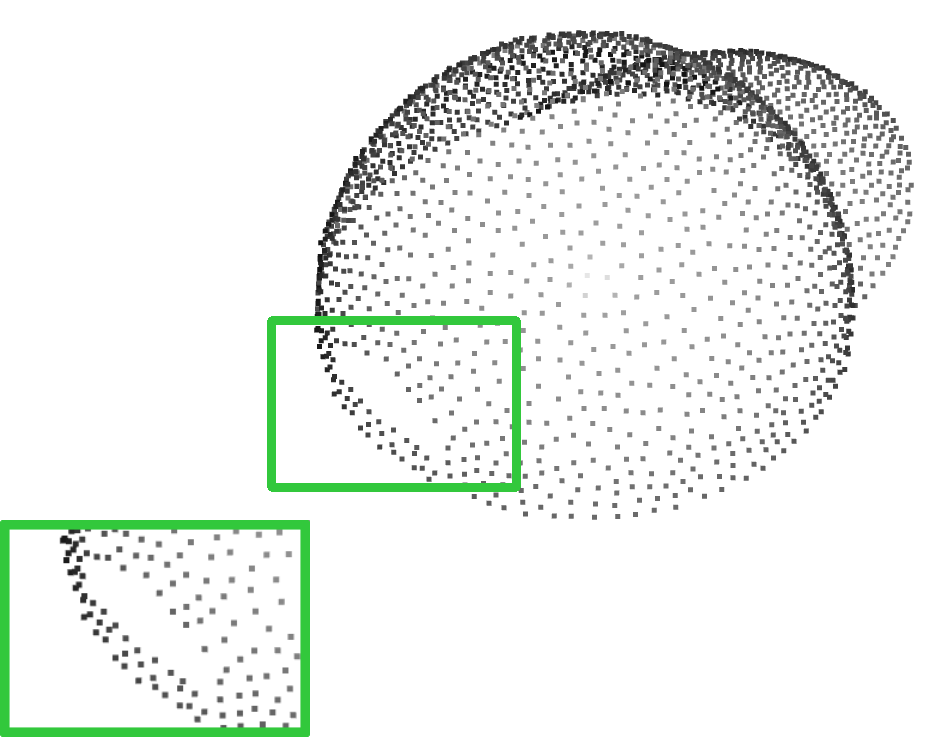} &
    \includegraphics[width=0.15\linewidth]{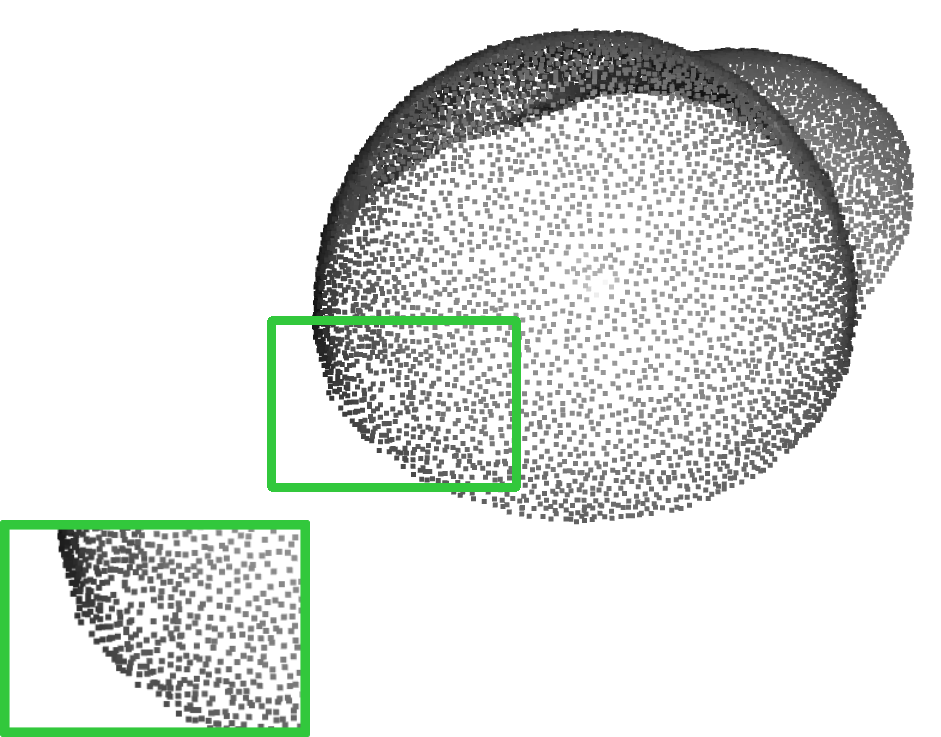} &
    \includegraphics[width=0.15\linewidth]{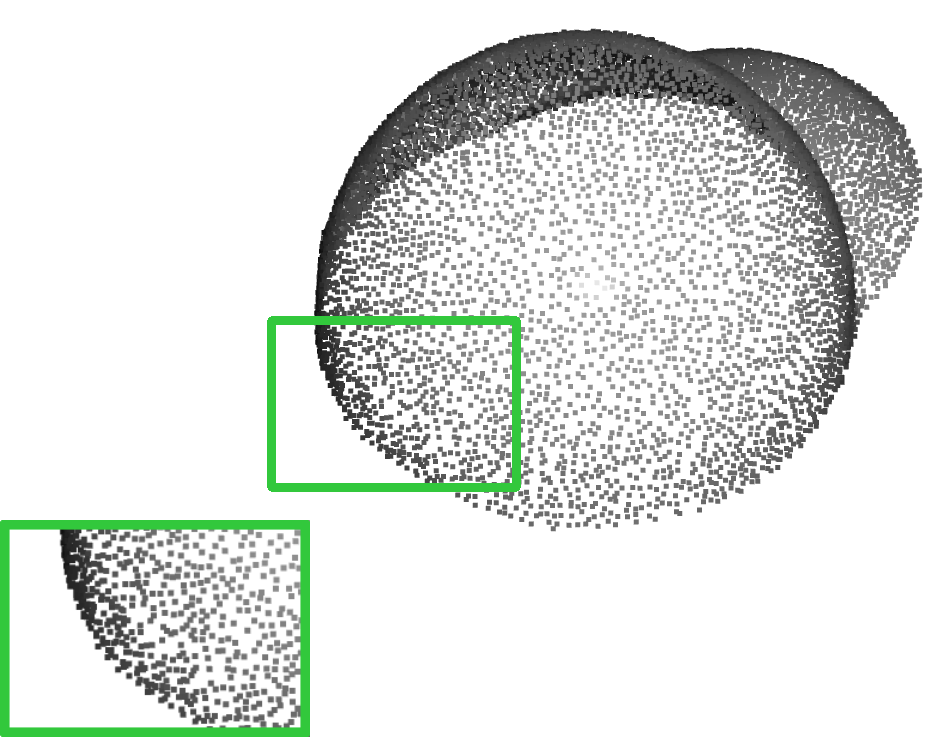} &
    \includegraphics[width=0.15\linewidth]{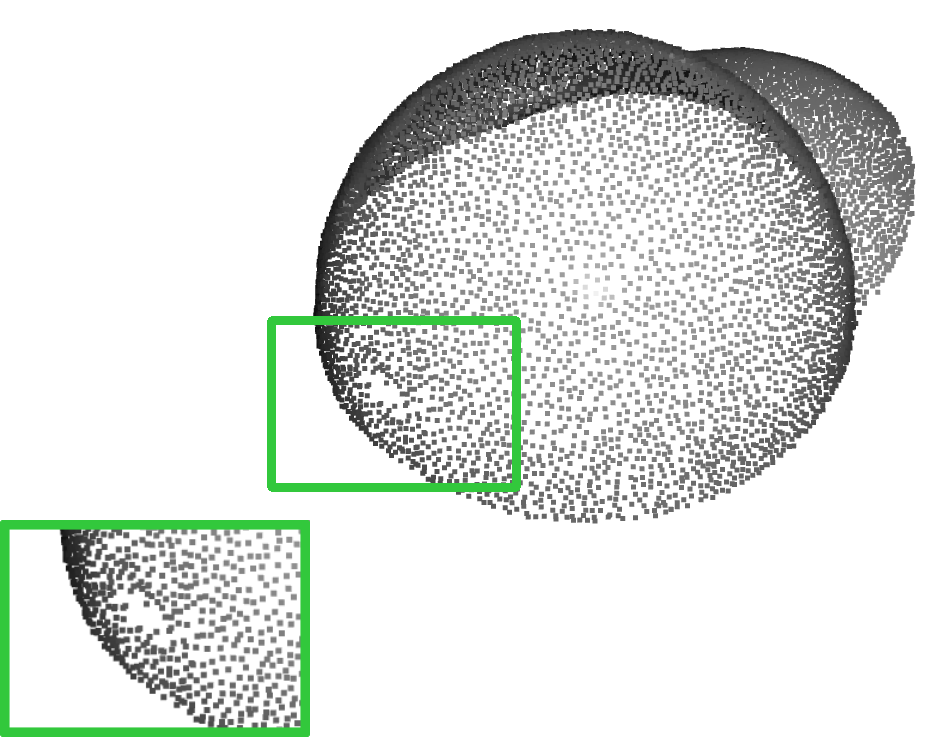} &
    \includegraphics[width=0.15\linewidth]{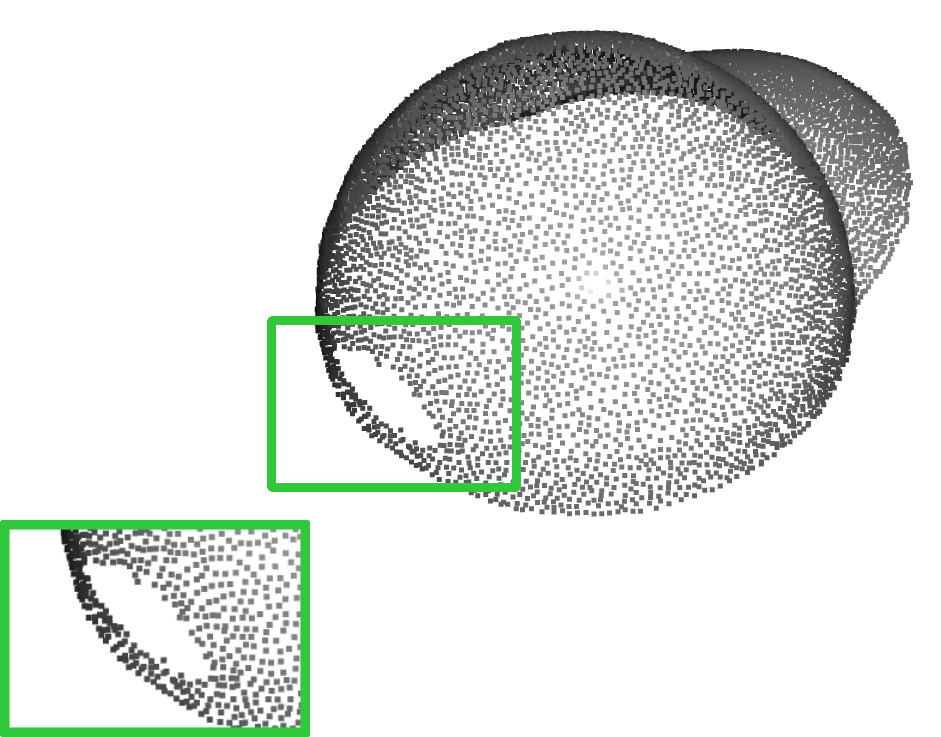} &
    \includegraphics[width=0.15\linewidth]{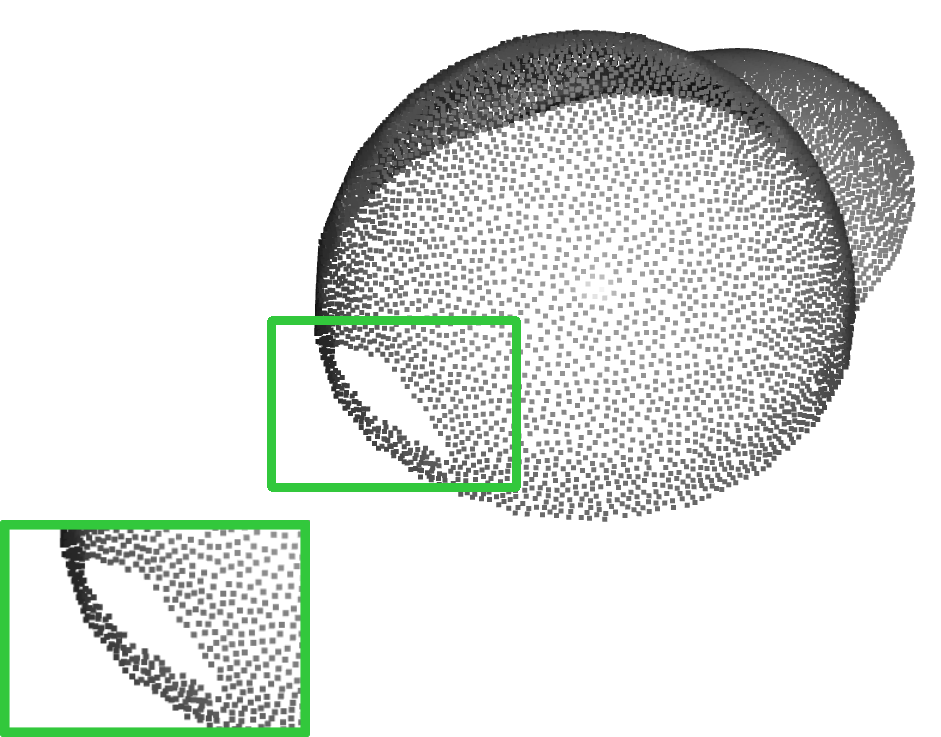} \\
    \includegraphics[width=0.15\linewidth]{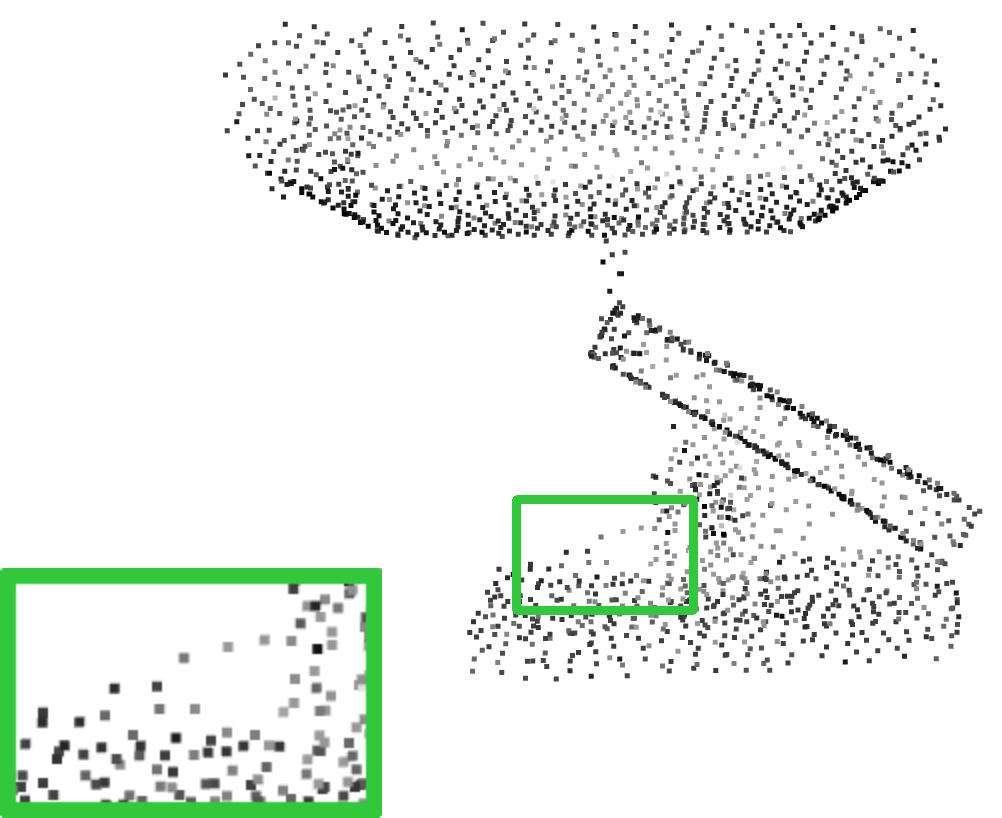} &
    \includegraphics[width=0.15\linewidth]{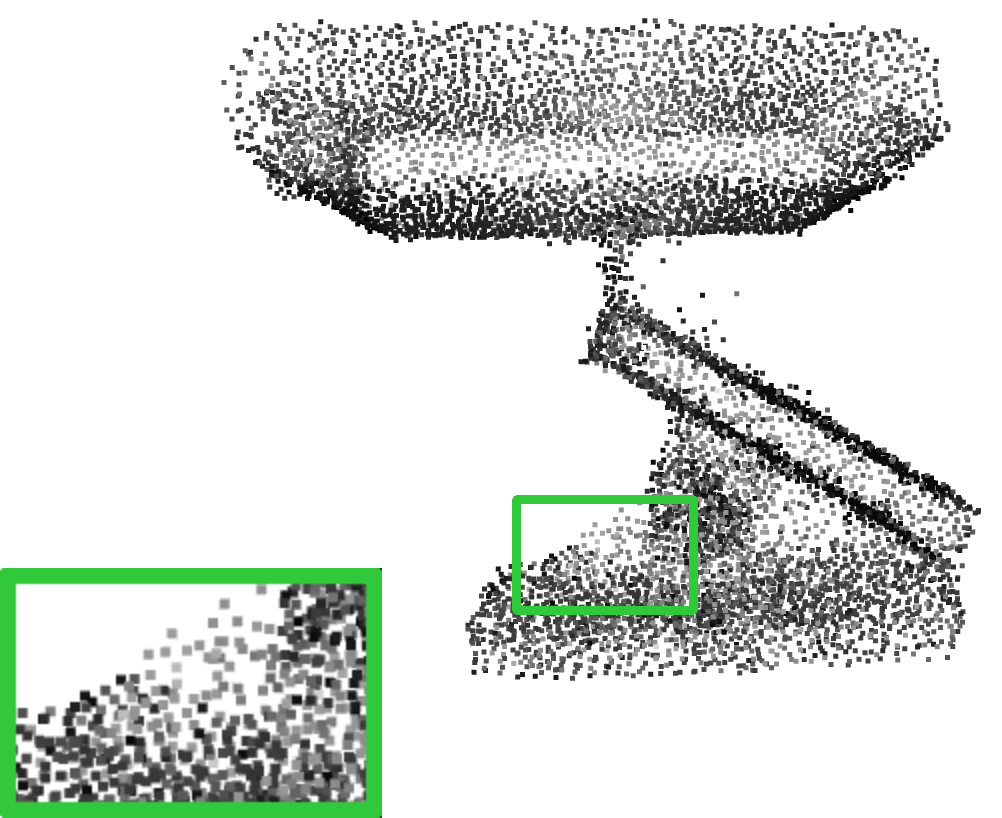} &
    \includegraphics[width=0.15\linewidth]{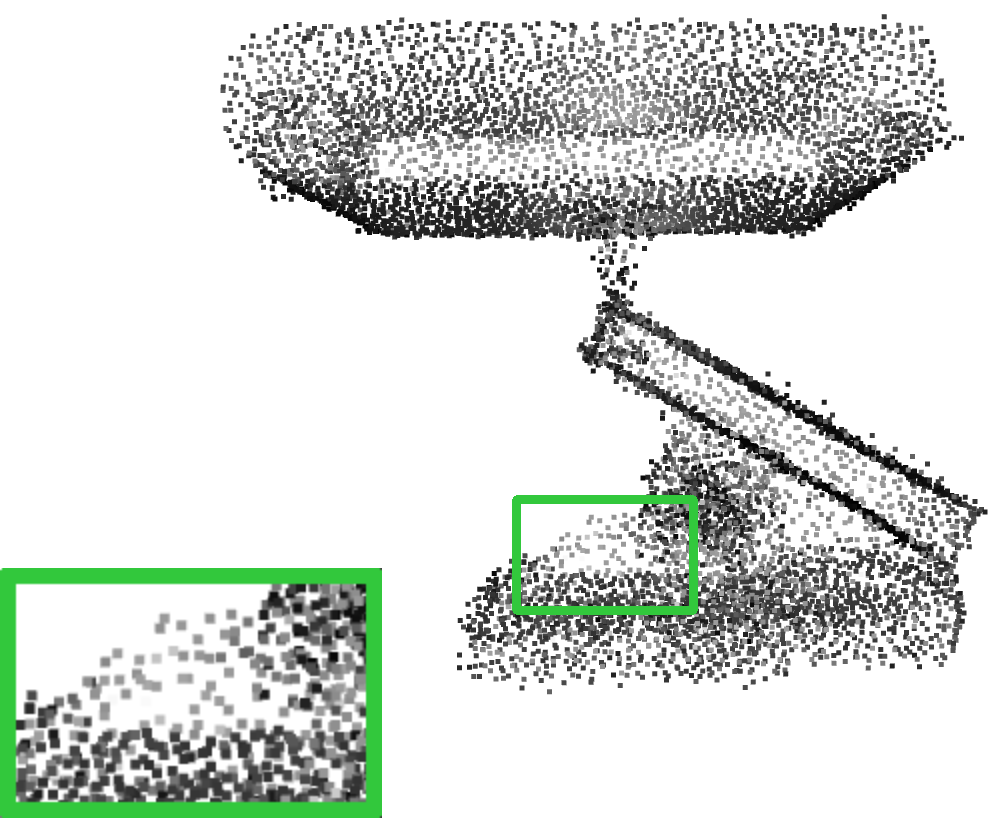} &
    \includegraphics[width=0.15\linewidth]{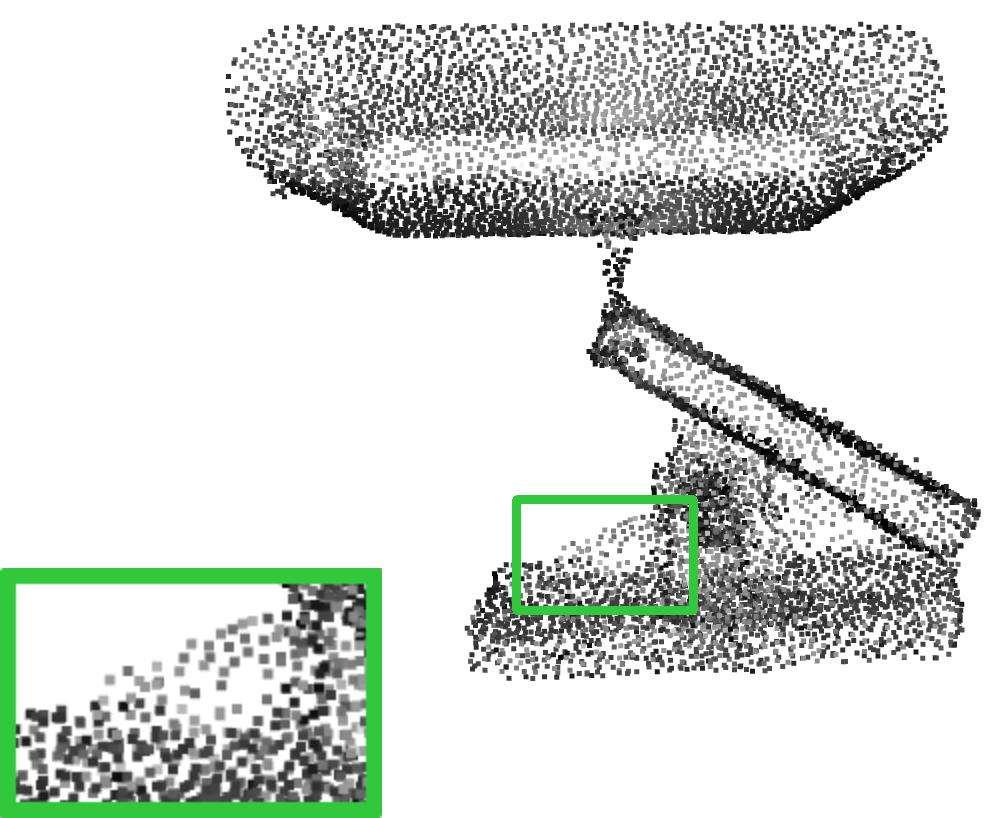} &
    \includegraphics[width=0.15\linewidth]{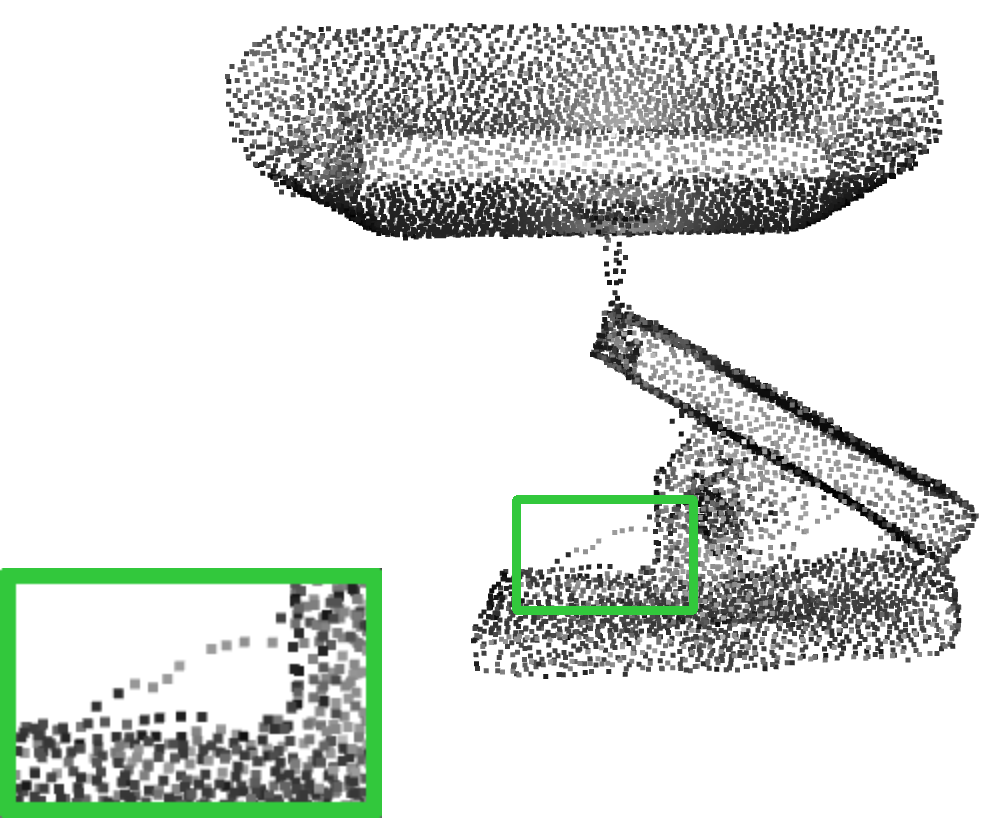} &
    \includegraphics[width=0.15\linewidth]{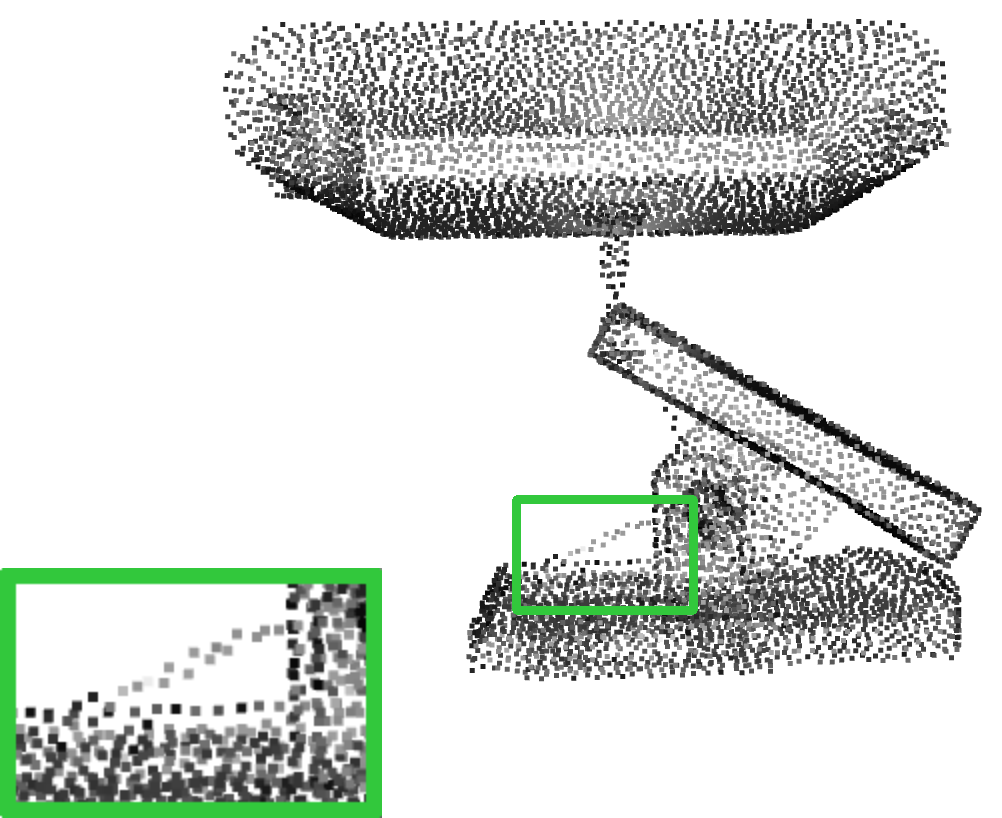} \\
    \includegraphics[width=0.15\linewidth]{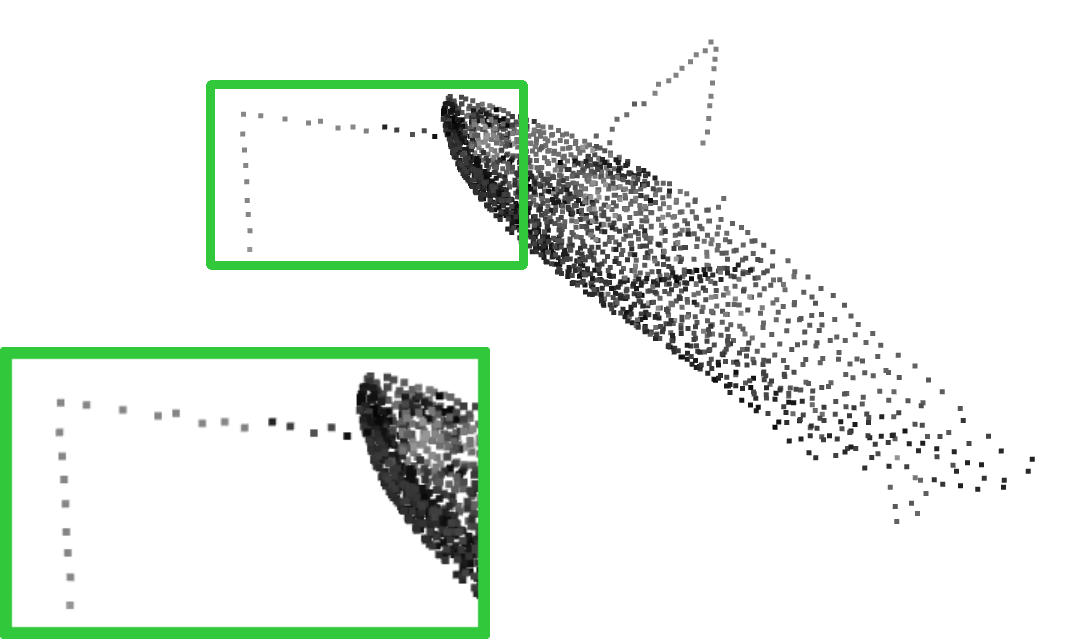} &
    \includegraphics[width=0.15\linewidth]{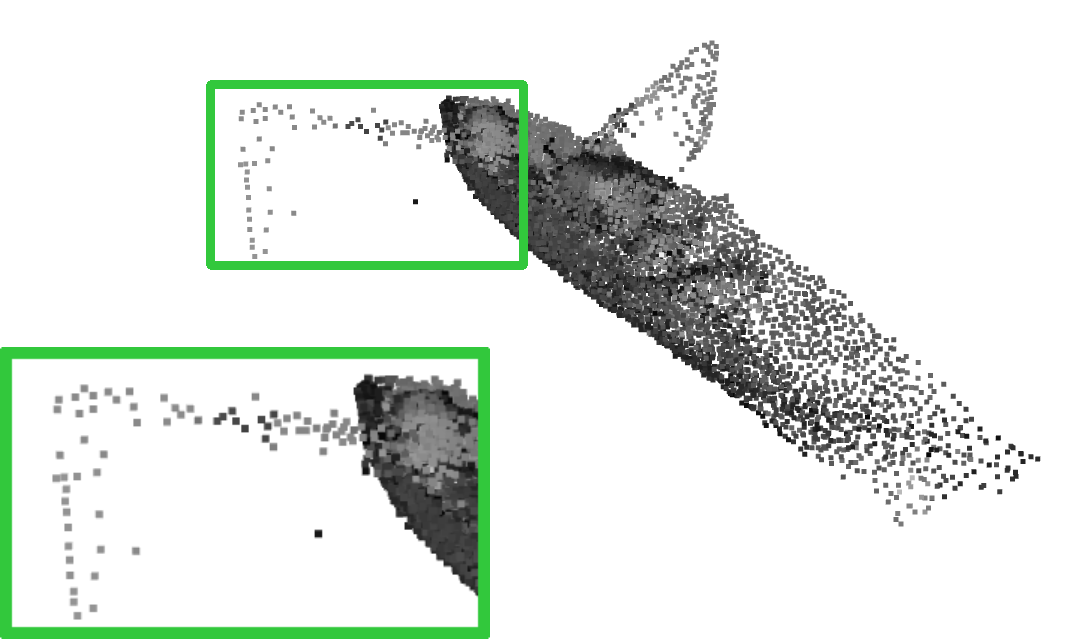} &
    \includegraphics[width=0.15\linewidth]{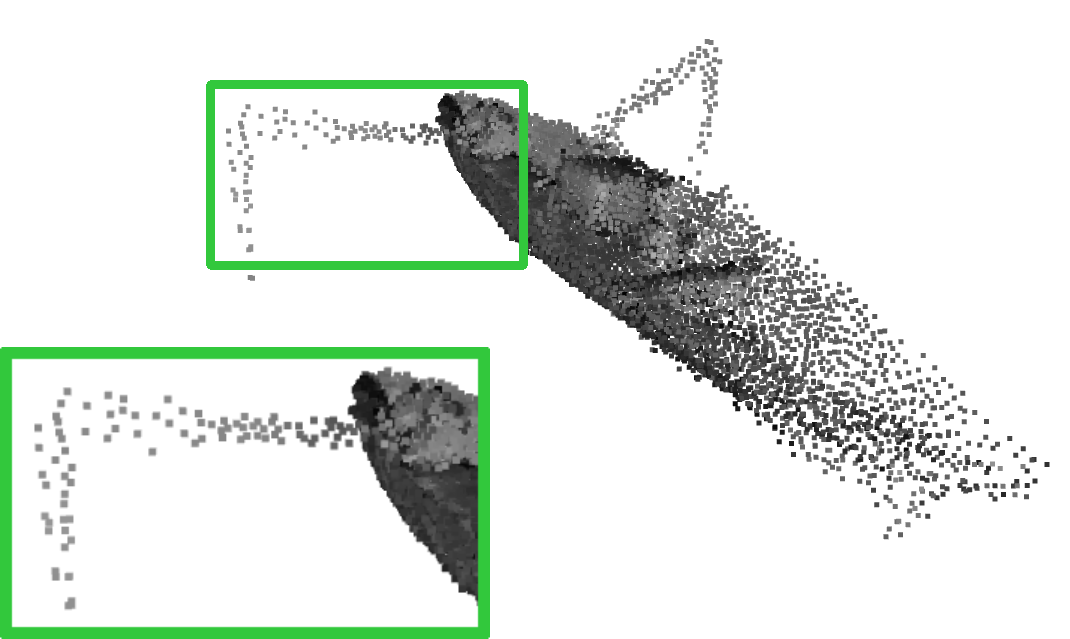} &
    \includegraphics[width=0.15\linewidth]{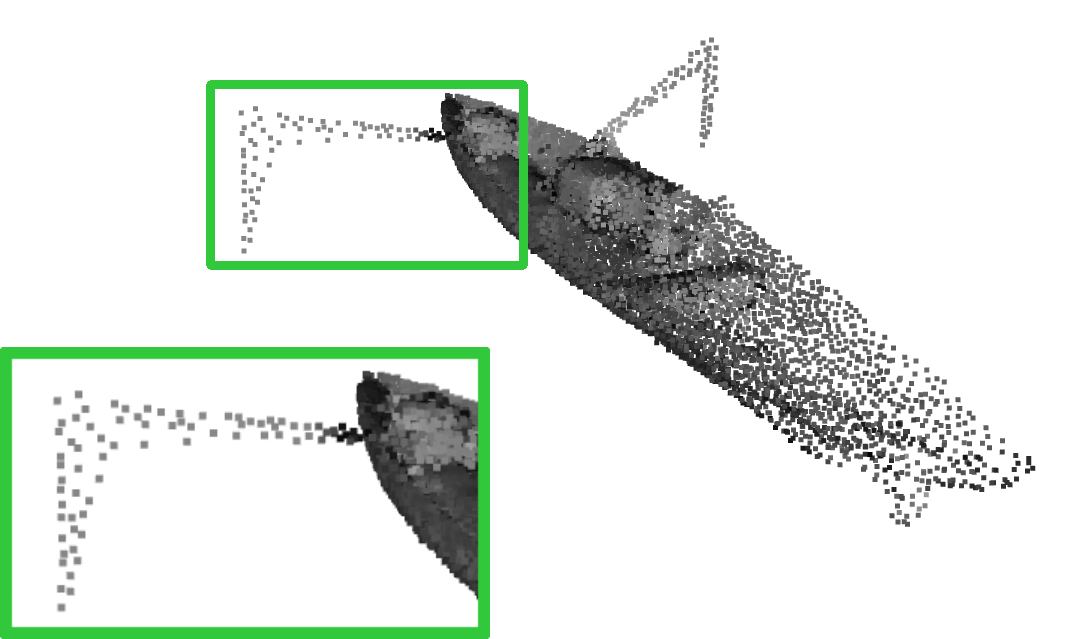} &
    \includegraphics[width=0.15\linewidth]{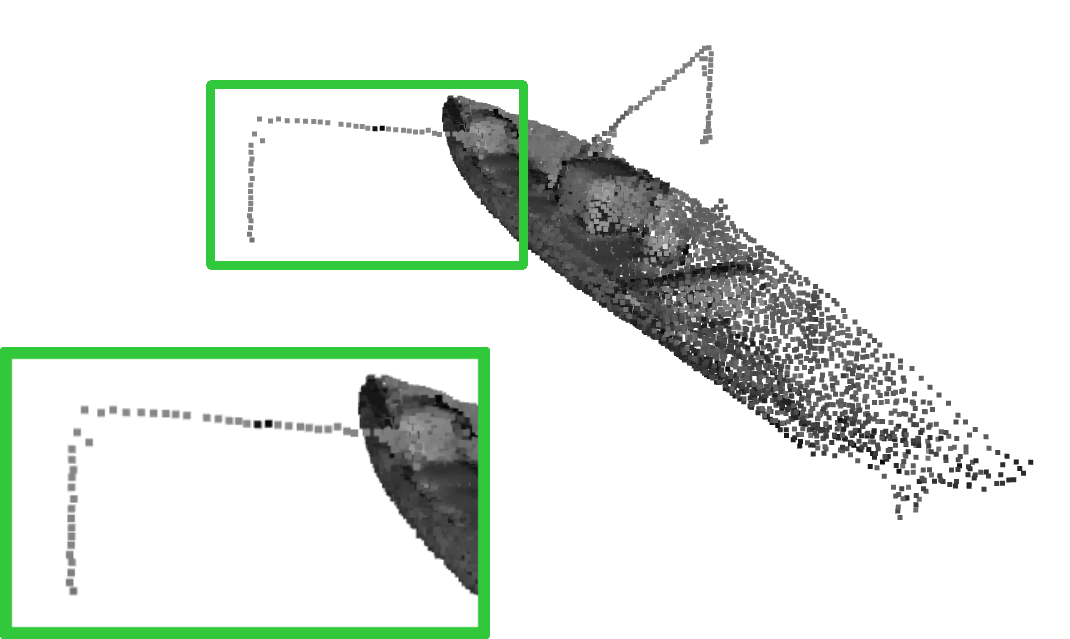} &
    \includegraphics[width=0.15\linewidth]{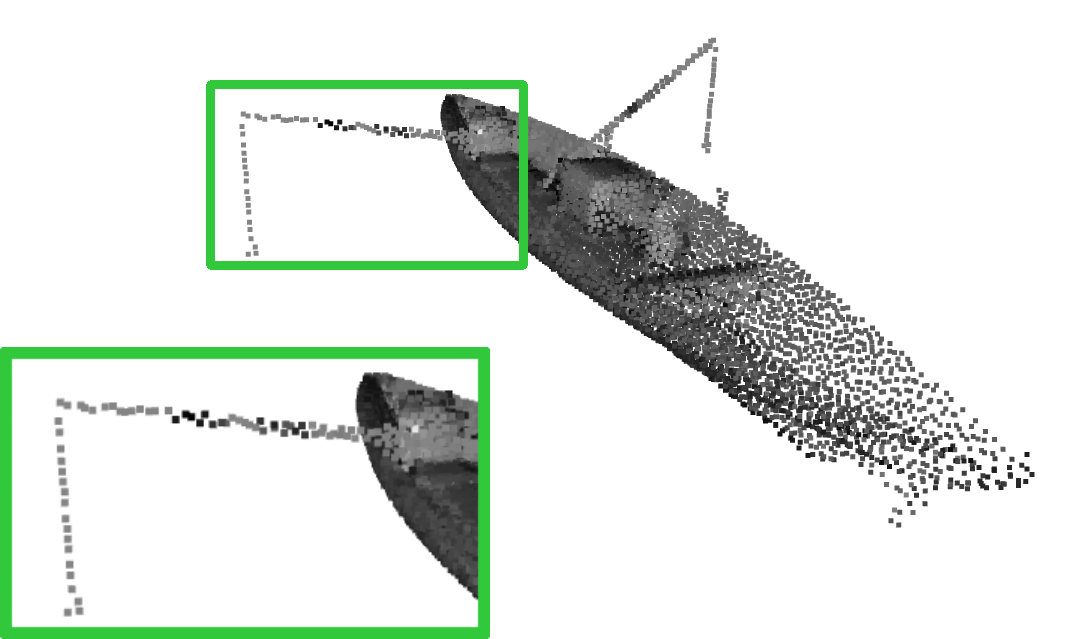} \\
    \includegraphics[width=0.15\linewidth]{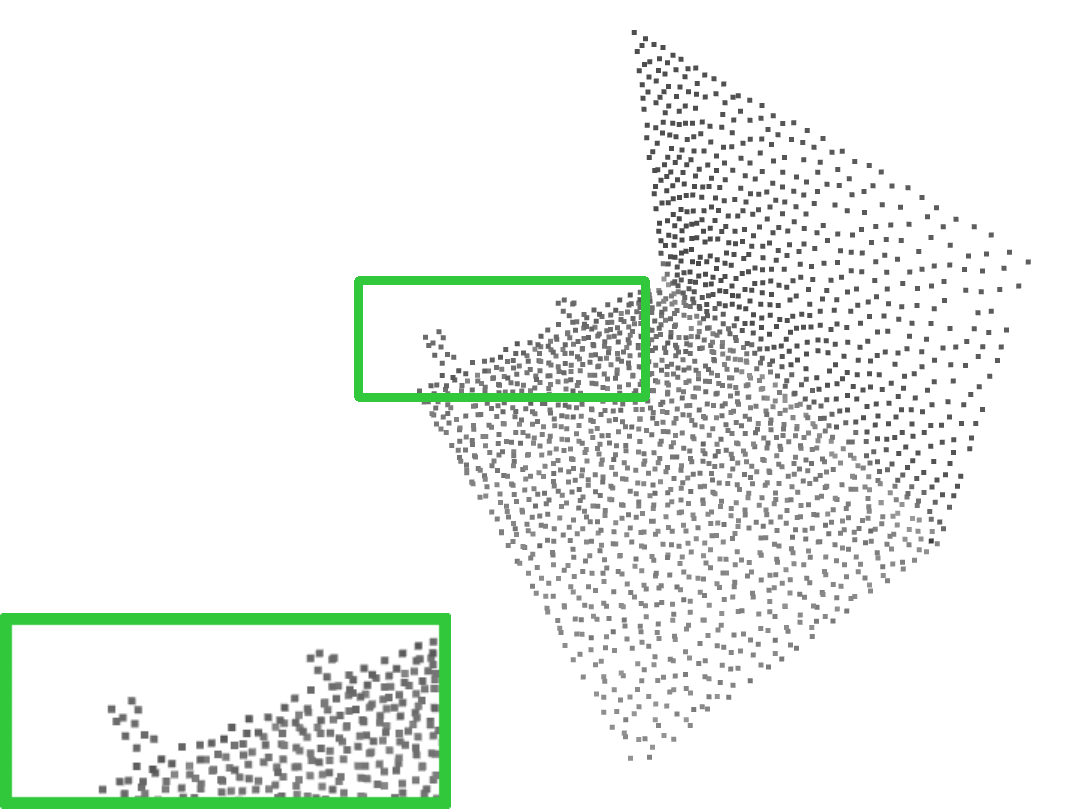} &
    \includegraphics[width=0.15\linewidth]{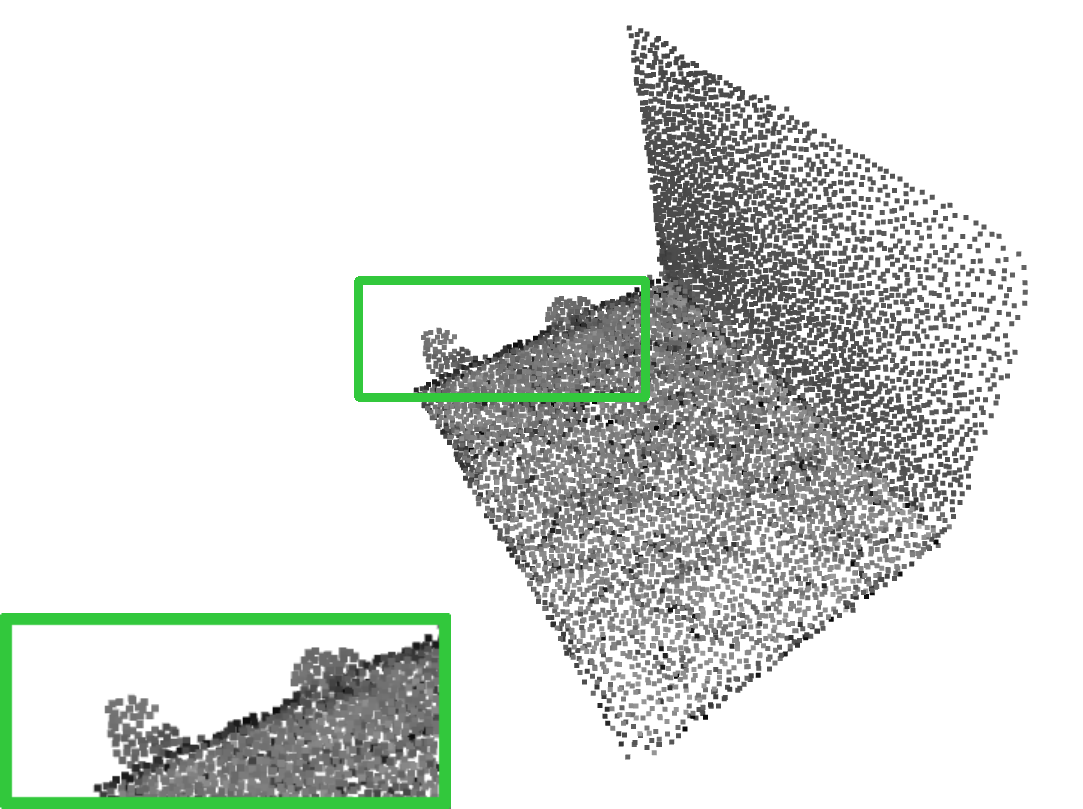} &
    \includegraphics[width=0.15\linewidth]{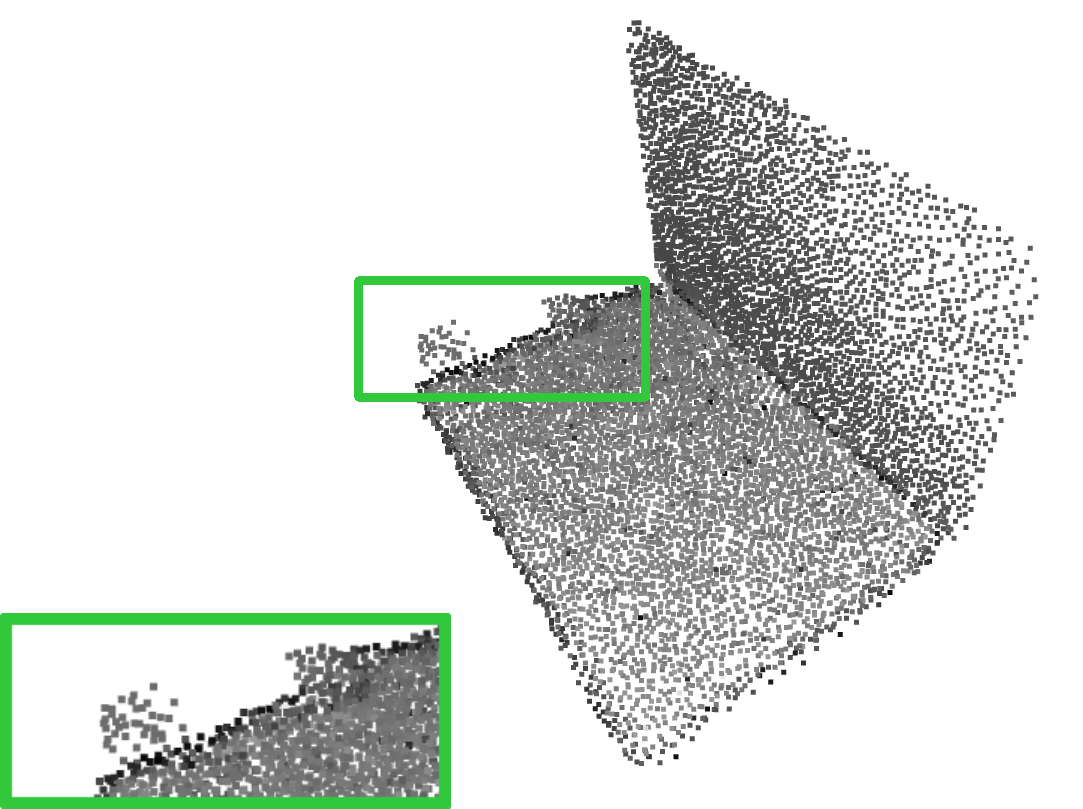} &
    \includegraphics[width=0.15\linewidth]{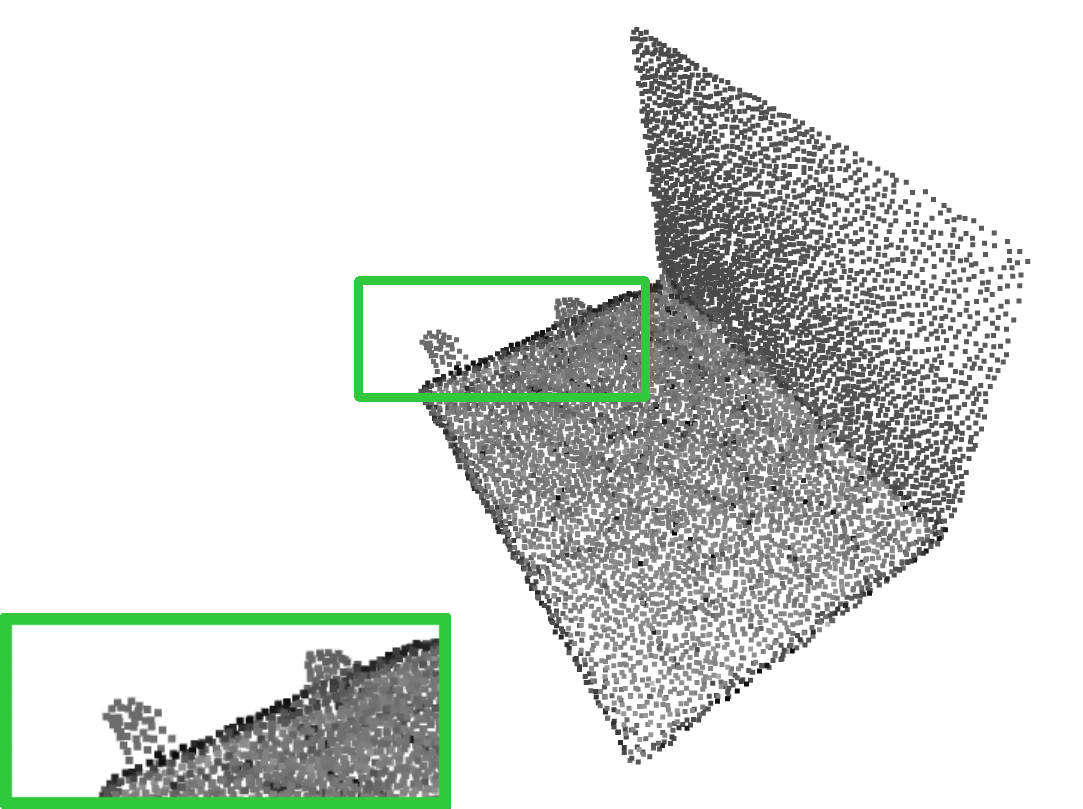} &
    \includegraphics[width=0.15\linewidth]{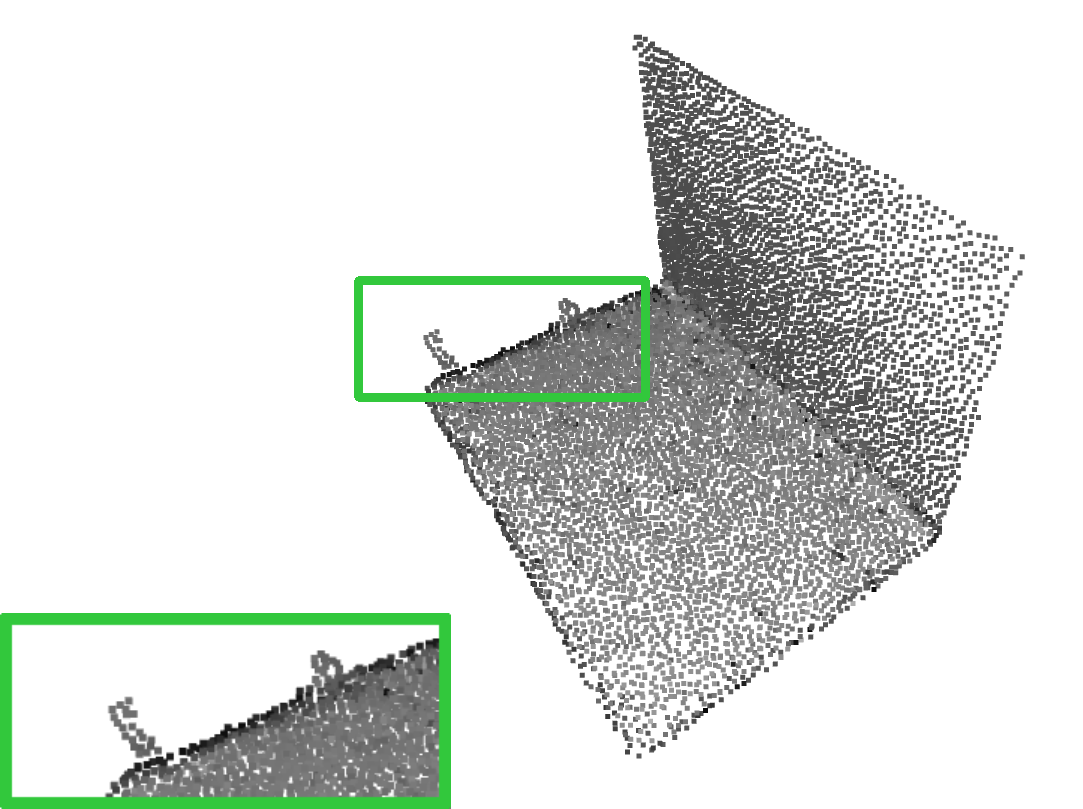} &
    \includegraphics[width=0.15\linewidth]{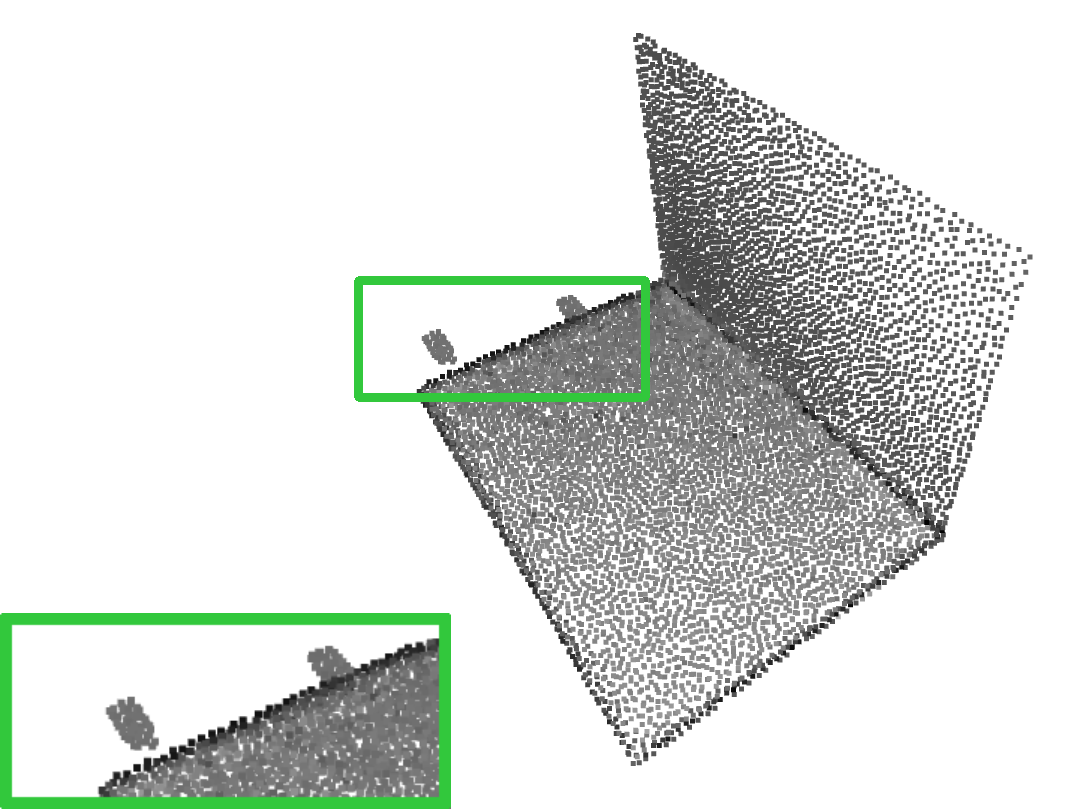} \\
    \includegraphics[width=0.15\linewidth]{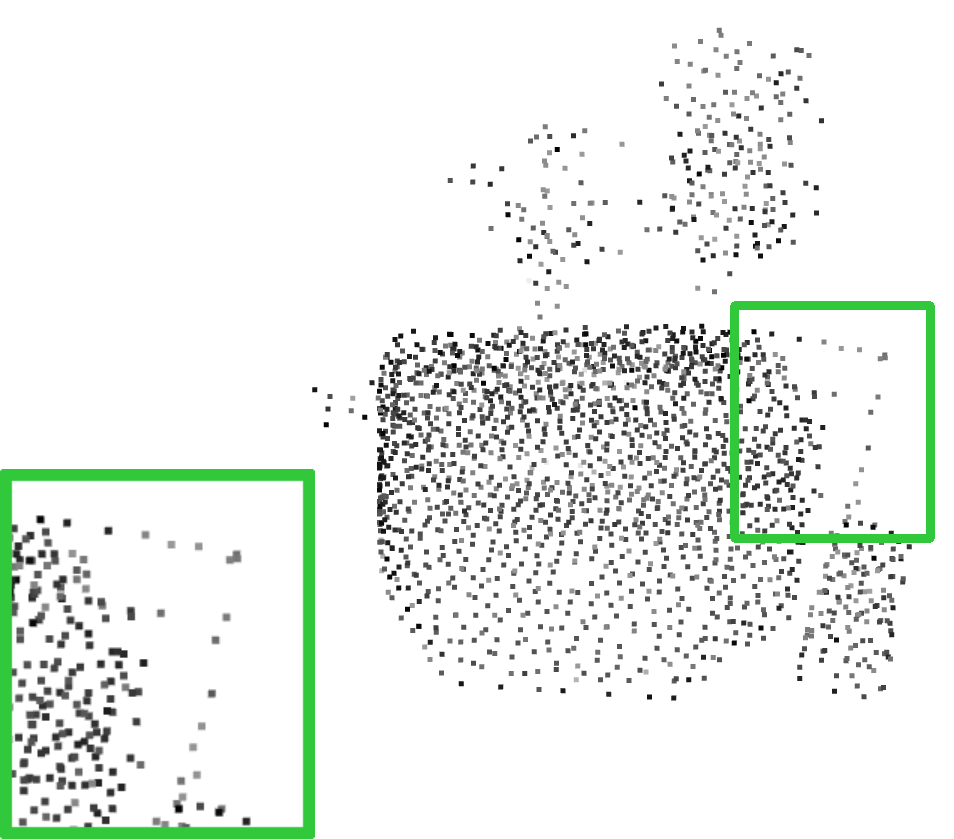} &
    \includegraphics[width=0.15\linewidth]{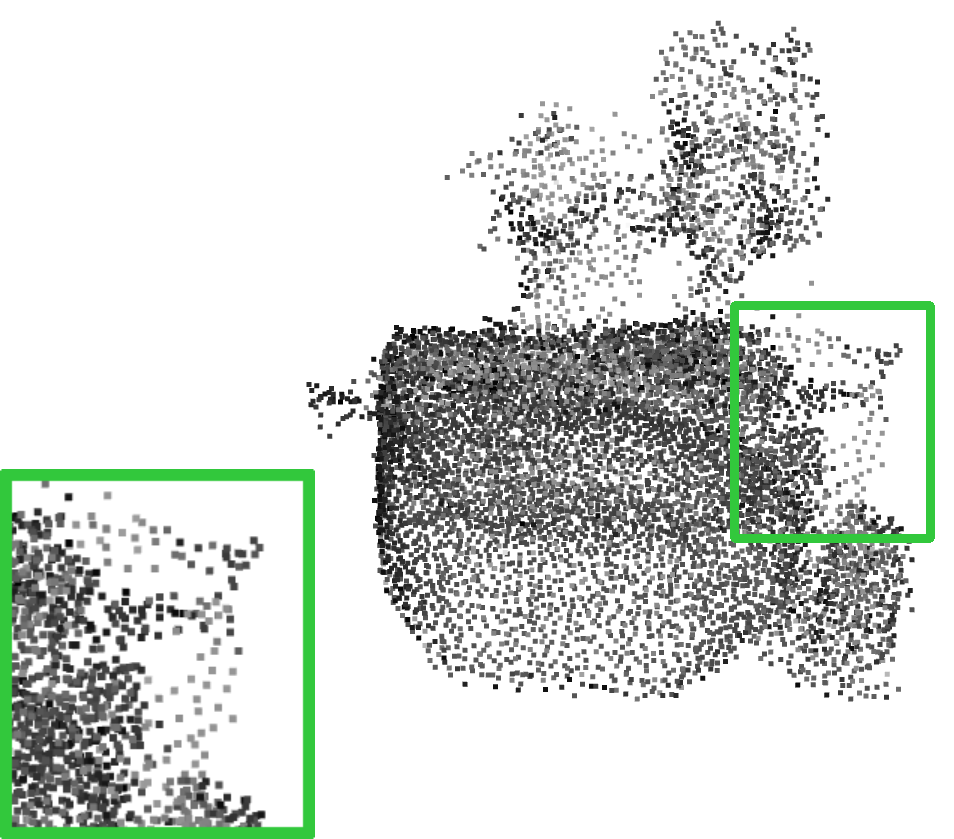} &
    \includegraphics[width=0.15\linewidth]{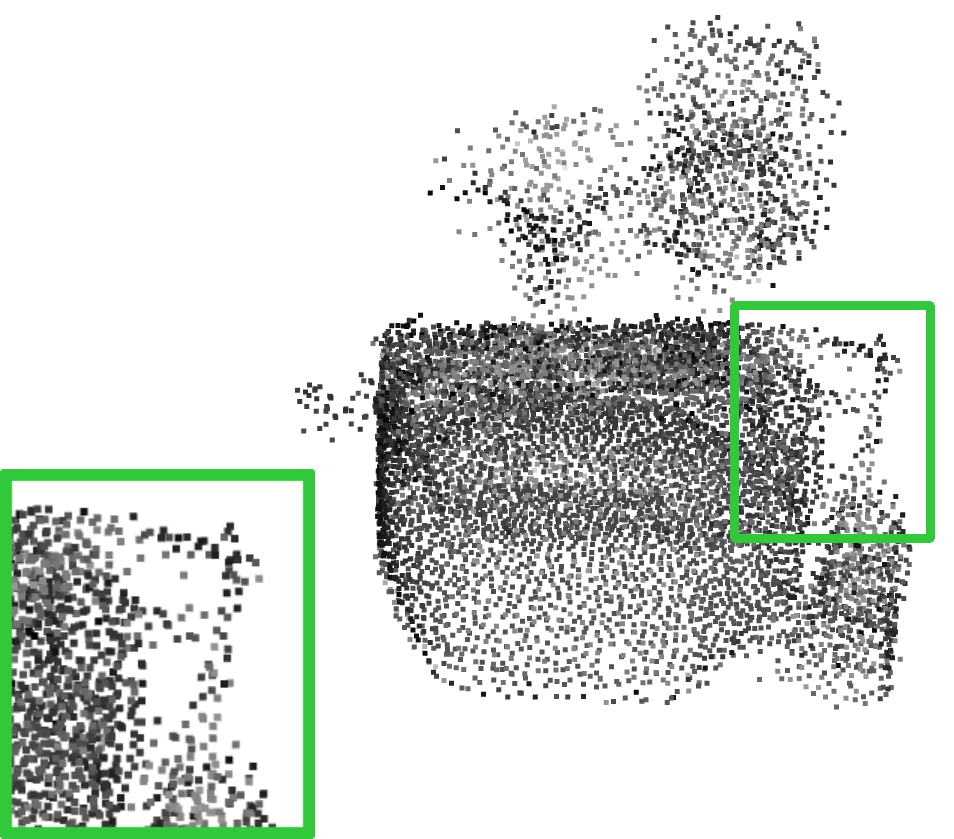} &
    \includegraphics[width=0.15\linewidth]{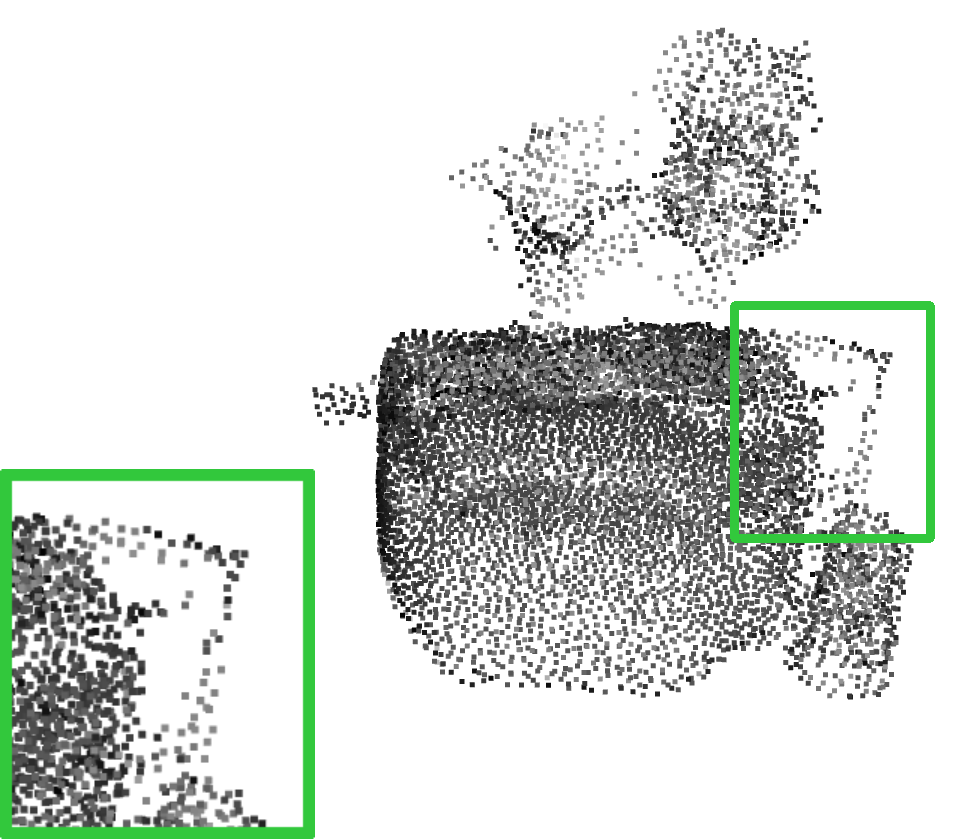} &
    \includegraphics[width=0.15\linewidth]{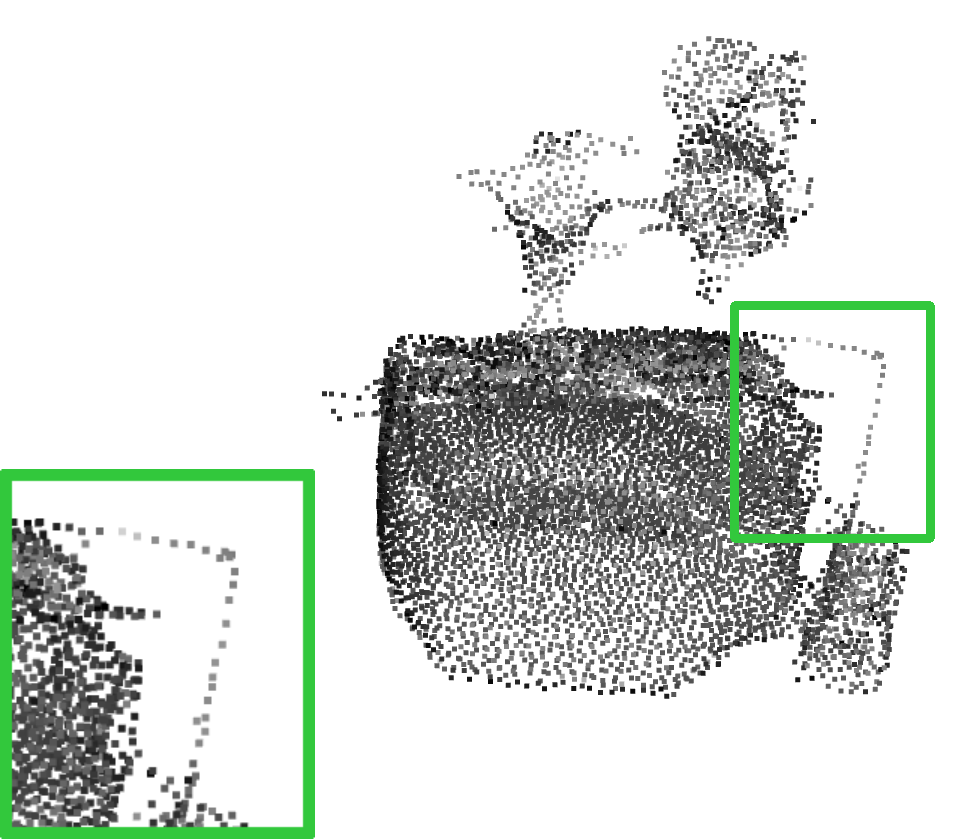} &
    \includegraphics[width=0.15\linewidth]{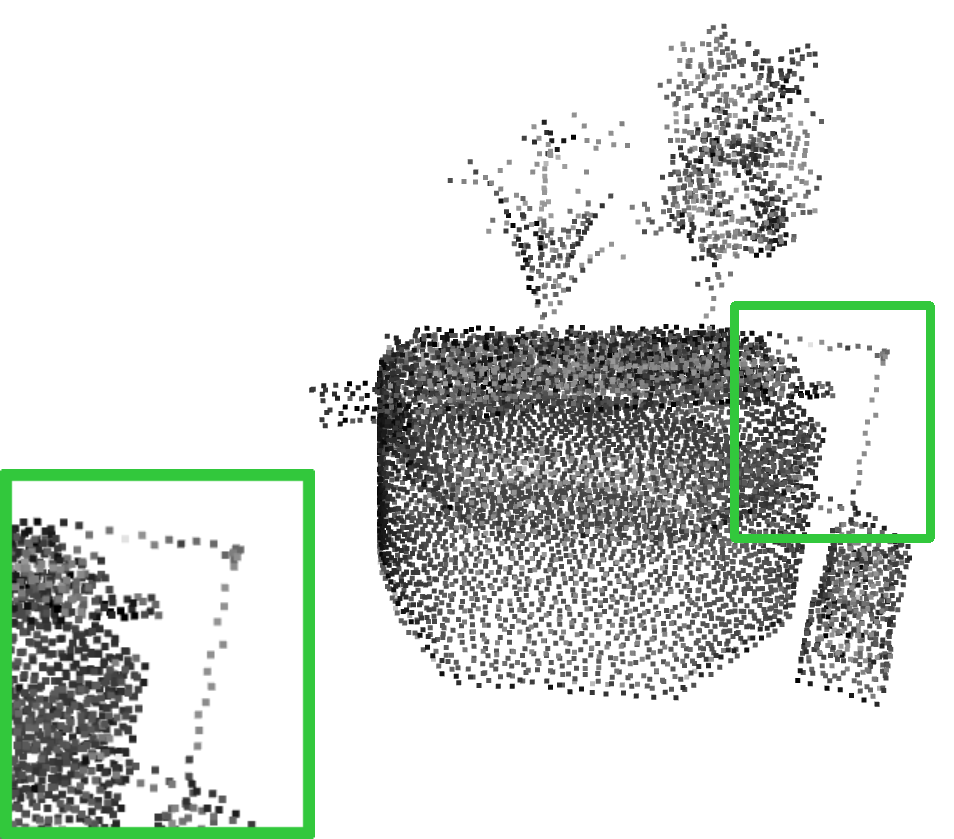} \\
    \includegraphics[width=0.15\linewidth]{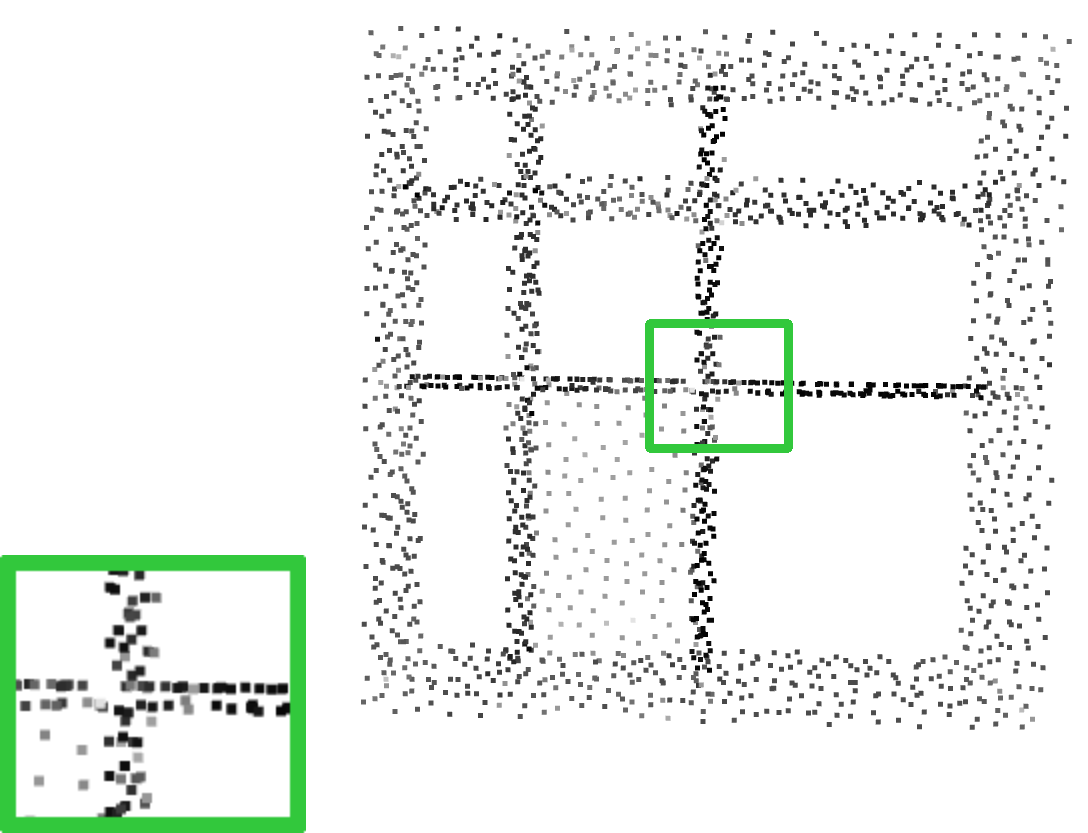} &
    \includegraphics[width=0.15\linewidth]{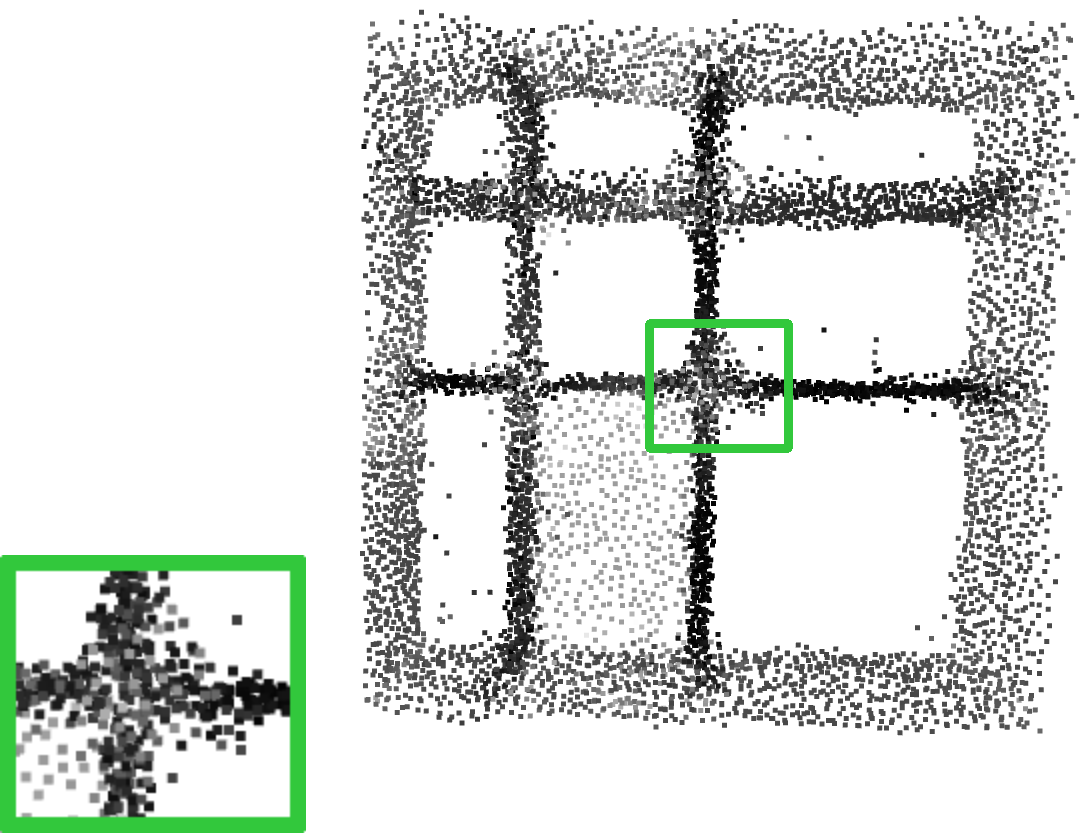} &
    \includegraphics[width=0.15\linewidth]{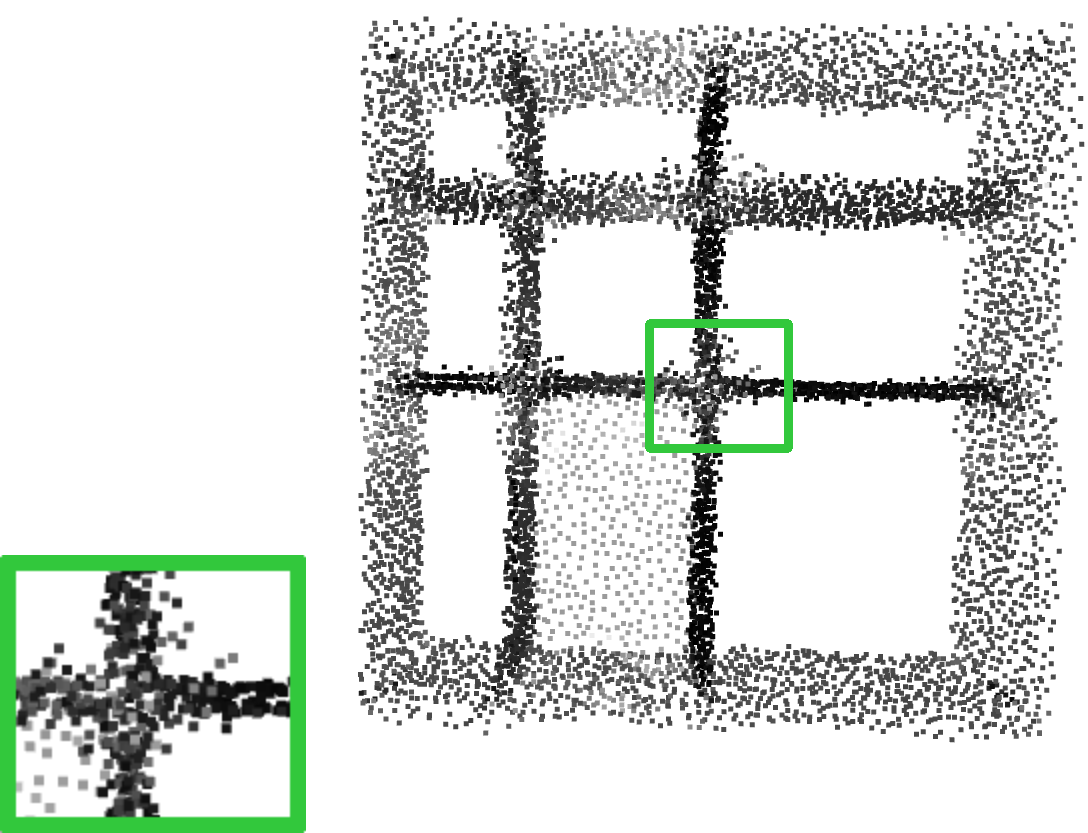} &
    \includegraphics[width=0.15\linewidth]{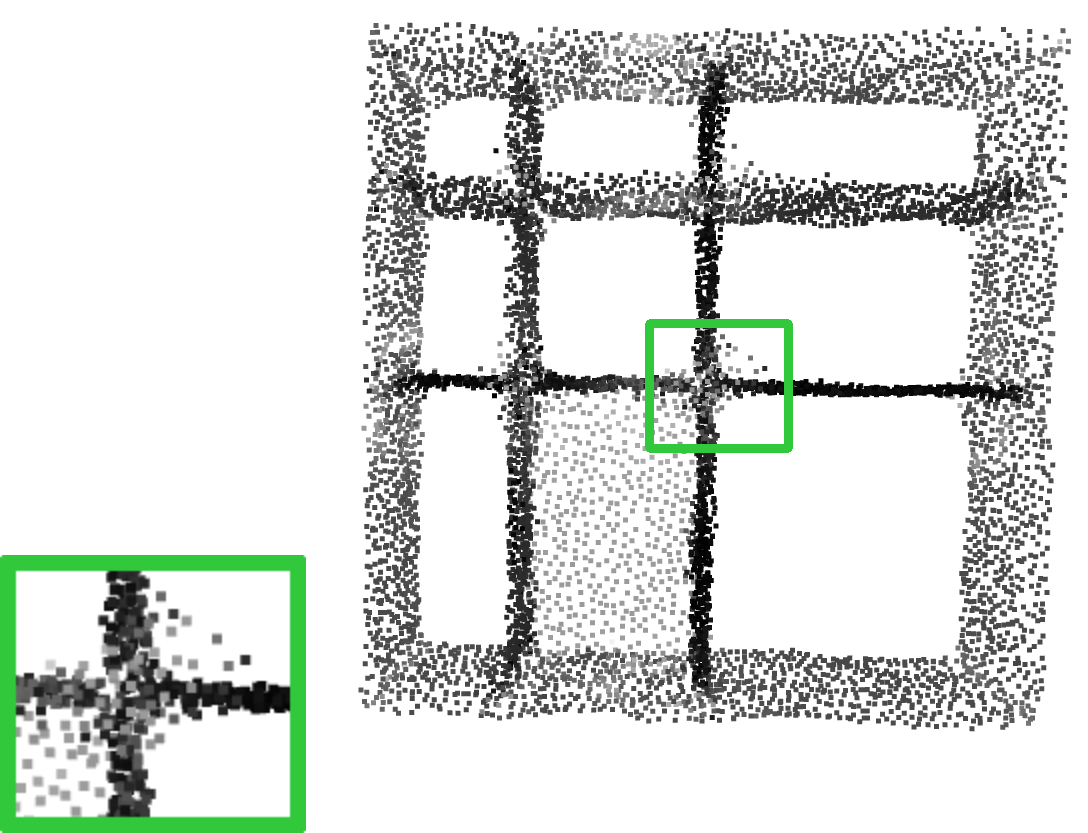} &
    \includegraphics[width=0.15\linewidth]{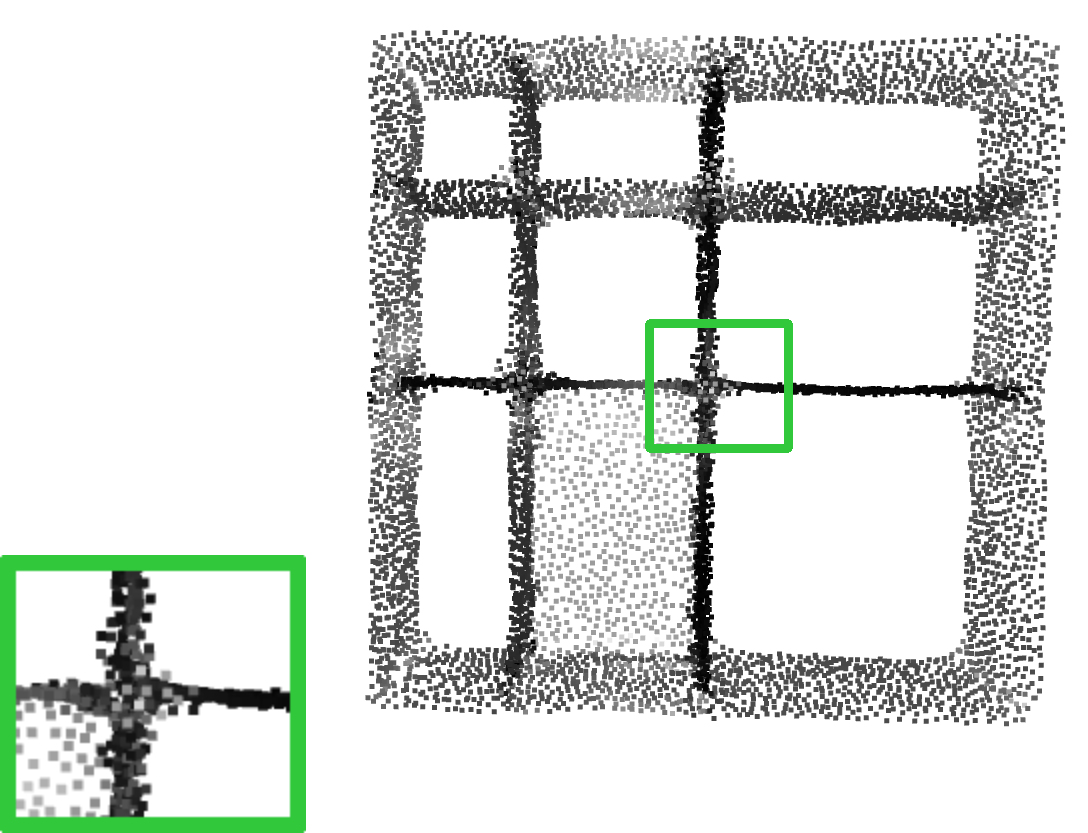} &
    \includegraphics[width=0.15\linewidth]{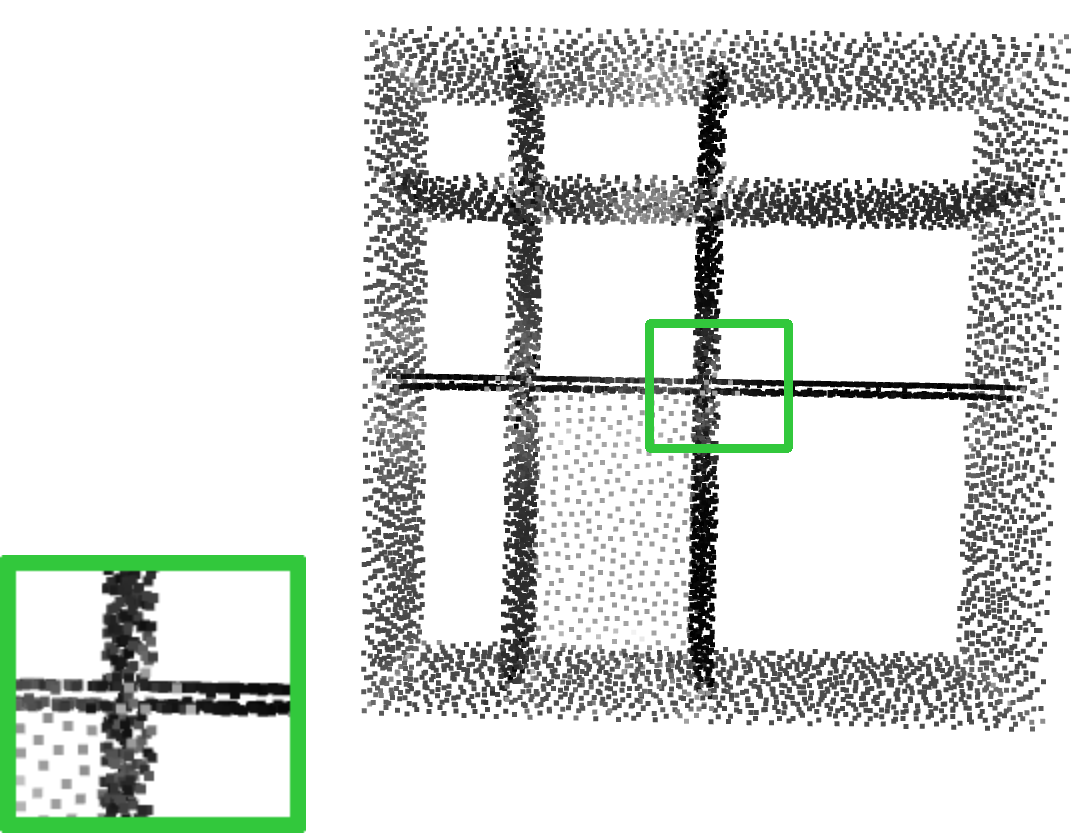} \\
    \includegraphics[width=0.15\linewidth]{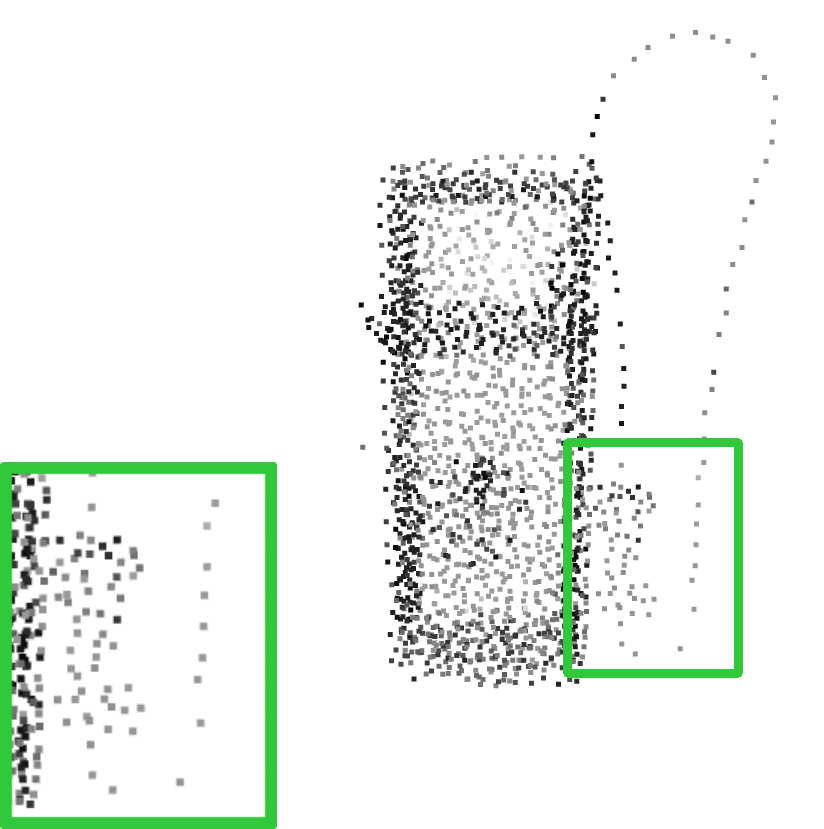} &
    \includegraphics[width=0.15\linewidth]{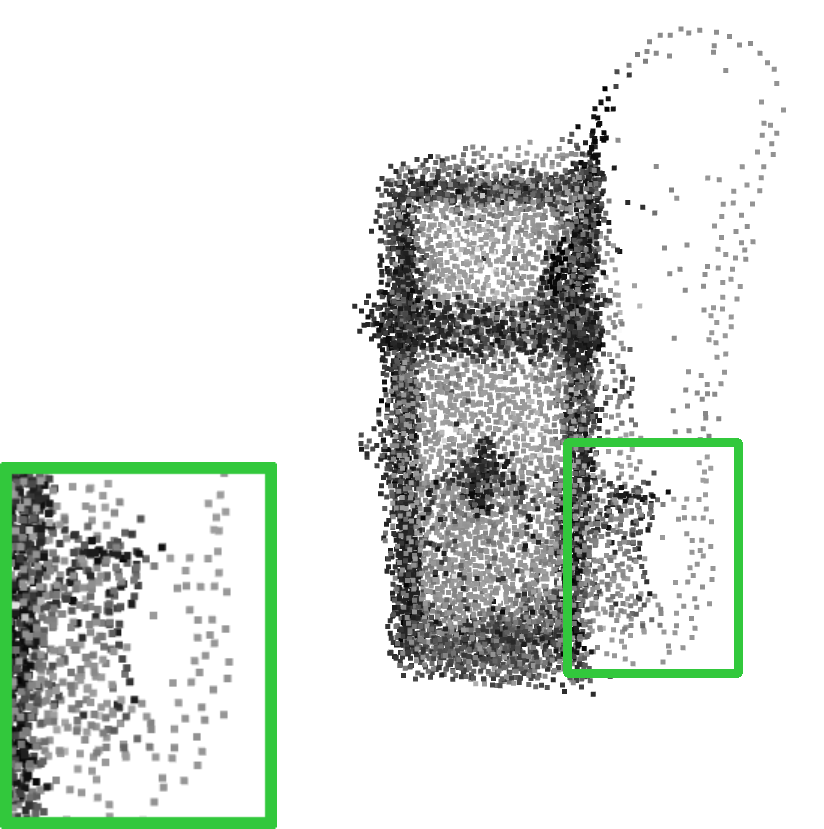} &
    \includegraphics[width=0.15\linewidth]{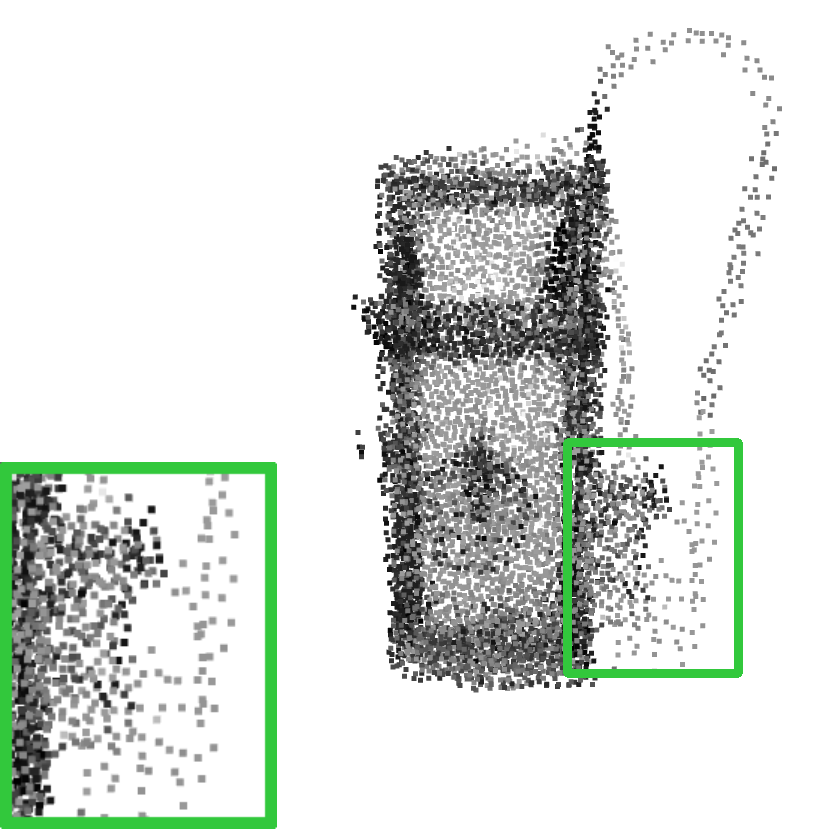} &
    \includegraphics[width=0.15\linewidth]{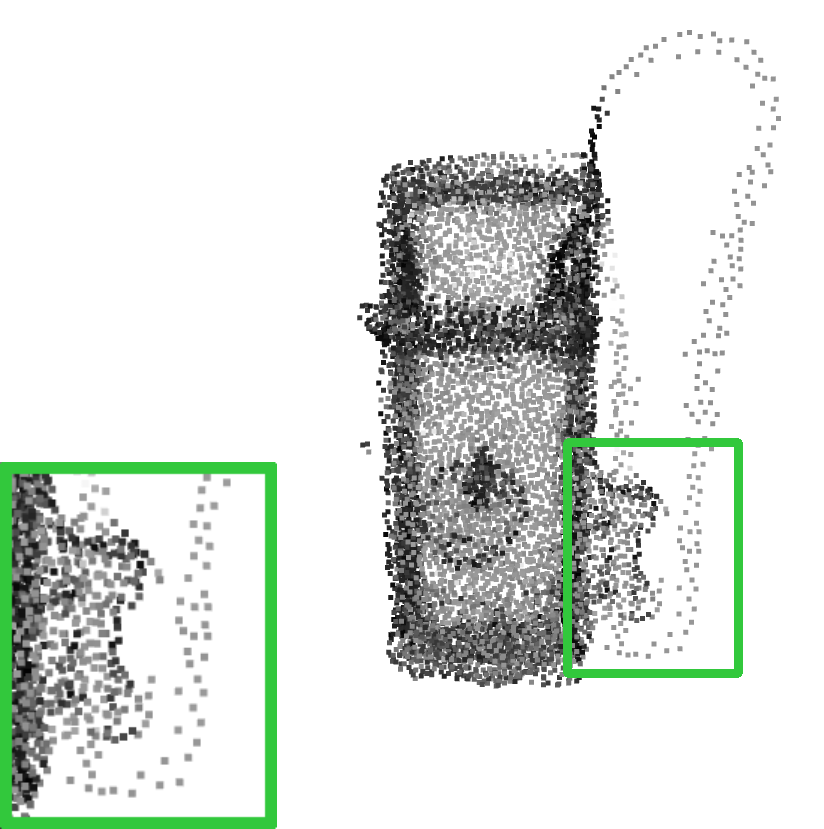} &
    \includegraphics[width=0.15\linewidth]{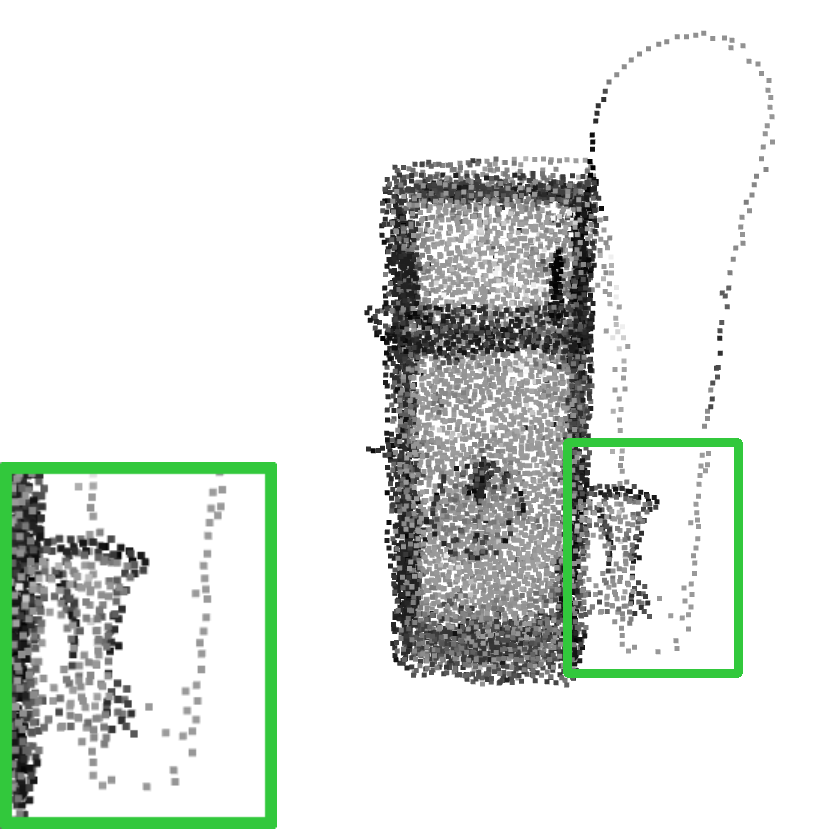} &
    \includegraphics[width=0.15\linewidth]{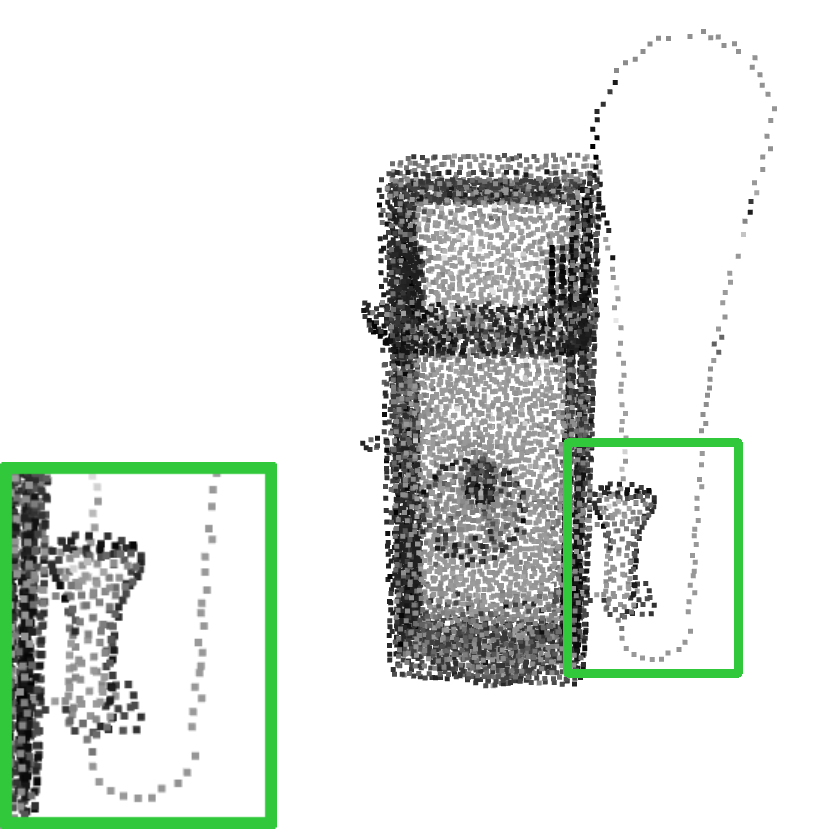} \\
    \includegraphics[width=0.15\linewidth]{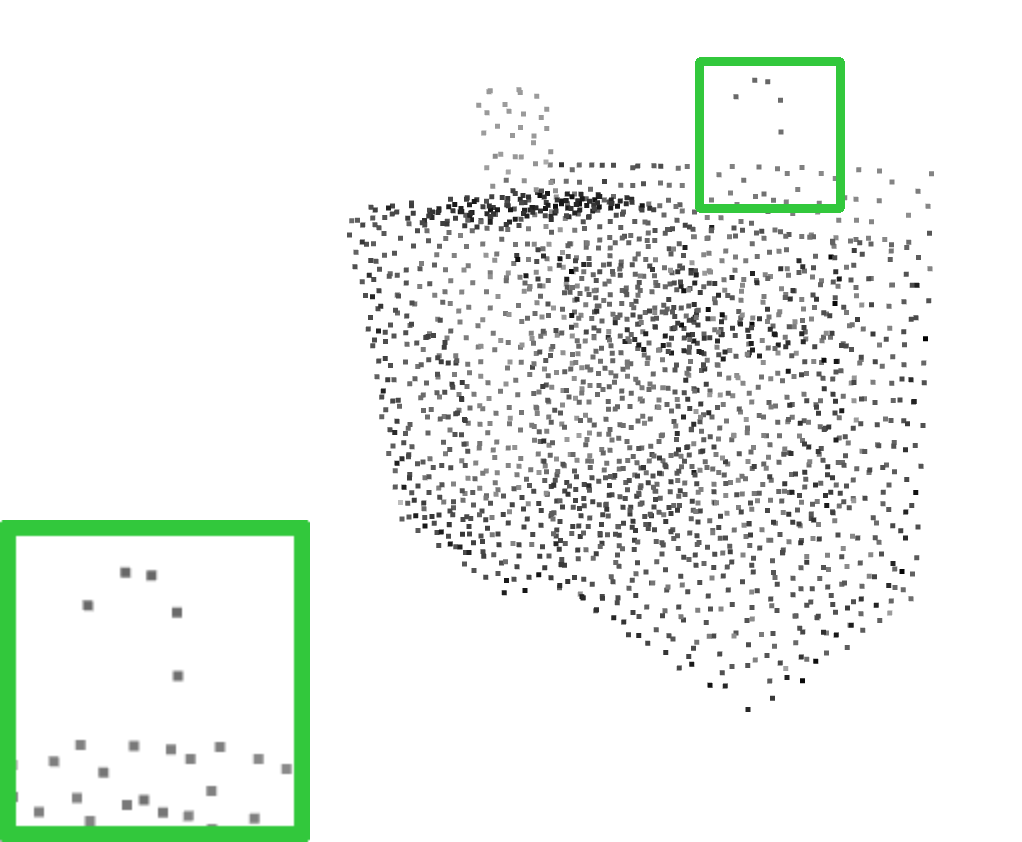} &
    \includegraphics[width=0.15\linewidth]{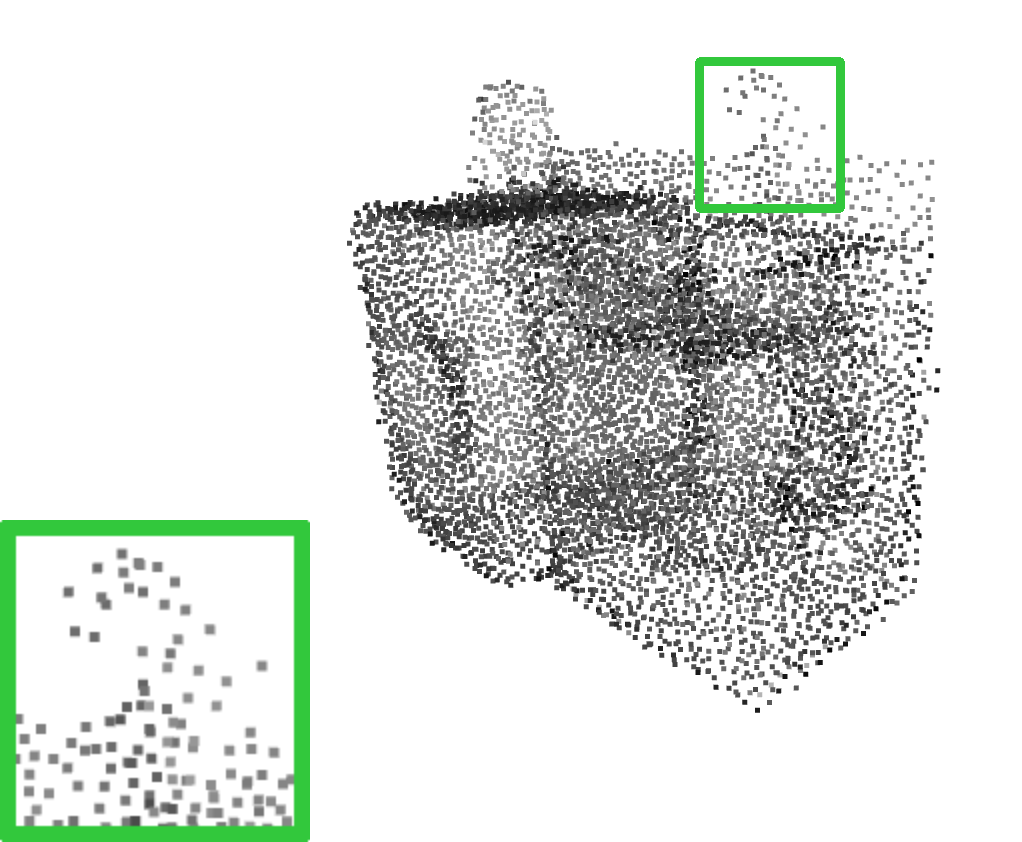} &
    \includegraphics[width=0.15\linewidth]{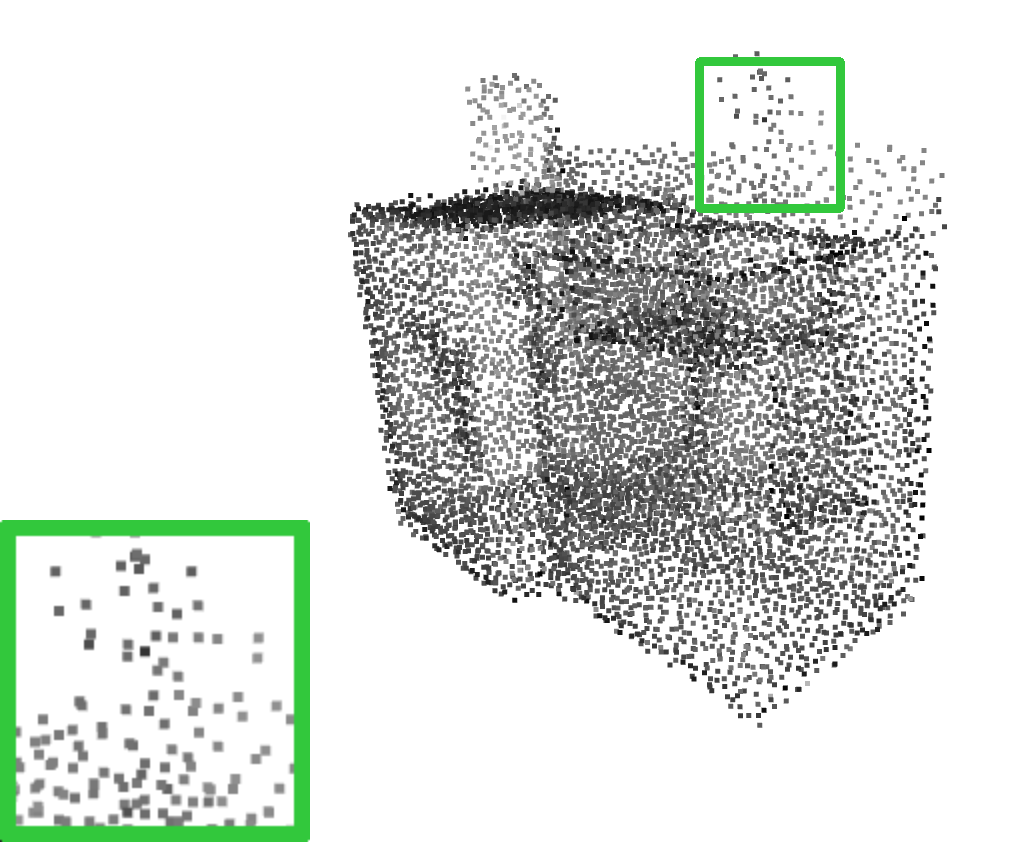} &
    \includegraphics[width=0.15\linewidth]{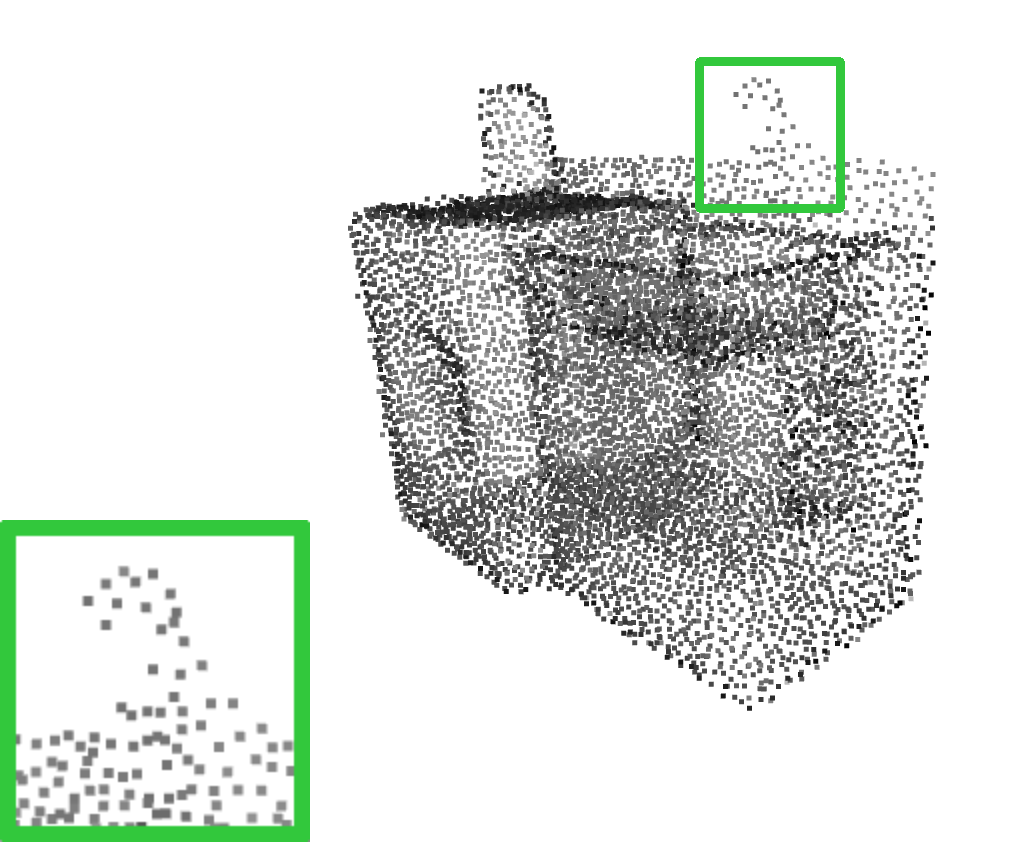} &
    \includegraphics[width=0.15\linewidth]{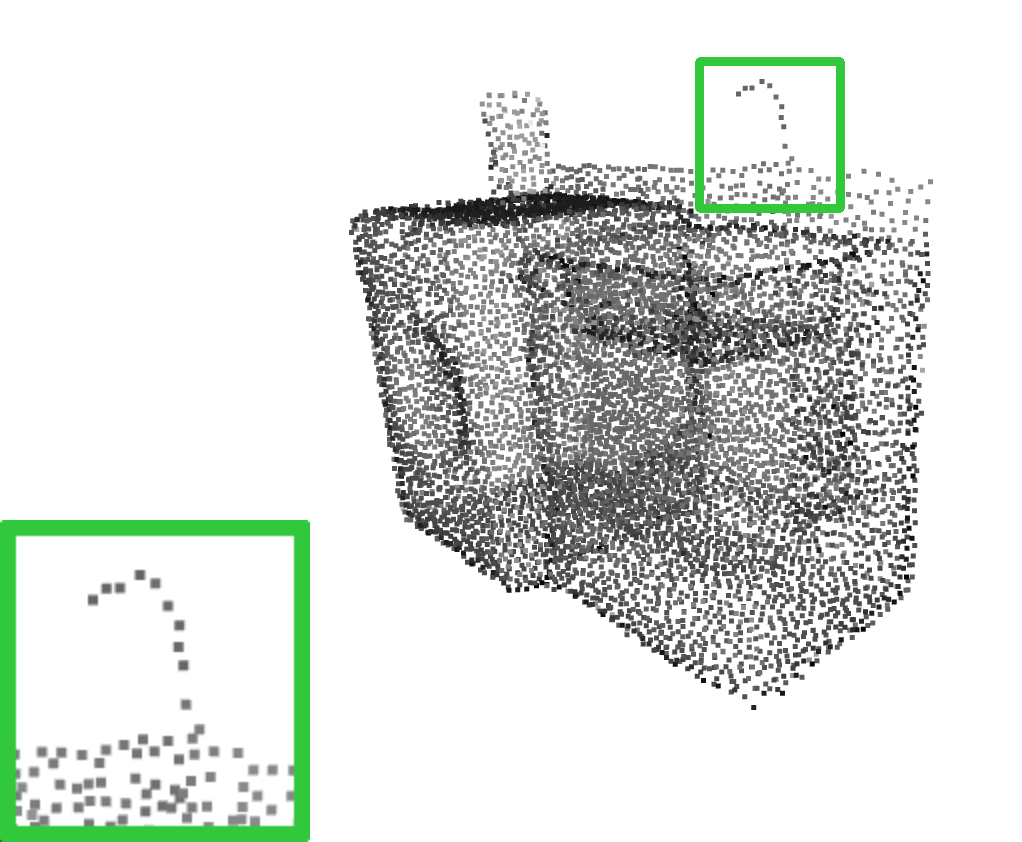} &
    \includegraphics[width=0.15\linewidth]{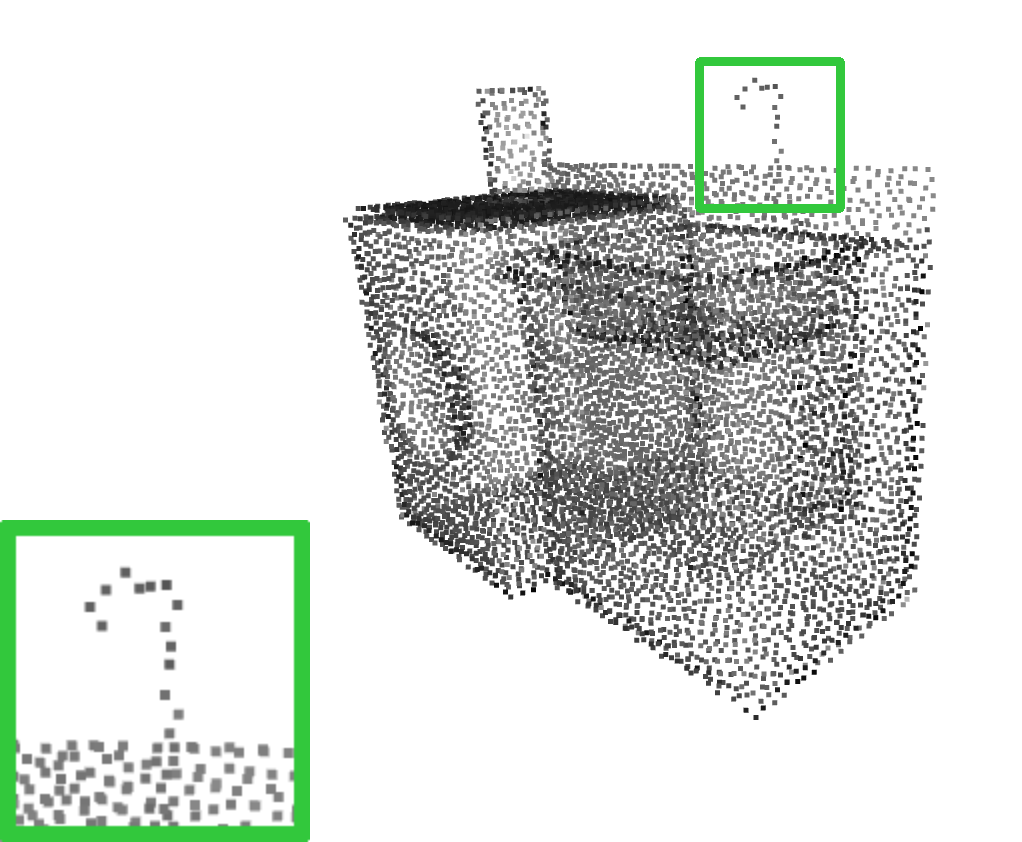} \\
    Input & PU-GAN \cite{pugan} & PU-GCN \cite{pugcn} & Dis-PU \cite{dispu} & Ours & Ground Truth
\end{tabular}
\addtolength{\tabcolsep}{2pt}
\vspace{0.2cm}
\captionof{figure}{Qualitative comparison with state-of-the-art methods on the PU1K dataset. Inputs with $2048$ points (left) are upsampled to $8192$ points (right), with upsampling ratio $r = 4$. Details are best viewed when zoomed in.}
\label{fig:supp_qual_1}
\end{table}

\begin{table}
\centering
\addtolength{\tabcolsep}{10pt} 
\begin{tabular}{ c c c c c }
    & $r = 4$ & $r = 8$ & $r = 12$ & $r = 16$ \\
    \multirow{1}{*}[6em]{\rotatebox{90}{Input}} &
    \includegraphics[width=0.175\linewidth]{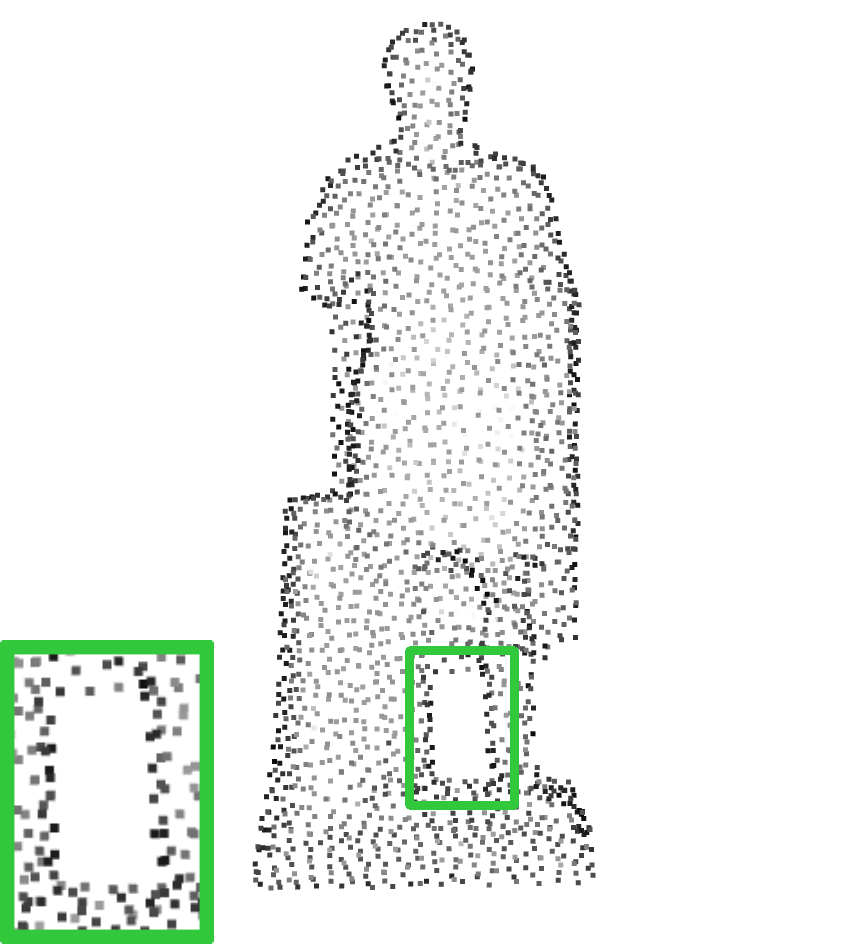} &
    \includegraphics[width=0.175\linewidth]{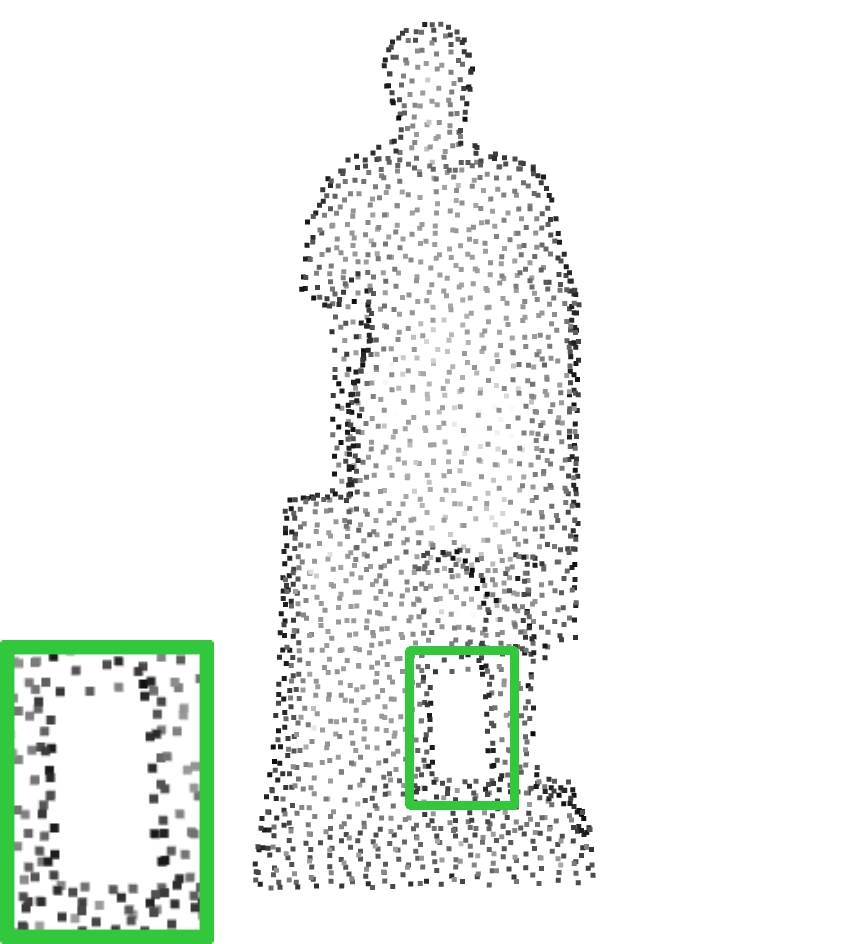} &
    \includegraphics[width=0.175\linewidth]{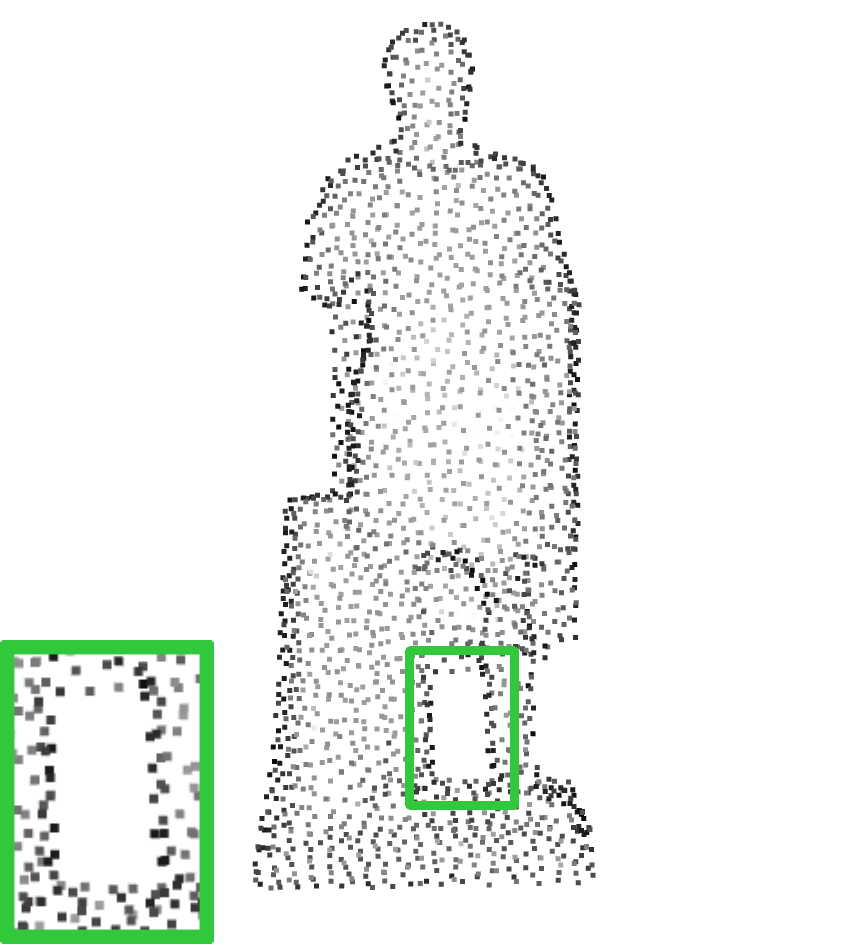} &
    \includegraphics[width=0.175\linewidth]{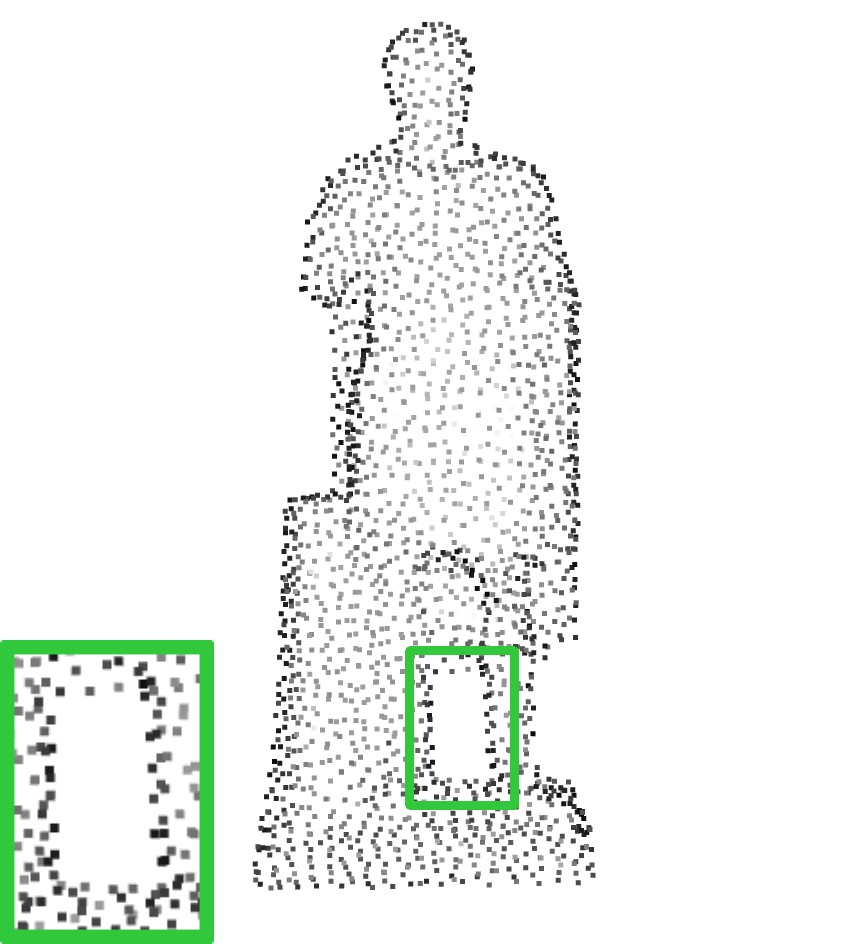} \\
    \multirow{1}{*}[6em]{\rotatebox{90}{MAFU \cite{mafu}}} &
    \includegraphics[width=0.175\linewidth]{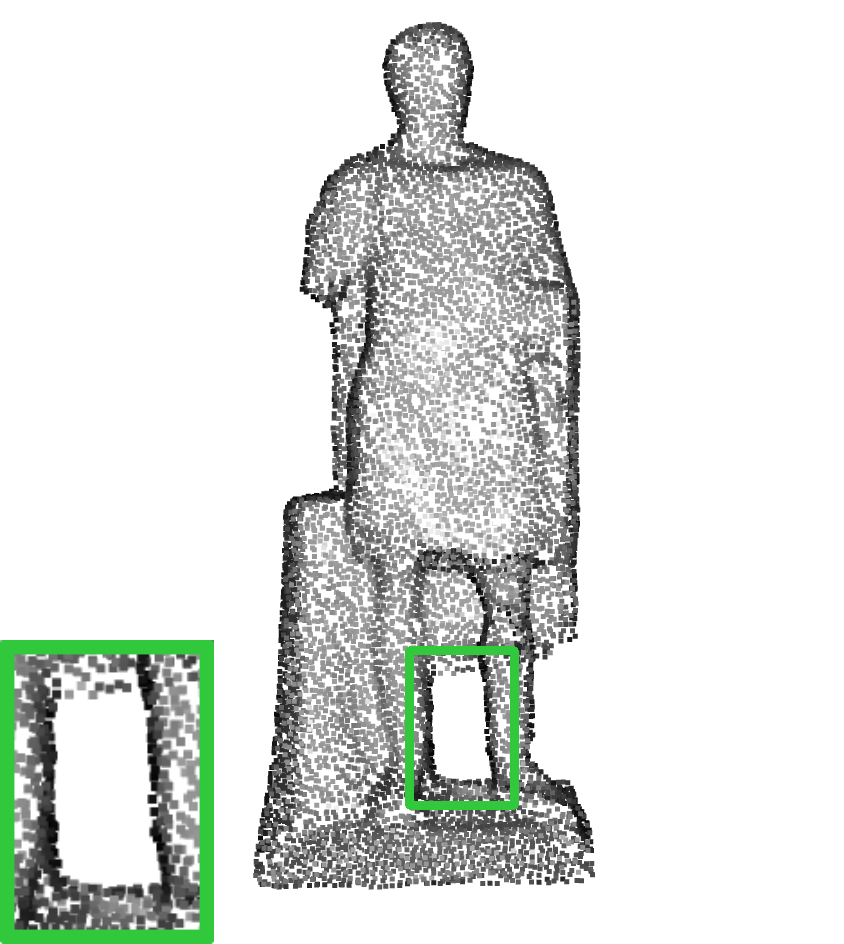} &
    \includegraphics[width=0.175\linewidth]{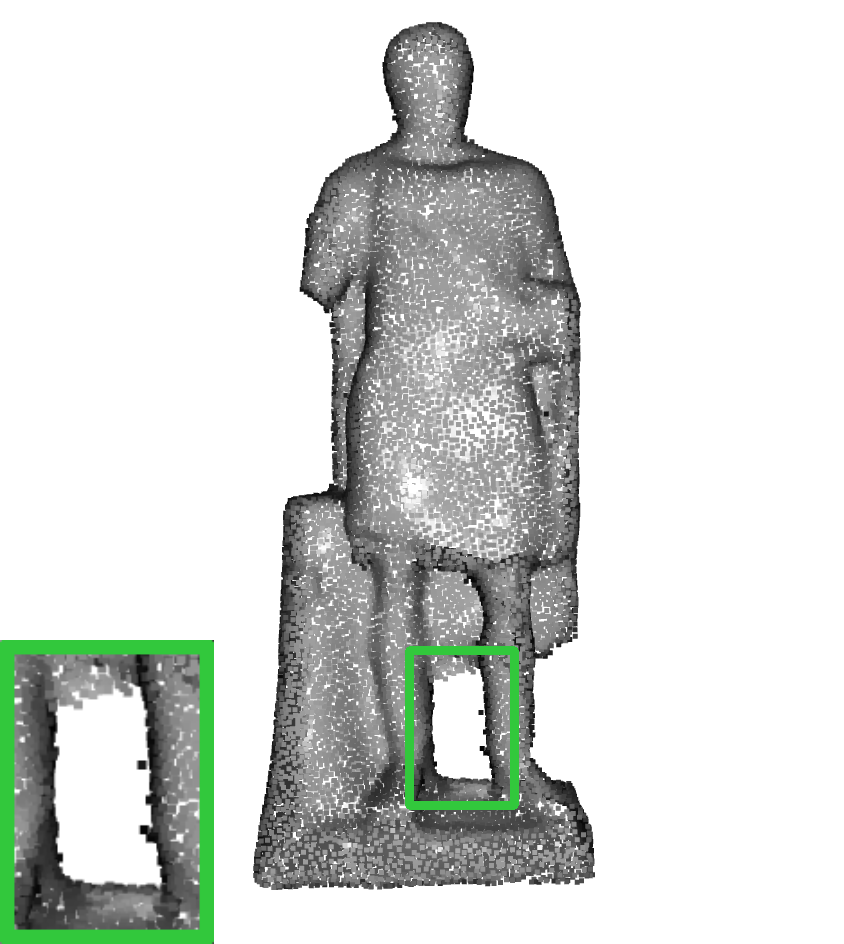} &
    \includegraphics[width=0.175\linewidth]{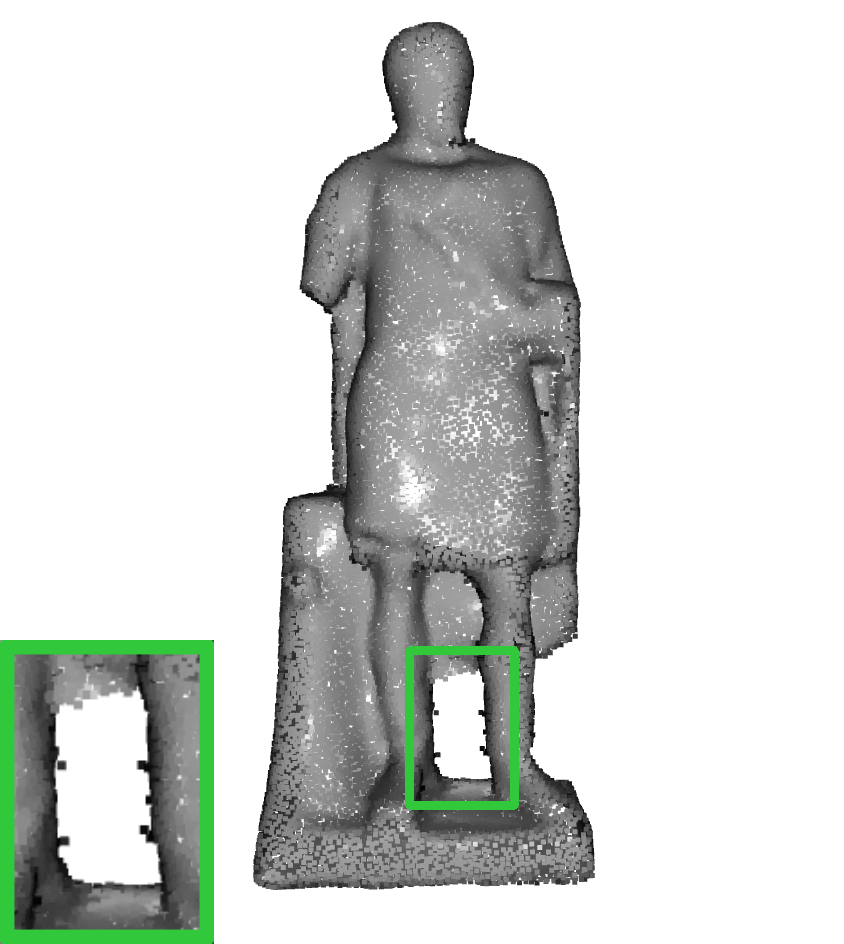} &
    \includegraphics[width=0.175\linewidth]{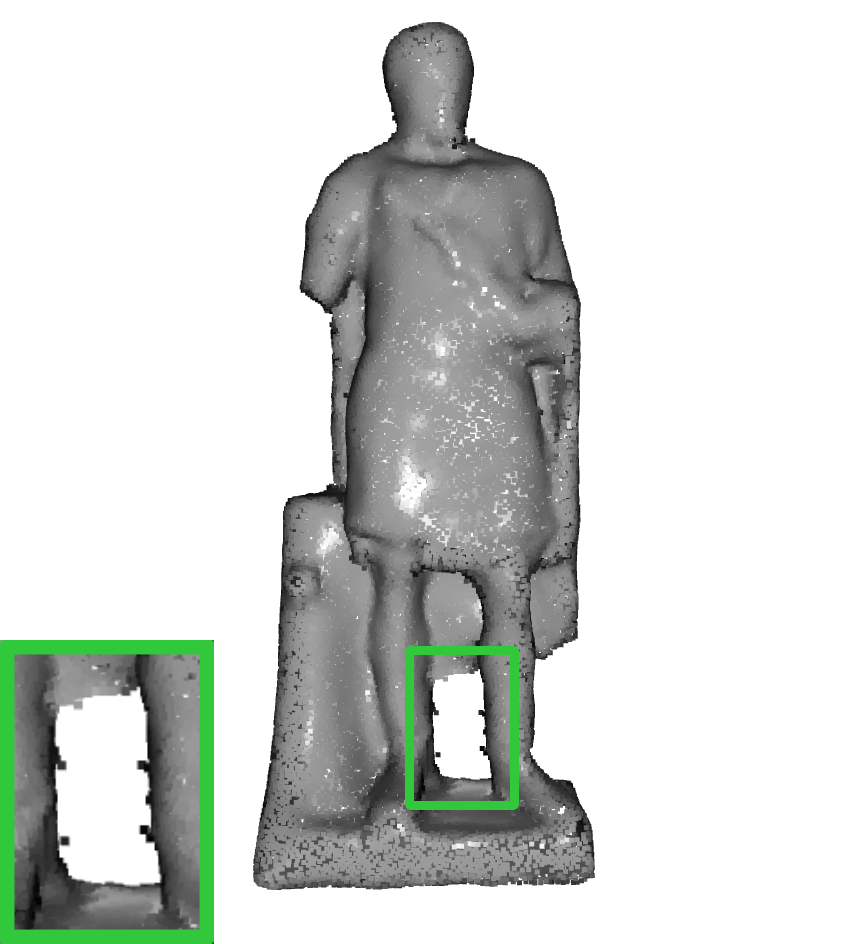} \\
    \multirow{1}{*}[6em]{\rotatebox{90}{PU-EVA \cite{pueva}}} &
    \includegraphics[width=0.175\linewidth]{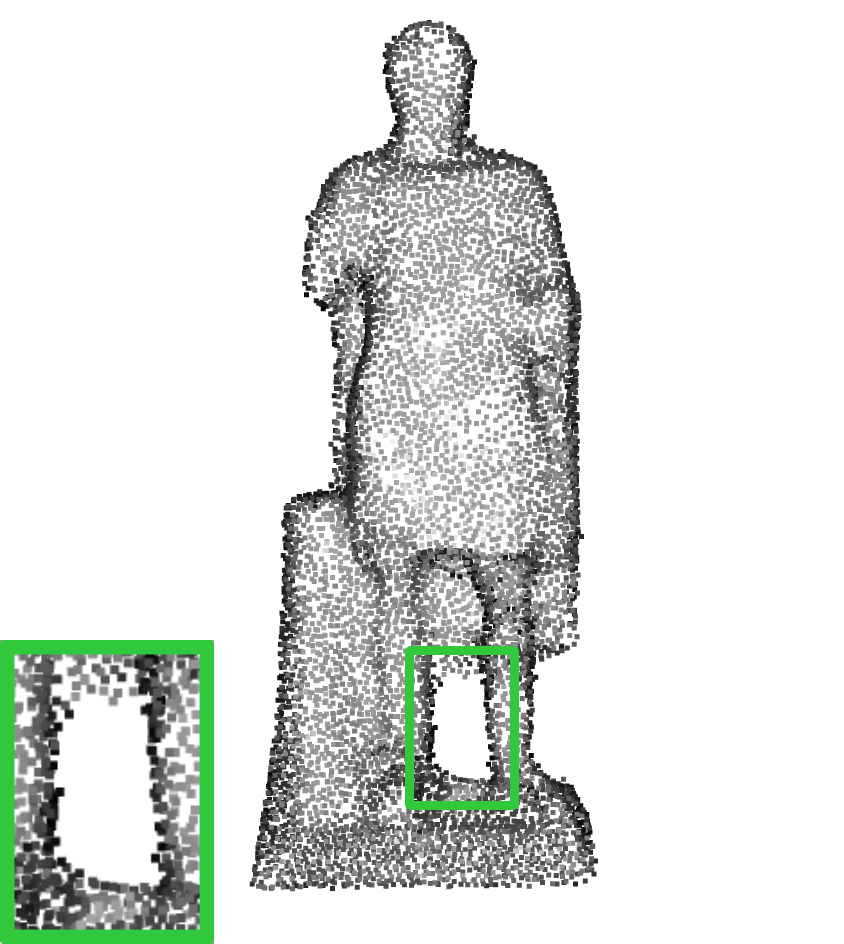} &
    \includegraphics[width=0.175\linewidth]{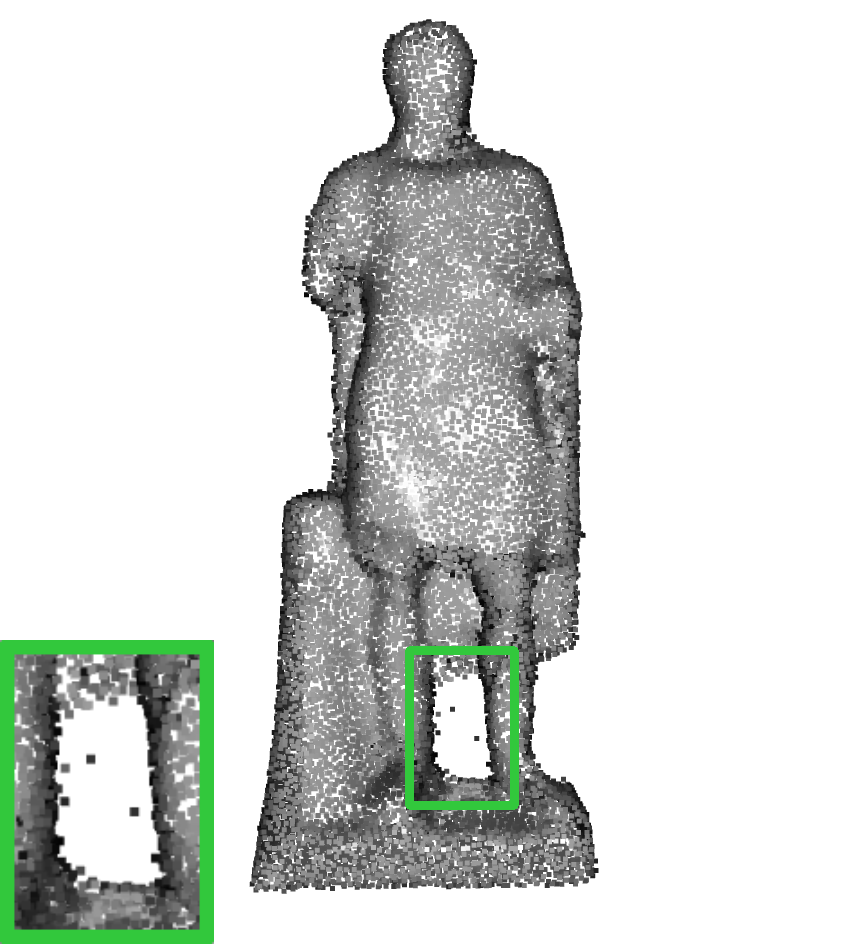} &
    \includegraphics[width=0.175\linewidth]{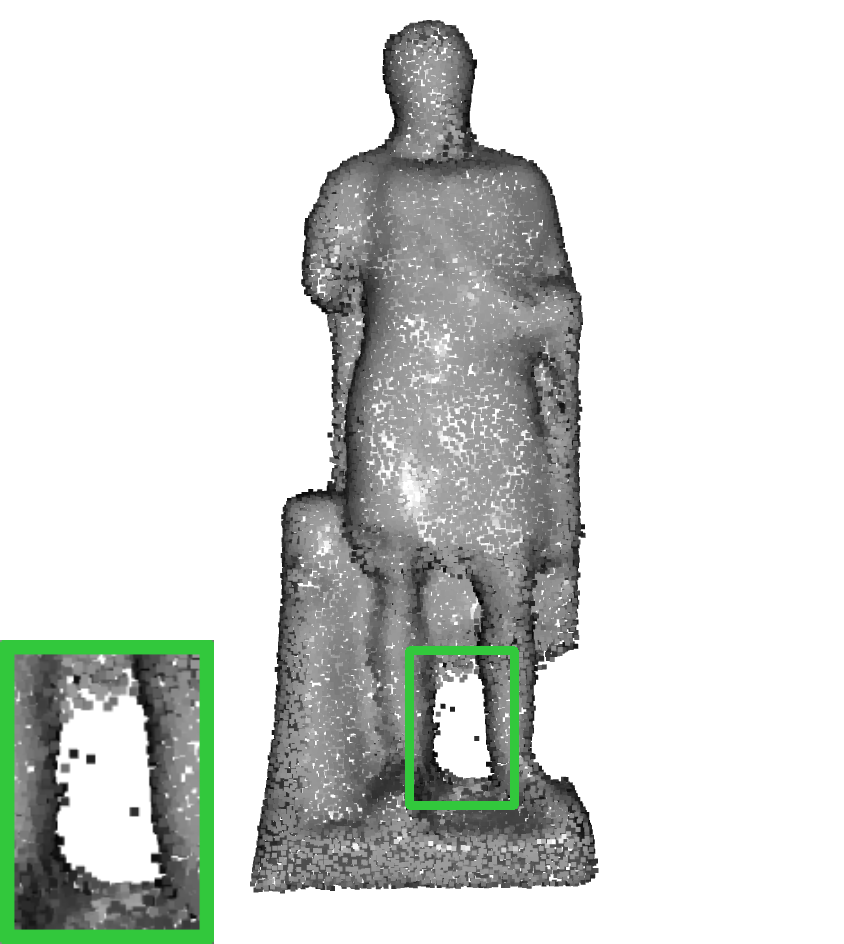} &
    \includegraphics[width=0.175\linewidth]{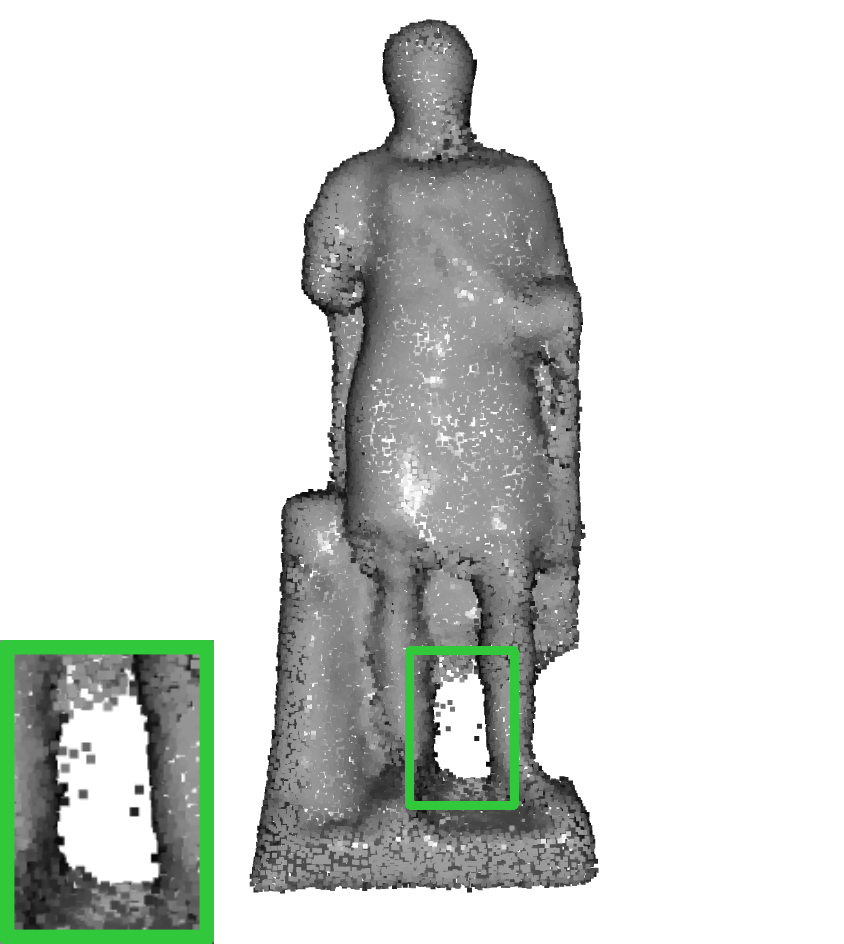} \\
    \multirow{1}{*}[6em]{\rotatebox{90}{NePs \cite{neuralpts}}} &
    \includegraphics[width=0.175\linewidth]{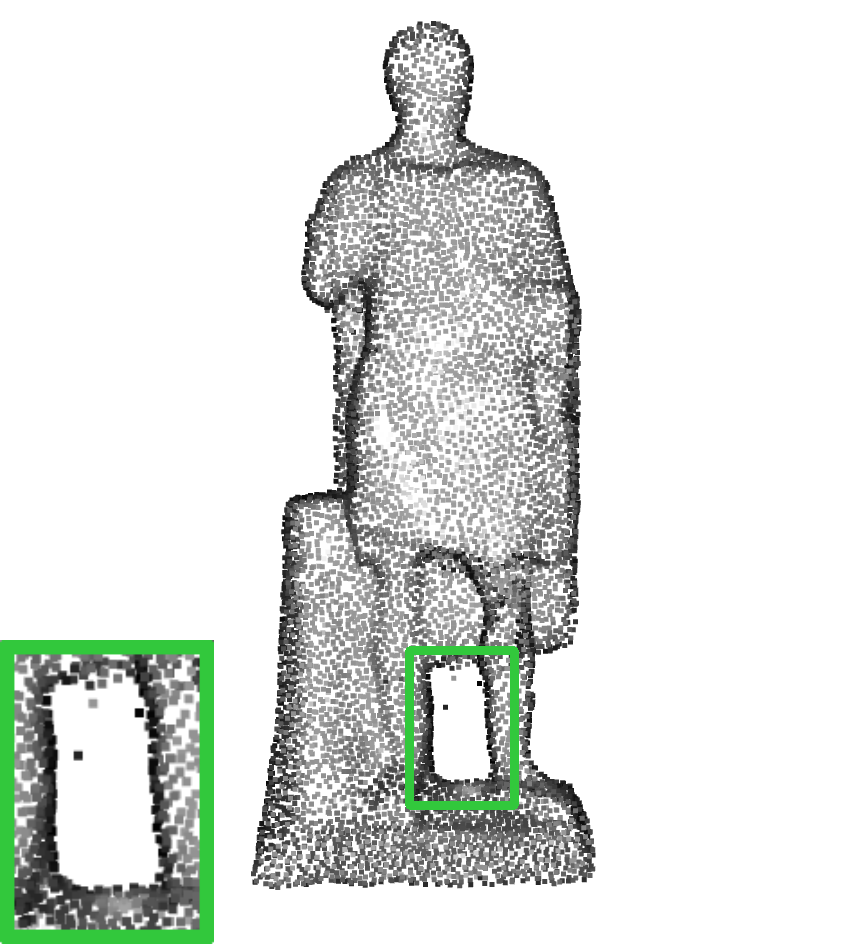} &
    \includegraphics[width=0.175\linewidth]{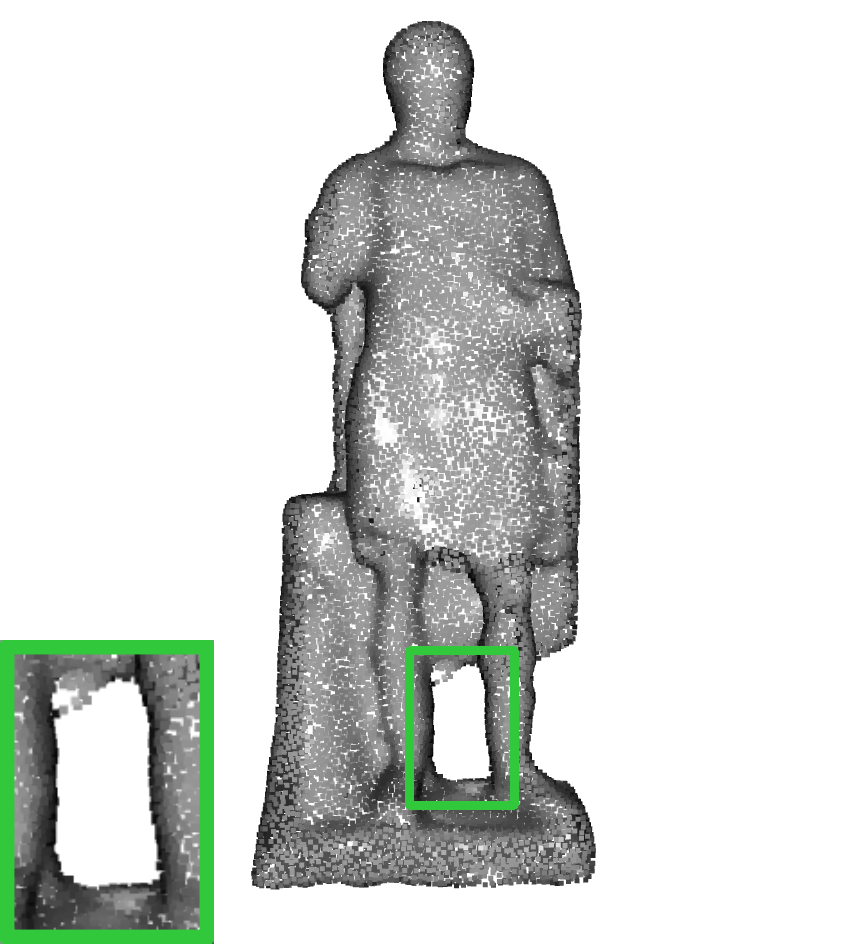} &
    \includegraphics[width=0.175\linewidth]{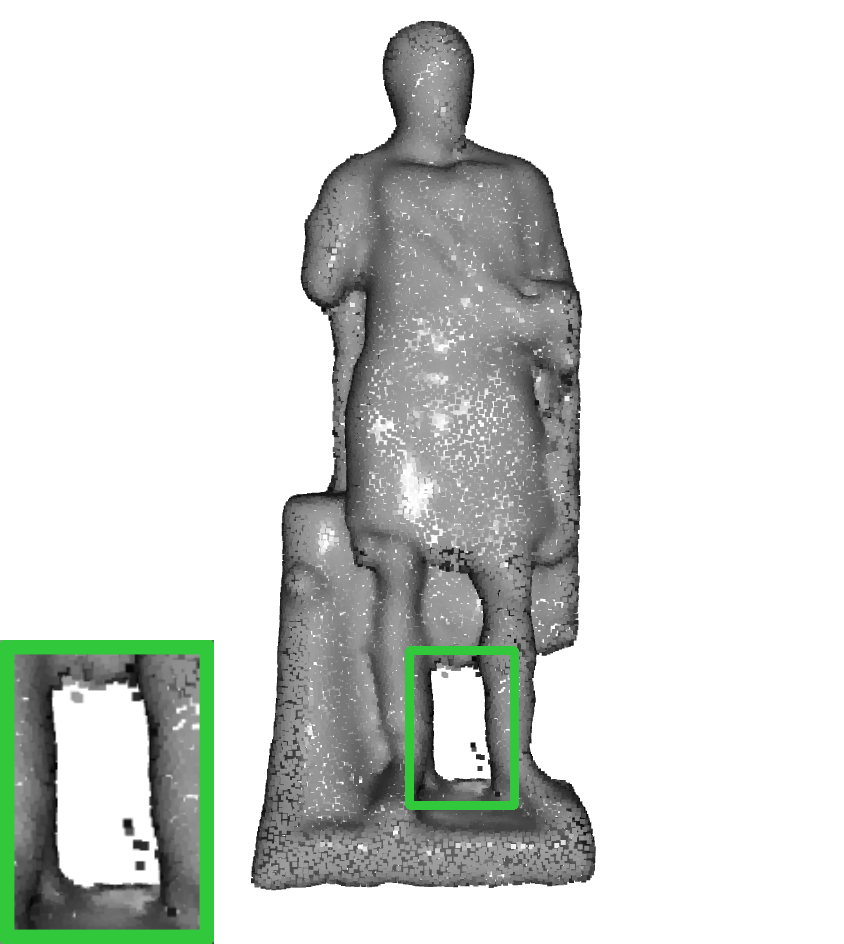} &
    \includegraphics[width=0.175\linewidth]{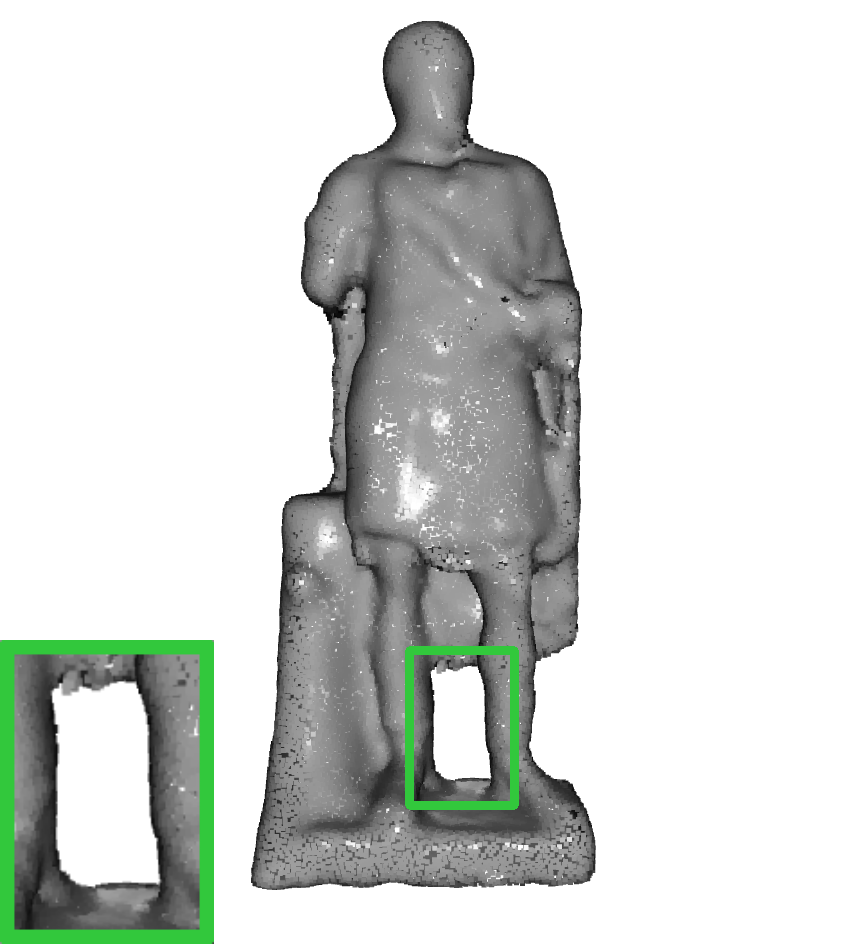} \\
    \multirow{1}{*}[6em]{\rotatebox{90}{Ours}} &
    \includegraphics[width=0.175\linewidth]{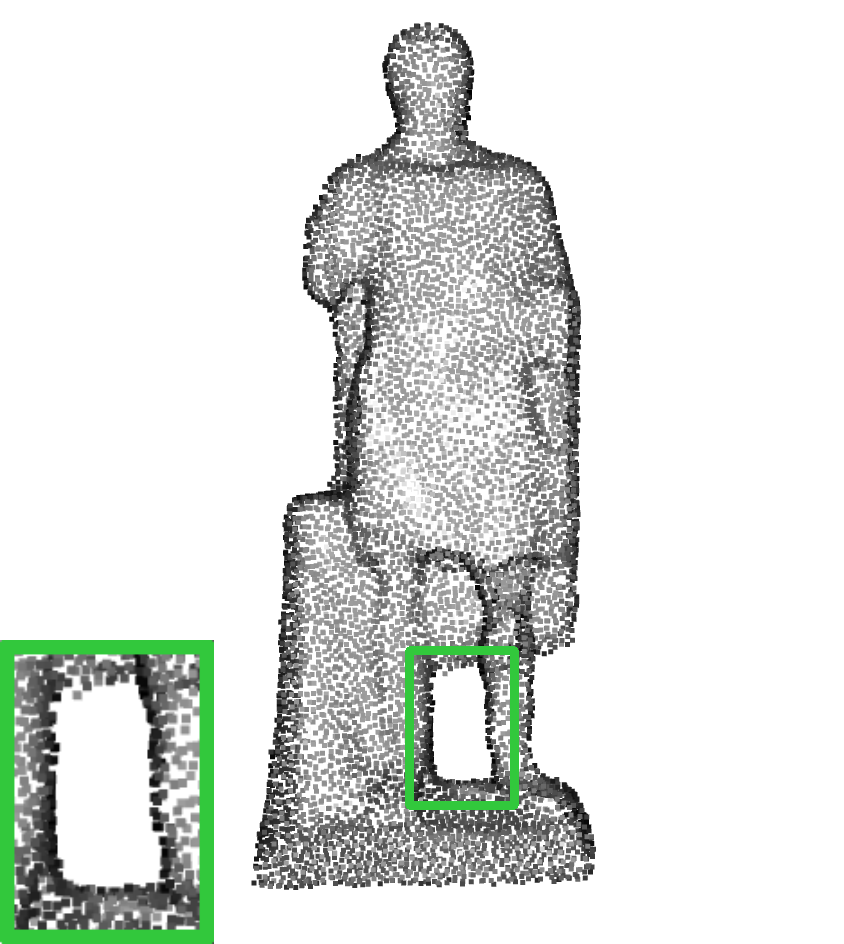} &
    \includegraphics[width=0.175\linewidth]{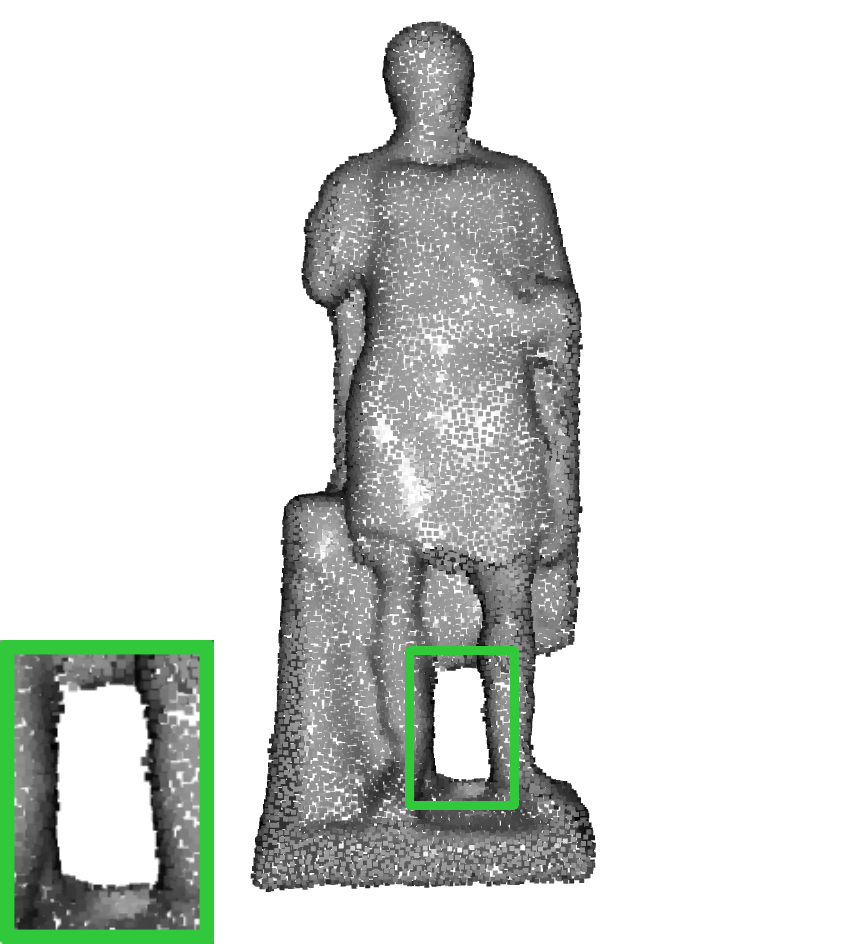} &
    \includegraphics[width=0.175\linewidth]{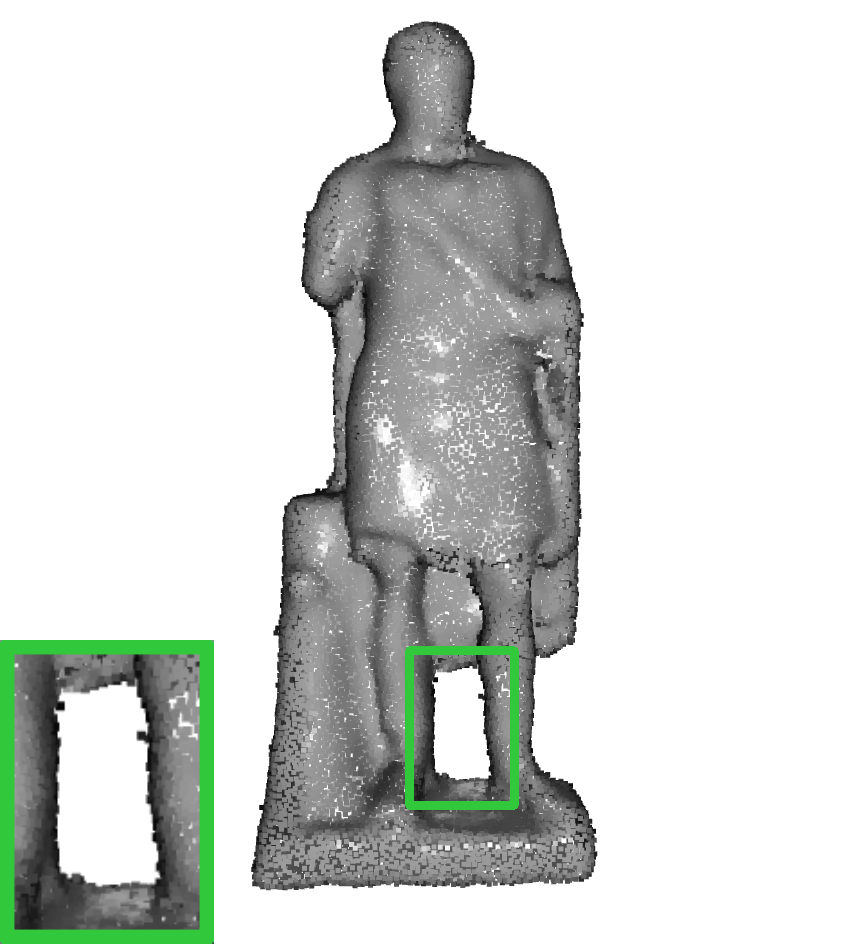} &
    \includegraphics[width=0.175\linewidth]{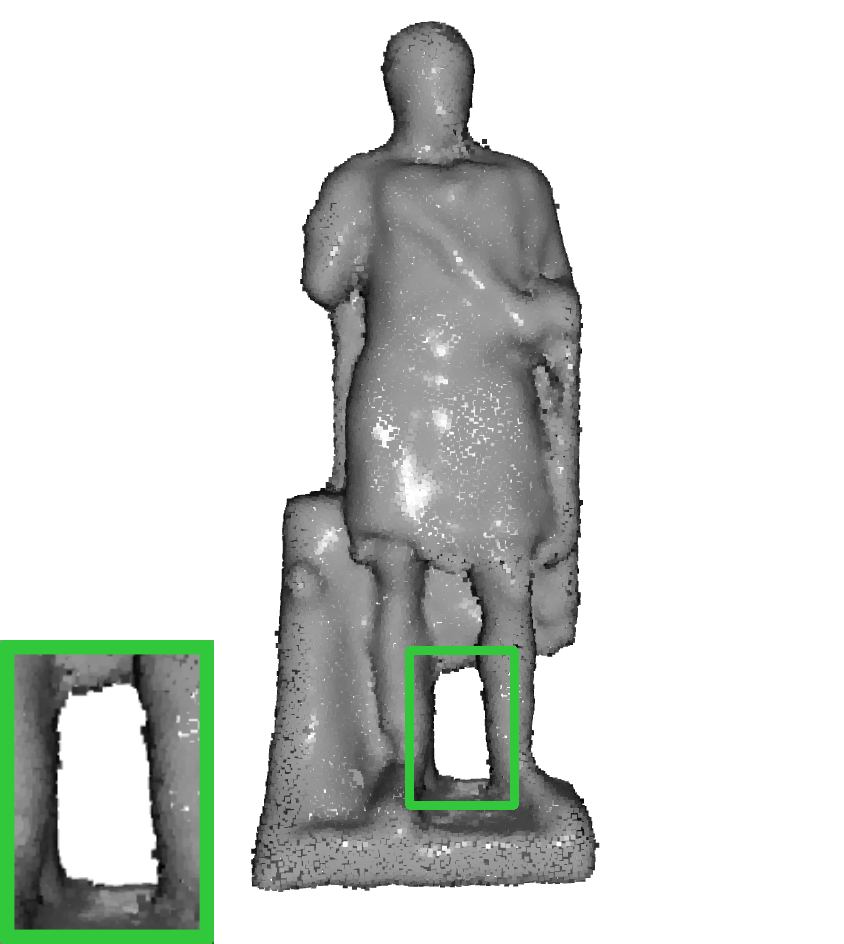} \\
    \multirow{1}{*}[6em]{\rotatebox{90}{Ground Truth}} &
    \includegraphics[width=0.175\linewidth]{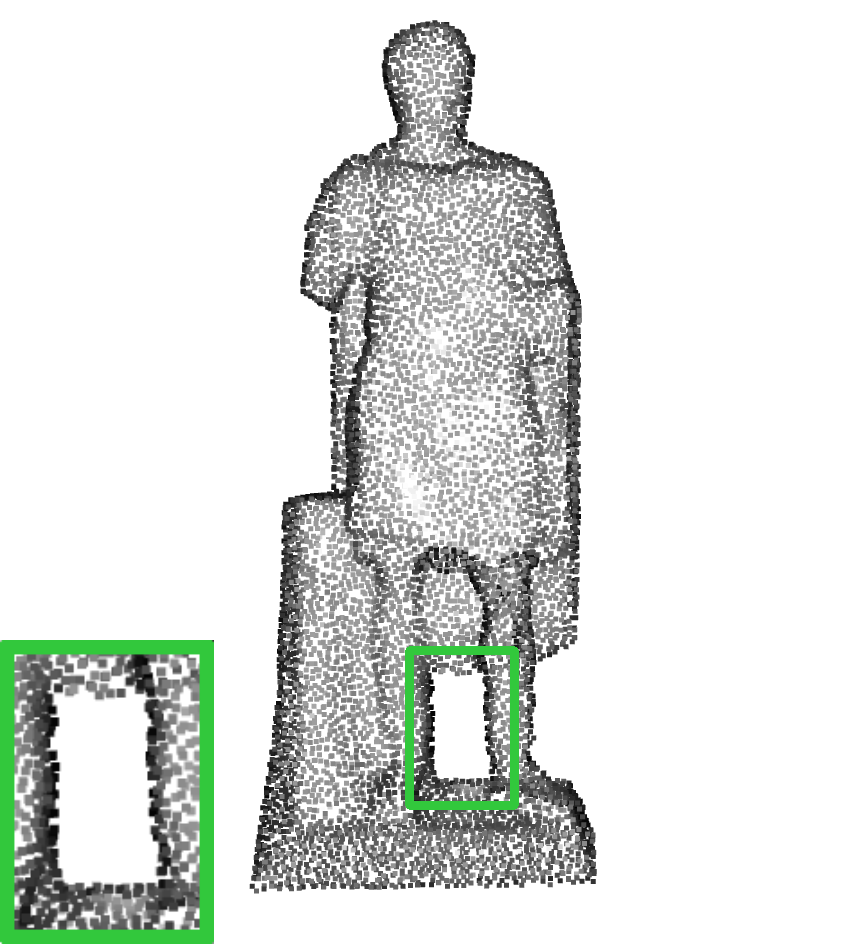} &
    \includegraphics[width=0.175\linewidth]{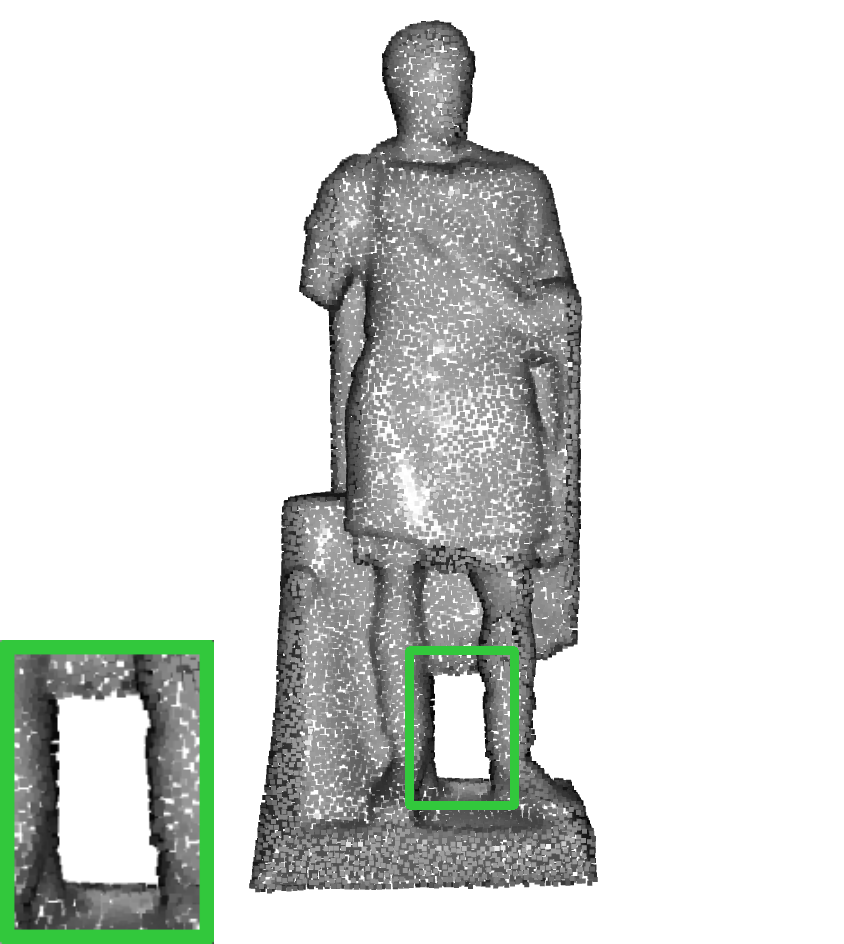} &
    \includegraphics[width=0.175\linewidth]{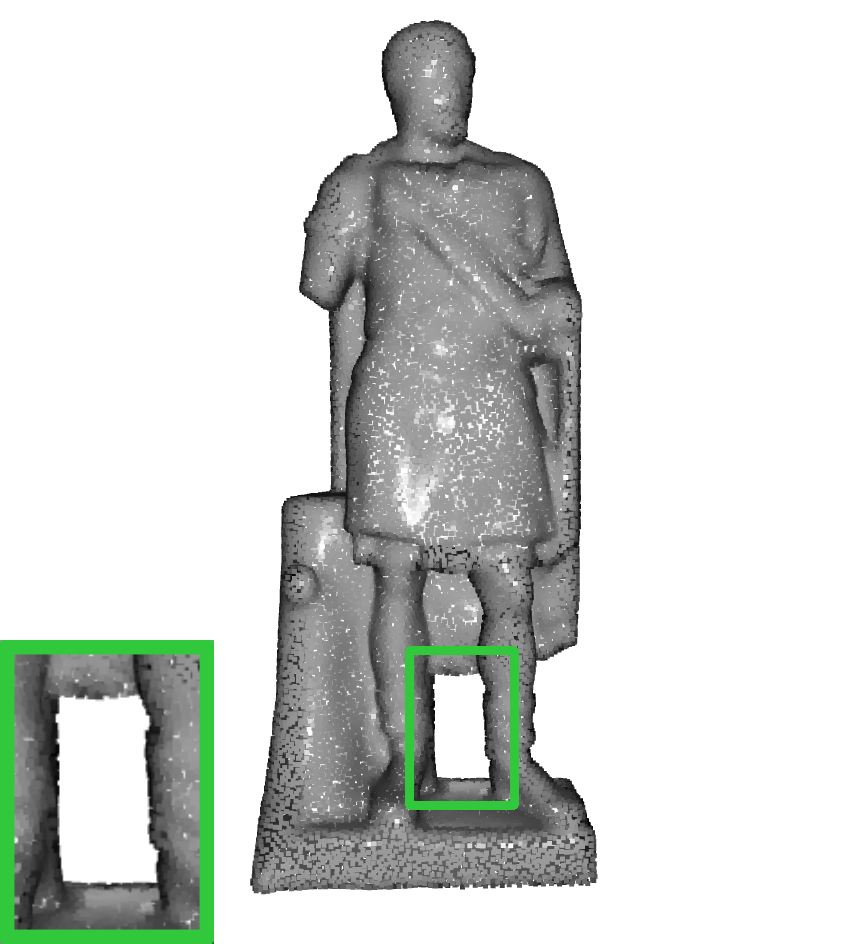} &
    \includegraphics[width=0.175\linewidth]{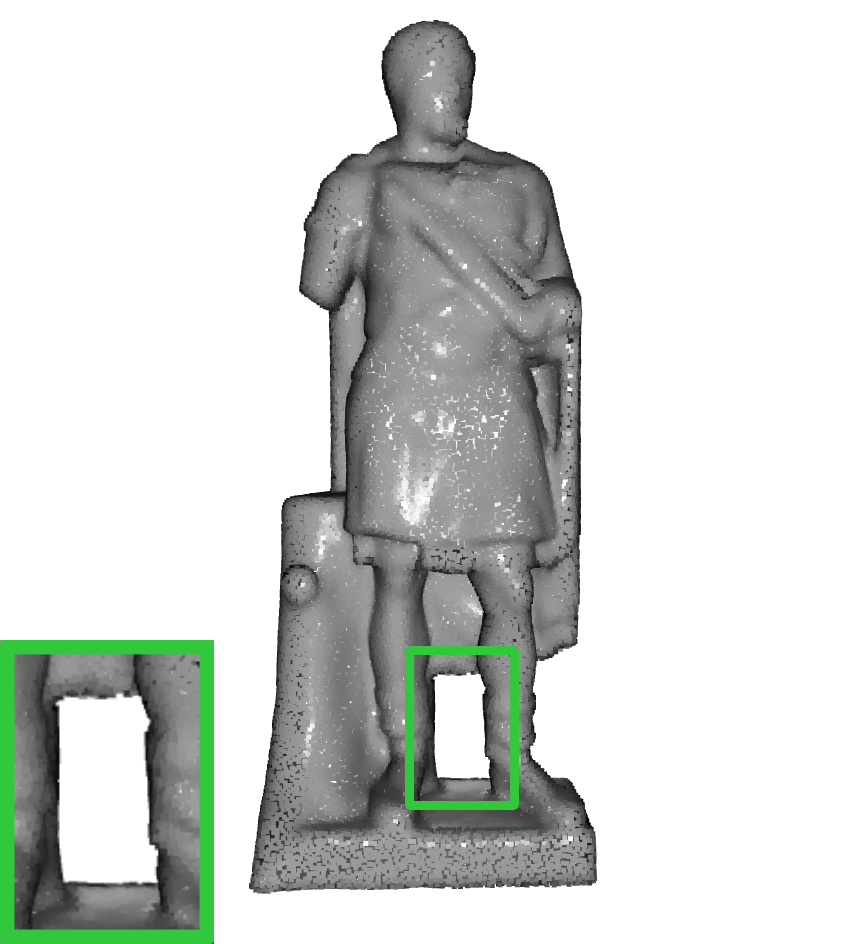} \\
\end{tabular}
\addtolength{\tabcolsep}{-10pt}
\vspace{0.2cm}
\captionof{figure}{Qualitative comparison with state-of-the-art methods on the PU-GAN dataset and flexible upsampling ratios. Inputs with $2048$ points are upsampled with $r \in \{4,8,12,16\}$. Details are best viewed when zoomed in.}
\label{fig:supp_qual_2}
\end{table}

\begin{table}[h!]
\centering
\begin{tabular}{ c c c c c }
 \multirow{1}{*}[7em]{\rotatebox{90}{Input}} &
 \includegraphics[width=0.18\linewidth]{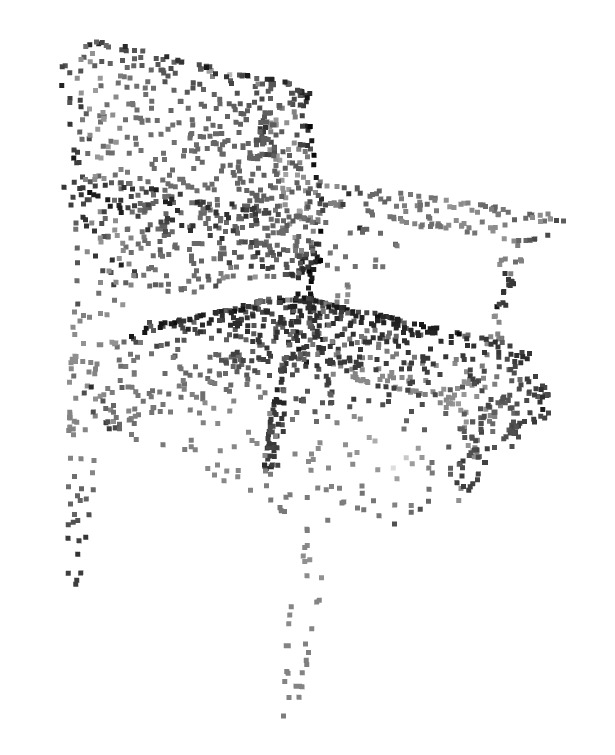} & 
 \includegraphics[width=0.32\linewidth]{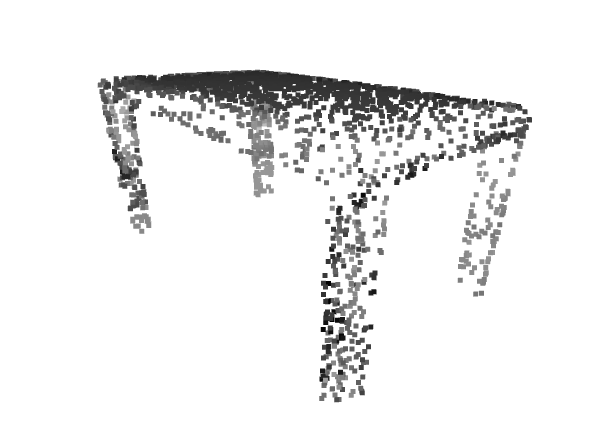} & 
 \includegraphics[width=0.18\linewidth]{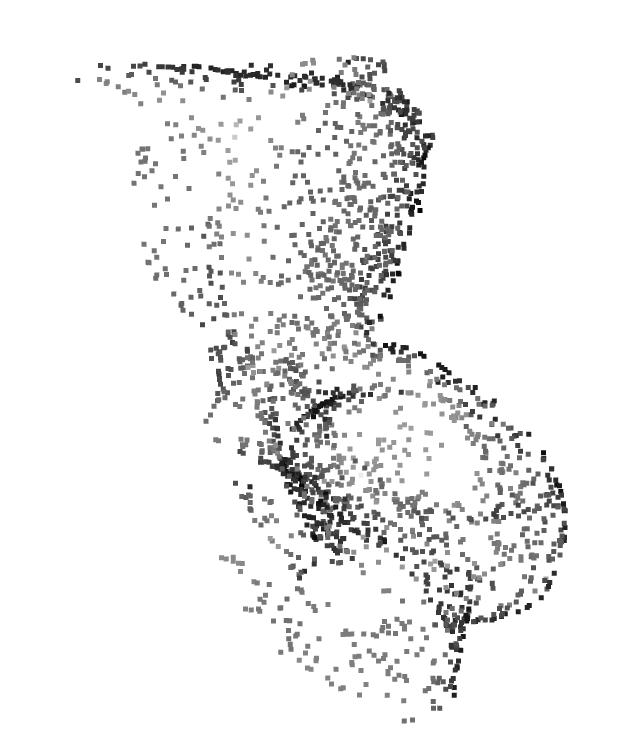} &
 \includegraphics[width=0.22\linewidth]{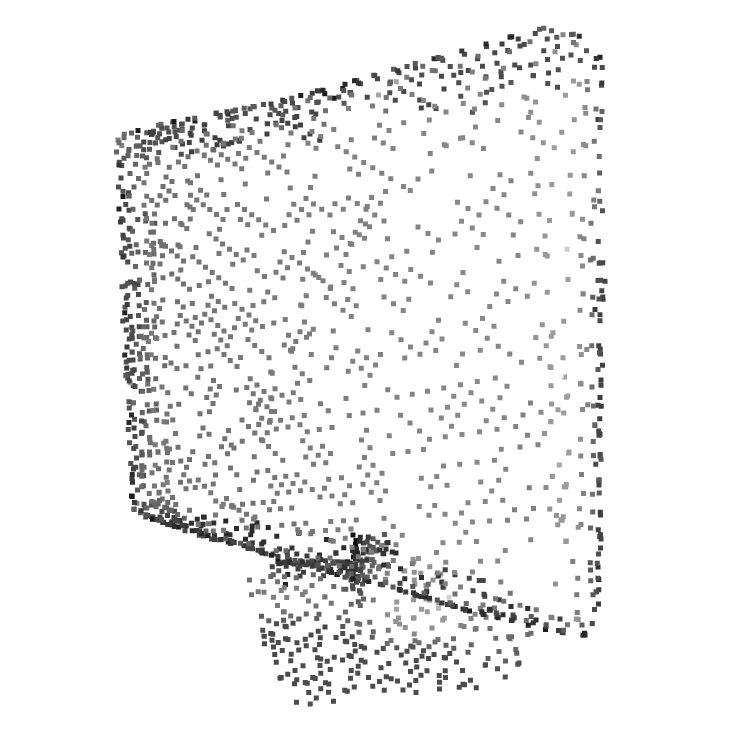}\\
 \multirow{1}{*}[7em]{\rotatebox{90}{Output}} &
 \includegraphics[width=0.18\linewidth]{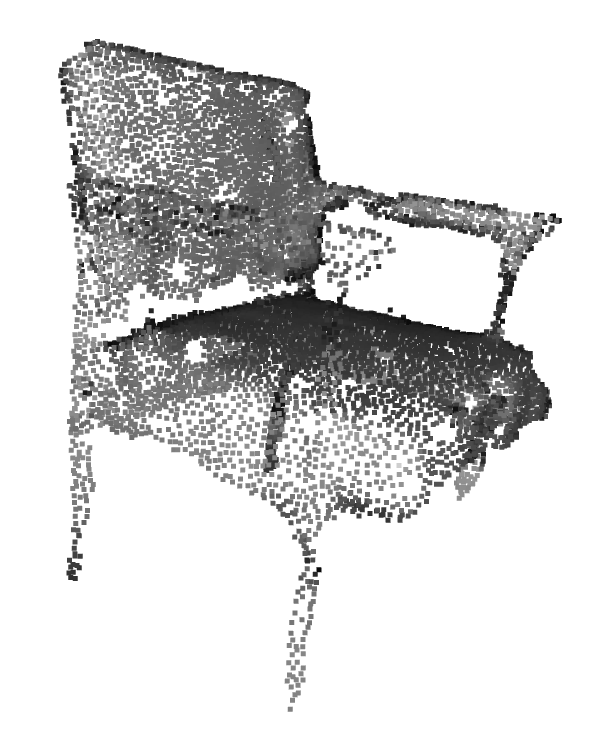} & 
 \includegraphics[width=0.32\linewidth]{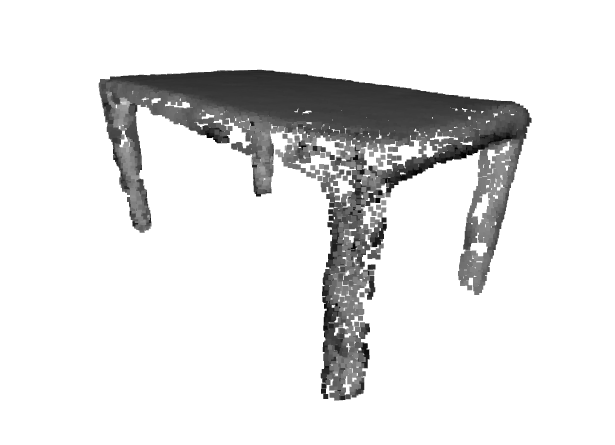} & 
 \includegraphics[width=0.18\linewidth]{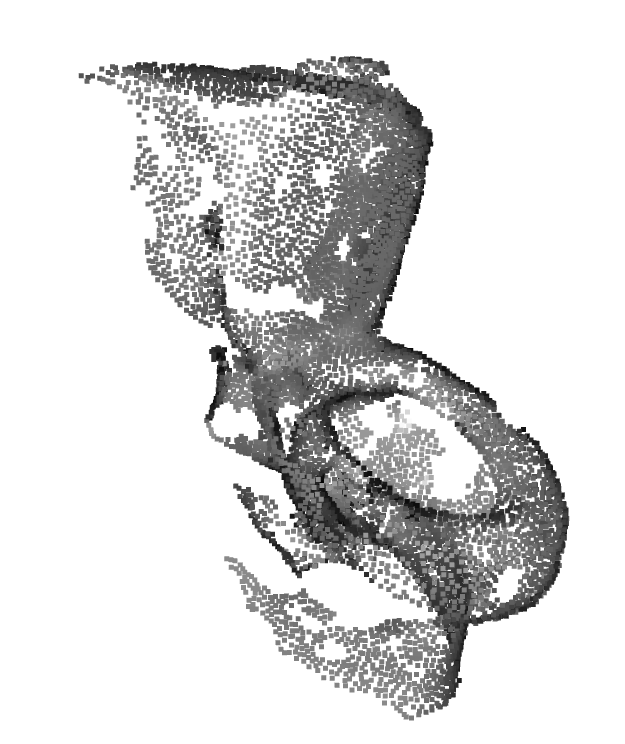} &
 \includegraphics[width=0.22\linewidth]{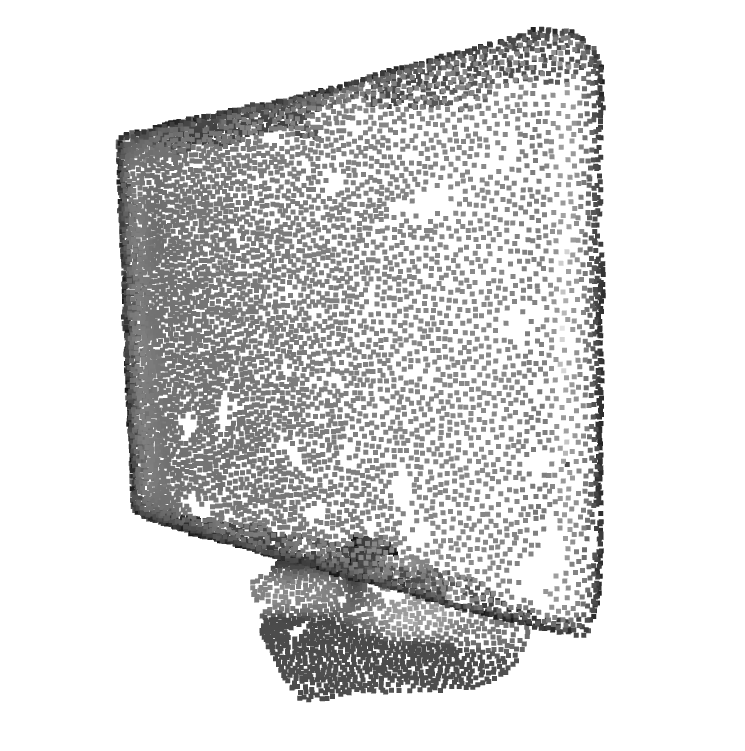}
\end{tabular}
\vspace{0.2cm}
\captionof{figure}{Qualitative upsampling results on the real-world ScanObjectNN dataset \cite{scanobj}.}
\label{fig:scanobj}
\end{table}

\begin{table}[h!]
\centering
\begin{tabular}{ c c c c }
 & 512 $\times$ 16 & 1024 $\times$ 8 & 2048 $\times$ 4 \\
 \multirow{1}{*}[5em]{\rotatebox{90}{Input}} &
 \includegraphics[width=0.31\linewidth]{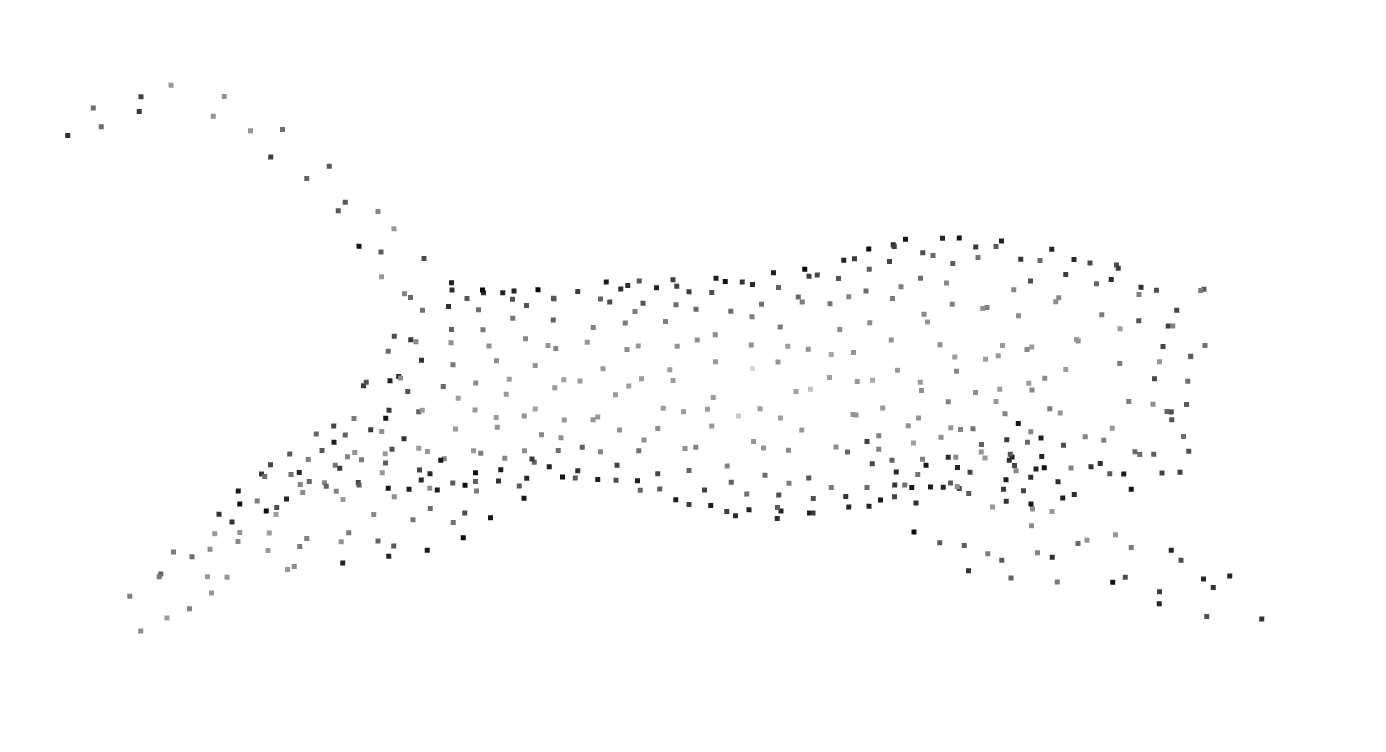} & 
 \includegraphics[width=0.31\linewidth]{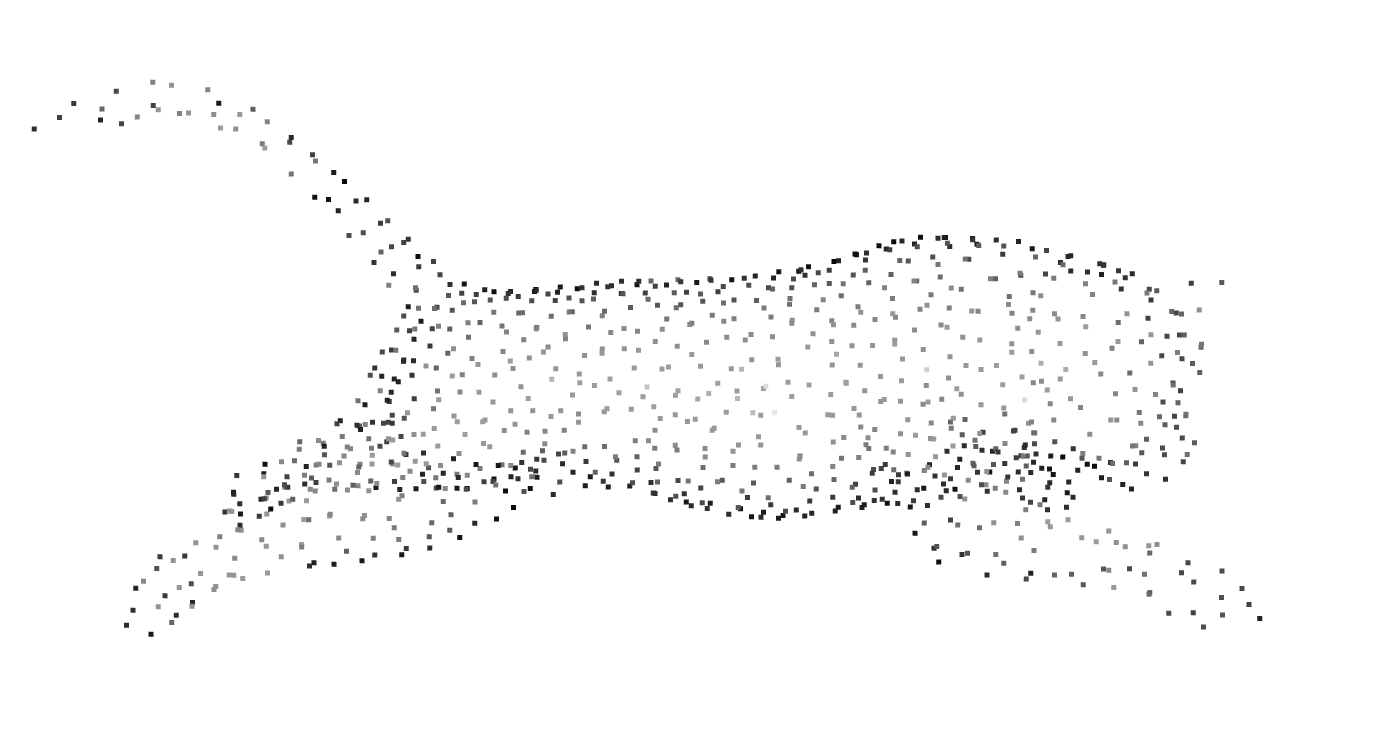} & 
 \includegraphics[width=0.31\linewidth]{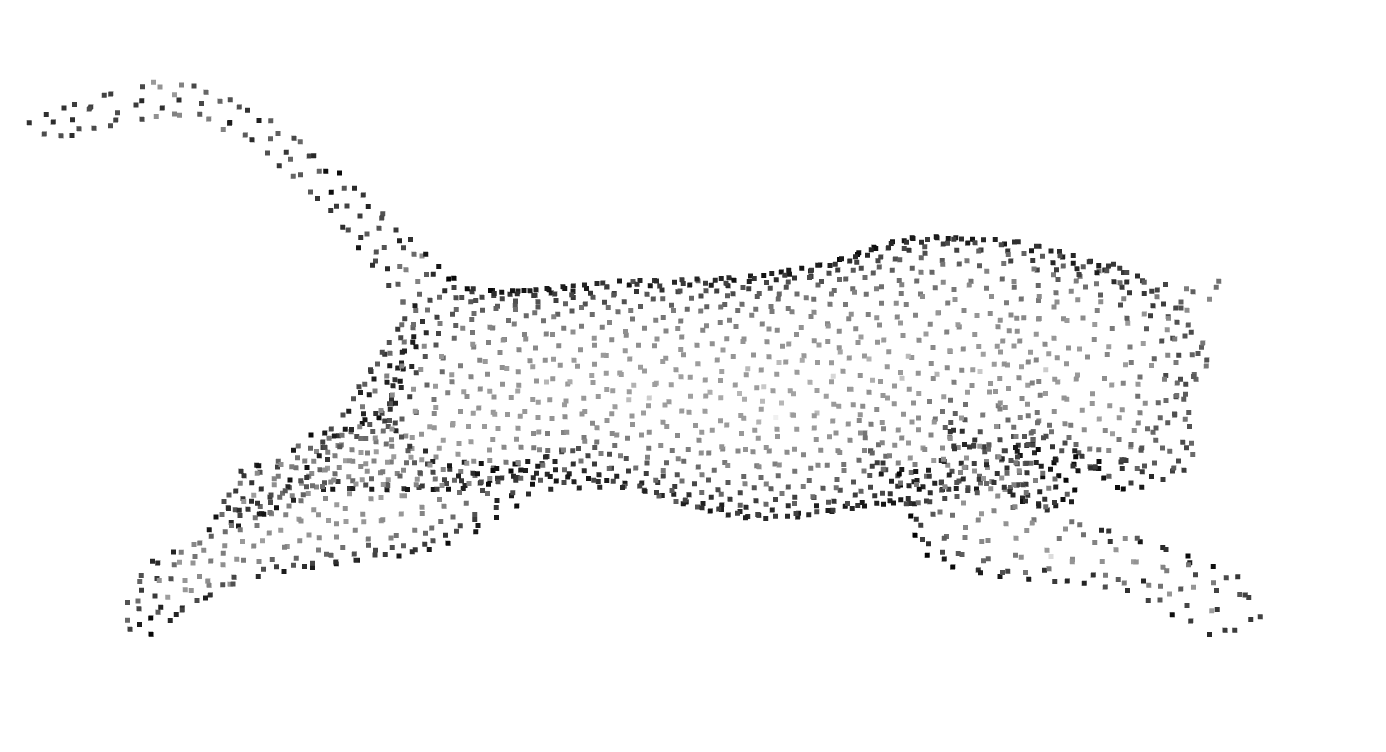} \\
 \multirow{1}{*}[5em]{\rotatebox{90}{Output}} &
 \includegraphics[width=0.31\linewidth]{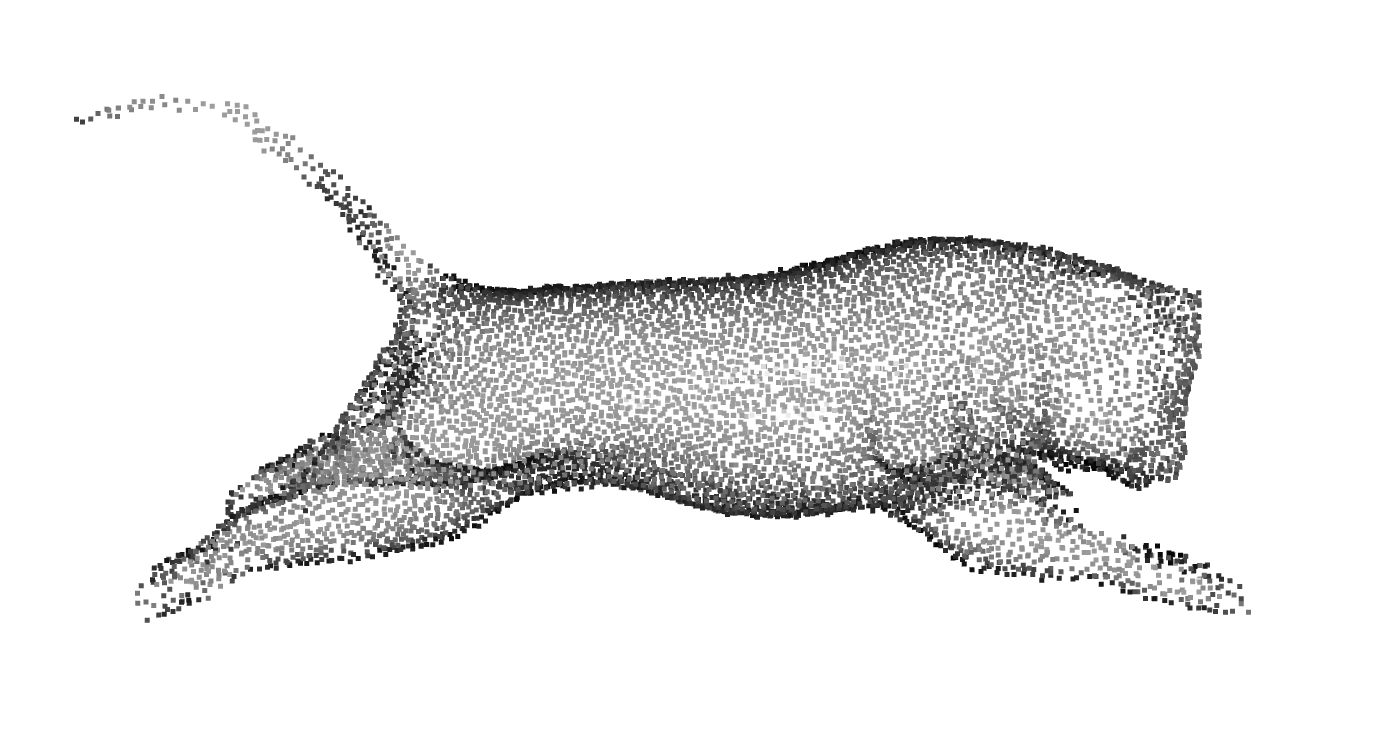} & 
 \includegraphics[width=0.31\linewidth]{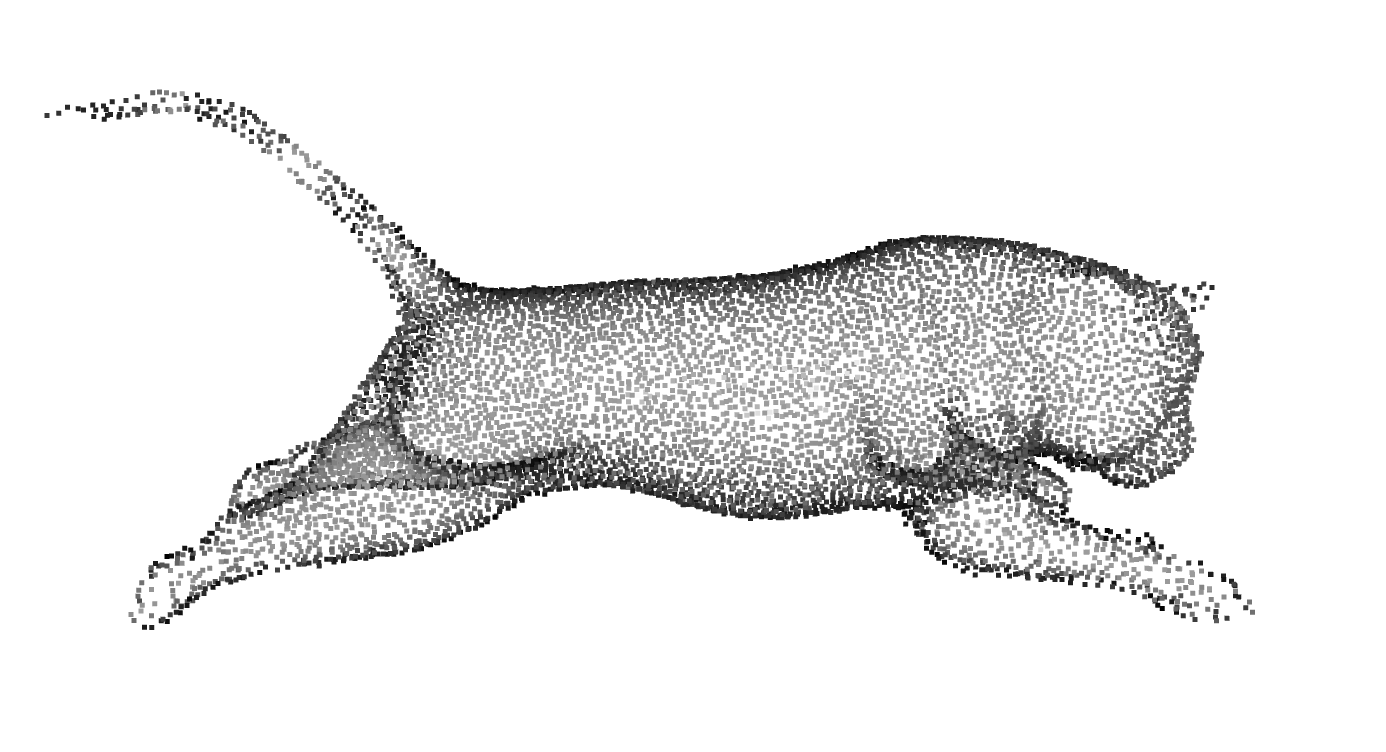} & 
 \includegraphics[width=0.31\linewidth]{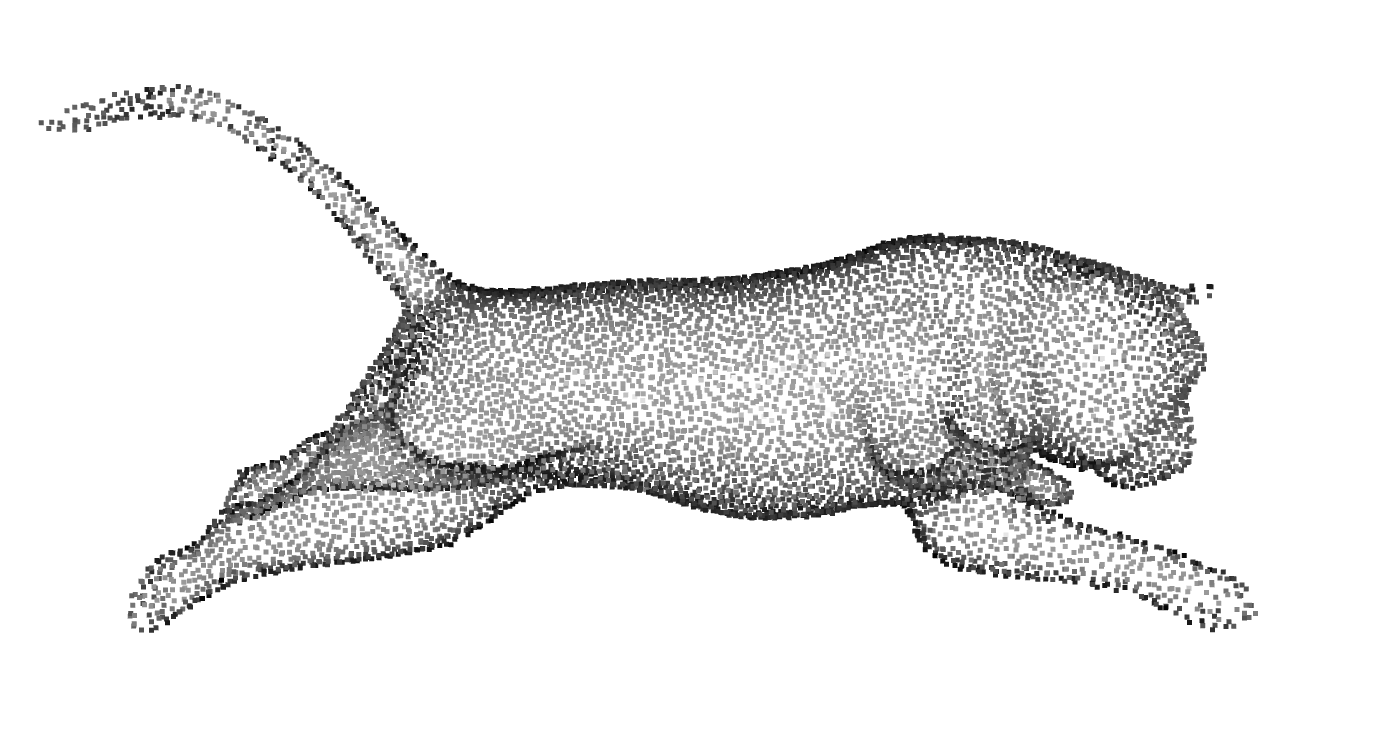}
\end{tabular}
\vspace{0.2cm}
\captionof{figure}{Effect of input point cloud size on the upsampling results while keeping the output size fixed.}
\label{fig:max_back}
\end{table}

\section{Details on SMOG Sampling}

In this section, we discuss the implementation details of the sampling process from the SMOG distribution. More specifically, we have to ensure that the covariance matrix $\mathbf{\Sigma}$ is positive semi-definite, in order for it to be well-defined. The simplest way to impose this condition is to clamp the covariance value $\sigma_{\theta\phi}$ between the following bounds:
\begin{equation}
    -\sigma_\theta \sigma_\phi \leq \sigma_{\theta\phi} \leq \sigma_\theta \sigma_\phi
\end{equation}
To prove this, consider a real symmetric $2 \times 2$ matrix:
\begin{equation}
    A = \begin{bmatrix} a & b \\ b & c \end{bmatrix}
\end{equation}
Such a matrix is positive semi-definite if and only if its eigenvalues are non-negative. The eigenvalues of a $2 \times 2$ matrix can be computed in closed form as the solutions to the following equation:
\begin{equation}
    \det (A - \lambda I) = (a - \lambda)(c - \lambda) - b^2 = 0
\end{equation}
We can immediately apply the well-known formula for solving quadratic equations:
\begin{equation}
    \lambda_{1,2} = \frac{a + c \pm \sqrt{(a+c)^2 - 4(ac - b^2)}}{2}
\end{equation}
The argument of the square root is always non-negative, as it can be rewritten as the sum of two positive numbers:
\begin{equation}
    4b^2 + (a - c)^2 \geq 0
\end{equation}
Since $a > 0$ and $c > 0$ by construction, the first solution is also guaranteed to be non-negative as the sum of three non-negative components:
\begin{equation}
    \lambda_1 = \frac{a + c + \sqrt{(a+c)^2 - 4(ac - b^2)}}{2} \geq 0
\end{equation}
Let's analyze the second solution:
\begin{equation}
    \lambda_2 = \frac{a + c - \sqrt{(a+c)^2 - 4(ac - b^2)}}{2} \geq 0
\end{equation}
It can be rewritten as follows:
\begin{equation}
    a + c \geq \sqrt{(a+c)^2 - 4(ac - b^2)}
\end{equation}
Since both sides of the inequality are non-negative, we can square them and we can expand the right-hand side:
\begin{equation}
    a^2 + 2ac + c^2 \geq a^2 + c^2 - 2ac + 4b^2
\end{equation}
With simple computations, this becomes:
\begin{equation}
    4ac \geq 4 b^2 \implies |b| \leq \sqrt{ac} \implies -\sqrt{ac} \leq b \leq \sqrt{ac} 
\end{equation}
Note that this finding is confirmed by noting that the correlation value $\rho_{\theta\phi}$ is bounded between $[-1, 1]$ and by definition:
\begin{equation}
    \rho_{\theta\phi} = \frac{\sigma_{\theta\phi}}{\sqrt{\sigma_\theta^2 \sigma_\phi^2}} \implies
     -1 \leq \frac{\sigma_{\theta\phi}}{\sqrt{\sigma_\theta^2 \sigma_\phi^2}} \leq 1 \implies -\sigma_\theta \sigma_\phi \leq \sigma_{\theta\phi} \leq \sigma_\theta \sigma_\phi
\end{equation}
From an implementation point of view, we ensure this condition with the \texttt{torch.clamp} function. Even if this naive approach sets the gradients and the related update signal to zero outside the acceptance region, we have found experimentally that very few values lie outside this range during training. An alternative approach for imposing the same constraint is to decompose the covariance matrix as follows:
\begin{equation}
    \mathbf{\Sigma = R(\alpha)DR(\alpha)^\top}
\end{equation}
where $\mathbf{R}(\alpha)$ is the 2D rotation matrix for a given angle $\alpha$ and $\mathbf{D}$ is a diagonal matrix with variances as diagonal elements. The resulting covariance matrix is symmetric and positive definite as required, with the additional advantage of predicting a continuous value for the angle $\alpha$ as the output of the network, instead of clamping and zeroing the gradients. We have implemented and tested this strategy, but we did not notice any significant improvement with respect to the previous approach.

\end{document}